\documentclass{article}

\PassOptionsToPackage{numbers, compress}{natbib}

\usepackage[final]{neurips_2025}

\usepackage[utf8]{inputenc} 
\usepackage[T1]{fontenc}    
\usepackage{hyperref}       
\usepackage{url}            
\usepackage{booktabs}       
\usepackage{amsfonts}       
\usepackage{nicefrac}       
\usepackage{microtype}      
\usepackage{xcolor}         
\usepackage[table]{xcolor}

\usepackage{graphicx}
\usepackage{subcaption}
\usepackage{amsmath}
\usepackage{amssymb}
\usepackage{mathtools}
\usepackage{amsthm}
\usepackage{booktabs} 
\usepackage{multirow}
\usepackage{xspace}
\usepackage{multirow}
\usepackage{pifont}
\usepackage{wrapfig}
\usepackage{caption}
\usepackage{ulem}
\usepackage{siunitx}
\usepackage{array}
\usepackage{algorithm}
\usepackage{algorithmicx}
\usepackage[noend]{algpseudocode}

\usepackage{xcolor}
\usepackage{listings}
\usepackage{upquote}
\usepackage[scaled=0.85]{beramono} 

\lstdefinelanguage{ColabPython}{
    language=Python,
    commentstyle=\color[HTML]{008000},
    numberstyle=\tiny\color[rgb]{0.5,0.5,0.5},
    stringstyle=\color[HTML]{a31515},
    keywordstyle=[1]{\color[HTML]{af00db}},
    keywords=[1]{and, as, assert, break, continue, del, elif, else, except, finally, for, from, global, if, import, in, is, lambda, nonlocal, not, or, pass, raise, return, try, while, with, yield},
    keywordstyle=[2]{\color[HTML]{0000ff}},
    keywords=[2]{class, def},
    keywordstyle=[3]{\color[HTML]{257693}},
    keywords=[3]{abs, all, any, ascii, bin, bool, bytearray, bytes, callable, chr, classmethod, compile, complex, delattr, dict, dir, divmod, enumerate, eval, exec, exit, filter, float, format, frozenset, getattr, globals, hasattr, hash, help, hex, id, input, int, isinstance, issubclass, iter, len, list, locals, map, max, min, next, object, oct, open, ord, pow, print, property, range, repr, reversed, round, set, setattr, slice, sorted, staticmethod, str, sum, super, tuple, type, vars, zip},
}

\lstdefinestyle{colab}{
    language=ColabPython,
    backgroundcolor=\color[HTML]{f7f7f7},   
    basicstyle=\ttfamily\footnotesize,
    breakatwhitespace=false,         
    breaklines=true,                 
    captionpos=b,                    
    keepspaces=true,                 
    numbers=left,                    
    numbersep=5pt,                  
    showspaces=false,                
    showstringspaces=false,
    showtabs=false,                  
    tabsize=2,
    frame=ltb,
    framerule=0pt,
    xleftmargin=0.5em,
    xrightmargin=0em,
    framexleftmargin=0.0em,
    framextopmargin=0.5em,
    framexbottommargin=0.5em,
}

\algnewcommand{\Input}{\item[\textbf{Input:}]}
\algnewcommand{\Output}{\item[\textbf{Output:}]}
\makeatletter
\DeclareRobustCommand\onedot{\futurelet\@let@token\@onedot}
\def\@onedot{\ifx\@let@token.\else.\null\fi\xspace}

\def\eg{{e.g}\onedot} 
\def\ie{{i.e}\onedot}

\def\wrt{w.r.t\onedot}

\makeatother

\title{AccuQuant: Simulating Multiple Denoising Steps for Quantizing Diffusion Models}

\hypersetup{
    colorlinks=true,   
    urlcolor=magenta,
}

\author{%
  \makebox[\textwidth][c]{%
    \small
    Seunghoon Lee\thanks{Equal contribution.}\hspace{1em}%
    Jeongwoo Choi\footnotemark[1]\hspace{1em}%
    Byunggwan Son\hspace{1em}%
    Jaehyeon Moon\hspace{1em}%
    Jeimin Jeon\hspace{1em}%
    Bumsub Ham\thanks{Corresponding author.}%
  }\\[10pt]
  School of Electrical and Electronic Engineering, Yonsei University\\
  \url{https://cvlab.yonsei.ac.kr/projects/AccuQuant}
}

\begin{document}

\maketitle

\begin{abstract}
    We present in this paper a novel post-training quantization~(PTQ) method, dubbed AccuQuant, for diffusion models. We show analytically and empirically that quantization errors for diffusion models are accumulated over denoising steps in a sampling process. To alleviate the error accumulation problem, AccuQuant minimizes the discrepancies between outputs of a full-precision diffusion model and its quantized version within a couple of denoising steps. That is, it simulates multiple denoising steps of a diffusion sampling process explicitly for quantization, accounting the accumulated errors over multiple denoising steps, which is in contrast to previous approaches to imitating a training process of diffusion models, namely, minimizing the discrepancies independently for each step. We also present an efficient implementation technique for AccuQuant, together with a novel objective, which reduces a memory complexity significantly from $\mathcal{O}(n)$ to $\mathcal{O}(1)$, where $n$ is the number of denoising steps. We demonstrate the efficacy and efficiency of AccuQuant across various tasks and diffusion models on standard benchmarks. 
\end{abstract}

\vspace{-3mm}
\section{Introduction}
\label{sec:introduction}
Diffusion models~\cite{ho2020denoising,song2020denoising} have shown the effectiveness for various generation tasks, including text-to-image generation~\cite{rombach2022high,ruiz2023dreambooth}, audio generation~\cite{liu2023audioldm}, and video generation~\cite{blattmann2023align,blattmann2023stable,guo2023animatediff}. In the context of image generation, diffusion models train neural networks to denoise images progressively, corrupted by Gaussian noise, reversing the noise-adding process, and recovering original images. At test time, starting from a random noise, diffusion models perform a sampling process, where the trained neural networks denoise the corrupted image gradually. To generate realistic images, the sampling process typically involves lots of denoising steps, which is computationally demanding. To overcome this problem, many approaches attempt to reduce the number of denoising steps~\cite{meng2023distillation,yin2024one}, providing an efficient image generation process. Another line of research focuses on compressing neural networks themselves~(\eg, using network quantization~\cite{li2023q,huang2024tfmq} or pruning~\cite{castells2024ld}) to reduce the computational cost for each denoising step.

Network quantization lowers bit-widths of full-precision weights and activations into lower ones, enabling a fixed-point computation for efficient inference. There are mainly two approaches to quantizing neural networks: Quantization-aware training~(QAT) and post-training quantization~(PTQ). QAT optimizes network weights and quantization parameters~(\eg,~step-sizes and zero-points) jointly using entire training samples, which is computationally expensive, making it hard to apply QAT for large models~(\eg, ViTs~\cite{dosovitskiy2020image}). PTQ has recently gained significant attention across various models~\cite{nagel2020up,li2021brecq,lv2024ptq4sam,moon2024instance} due to its efficiency. In contrast to QAT, PTQ calibrates quantization parameters only without retraining the network weights, using a small subset of training samples.

\begin{figure}[t]
    \centering
    \vspace{2mm} 
    \includegraphics[width=\textwidth]{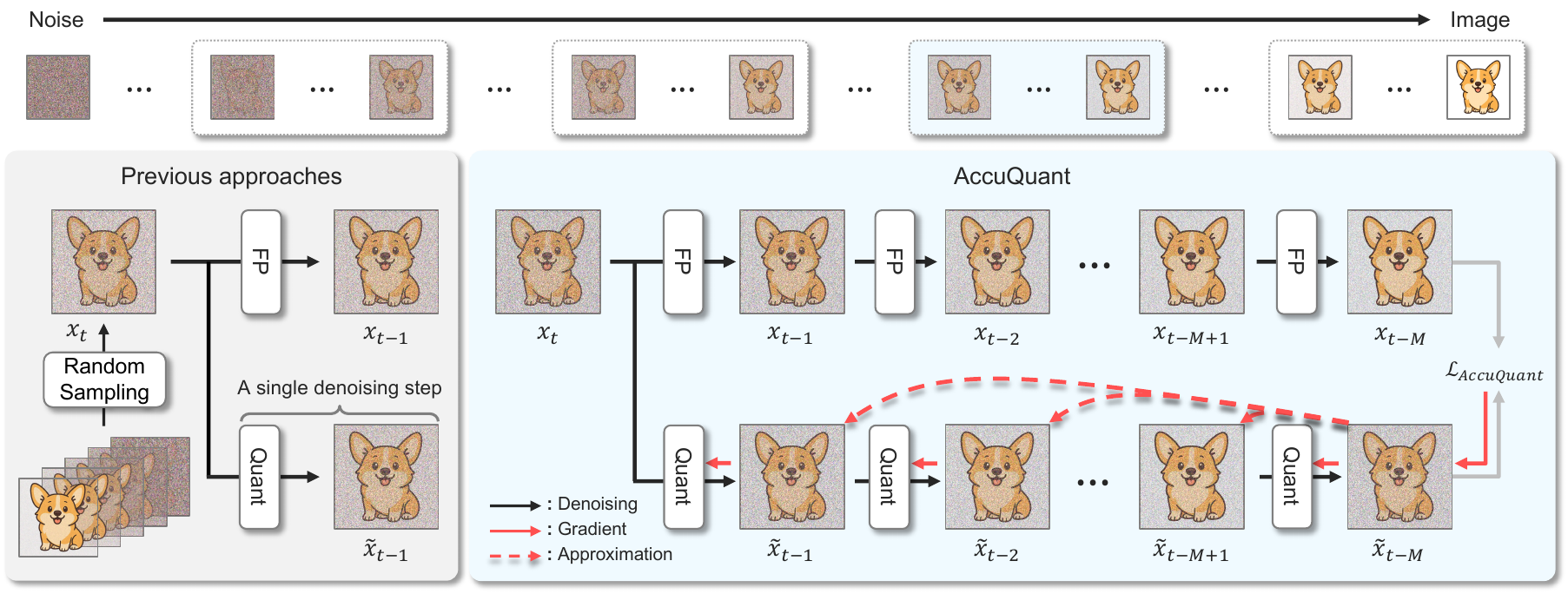}
    \caption{Calibration processes of previous approaches and AccuQuant. \textbf{Left}: Previous methods minimize the quantization error at each denoising step individually, failing to account for accumulated quantization errors during calibration. \textbf{Right}: AccuQuant addresses this problem effectively and efficiently by simulating multiple denoising steps of diffusion models, that is, aligning generated images of full-precision and quantized models over multiple denoising steps, with a memory complexity of~$\mathcal{O}(1)$, independent of the number of steps.}
    \label{fig:teaser}
    \vspace{-3mm}
\end{figure}

Albeit efficient, it is challenging to directly apply PTQ methods, designed for neural networks for discriminative tasks, such as image classification, to diffusion models, since they involve a sequential process to denoise images. As the sampling process goes on, the quantization error at each denoising step is propagated. An overall quantization error can then be split into two parts: Quantization error at the current denoising step and the error accumulated along the previous steps. Most methods~\cite{shang2023post,li2023q,so2024temporal,huang2024tfmq} neglect the accumulated error, and try to minimize the quantization error at each denoising step~(Fig.~\ref{fig:teaser}(\textbf{Left})). Specifically, they exploit the same error-free image as inputs at every step for both full-precision and quantized models, while not considering that the input of the quantized model could contain quantization error accumulated along previous sampling steps. They then calibrate quantization parameters to minimize the difference between the outputs of these models at every step. This is not enough to reduce the overall quantization error, resulting in substantial performance degradation. Although recent works attempt to address the accumulated errors, they are able to handle single-step errors only~\cite{tang2025post} or require additional parameters for error correction~\cite{yao2024timestep,he2023ptqd}.

We introduce in this paper a novel PTQ method for diffusion models, dubbed AccuQuant, that minimizes an overall quantization error, including the ones accumulated over previous denoising steps effectively and efficiently~(Fig.~\ref{fig:teaser}(\textbf{Right})). To this end, AccuQuant groups a couple of denoising steps, and minimizes the difference between the outputs of quantized and full-precision models for each group to calibrate quantization parameters. That is, AccuQuant considers multiple denoising steps in the diffusion process, in contrast to previous approaches to simulating a single denoising step only~\cite{li2023q,he2023ptqd,so2024temporal,huang2024tfmq}, namely, calibrating each step independently. This enables considering the accumulated quantization errors explicitly within each group to minimize an overall quantization error. A naive implementation of AccuQuant, however, requires substantial memory storing intermediate activations for all denoising steps within each group in order to compute gradients, resulting in a memory complexity of $\mathcal{O}(n)$ \wrt the number of denoising steps in each group. To address this, we propose a novel gradient approximation technique, together with a new objective, that reduces the memory complexity significantly from $\mathcal{O}(n)$ to $\mathcal{O}(1)$, enabling applying AccuQuant to large-scale diffusion models efficiently. We show that our approach achieves state-of-the-art performance across various settings, especially in terms of FID2FP32~\cite{tang2025post}, which evaluates how closely the outputs of the quantized model match those of the full-precision model. This suggest that AccuQuant aligns the quantized model with its full-precision counterpart more effectively than previous methods~\cite{li2023q,so2024temporal,huang2024tfmq,he2023ptqd,tang2025post}.

We summarize the main contributions as follows: \vspace{-2mm}
\begin{itemize}
    \item We introduce a novel calibration method for quantizing diffusion models that aligns multiple denoising steps in the sampling processes of quantized and full-precision diffusion models, reducing accumulated quantization errors effectively.
    \item We present a gradient approximation technique for an efficient implementation of AccuQuant, together with a new objective, reducing the memory complexity significantly to $\mathcal{O}(1)$. We also provide a detailed analysis on quantization errors of diffusion models.
    \item We demonstrate the effectiveness of AccuQuant through extensive experiments across various models on standard benchmarks~\cite{heusel2017gans,tang2025post,hessel2022clipscore,zhang2018unreasonable,nash2021generating,salimans2016improved}.
\end{itemize}

\vspace{-2mm}
\section{Related work}
\label{sec:relatedwork}

\label{sec:network_quantization}
\subsection{Quantization for neural networks}
Network quantization reduces the bit-width of weights and/or activations in a neural network. QAT methods~\cite{choi2018pact,yang2019quantization,lee2021network,kim2021distance} simulate the quantization process at training time by converting the full-precision weights/activations into lower-precision representations through a rounding function. This requires retraining the neural network to quantize, which is computationally expensive. On the other hand, PTQ calibrates quantization parameters only with a small number of calibration samples. It quantizes neural networks efficiently, without involving a retraining process, but it is limited to handle outliers in weights/activations. This problem can be alleviated  by clipping the outliers, adopting a outlier channel splitting technique~\cite{zhao2019improving}, or assigning different quantization step sizes for weights/activations with large magnitudes~\cite{cai2017deep,fang2020post}. Recent PTQ methods have demonstrated the effectiveness across various architectures, including CNNs~\cite{liu2023pd,shomron2021post,jeon2022mr} and transformers~\cite{yuan2021ptq4vit,li2023repq,moon2024instance}. In particular, they optimize a rounding function for network weights (\textit{\ie},~determining each weight to be rounded up or down), by exploiting the output differences of each layer~\cite{nagel2020up,wei2022qdrop} or a Hessian-based reconstruction metric~\cite{li2021brecq}, before and after quantization.

\subsection{Quantization for diffusion models}
\label{sec:diffusion_quantization}
Most approaches to quantizing diffusion models adopt a PTQ technique, mainly due to its efficiency. Architectural characteristics of diffusion models make it difficult to directly apply existing PTQ methods, especially for extremely low bit levels. In particular, Q-Diffusion~\cite{li2023q} shows that residual connections in diffusion models, such as U-Net~\cite{ronneberger2015u}, cause significantly different distributions for concatenated activations, and introduces a split quantization technique that performs quantization prior to the concatenation. TFMQ-DM~\cite{huang2024tfmq} shows that previous PTQ methods, which are not designed for diffusion models, could disturb temporal features along denoising steps from original ones, and proposes to quantize temporal embedding layers of diffusion models separately.

Diffusion models apply a denoising operation iteratively over time steps to generate images, providing different distributions of activations across the denoising steps. In order to consider the time-varying characteristics of diffusion models for quantization, training samples in calibration datasets would be carefully chosen. To this end, Q-Diffusion~\cite{li2023q} proposes to sample images uniformly along denoising steps. PTQ4DM~\cite{shang2023post} exploits a skewed normal distribution to sample more images at later denoising steps, which typically provide more realistic images.

Related to ours, PTQD~\cite{he2023ptqd}, TAC~\cite{yao2024timestep} and PCR~\cite{tang2025post} attempt to alleviate the effect of accumulated quantization errors. PTQD~\cite{he2023ptqd} and TAC~\cite{yao2024timestep} analyze the relationship between the outputs of a full-precision network and its quantized counterpart. Assuming that the outputs from these networks are related at each denoising step, they correct the accumulated errors by computing the correlation coefficient and bias~\cite{he2023ptqd} or the reconstruction coefficient and bias~\cite{yao2024timestep}. However, these approaches require additional memory to store these parameters for each denoising setp, and incur computational overhead for the error-correction stage. PCR~\cite{tang2025post} instead tries to reduce the accumulated quantization error directly in a calibration phase. Similar to ours, PCR~\cite{tang2025post} calibrates quantized diffusion model at each denoising step progressively, but it exploits the generated image of a quantized model in a previous step as inputs for both full-precision and quantized models. This could not account for the differences between quantized and full-precision models across multiple denoising steps effectively, mitigating quantization errors within a single denoising step only. On the contrary, our approach exploits separate inputs for full-precision and quantized models. That is, the inputs for full-precision and quantized models come from the corresponding models in a previous denoising step, respectively. This enables minimizing the discrepancies between the quantized and full-precision models across multiple denoising steps, reducing accumulated quantization errors explicitly. Although a memory complexity of our approach is $\mathcal{O}(n)$ \wrt the number of denoising steps, we provide an efficient alternative with a complexity of~$\mathcal{O}(1)$.

\section{Method}
\label{sec:method}

In this section, we briefly describe diffusion models and network quantization~(Sec.\ref{sec:preliminaries}). We then provide an analysis on quantization error in detail~(Sec.\ref{sec:problem_statement}). Finally, we present a detailed description of AccuQuant, including a gradient approximation technique~(Sec.\ref{sec:AccuQuant}).

\subsection{Preliminaries}
\label{sec:preliminaries}

\paragraph{Diffusion models.}
During a forward diffusion process, Gaussian noise $\epsilon$, sampled from a normal distribution $\mathcal{N}(0, 1)$, is added to an input image $x_0$ progressively over time steps. Specifically, a noisy image $x_t$ at step $t$ can be represented as follows: 
\begin{equation}
    x_t = \sqrt{\alpha_t} x_0 + \sqrt{1-\alpha_t} \epsilon, \quad \epsilon\sim\mathcal{N}(0,I),
    \label{eq:diffusion_forward}
\end{equation}
where $\alpha_t$ is a noise scheduling coefficient at the denoising step $t$. Diffusion models reverse this process, gradually removing the noise from the noisy image $x_t$ to recover the original one~$x_0$. To this end, a neural network estimates and removes the noise $\epsilon$ from the corrupted image $x_t$ iteratively, until a clean image is obtained. For example, for a deterministic sampling of the DDIM sampler~\cite{song2020denoising}, a sampled image of $x_{t-1}$ at step $t-1$ is computed as:
\begin{equation}
    x_{t-1}=\sqrt{\alpha_{t-1}}\left( \frac{x_t-\sqrt{1-\alpha_{t-1}}\epsilon_\theta(x_t,t)}{\sqrt{\alpha_{t}}} \right) +\sqrt{1-\alpha_{t-1}}\epsilon_\theta(x_t,t),
    \label{eq:ddim_denoising}
\end{equation}
where we denote by $\epsilon_\theta(x_t,t)$ the noise predicted by the neural network, parameterized by $\theta$, for the image~$x_t$ at step~$t$.

\paragraph{Network quantization.}
Given a floating-point value $v$ and a target bit-width $b$, network quantization converts the value $v$ into a low-precision integer~$\bar{v}$ as follows:
\begin{equation}
    \bar{v} = \text{clip}\left(\text{round}\left(\frac{v}{s}\right)+z,0,2^b-1\right),
    \label{eq:quantization}
\end{equation}
where we denote by $s$ and $z$ a step-size and a zero-point, respectively. $\text{round}(\cdot)$ is a rounding function~(\eg,~nearest rounding or adaptive rounding~\cite{nagel2020up,li2021brecq}), and $\text{clip}(\cdot, v_{\text{min}}, v_{\text{max}})$ is a clipping function that maps an input value within a range of $[v_{\text{min}}, v_{\text{max}}]$. The integer value $\bar{v}$ is then re-scaled to obtain a quantized value $\hat{v}$ as follows:
\begin{equation}
    \hat{v} = s(\bar{v}-z).
    \label{eq:dequantization}
\end{equation} 
Note that the quantization parameters of $s$ and $z$ are trained with all training samples for QAT, while they are computed using a small set of calibration samples for PTQ.

\subsection{Quantization errors for diffusion models}
\label{sec:problem_statement}
In this section, we show that quantized diffusion models suffer from an error accumulation problem, where quantization error at each step is accumulated along the sampling process progressively, degrading the quality of generated images. 

\begin{figure}
    \begin{subfigure}{0.49\textwidth}
        \centering
        \includegraphics[width=0.98\textwidth]{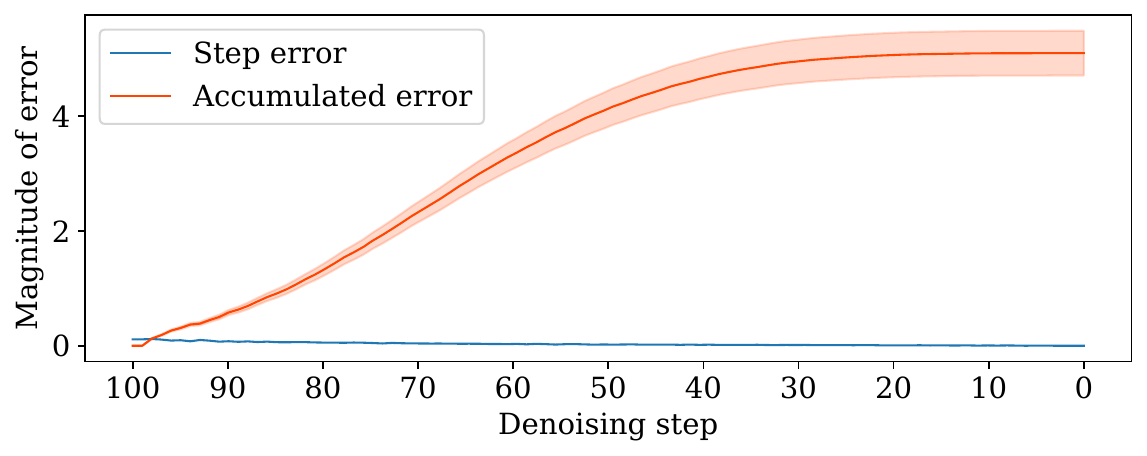}
        \vspace{-2mm}
        \caption{
            Quantization errors on CIFAR-10~\cite{krizhevsky2009learning}.
        }
            \label{fig:accumulatederror_theoretical}
    \end{subfigure}
    \hfill
    \begin{subfigure}{0.49\textwidth}
        \centering
        \includegraphics[width=0.98\textwidth]{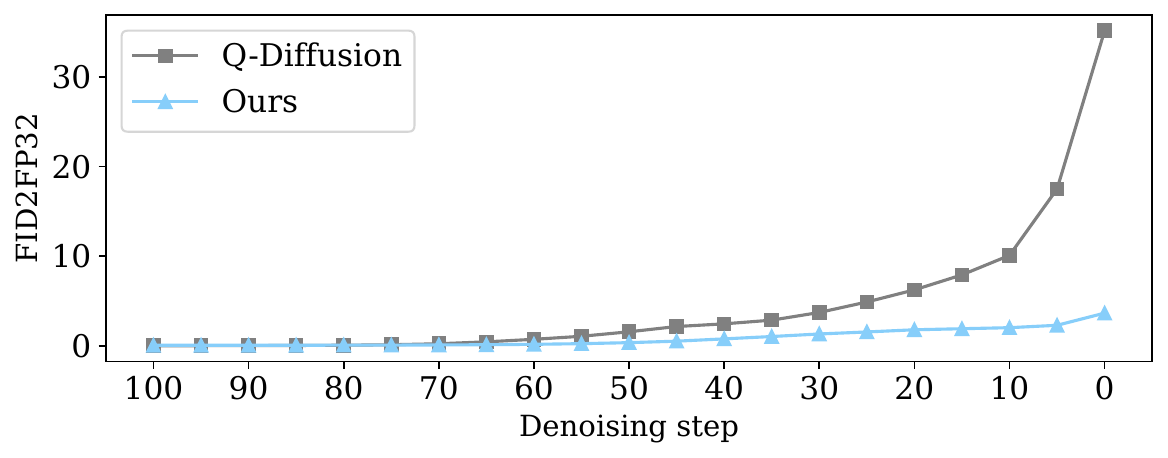}
        \vspace{-2mm}
        \caption{
            FID2FP32~\cite{tang2025post} comparisons on CIFAR-10~\cite{krizhevsky2009learning}.
            }
        \label{fig:fid2fp32}
    \end{subfigure}
    \vspace{-1mm}
    \caption{
        (a) We show the magnitude of the step error $c_t\delta_t$ and the accumulated error $d_t\Delta_t$ at each denoising step. The accumulated error increases drastically, while the step error remains relatively constant, suggesting that reducing the accumulated error is crucial for generating better images, compared to the step error.
        (b) Q-Diffusion accumulates quantization errors according to denoising steps, resulting in a very high FID2FP32 score. On the contrary, our method effectively mitigates the problem,  maintaining a low score.
        }
    \vspace{-4mm}
\end{figure}

Let us denote by $x_t$ and $\epsilon_\theta$ an image and an estimated noise at the denoising step $t$ for a full-precision model, respectively. We define $\tilde{x}_t$ and $\tilde{\epsilon}_\theta$ similarly for a quantized diffusion model. We first categorize the overall quantization error into two parts: a step error $\delta_t$ representing the quantization error for the estimated noise $\tilde{\epsilon}_\theta$ at step~$t$, and an accumulated error $\Delta_t$ representing the accumulated quantization error for the image $\tilde{x}_t$ along all the previous steps. We can then represent $\tilde{x}_t$ and $\tilde{\epsilon}_\theta$ as follows:
\begin{align}
    \tilde{x}_t &= x_t + \Delta_t \label{eq:accumulated_error}, \\
    \tilde{\epsilon}_\theta(\tilde{x}_t,t) &= \epsilon_\theta(x_t,t) + \delta_t.\label{eq:quantization_error}
\end{align}
By plugging Eqs.~(\ref{eq:accumulated_error}) and (\ref{eq:quantization_error}) into Eq.~(\ref{eq:ddim_denoising}), we can compute the output image~$\tilde{x}_{t-1}$ of the quantized model at step of~$t-1$ as follows:\vspace{-2mm}
\begin{equation}
    \tilde{x}_{t-1}=x_{t-1} +c_t \delta_t + d_t\Delta_t,
    \label{eq:ddim_denoising_quantized}
\end{equation}
where
\begin{equation}
	c_t = -\frac{\sqrt{\alpha_{t-1}}\sqrt{1-\alpha_{t}}}{\sqrt{\alpha_{t}}} + \sqrt{1-\alpha_{t-1}}, \quad
	d_t = \frac{\sqrt{\alpha_{t-1}}}{\sqrt{\alpha_{t}}}. \label{eq:d_t}
\end{equation}
We show in Fig.~\ref{fig:accumulatederror_theoretical} the step error of $c_t \delta_t$ and the accumulated error of $d_t \Delta_t$ in Eq.~(\ref{eq:ddim_denoising_quantized}) at each denoising step of DDIM~\cite{song2020denoising} on CIFAR-10~\cite{krizhevsky2009learning}. Specifically, we assume that $\delta_t$ is a Gaussian noise with a zero mean and unit variance, and calculate the accumulated error $\Delta_{t-1}$ recursively,\textit{~\ie},~$\Delta_{t-1}=c_t \delta_t + d_t \Delta_t$, starting from $\Delta_T$ being 0, where $T$ is a total number of denoising steps. We can see that the accumulated error~$d_t \Delta_t$ increases drastically according to denoising steps, while the step error~$c_t \delta_t$ does not. This suggests that the accumulated error for the image~$\tilde{x}_t$ have a much greater impact on the quality of a generated image, compared with the step error for the estimated noise, as the sampling process goes on.

To further validate our observation, we simulate in Fig.~\ref{fig:fid2fp32} the sampling process of DDIM~\cite{song2020denoising} with Q-Diffusion~\cite{li2023q}, and compute FID2FP32~\cite{tang2025post} with generated images from full-precision and quantized models at every 5 steps. Note that FID2FP32~\cite{tang2025post} measures the FID score~\cite{heusel2017gans} with the outputs of quantized and full-precision models, evaluating how closely the quantized model approximates the output from its full-precision counterpart. We can see from the blue line in Fig.~\ref{fig:fid2fp32} that FID2FP32~\cite{tang2025post} increases accordingly along denoising steps, demonstrating once again that quantization errors are accumulated in the sampling process. Note that all diffusion models, generating images through multiple denoising steps, suffer from the error accumulation, similar to DDIM~\cite{song2020denoising} in Fig.~\ref{fig:fid2fp32}, regardless of types of sampling methods.  Therefore, reducing the accumulated error would be a key to maintain the quality of generated images for quantized diffusion models.

\subsection{AccuQuant}
\label{sec:AccuQuant}

To address the error accumulation problem, we introduce a novel PTQ method for quantizing diffusion models, dubbed AccuQuant, that imitates multiple denoising steps of a full-precision diffusion model, reducing the accumulated error effectively. Unlike previous approaches to focusing on individual denoising steps~\cite{shang2023post,li2023q,so2024temporal,huang2024tfmq,yao2024timestep}, our framework simulates the multiple denoising steps in sampling process of full-precision diffusion model for quantization. This enables considering the accumulated error during a calibration process~(\textit{\ie},~optimizing quantization parameters), reducing the error for quantizing diffusion models.

Specifically, we group $M$ consecutive denoising steps, splitting an entire denoising sequence into a total of $T/M$ groups. That is, a denoising step for the $l^{\text{th}}$ group starts with step of~$T-M(l-1)$. Given an image obtained from the full-precision model at step $t$, denoted by~$x_{t}$, we apply a denoising process $M$ times to generate images of $x_{{t}-M}$ and $\tilde{x}_{{t}-M}$ from full-precision and quantized models, respectively, as follows:
\begin{equation}
    x_{t-M} = D_M(x_{t},t), \quad
    \tilde{x}_{t-M} = \tilde{D}_M(x_{t},t;s_l), \label{eq:group_output_quantized}
\end{equation}
where $D_M$ represents a denoising process for the full-precision model over $M$ steps. $\tilde{D}_M$ is defined similarly for the quantized model with the step size of~$s_l$ for quantization. To calibrate the step-size $s_l$ for the $l^{\text{th}}$ group, we minimize the mean squared error between $x_{{t}-M}$ and $\tilde{x}_{{t}-M}$ as follows:
\begin{equation}
    s_l^* = \underset{s_l}{\arg\min}~\mathcal{L}_{\text{MSE}}(x_{t},t;s_l),
    \label{eq:s_l}
\end{equation}
where
\begin{equation}
    \mathcal{L}_{\text{MSE}}(x_{t};s_l;t) = \left\|D_M(x_{t},t) - \tilde{D}_M(x_{t},t;s_l)\right\|_2^2.
    \label{eq:mseloss}
\end{equation}
The calibration process is then repeated sequentially for each group. 
By doing so, AccuQuant accounts for the overall quantization error~(\textit{\ie},~both quantization errors at individual steps and accumulated ones within multiple steps), without introducing additional parameters or extra computational costs during a sampling process.

\begin{wrapfigure}{r}{0.5\textwidth}
    \vspace{-6mm}
    \begin{center}
        \includegraphics[width=0.5\columnwidth]{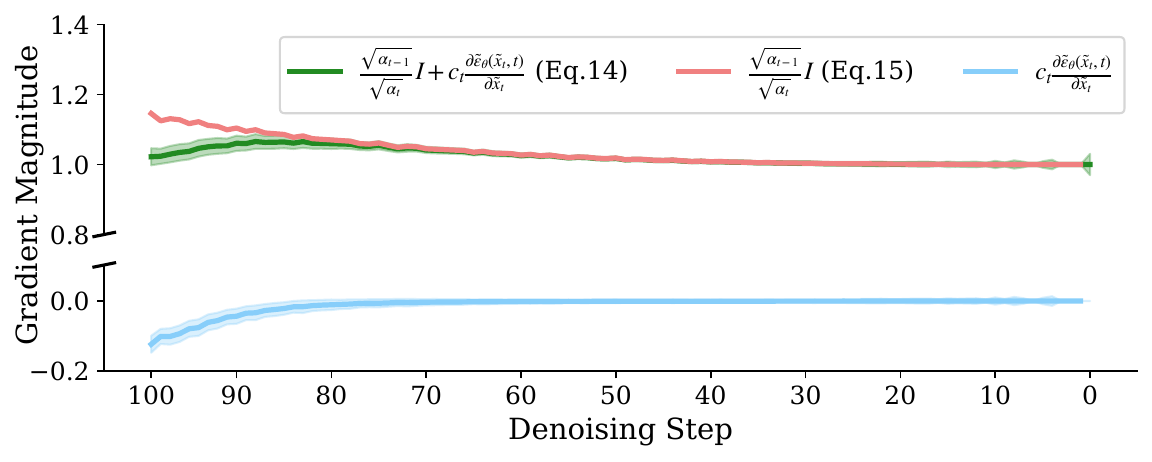}
       \vspace{-5mm}
        \caption{Plots of the gradient in Eq.~\eqref{eq:real_gradient} and its components over denoising steps. The gradients are calculated during a calibration process of AccuQuant for DDIM~\cite{song2020denoising} on CIFAR-10~\cite{krizhevsky2009learning}. Shade indicates the standard deviation.}
        \label{fig:gradient_approximation}
    \end{center}
    \vspace{-14mm}
\end{wrapfigure}

\paragraph{Gradient approximation.}
\label{sec:gradient_approximation}
In order to compute gradients for calibrating quantization parameters, AccuQuant requires lots of memory to store feature maps for all denoising steps within a group. For the $l^{th}$ group, the gradient of the objective $\mathcal{L}_{\text{MSE}}$ \wrt the step-size~$s_l$ is computed as follows:
\begingroup
\small
\begin{align}
    \frac{\partial \mathcal{L}_{\text{MSE}}}{\partial s_l}
    &= \frac{\partial\mathcal{L}_{\text{MSE}}}{\partial\tilde{x}_{{t}-M}}\frac{\partial \tilde{x}_{{t}-M}}{\partial s_l} \nonumber\\
    &=\frac{\partial\mathcal{L}_{\text{MSE}}}{\partial \tilde{x}_{{t}-M}}\left[\sum_{m=1}^{M} g_m\frac{\partial \tilde{x}_{{t}-m}}{\partial s_l}\right],
    \label{eq:group_gradient}
\end{align}
\endgroup
\vspace{1mm}

where $g_m$ is a cumulative product of partial derivatives, across multiple denoising steps, between input and output images, as follows:
\begin{equation}
    g_m = 
    \begin{cases}
        1 &  m = M \\
        \prod_{j=1}^{M-m}\frac{\partial \tilde{x}_{{t}-M-1+j}}{\partial \tilde{x}_{{t}-M+j}}, & m \neq M.
    \end{cases}
    \label{eq:g_m}
\end{equation}
This suggests that we should store all intermediate feature maps, calculated across denoising steps within each group, to compute $g_m$. This results in a memory complexity of $\mathcal{O}(M)$ \wrt the number of denoising steps, which limits the scalability of AccuQuant.

To address this, we propose an efficient implementation technique to reduce the memory complexity. We can represent the gradient between consecutive denoising steps for the quantized diffusion model as follows:
    \begin{equation}
        \frac{\partial \tilde{x}_{t-1}}{\partial \tilde{x}_{t}} = \frac{\sqrt{\alpha_{t-1}}}{\sqrt{\alpha_{t}}}I+c_t\frac{\partial \tilde{\epsilon}_\theta(\tilde{x}_t,t)}{\partial \tilde{x}_{t}}.
        \label{eq:real_gradient}
    \end{equation}
    We have empirically observed in Fig.~\ref{fig:gradient_approximation} that the first term in~Eq.~\eqref{eq:real_gradient} dominates the entire gradient, whereas the second one is negligible (see red and blue lines). Similar findings are reported in~\cite{poole2022dreamfusion,zhang2023redi}. Based on this observation, we approximate the gradient in Eq.~\eqref{eq:real_gradient}:
        \begin{equation}
        \frac{\partial \tilde{x}_{t-1}}{\partial \tilde{x}_{t}} \approx \frac{\sqrt{\alpha_{t-1}}}{\sqrt{\alpha_{t}}}I.
        \label{eq:gradient_approximation}
    \end{equation}
Using the approximation in Eq.~(\ref{eq:gradient_approximation}), we can represent $g_m$ in Eq.~(\ref{eq:g_m}) as:
\begin{equation}
    g_m \approx \frac{\sqrt{\alpha_{t-M}}}{\sqrt{\alpha_{t-m}}}I, \quad \text{for } m = 1,2,\dots,M.
    \label{eq:g_m_approximation}
\end{equation}
That is, we can simplify the gradient, computed by backpropagating through multiple denoising steps, into a scalar ratio of noise scheduling coefficients, reducing the memory complexity from $\mathcal{O}(M)$ to $\mathcal{O}(1)$. 
Using the approximation in Eq.~\eqref{eq:g_m_approximation}, the gradient of the objective in  Eq.~\eqref{eq:group_gradient} can then be represented as follows:
\vspace{-2mm}
\begin{equation}
    \frac{\partial \mathcal{L}_{\text{MSE}}}{\partial s_l} \approx \frac{\partial\mathcal{L}_{\text{MSE}}}{\partial \tilde{x}_{{t}-M}} \left[\sum_{m=1}^{M} \frac{\sqrt{\alpha_{t-M}}}{\sqrt{\alpha_{t-m}}} \frac{\partial \tilde{x}_{{t}-m}}{\partial s_l} \right].
    \label{eq:group_gradient_approx}
\end{equation}
To fully exploit the approximated gradient of Eq.~\eqref{eq:group_gradient_approx}, we introduce a new loss function:
\begin{equation} 
    \mathcal{L}_{\text{AccuQuant}} = \sum_{m=1}^{M} \frac{\sqrt{\alpha_{t-M}}}{\sqrt{\alpha_{t-m}}} \lVert \texttt{sg}(x_{t-M}-\tilde{x}_{t-M}) + \tilde{x}_{t-m} - \texttt{sg}(\tilde{x}_{t-m}) \rVert_2^2,
    \label{eq:AccuQuantloss}
\end{equation}
such that its gradient becomes Eq.~\eqref{eq:group_gradient_approx}. Here, $\texttt{sg}$ is a stop-gradient operator preventing computing gradients. Accordingly, we calibrate the quantization parameters of quantized diffusion models by optimizing the objective of Eq.~\eqref{eq:AccuQuantloss}, while considering the accumulated errors with a memory complexity of $\mathcal{O}(1)$. Detailed derivations for Eqs.~\eqref{eq:g_m_approximation} and~\eqref{eq:AccuQuantloss} and the algorithm of AccuQuant can be found in the section~\ref{appendix:algorithm}.

\begin{table}[!t]
    \sisetup{
    detect-weight   = true,  
    detect-family   = true,   
    mode            = text,  
    }
    \centering
    \scriptsize
    \setlength{\tabcolsep}{2pt}  
    \caption{Quantization results for unconditional image generation on CIFAR-10 ($32\times32$)~\cite{krizhevsky2009learning}, LSUN-Churches ($256\times256$) and LSUN-Bedrooms ($256\times256$)~\cite{yu2015lsun}. $\dagger$: The official implementation of TFMQ-DM uses a 32-bit for certain layers, such as attention modules. We modified these components, such that all layers are quantized with the same bit-width, for fair comparisons with other methods. TAC~\cite{yao2024timestep} does not provide source codes, and thus its performance cannot be measured.}
    \vspace{3mm}
    \label{tab:uncond}
    \begin{subtable}[t]{0.49\textwidth}
        \centering

        \renewcommand{\arraystretch}{1.4}  

        \resizebox{\linewidth}{!}{
        \begin{tabular}{@{}c l c S[table-format=1.2] S[table-format=2.2] S[table-format=2.2]@{}}
        \toprule
        \textbf{Model} & \textbf{Method} & \textbf{Bits(W/A)} & \textbf{IS}$\uparrow$ & \textbf{FID}$\downarrow$ & \textbf{FID2FP32}$\downarrow$ \\
        \midrule
        \multirow{21}{*}{\parbox{1.8cm}{\centering DDIM\\CIFAR-10\\(steps=100)}} 
            & Full-Precision                    & 32/32 &  9.09 &  4.26 &  0.00 \\
        \cmidrule(lr){2-6}
            & Q-Diffusion~\cite{li2023q}        &  6/6  &  8.63 & 30.46 & 35.24 \\
            & TFMQ-DM{$^\dagger$}~\cite{huang2024tfmq}      &  6/6  &  8.94 &  7.84 &  3.64 \\
            & \cellcolor[HTML]{ECF9FF}Ours                  &  \cellcolor[HTML]{ECF9FF}6/6  & \cellcolor[HTML]{ECF9FF}\bfseries\tablenum{9.18} &\cellcolor[HTML]{ECF9FF}\bfseries\tablenum{5.79} &\cellcolor[HTML]{ECF9FF}\bfseries\tablenum{3.30} \\
        \cmidrule(lr){2-6}
            & Q-Diffusion~\cite{li2023q}        &  4/8  &  9.12 &  4.93 &  3.26 \\
            & TFMQ-DM{$^\dagger$}~\cite{huang2024tfmq}      &  4/8  &  8.89 &  5.52 &  1.52 \\
            & TAC~\cite{yao2024timestep}        &  4/8  & \bfseries\tablenum{9.15} &  4.89 &    {{--}} \\
            & \cellcolor[HTML]{ECF9FF}Ours                  &  \cellcolor[HTML]{ECF9FF}4/8  & \cellcolor[HTML]{ECF9FF}9.06 &\cellcolor[HTML]{ECF9FF}\bfseries\tablenum{4.75} &\cellcolor[HTML]{ECF9FF}\bfseries\tablenum{1.15} \\
        \cmidrule(lr){2-6}
            & Q-Diffusion~\cite{li2023q}        &  4/6  & \bfseries\tablenum{9.47} & 25.35 & 30.66 \\
            & TFMQ-DM{$^\dagger$}~\cite{huang2024tfmq}      &  4/6  &  9.15 & 11.04 &  7.39 \\
            & \cellcolor[HTML]{ECF9FF}Ours                  &  \cellcolor[HTML]{ECF9FF}4/6  & \cellcolor[HTML]{ECF9FF}8.99 & \cellcolor[HTML]{ECF9FF}\bfseries\tablenum{7.07} &\cellcolor[HTML]{ECF9FF}\bfseries\tablenum{3.94} \\
        \cmidrule(lr){2-6}
            & Q-Diffusion~\cite{li2023q}        &  3/8  &  8.53 & 17.31 & 13.94 \\
            & TFMQ-DM{$^\dagger$}~\cite{huang2024tfmq}      &  3/8  &  8.40 & 28.47 & 26.96 \\
            & TAC~\cite{yao2024timestep}        &  3/8  & \bfseries\tablenum{8.86} & 9.55 &    {{--}} \\
            & \cellcolor[HTML]{ECF9FF}Ours                  &  \cellcolor[HTML]{ECF9FF}3/8  & \cellcolor[HTML]{ECF9FF}8.76 & \cellcolor[HTML]{ECF9FF}\bfseries\tablenum{9.03} & \cellcolor[HTML]{ECF9FF}\bfseries\tablenum{5.15} \\
        \cmidrule(lr){2-6}
            & Q-Diffusion~\cite{li2023q}        &  3/6  &  7.51 & 40.94 & 42.74 \\
            & TFMQ-DM{$^\dagger$}~\cite{huang2024tfmq}      &  3/6  &  8.34 & 29.29 & 27.11 \\
            & TAC~\cite{yao2024timestep}        &  3/6  &  8.27 & 31.88 &    {{--}} \\
            & \cellcolor[HTML]{ECF9FF}Ours                  &  \cellcolor[HTML]{ECF9FF}3/6  & \cellcolor[HTML]{ECF9FF}\bfseries\tablenum{8.65} &\cellcolor[HTML]{ECF9FF}\bfseries\tablenum{9.89} &\cellcolor[HTML]{ECF9FF}\bfseries\tablenum{6.69} \\
        \bottomrule
        \end{tabular}
        }
    \end{subtable}%
    \hfill
    \begin{subtable}[t]{0.49\textwidth}
        \centering

        {\tiny                   
        \renewcommand{\arraystretch}{1.035} 
        \resizebox{\linewidth}{!}{
        \begin{tabular}{@{}c l c S S S@{}}
        \toprule
        \textbf{Model} & \textbf{Method} & \textbf{Bits(W/A)} & \textbf{FID}$\downarrow$ & \textbf{sFID}$\downarrow$ & \textbf{FID2FP32}$\downarrow$ \\ 
        \midrule
        \multirow{21}{*}{\parbox{1.5cm}{\centering LDM-8\\LSUN-Churches\\(steps=500)}} 
            & Full-Precision                    & 32/32 &  3.91 & 10.10 &  0.00 \\
        \cmidrule(lr){2-6}
            & Q-Diffusion~\cite{li2023q}        &  4/8  &  4.49 & \bfseries\tablenum{10.36} &  0.79 \\
            & TFMQ-DM{$^\dagger$}~\cite{huang2024tfmq}      &  4/8  &  8.43 & 19.83         & 10.14 \\
            & TAC~\cite{yao2024timestep}        &  4/8  & \bfseries\tablenum{3.81} & {--}            & {--}    \\
            & \cellcolor[HTML]{ECF9FF}Ours                  &  \cellcolor[HTML]{ECF9FF}4/8  &  \cellcolor[HTML]{ECF9FF}3.99 & \cellcolor[HTML]{ECF9FF}10.67         & \cellcolor[HTML]{ECF9FF}\bfseries\tablenum{0.66} \\
        \cmidrule(lr){2-6}
            & Q-Diffusion~\cite{li2023q}        &  4/6  & 45.24 & 25.96         & 40.58 \\
            & TFMQ-DM{$^\dagger$}~\cite{huang2024tfmq}      &  4/6  &  8.60 & 20.77         & 10.07 \\
            & \cellcolor[HTML]{ECF9FF}Ours                  &  \cellcolor[HTML]{ECF9FF}4/6  & \cellcolor[HTML]{ECF9FF}\bfseries\tablenum{7.47} & \cellcolor[HTML]{ECF9FF}\bfseries\tablenum{11.07}& \cellcolor[HTML]{ECF9FF}\bfseries\tablenum{4.84} \\
        \cmidrule(lr){2-6}
            & Q-Diffusion~\cite{li2023q}        &  3/8  &  5.68 & 11.25         &  1.49 \\
            & TFMQ-DM{$^\dagger$}~\cite{huang2024tfmq}      &  3/8  &  9.03 & 21.77         & 10.48 \\
            & TAC~\cite{yao2024timestep}        &  3/8  &  7.78 & {--}            & {--}    \\
            & \cellcolor[HTML]{ECF9FF}Ours                  &  \cellcolor[HTML]{ECF9FF}3/8  & \cellcolor[HTML]{ECF9FF}\bfseries\tablenum{5.04} & \cellcolor[HTML]{ECF9FF}\bfseries\tablenum{10.95}& \cellcolor[HTML]{ECF9FF}\bfseries\tablenum{1.26} \\
        \cmidrule(lr){2-6}
            & Q-Diffusion~\cite{li2023q}        &  3/6  &  47.49 & 27.97        &  42.27 \\
            & TFMQ-DM{$^\dagger$}~\cite{huang2024tfmq}      &  3/6  &  9.84  & 14.20        & 10.19 \\
            & \cellcolor[HTML]{ECF9FF}Ours                  &  \cellcolor[HTML]{ECF9FF}3/6  & \cellcolor[HTML]{ECF9FF}\bfseries\tablenum{8.91} & \cellcolor[HTML]{ECF9FF}\bfseries\tablenum{12.58}& \cellcolor[HTML]{ECF9FF}\bfseries\tablenum{5.79} \\
        \midrule
        \multirow{11}{*}{\parbox{1.5cm}{\centering LDM-4\\LSUN-Bedrooms\\(steps=200)}} 
            & Full-Precision                    & 32/32 &  3.04 &  7.07 & 0.00  \\
        \cmidrule(lr){2-6}
            & Q-Diffusion~\cite{li2023q}        &  4/8  &  5.46 &  7.92 & 1.92  \\
            & TFMQ-DM{$^\dagger$}~\cite{huang2024tfmq}      &  4/8  &  7.33 & 12.68 & 4.59  \\
            & TAC~\cite{yao2024timestep}        &  4/8  &  4.94 & {--}    & {--}    \\
            & \cellcolor[HTML]{ECF9FF}Ours                  &  \cellcolor[HTML]{ECF9FF}4/8  & \cellcolor[HTML]{ECF9FF}\bfseries\tablenum{4.03} & \cellcolor[HTML]{ECF9FF}\bfseries\tablenum{7.81} & \cellcolor[HTML]{ECF9FF}\bfseries\tablenum{1.07} \\
        \cmidrule(lr){2-6}
            & Q-Diffusion~\cite{li2023q}        &  3/8  & 11.98 & 14.02 & 7.54  \\
            & TFMQ-DM{$^\dagger$}~\cite{huang2024tfmq}      &  3/8  & 11.32 & 11.62 & 7.64  \\
            & TAC~\cite{yao2024timestep}        &  3/8  &  5.14 & {--}    & {--}    \\
            & \cellcolor[HTML]{ECF9FF}Ours                  &  \cellcolor[HTML]{ECF9FF}3/8  & \cellcolor[HTML]{ECF9FF}\bfseries\tablenum{4.18} & \cellcolor[HTML]{ECF9FF}\bfseries\tablenum{10.45} &\cellcolor[HTML]{ECF9FF} \bfseries\tablenum{1.14} \\
        \bottomrule
        \end{tabular}
        }
        } 

    \end{subtable}

\end{table}

\section{Experiments}
\label{sec:experiments}
We describe in this section implementation details~(Sec.\ref{sec:details}), and compare our approach with other quantization methods quantitatively and qualitatively~(Sec.\ref{sec:results}). We then present a detailed analysis of our method~(Sec.\ref{sec:results}). More quantitative and qualitative results can be found in the appendix.

\subsection{Implementation details}
\label{sec:details}

\paragraph{Datasets and models.}
We apply AccuQuant to various diffusion models and perform extensive experiments on standard benchmarks for unconditional, class-conditional, and text-to-image generation tasks. For the unconditional generation task, we exploit DDIM~\cite{song2020denoising} on CIFAR-10~\cite{krizhevsky2009learning}, and Latent Diffusion Model~(LDM)~\cite{rombach2022high} on LSUN-Bedrooms and LSUN-Churches~\cite{yu2015lsun}. For class-conditional generation, we perform experiments using LDM~\cite{rombach2022high} on ImageNet~\cite{deng2009imagenet}. We use Stable Diffusion~(SD) v1.4~\cite{rombach2022high} on MS-COCO~\cite{lin2014microsoft} for text-to-image generation.

\vspace{-2mm}

\paragraph{Quantization.}
We employ adaptive rounding~\cite{nagel2020up,li2021brecq} for weight quantizers, following prior approaches~\cite{shang2023post,li2023q,huang2024tfmq}. For activation quantization, we split a denoising process into 20 groups for unconditional image generation, 10 groups for class-conditional image generation and 25 groups for text-to-image generation. We perform the calibration process for 50, 20, and 10 epochs on DDIM~\cite{song2020denoising}, LDM, and SD~\cite{rombach2022high}, respectively, using the Adam optimizer~\cite{kingma2014adam}. Following the work of~\cite{li2023q}, we generate 256 calibration samples for each group with full-precision models, maintaining the total number of the samples consistent with that of~\cite{li2023q} across all experiments.  More detailed settings can be found in the section~\ref{sec:details_appendix}.
\vspace{-2mm}

\paragraph{Evaluation metrics.}
Following the previous approaches~\cite{li2023q,huang2024tfmq,yao2024timestep}, we evaluate 50K images for unconditional/class-conditional generation with Inception Score~(IS)~\cite{salimans2016improved}, Frechet Inception Distance~(FID)~\cite{heusel2017gans}, and sFID~\cite{nash2021generating}. We also measure PSNR, SSIM, and LPIPS~\cite{zhang2018unreasonable} between images generated by full-precision and quantized models for class-conditional image generation. For text-to-image generation, we generate 5K images, and evaluate them with FID~\cite{heusel2017gans} and CLIP scores~\cite{hessel2022clipscore}, following the work of~\cite{tang2025post}. The ViT-L/14 is used to compute CLIP scores~\cite{hessel2022clipscore}. We also use FID2FP32~\cite{tang2025post} computing FID of generated images by quantized models, but with the images from full-precision models as a reference, measuring accumulated errors from quantized models for all experiments.
\vspace{-2mm}
\begin{table}[!t]
    \sisetup{
      detect-weight   = true,  
      detect-family   = true,   
      mode            = text,  
    }
    \scriptsize
    \setlength{\tabcolsep}{3pt}  
    \caption{Quantization results for class-conditional image generation on ImageNet ($256\times256$)~\cite{deng2009imagenet}.}
    \vspace{2mm}
    \label{tab:class_cond}
    \centering
    \resizebox{0.9\linewidth}{!}{
      \begin{tabular}{@{}c l c S[table-format=2.2] S[table-format=2.2] S[table-format=3.2] S[table-format=2.2] S[table-format=1.4] S[table-format=2.4] S[table-format=1.4]@{}}
        \toprule
        \textbf{Model} & \textbf{Method} & \textbf{Bits(W/A)} & \textbf{FID}$\downarrow$ & \textbf{sFID}$\downarrow$ & \textbf{IS}$\uparrow$ & \textbf{FID2FP32}$\downarrow$ & \textbf{LPIPS}$\downarrow$ & \textbf{PSNR}$\uparrow$ & \textbf{SSIM}$\uparrow$ \\
        \midrule
        \multirow{15}{*}{\parbox{1.2cm}{\centering LDM-4\\ImageNet\\(steps=20)}}
          & Full-Precision                   & 32/32 & 11.13               &  7.85               & 368.19              &  0.00               & 0.00         & 0.00         & 0.00        \\
        \cmidrule(lr){2-10}
        & Q-Diffusion~\cite{li2023q}    &  4/8  &  9.55               & 13.50               & 339.44      &  1.71               & 0.1912    & 23.6703   & 0.8391   \\
        & PTQD~\cite{he2023ptqd}        &  4/8  &  9.99              & 8.43               & \bfseries\tablenum{361.95}     &  0.84               & 0.1598    & 24.0672   & 0.8614   \\  
        & TFMQ-DM~\cite{huang2024tfmq}     &  4/8  &  9.80               &  \bfseries\tablenum{7.19}               & 358.38     &  1.25               & 0.1702    & 23.6309   & 0.8567   \\
          & \cellcolor[HTML]{ECF9FF}Ours                              & \cellcolor[HTML]{ECF9FF}4/8  & \cellcolor[HTML]{ECF9FF}\cellcolor[HTML]{ECF9FF}\bfseries\tablenum{9.39}       & \cellcolor[HTML]{ECF9FF}7.41      & \cellcolor[HTML]{ECF9FF}356.48              & \cellcolor[HTML]{ECF9FF}\bfseries\tablenum{0.65}       &\cellcolor[HTML]{ECF9FF}\bfseries\tablenum{0.1585} & \cellcolor[HTML]{ECF9FF}\bfseries\tablenum{24.4338} & \cellcolor[HTML]{ECF9FF}\bfseries\tablenum{0.8713} \\
        \cmidrule(lr){2-10}
        & Q-Diffusion~\cite{li2023q}       &  3/8  &  7.57               & 12.43               & 266.65              & 11.26               & 0.3741    & 18.4588   & 0.6532   \\
        & PTQD~\cite{he2023ptqd}        &  3/8  &  7.75              & 9.51               & \bfseries\tablenum{308.47}     &  6.30               & 0.3446    & 19.2485   & 0.6728   \\    
        & TFMQ-DM~\cite{huang2024tfmq}     &  3/8  &  7.96               &  8.77               & 295.46              &  4.47               & 0.3942    & 16.8180   & 0.7200   \\
          & \cellcolor[HTML]{ECF9FF}Ours                              & \cellcolor[HTML]{ECF9FF}3/8  & \cellcolor[HTML]{ECF9FF}\bfseries\tablenum{6.61}       & \cellcolor[HTML]{ECF9FF}\bfseries\tablenum{8.65}       & \cellcolor[HTML]{ECF9FF}301.06     & \cellcolor[HTML]{ECF9FF}\bfseries\tablenum{3.86}       & \cellcolor[HTML]{ECF9FF}\bfseries\tablenum{0.2937} & \cellcolor[HTML]{ECF9FF}\bfseries\tablenum{19.9392} & \cellcolor[HTML]{ECF9FF}\bfseries\tablenum{0.7649} \\
        \cmidrule(lr){2-10}
        & Q-Diffusion~\cite{li2023q}       &  3/6  & 28.83               & 39.30               &  99.55              & 41.25               & 0.4666    & 17.9855   & 0.6358   \\
        & PTQD~\cite{he2023ptqd}          &  3/6  &  9.15              & 11.42               & 244.43     &  13.85               & 0.3852    & 18.1508   & 0.6662   \\    
        & TFMQ-DM~\cite{huang2024tfmq}     & 3/6  &  8.17               & 10.71               & 237.32              & 10.06               & 0.3928    & 17.4886   & 0.7183   \\
          & \cellcolor[HTML]{ECF9FF}Ours                              & \cellcolor[HTML]{ECF9FF}3/6  & \cellcolor[HTML]{ECF9FF}\cellcolor[HTML]{ECF9FF}\bfseries\tablenum{5.95}       & \cellcolor[HTML]{ECF9FF}\cellcolor[HTML]{ECF9FF}\bfseries\tablenum{8.61}       & \cellcolor[HTML]{ECF9FF}\bfseries\tablenum{282.87}     & \cellcolor[HTML]{ECF9FF}\bfseries\tablenum{6.31}       & \cellcolor[HTML]{ECF9FF}\bfseries\tablenum{0.3296} & \cellcolor[HTML]{ECF9FF}\bfseries\tablenum{19.0135} & \cellcolor[HTML]{ECF9FF}\bfseries\tablenum{0.7296} \\
        \bottomrule
      \end{tabular}
    }
  \vspace{-2mm}

  \end{table}
\begin{table}[!t]
    \sisetup{
      detect-weight   = true,  
      detect-family   = true,   
      mode            = text,  
    }
    \centering
    \scriptsize
    \setlength{\tabcolsep}{6pt}  
    \caption{Quantization results for text-to-image generation on MS-COCO ($512\times512$)~\cite{lin2014microsoft}. MP:~Mixed-precision quantization.}
    \vspace{2mm}
    \label{tab:text_cond}
    \resizebox{0.9\linewidth}{!}{
      \begin{tabular}{@{}c l c S S S@{}}
        \toprule
        \textbf{Model} & \textbf{Method} & \textbf{Bits(W/A)} & \textbf{FID}$\downarrow$ & \textbf{FID2FP32}$\downarrow$ & \textbf{CLIP Score}$\uparrow$ \\
        \midrule
        \multirow{6}{*}{\parbox{2.2cm}{\centering Stable Diffusion v1.4\\MS-COCO\\(steps=50)}}
          & Full-Precision   & 32/32        & 27.50            & 0.00            & 26.46            \\
        \cmidrule(lr){2-6}
          & Q-Diffusion~\cite{li2023q}      & 4/8          & 27.87            & 20.42           & 26.15            \\
          & PTQ4DM~\cite{shang2023post}           & 4/8          & 25.64            & 17.73           & 26.25            \\
          & PCR~\cite{tang2025post}              & 4/8.1 (MP)   & 23.86            & 14.39           & 26.35            \\
          & PCR~\cite{tang2025post}              & 4/8.4 (MP)   & \bfseries\tablenum{22.04}   & 14.25           & 26.48            \\
          & \cellcolor[HTML]{ECF9FF}Ours             & \cellcolor[HTML]{ECF9FF}4/8          & \cellcolor[HTML]{ECF9FF}22.73            & \cellcolor[HTML]{ECF9FF}\bfseries\tablenum{10.99}  & \cellcolor[HTML]{ECF9FF}\bfseries\tablenum{26.85}   \\
        \bottomrule
      \end{tabular}
    }
  \vspace{-2mm}

  \end{table}
\subsection{Results}
\label{sec:results}
{\textbf{Quantitative results.}}  We show in Tables~\ref{tab:uncond}-\ref{tab:text_cond} quantitative comparisons of our method and the state of the art~\cite{li2023q,huang2024tfmq,yao2024timestep} for unconditional image generation~(\textit{\ie}, CIFAR-10~\cite{krizhevsky2009learning}, LSUN-Bedrooms, and LSUN-Churches~\cite{yu2015lsun}), class-conditional image generation~(\textit{\ie}, ImageNet~\cite{deng2009imagenet}) and text-to-image generation~(\textit{\ie} MS-COCO~\cite{lin2014microsoft}), respectively. For fair comparison, we report the results of TFMQ-DM~\cite{huang2024tfmq}, with official source codes provided by authors, under the same settings as ours. We refer to the results from the papers for other methods that do not provide official implementations. 

We summarize our findings as follows: 
(1) AccuQuant outperforms all previous approaches~\cite{shang2023post,li2023q,he2023ptqd,huang2024tfmq,yao2024timestep,yao2024timestep}, specially designed to quantize diffusion models, by significant margins in terms of FID2FP32~\cite{tang2025post}. Specifically, it provides better results than PCR~\cite{tang2025post}, even with lower bit-widths. This demonstrates that AccuQuant reduces accumulated errors effectively, and better maintains the behavior of full-precision models compared to other methods.
(2) AccuQuant achieves significant performance gains, especially in low-bit settings for activation quantization. Note that low-bit settings are more vulnerable to accumulated errors, due to the limited representational capacity. This demonstrates the robustness of AccuQuant and its ability to maintain high performance, even under constrained conditions.
(3) AccuQuant outperforms the state-of-the-art methods~\cite{li2023q,huang2024tfmq,yao2024timestep} in various standard benchmarks, verifying that considering multiple denoising steps in a sampling process is effective to reduce accumulated errors for quantizing diffusion models.

\begin{figure*}[!t]
   \vspace{2mm}
   \captionsetup[subfigure]{font=small, labelformat=empty}
   \begin{center}
   \begin{subfigure}[c]{0.24\linewidth}
      \centering
      \begin{tabular}{c @{\hspace{0.5mm}} c @{\hspace{0.5mm}} c}
         \includegraphics[width=0.3\columnwidth]{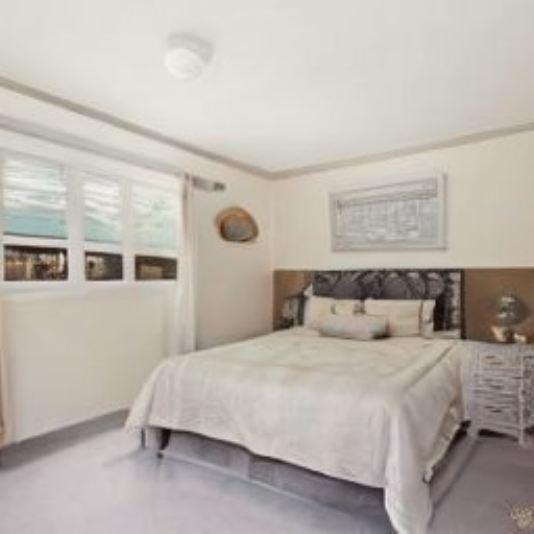}&
         \includegraphics[width=0.3\columnwidth]{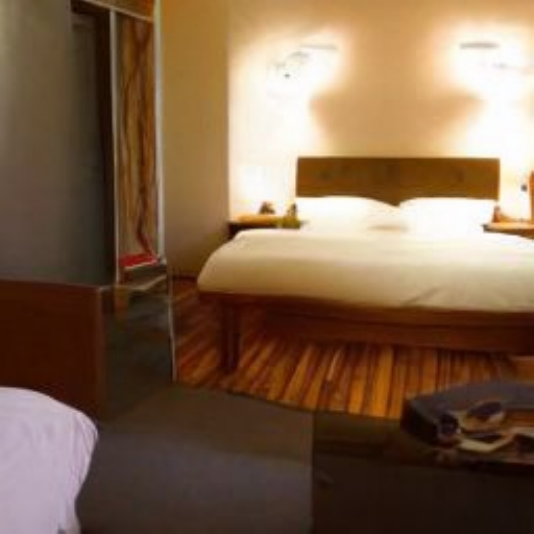}&
         \includegraphics[width=0.3\columnwidth]{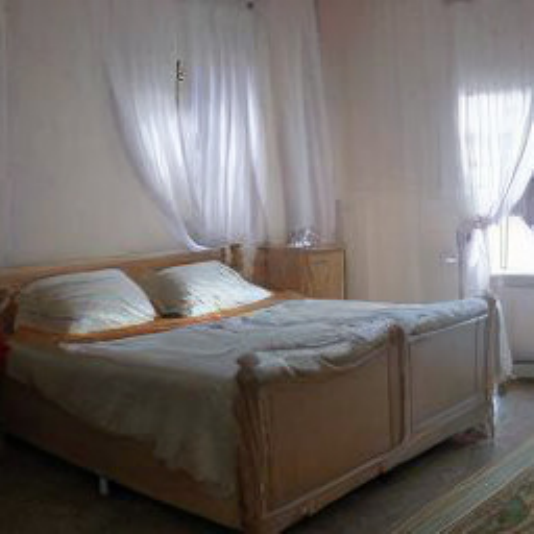}\\
         \includegraphics[width=0.3\columnwidth]{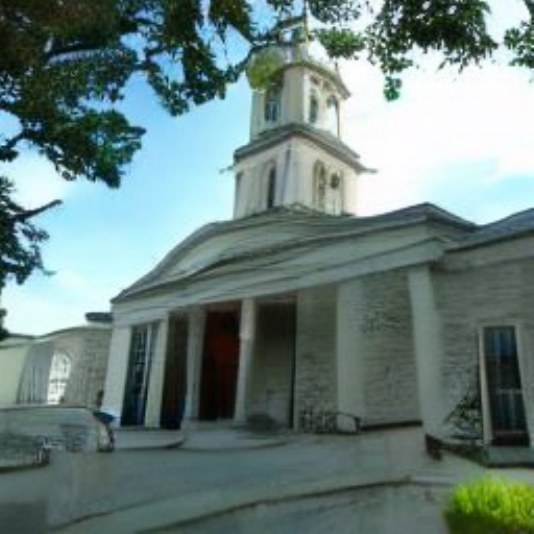}&
         \includegraphics[width=0.3\columnwidth]{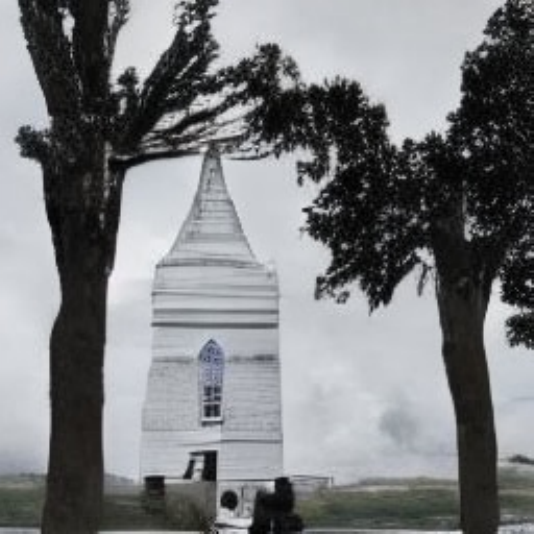}&
         \includegraphics[width=0.3\columnwidth]{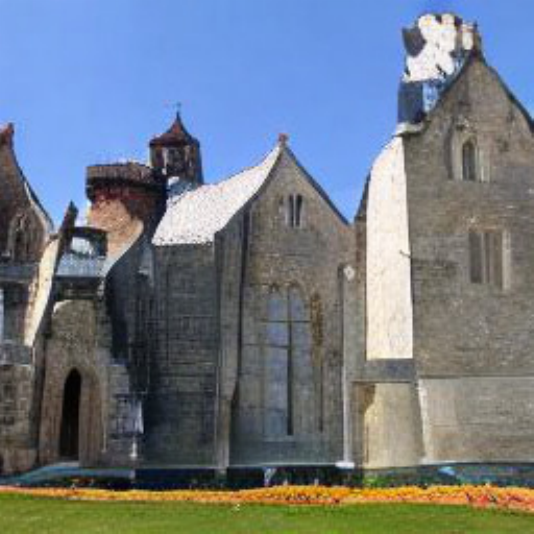}\\
      \end{tabular}
      \label{fig:lsun_fp}
      \vspace{-3mm}
      \caption{Full precision}
   \end{subfigure}
   \begin{subfigure}[c]{0.24\linewidth}
      \centering
      \begin{tabular}{c @{\hspace{0.5mm}} c @{\hspace{0.5mm}} c}
         \includegraphics[width=0.3\columnwidth]{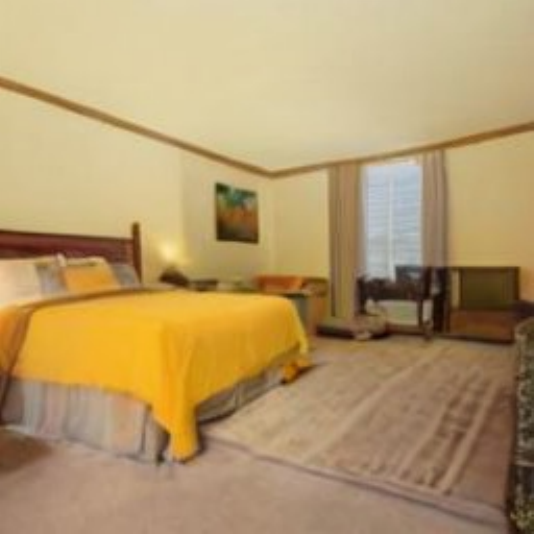}&
         \includegraphics[width=0.3\columnwidth]{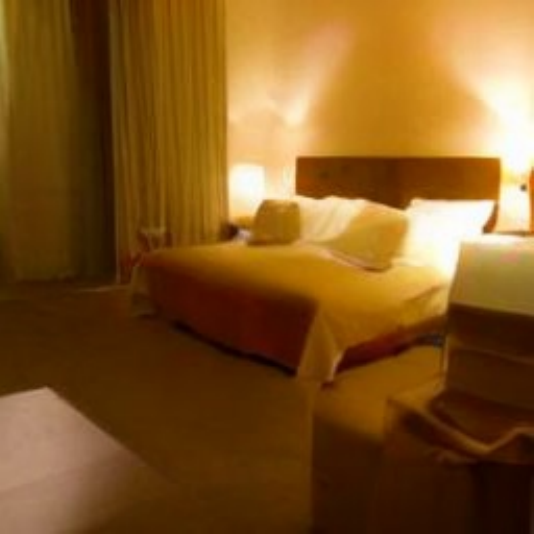}&
         \includegraphics[width=0.3\columnwidth]{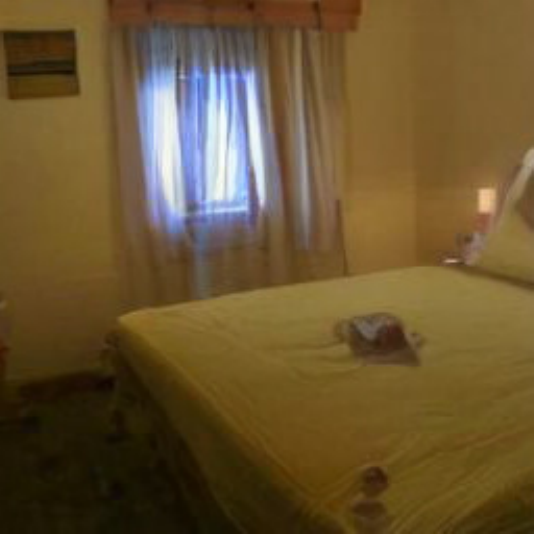}\\
         \includegraphics[width=0.3\columnwidth]{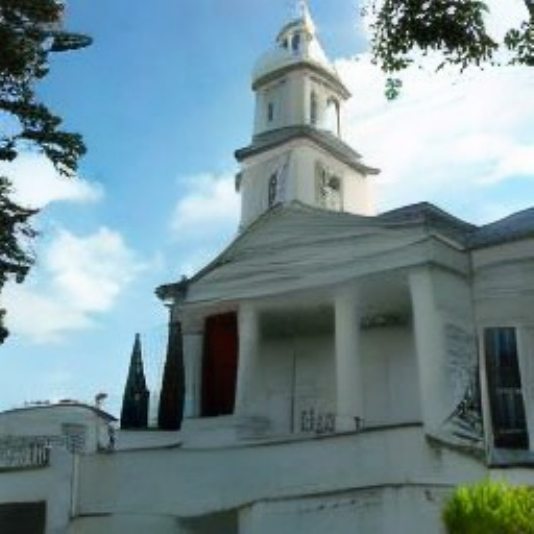}&
         \includegraphics[width=0.3\columnwidth]{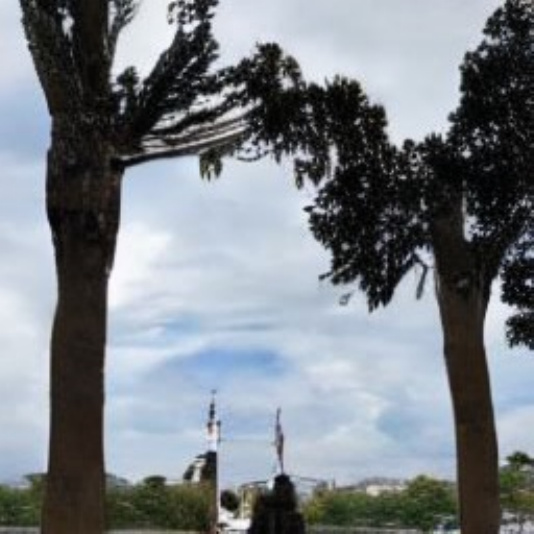}&
         \includegraphics[width=0.3\columnwidth]{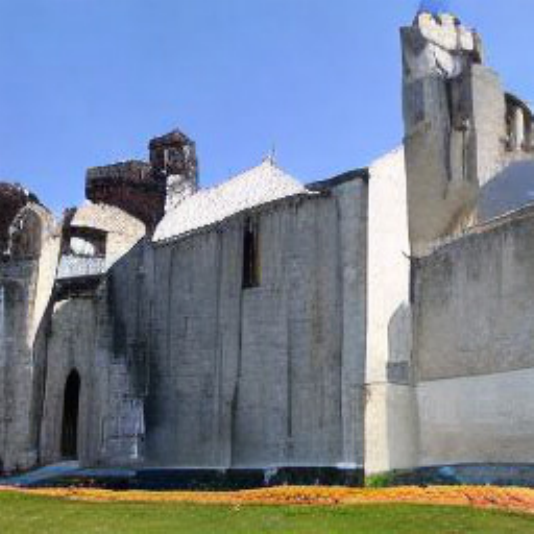}\\
      \end{tabular}
      \label{fig:lsun_qdiff}
      \vspace{-3mm}
      \caption{Q-Diffusion~\cite{li2023q}}
   \end{subfigure}
   \begin{subfigure}[c]{0.24\linewidth}
      \centering
      \begin{tabular}{c @{\hspace{0.5mm}} c @{\hspace{0.5mm}} c}
         \includegraphics[width=0.3\columnwidth]{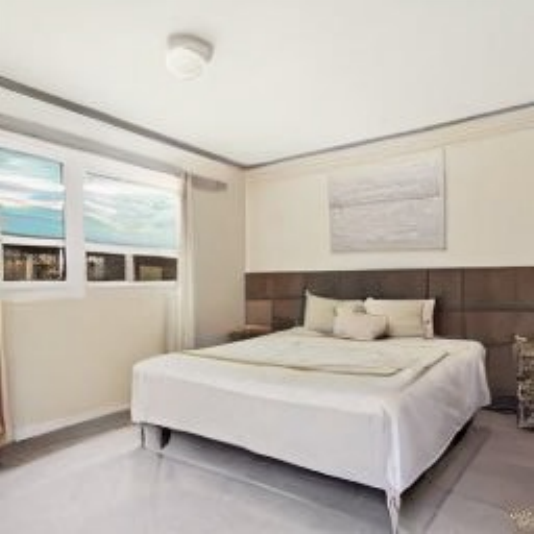}&
         \includegraphics[width=0.3\columnwidth]{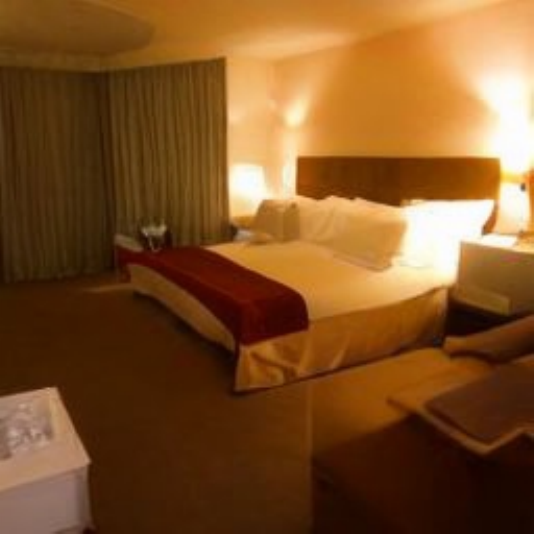}&
         \includegraphics[width=0.3\columnwidth]{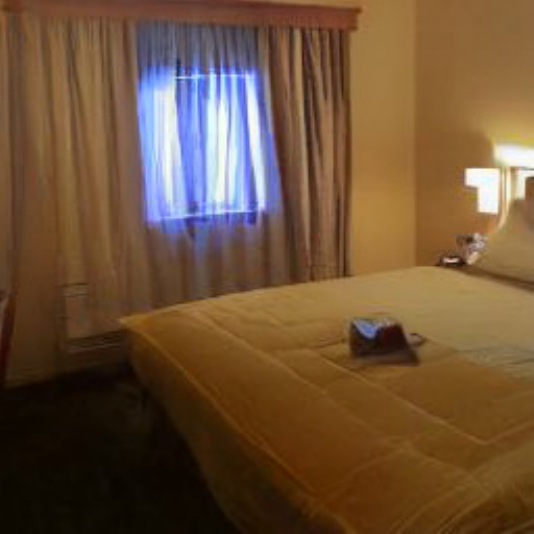}\\
         \includegraphics[width=0.3\columnwidth]{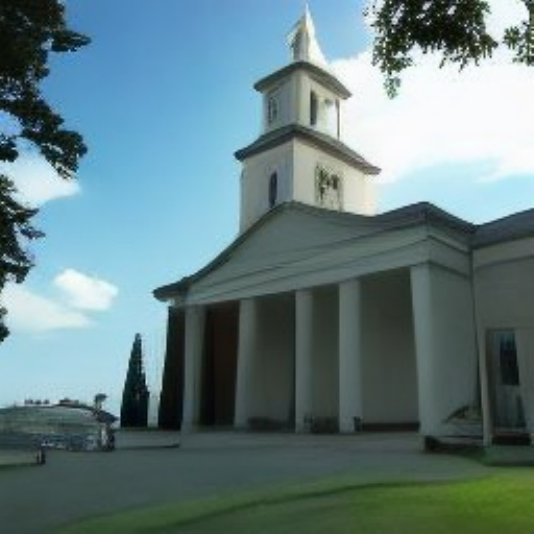}&
         \includegraphics[width=0.3\columnwidth]{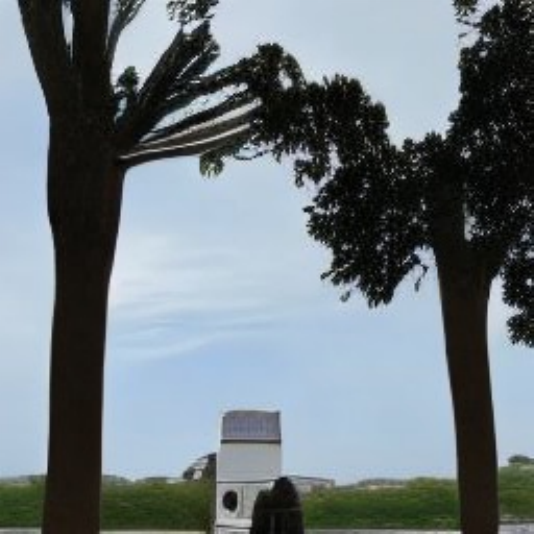}&
         \includegraphics[width=0.3\columnwidth]{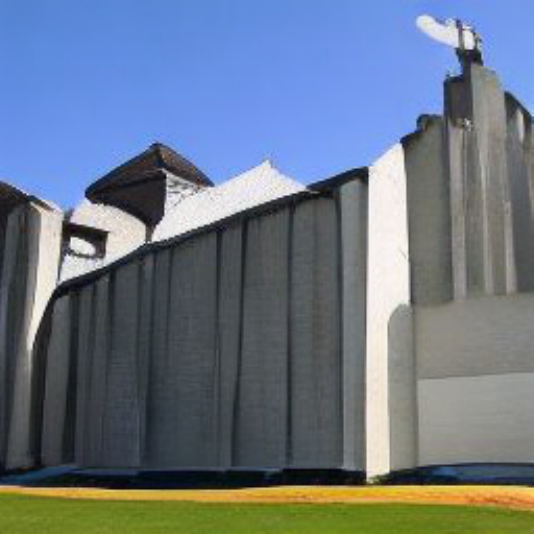}\\
      \end{tabular}
      \label{fig:lsun_tfmq}
      \vspace{-3mm}
      \caption{TFMQ-DM~\cite{huang2024tfmq}}
   \end{subfigure}
   \begin{subfigure}[c]{0.24\linewidth}
      \centering
      \begin{tabular}{c @{\hspace{0.5mm}} c @{\hspace{0.5mm}} c}
         \includegraphics[width=0.3\columnwidth]{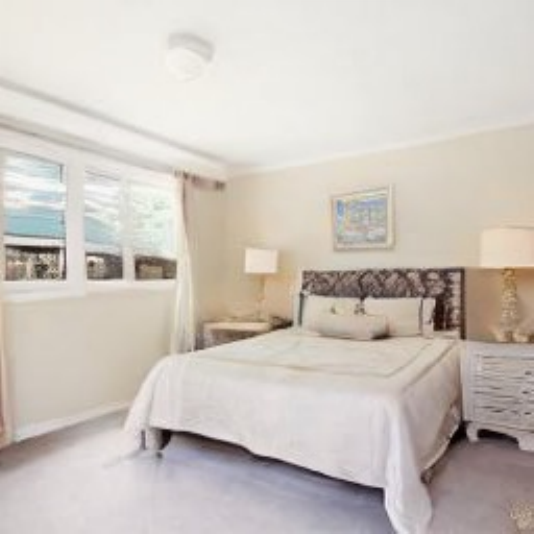}&
         \includegraphics[width=0.3\columnwidth]{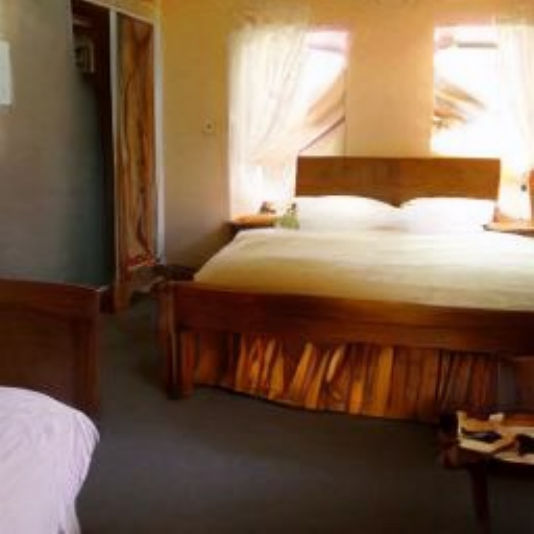}&
         \includegraphics[width=0.3\columnwidth]{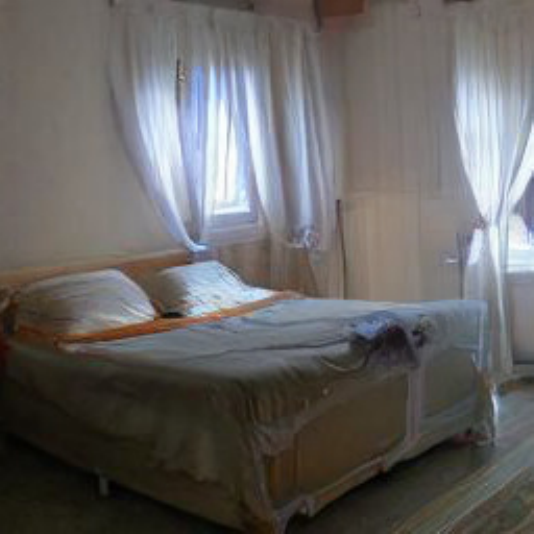}\\
         \includegraphics[width=0.3\columnwidth]{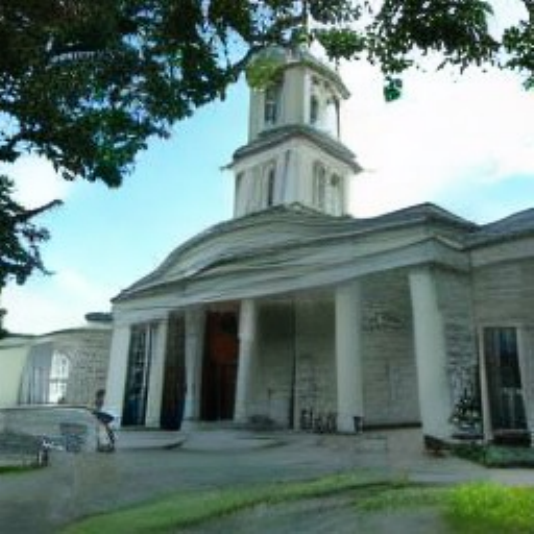}&
         \includegraphics[width=0.3\columnwidth]{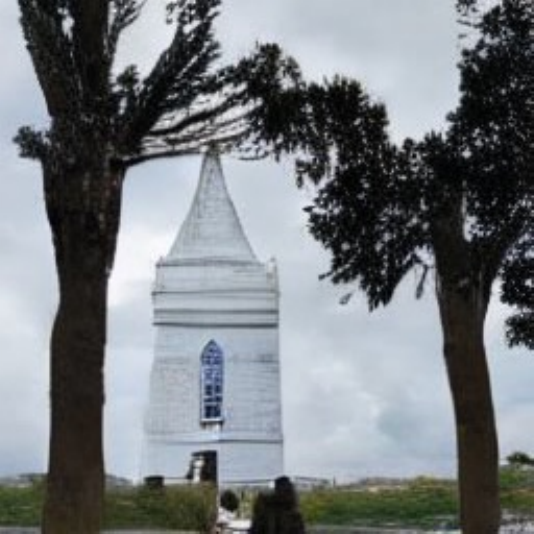}&
         \includegraphics[width=0.3\columnwidth]{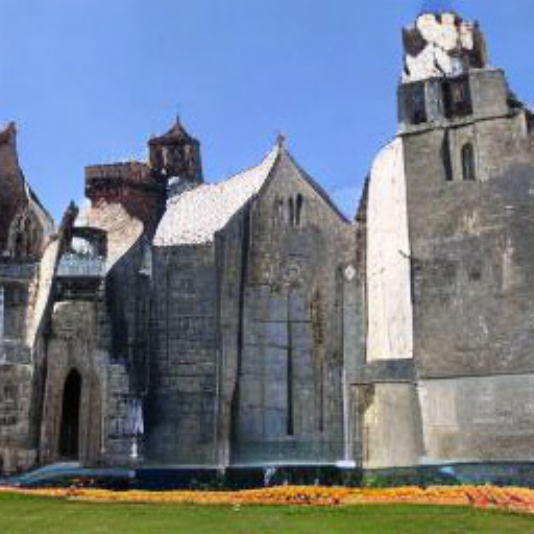}\\
      \end{tabular}
      \label{fig:lsun_ours}
      \vspace{-3mm}
      \caption{Ours}
   \end{subfigure}
   \vspace{-2mm}
   \caption{Visual comparisons of generated images on (\textbf{Top})~LSUN-Bedrooms~\cite{yu2015lsun} and (\textbf{Bottom})~LSUN-Churches~\cite{yu2015lsun} for unconditional image generation under a 3/8-bit setting.}
   \label{fig:lsun_results}
   \end{center}
   \vspace{-3mm}
 \end{figure*}

\begin{figure*}[!t]
  \vspace{2mm}

   \captionsetup[subfigure]{font=small, labelformat=empty}
   \begin{center}
     \begin{subfigure}[c]{0.24\linewidth}
       \centering
       \begin{tabular}{c @{\hspace{0.5mm}} c}
         \includegraphics[width=0.48\columnwidth]{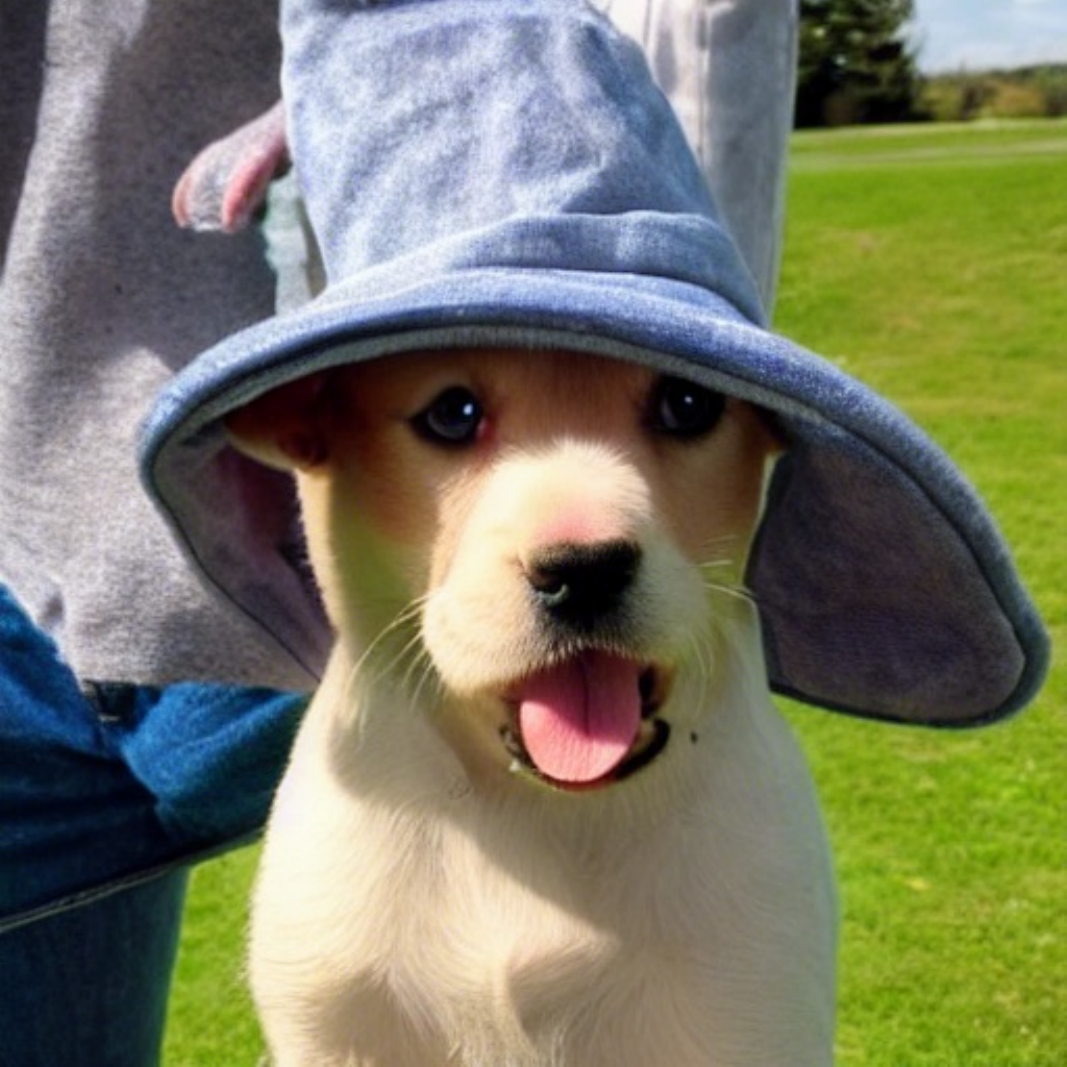} &
         \includegraphics[width=0.48\columnwidth]{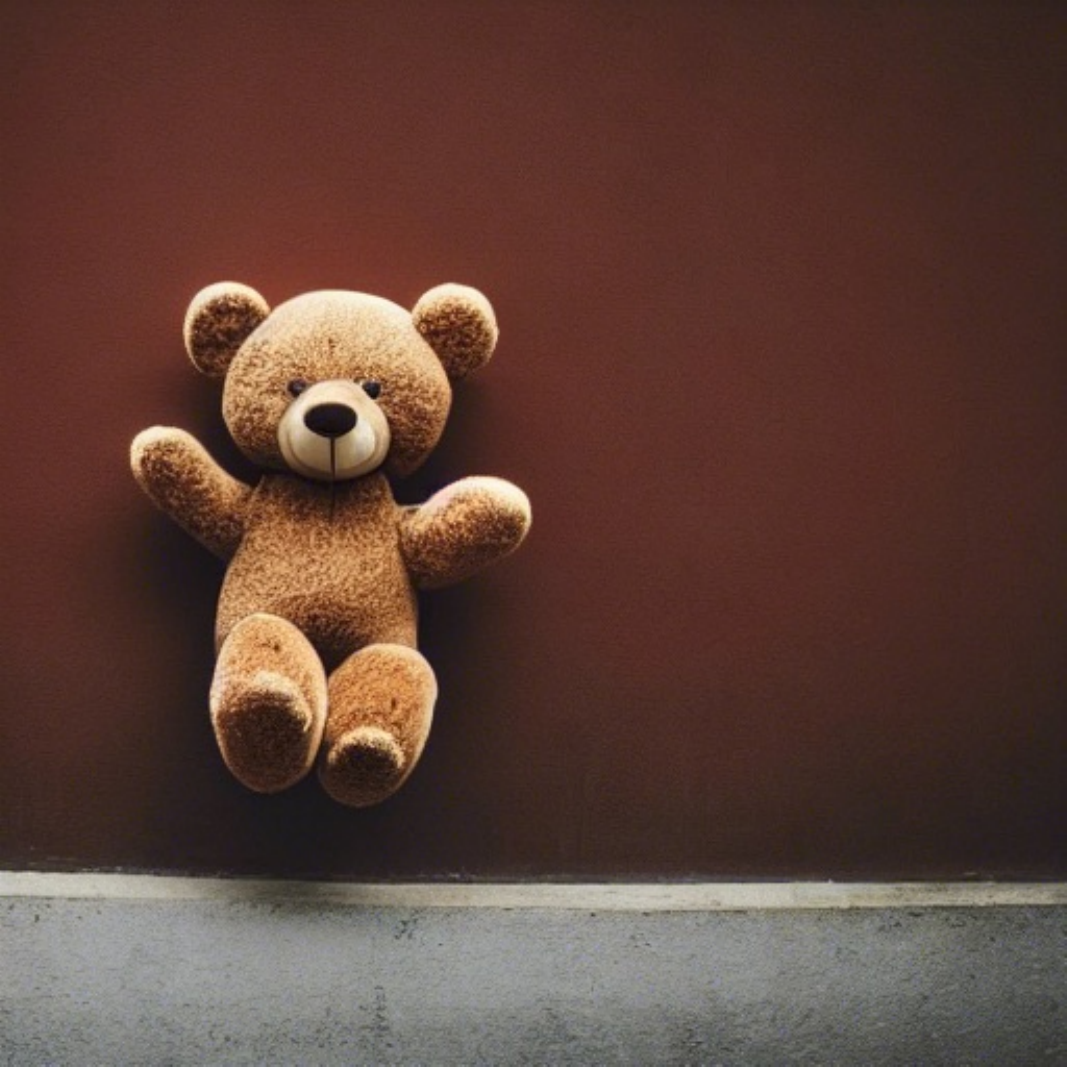} \\
         \includegraphics[width=0.48\columnwidth]{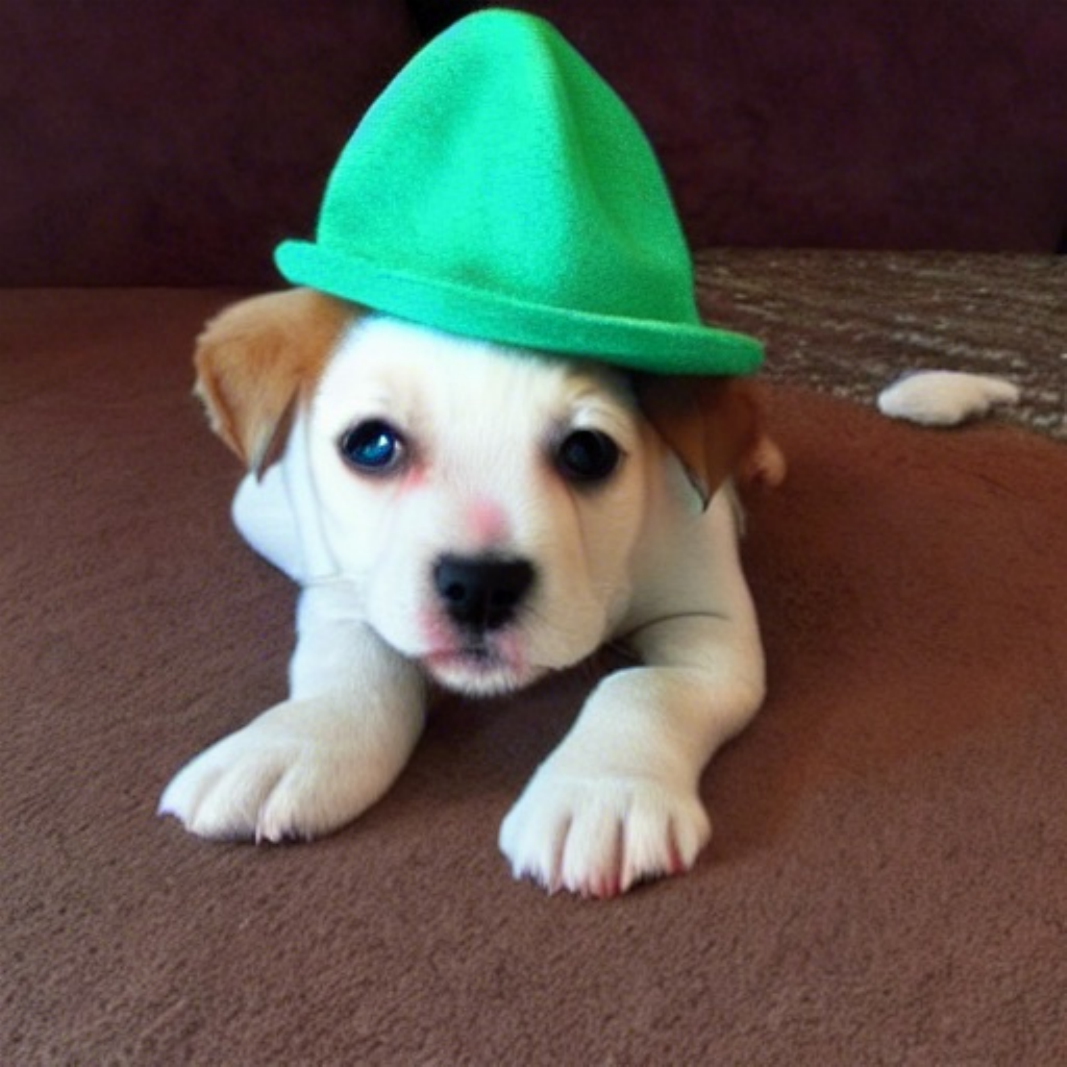} &
         \includegraphics[width=0.48\columnwidth]{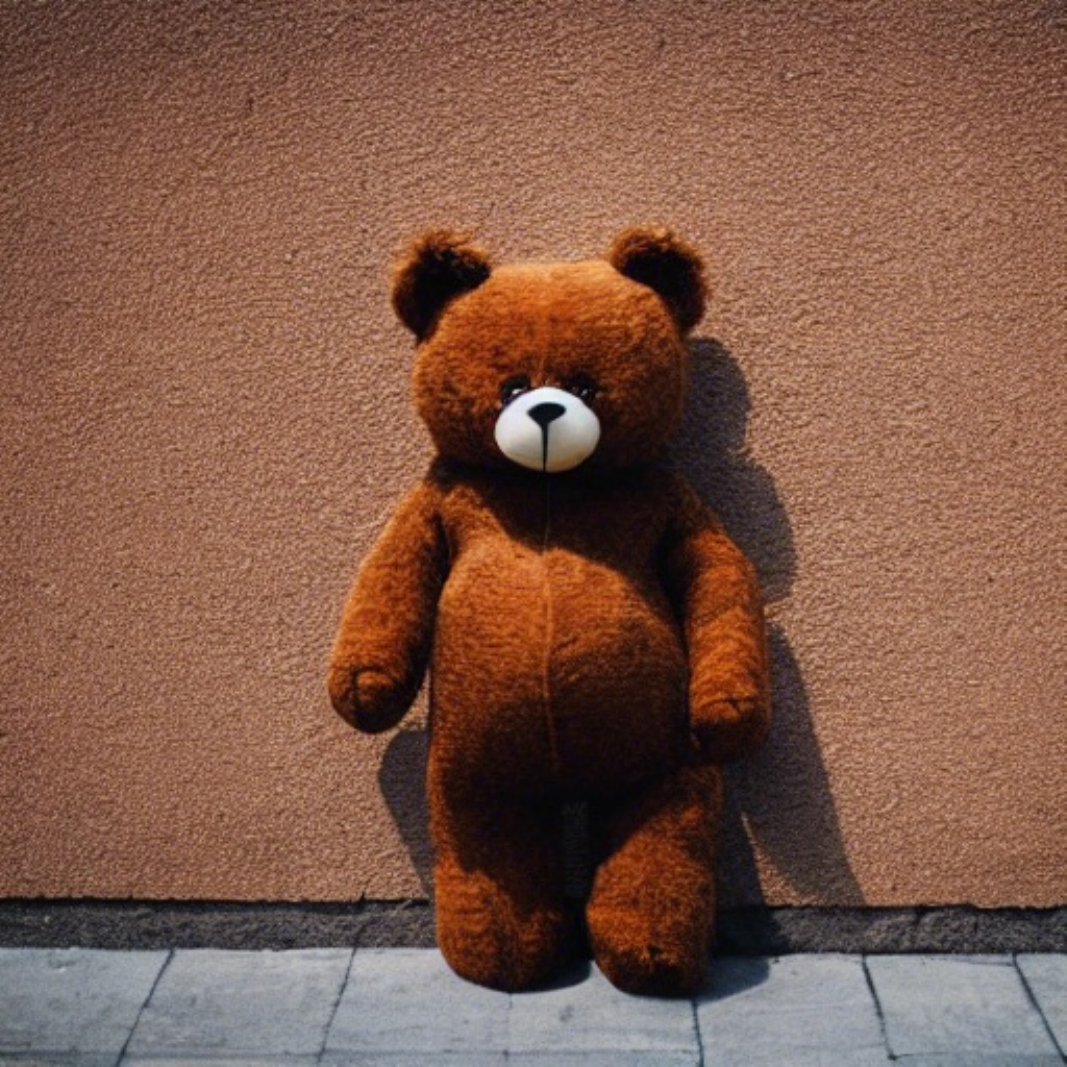} \\
       \end{tabular}
       \label{fig:fp}
       \vspace{-3mm}
       \caption{Full-Precision}
     \end{subfigure}
     \hfill
     \begin{subfigure}[c]{0.24\linewidth}
      \centering
      \begin{tabular}{c @{\hspace{0.5mm}} c}
        \includegraphics[width=0.48\columnwidth]{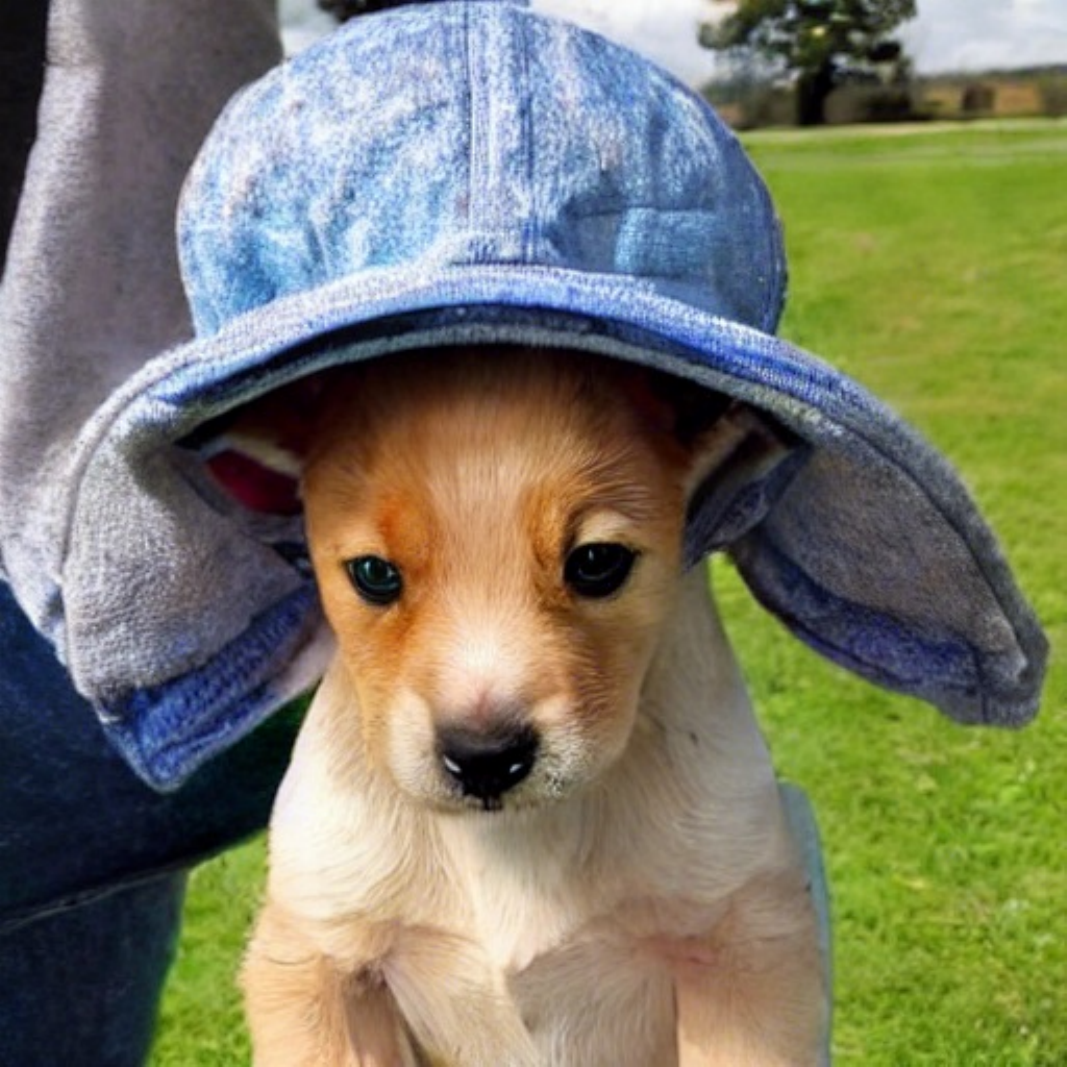} &
        \includegraphics[width=0.48\columnwidth]{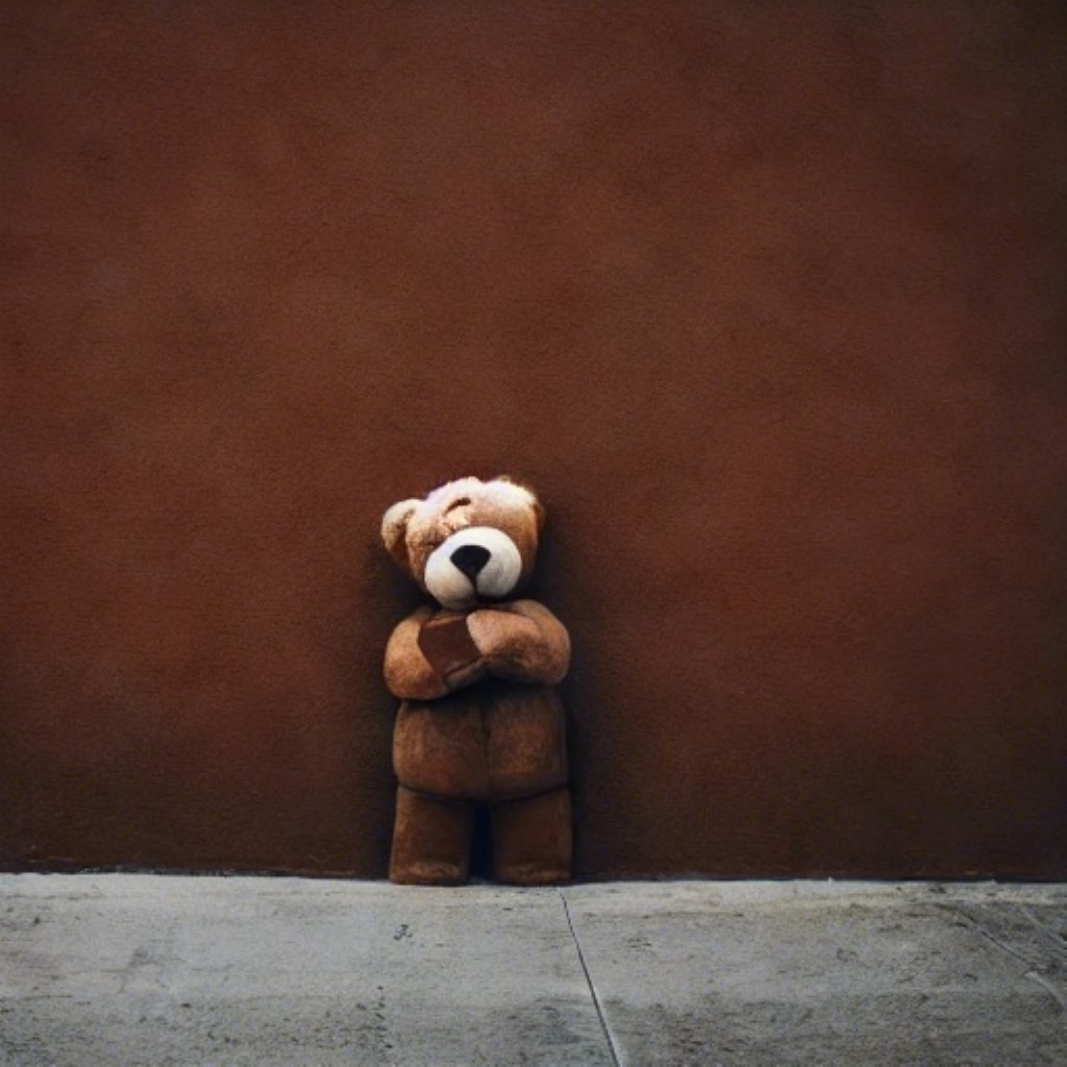} \\
        \includegraphics[width=0.48\columnwidth]{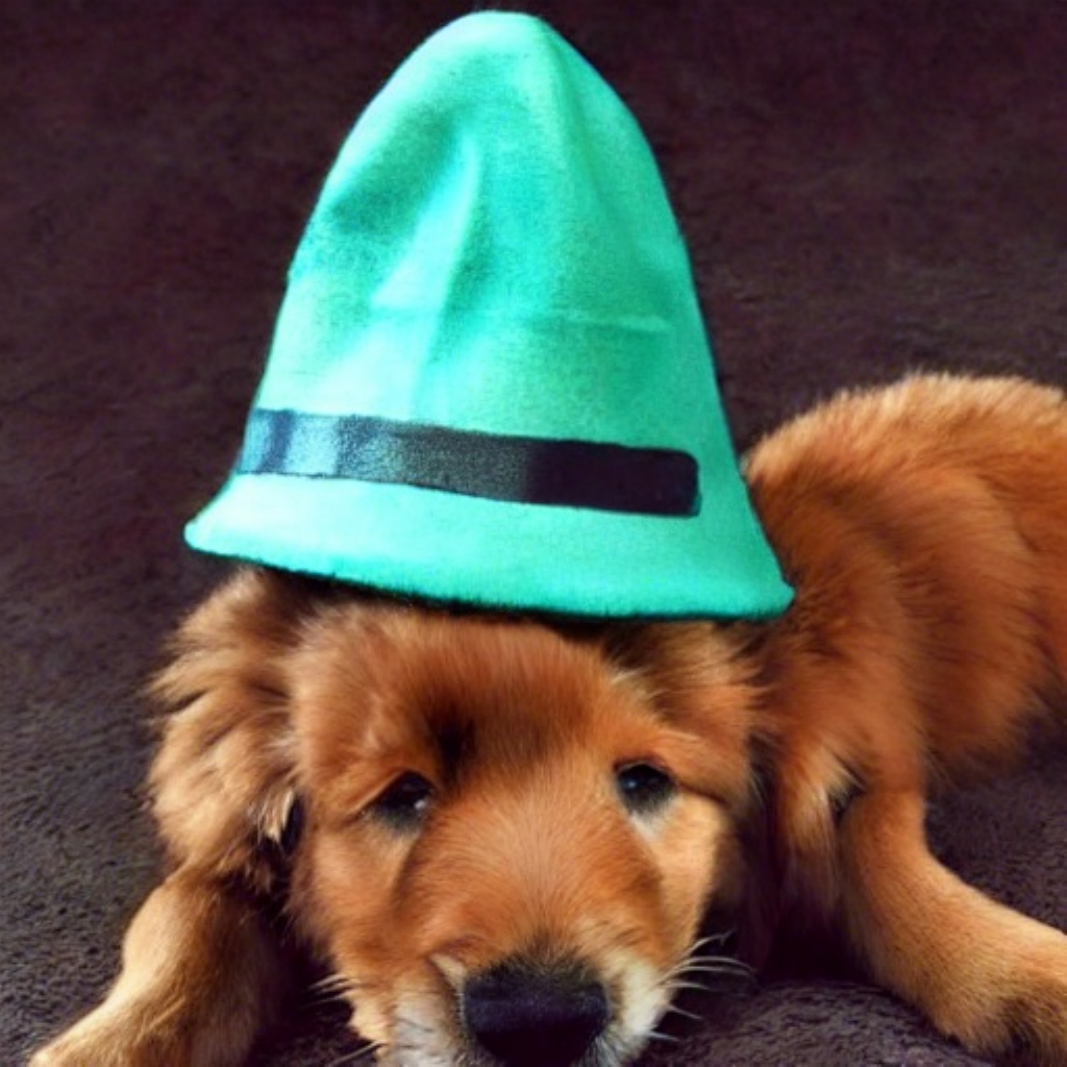} &
        \includegraphics[width=0.48\columnwidth]{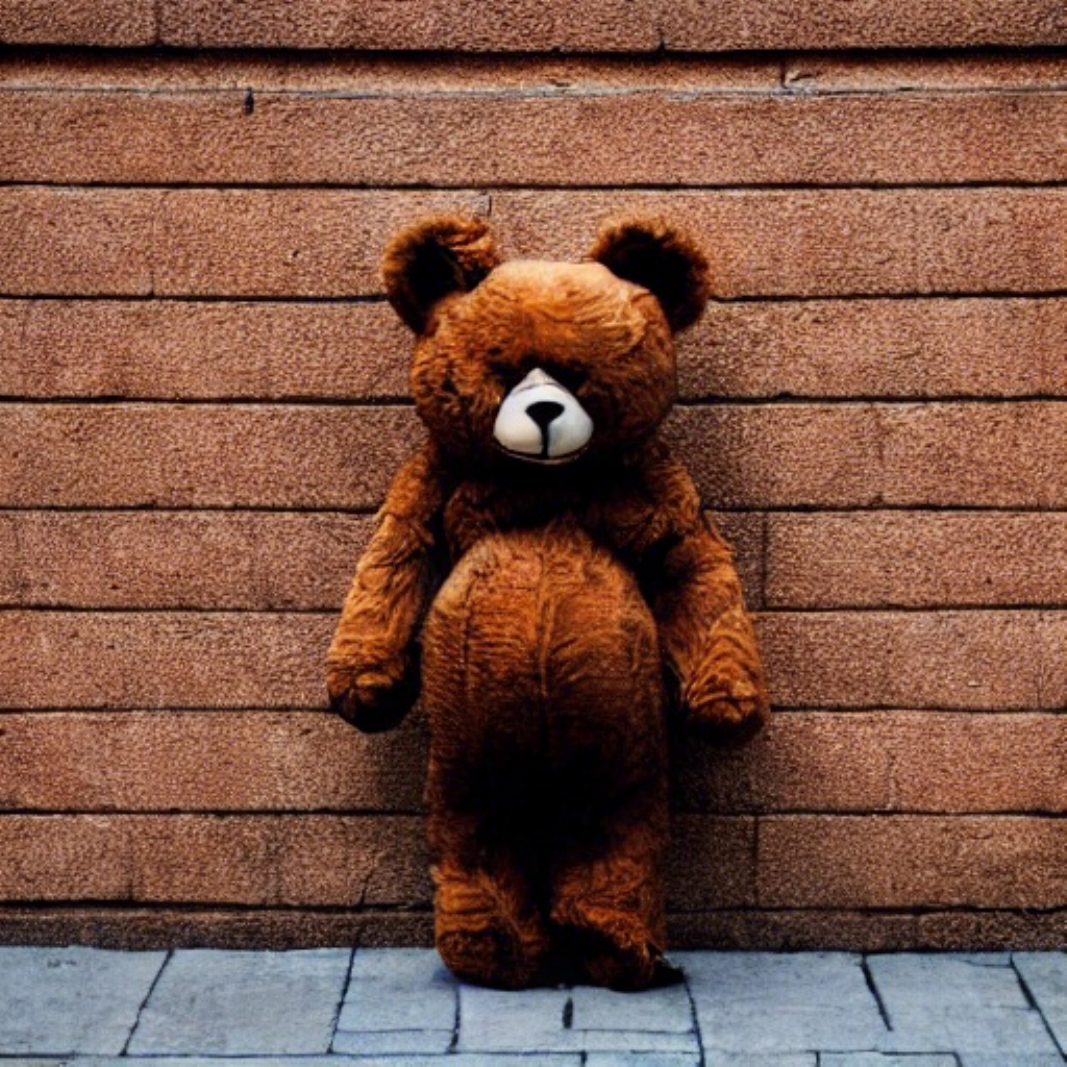} \\
      \end{tabular}
      \label{fig:qdiff}
      \vspace{-3mm}
      \caption{Q-Diffusion~\cite{li2023q}}
    \end{subfigure}
    \hfill
     \begin{subfigure}[c]{0.24\linewidth}
       \centering
       \begin{tabular}{c @{\hspace{0.5mm}} c}
         \includegraphics[width=0.48\columnwidth]{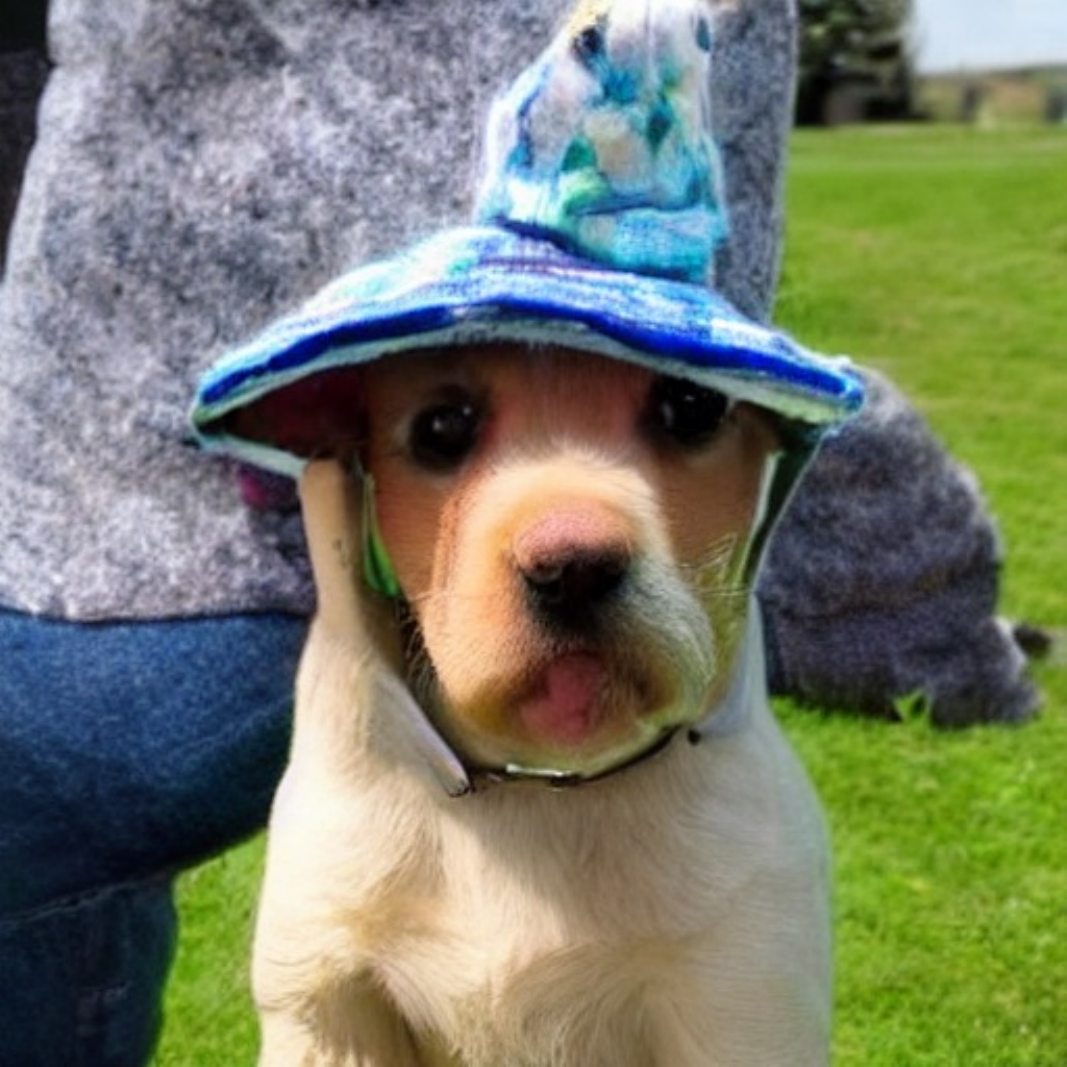} &
         \includegraphics[width=0.48\columnwidth]{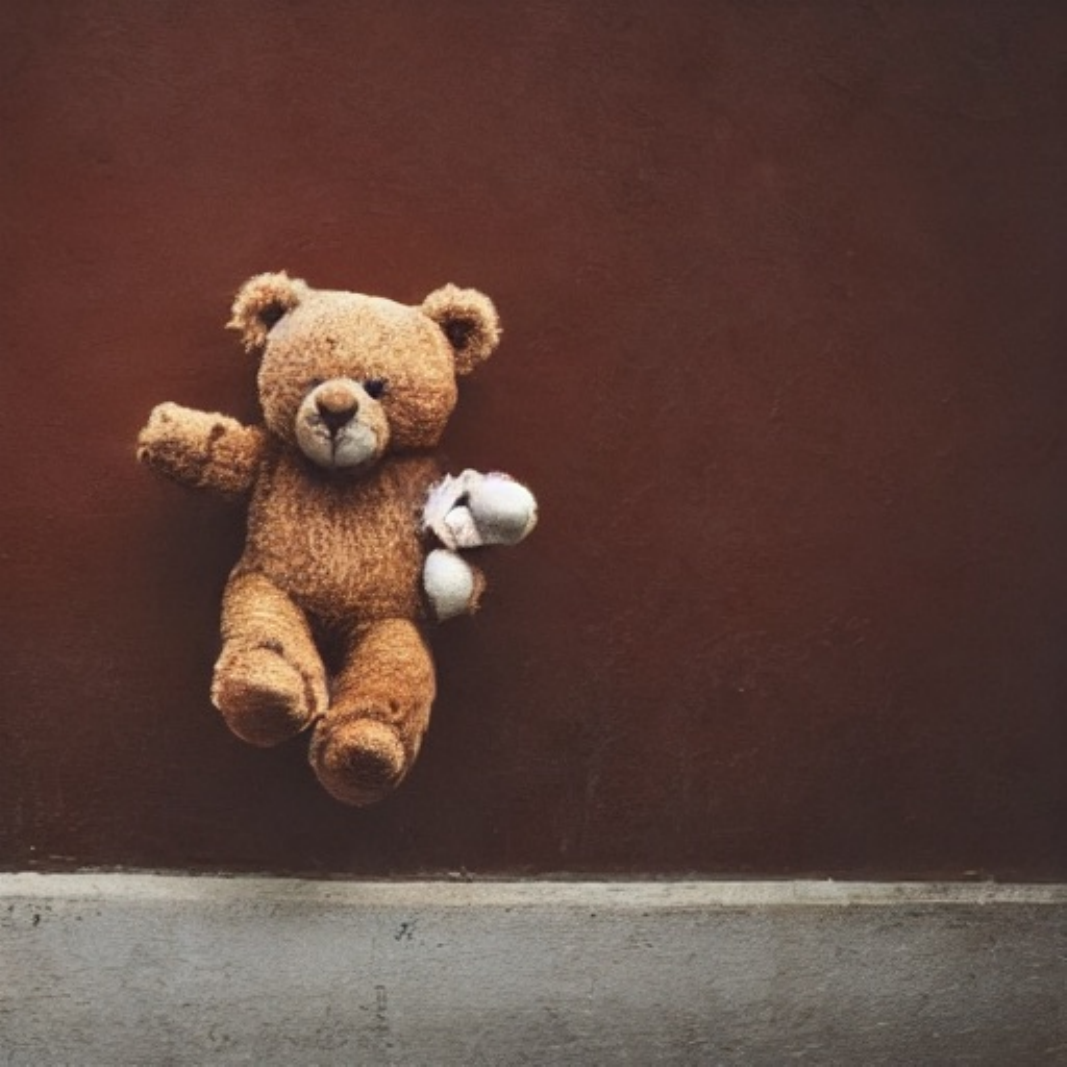} \\
         \includegraphics[width=0.48\columnwidth]{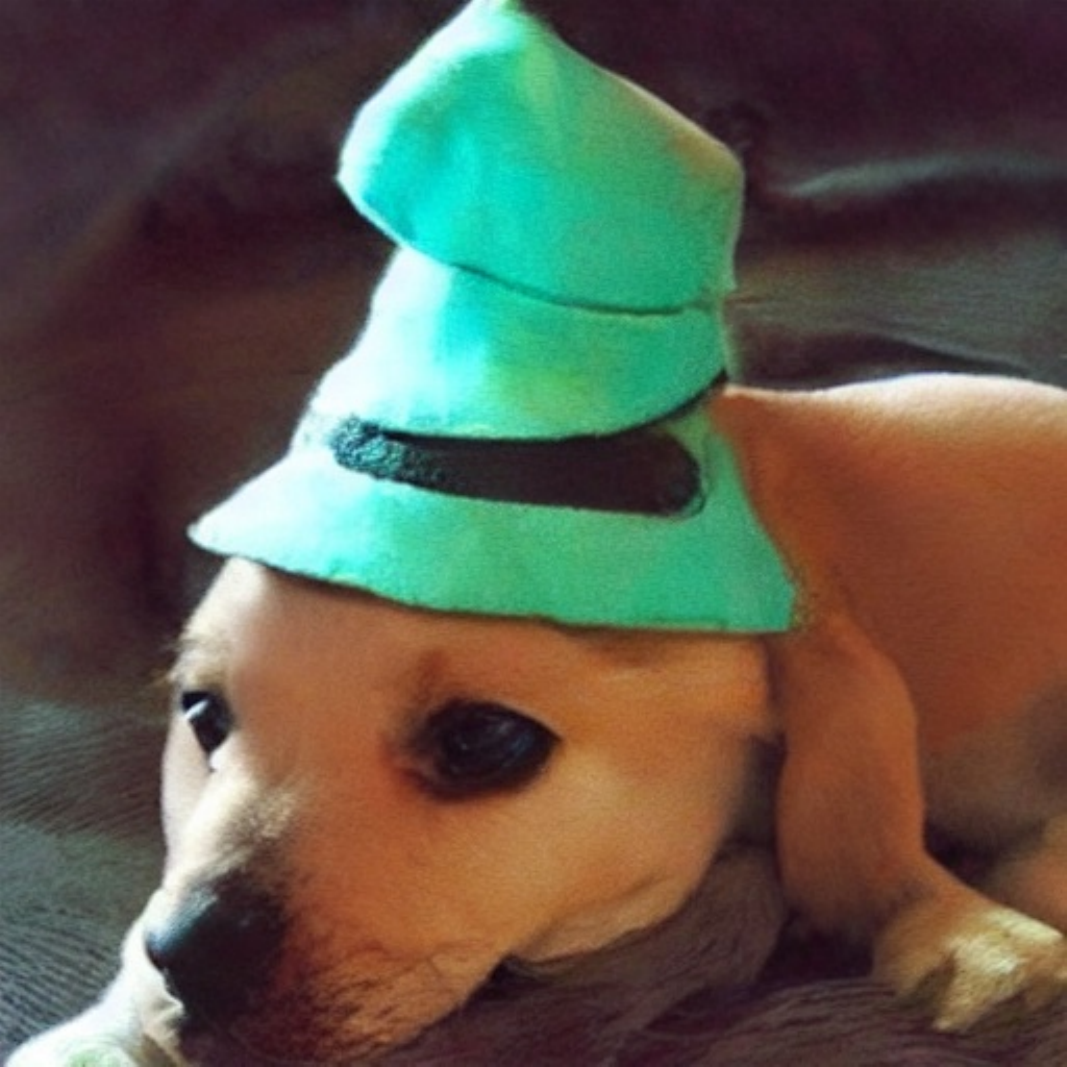} &
         \includegraphics[width=0.48\columnwidth]{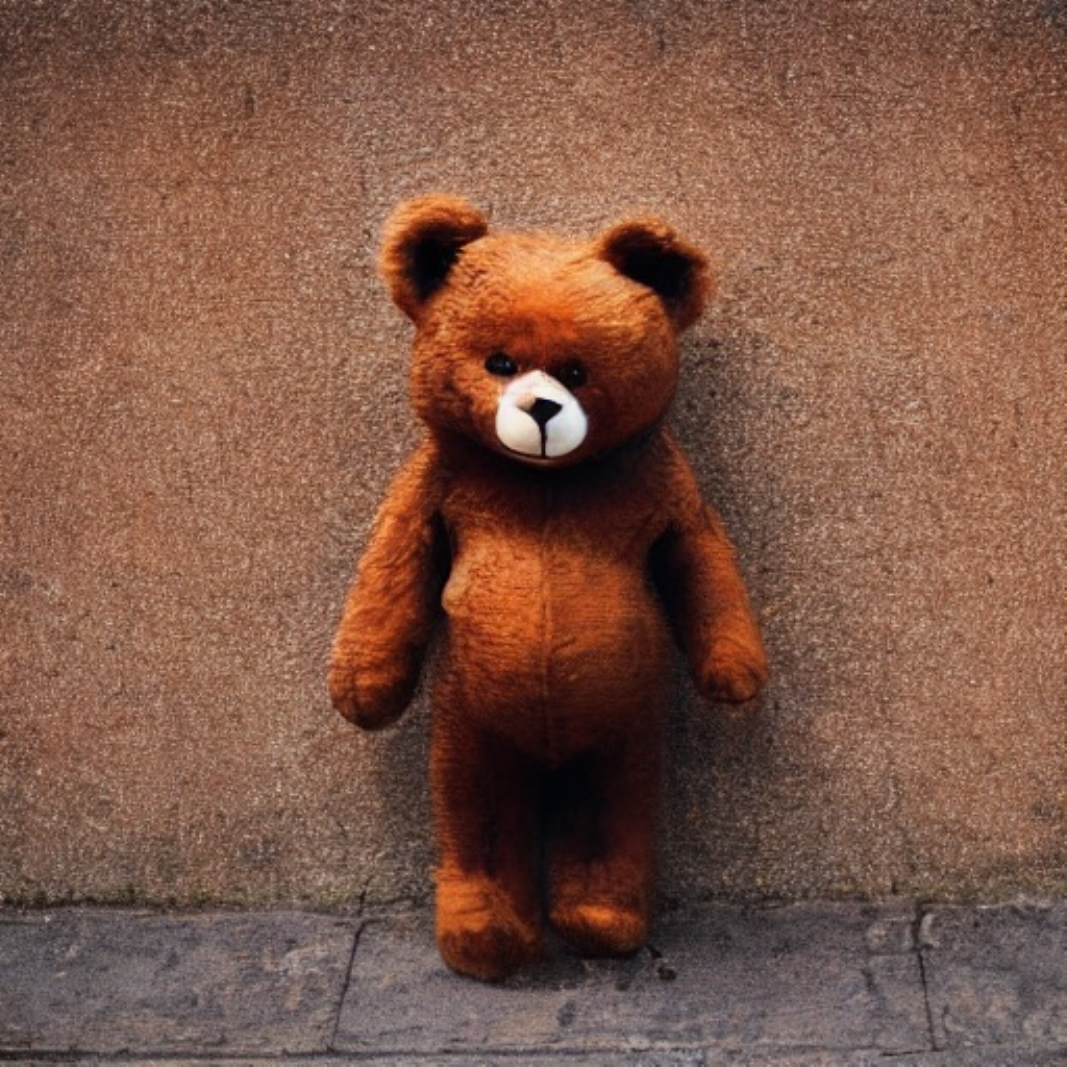} \\
       \end{tabular}
       \label{fig:pcr}
       \vspace{-3mm}
       \caption{PCR ($\tau=0.20$)~\cite{tang2025post}}
     \end{subfigure}
     \hfill
     \begin{subfigure}[c]{0.24\linewidth}
      \centering
      \begin{tabular}{c @{\hspace{0.5mm}} c}
        \includegraphics[width=0.48\columnwidth]{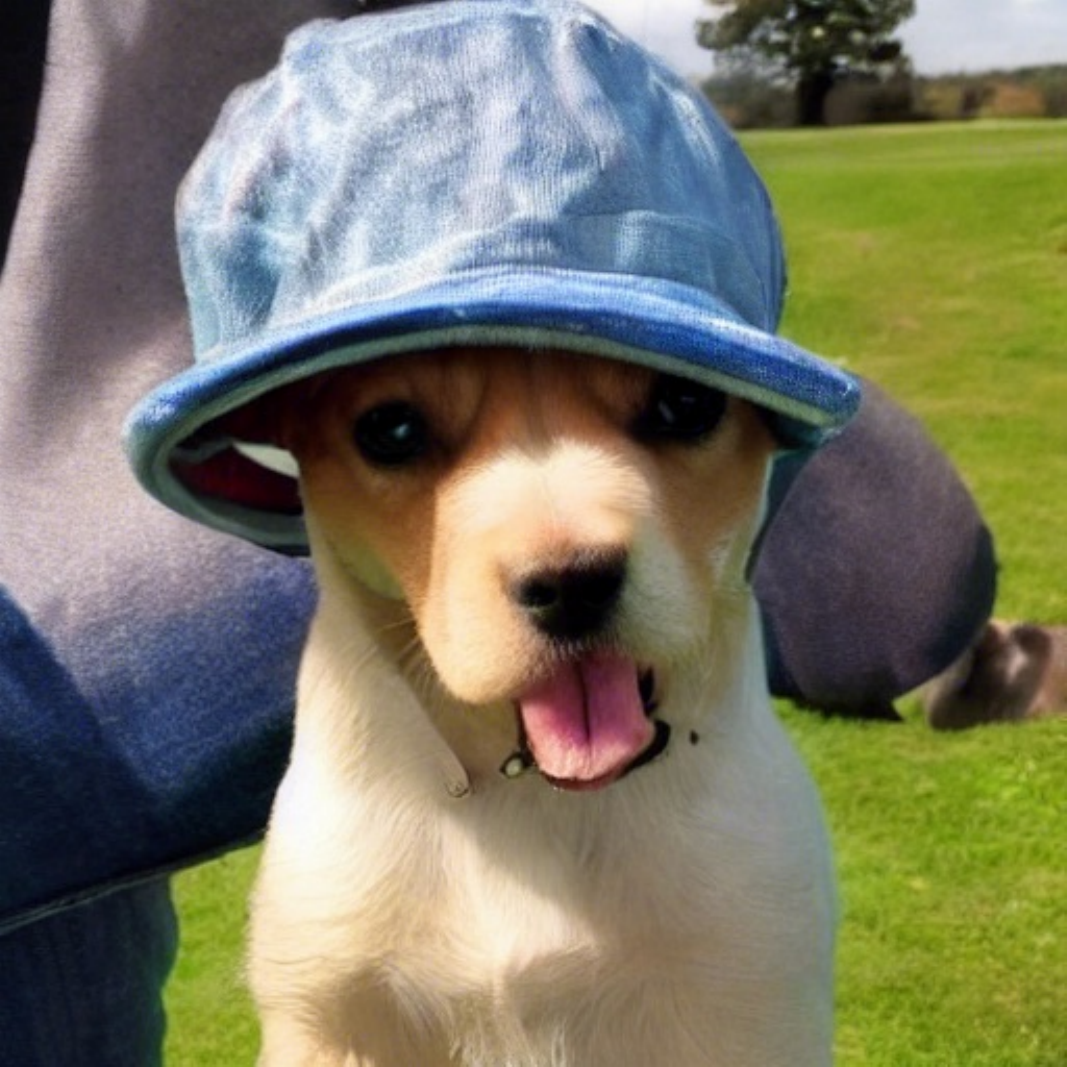} &
        \includegraphics[width=0.48\columnwidth]{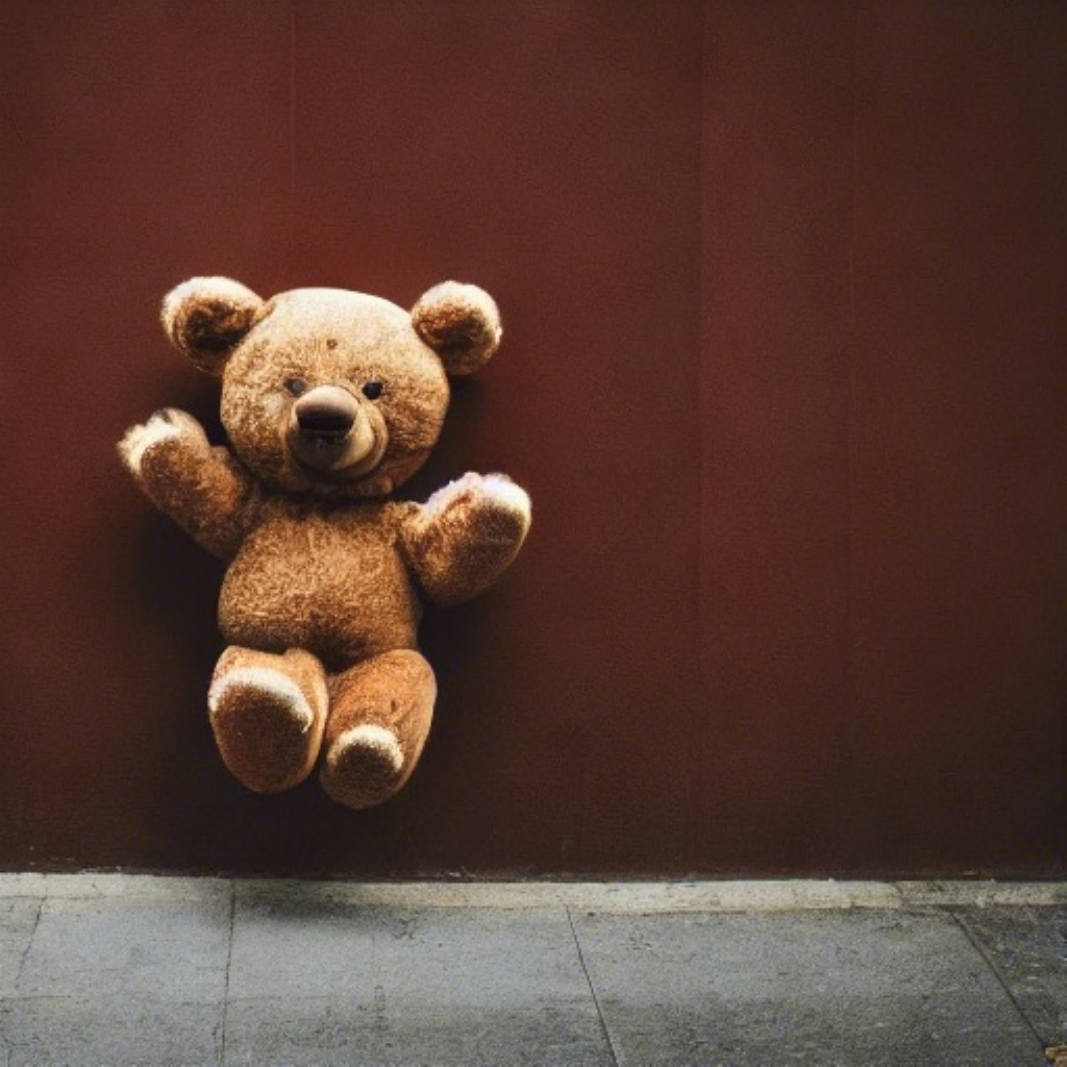} \\
        \includegraphics[width=0.48\columnwidth]{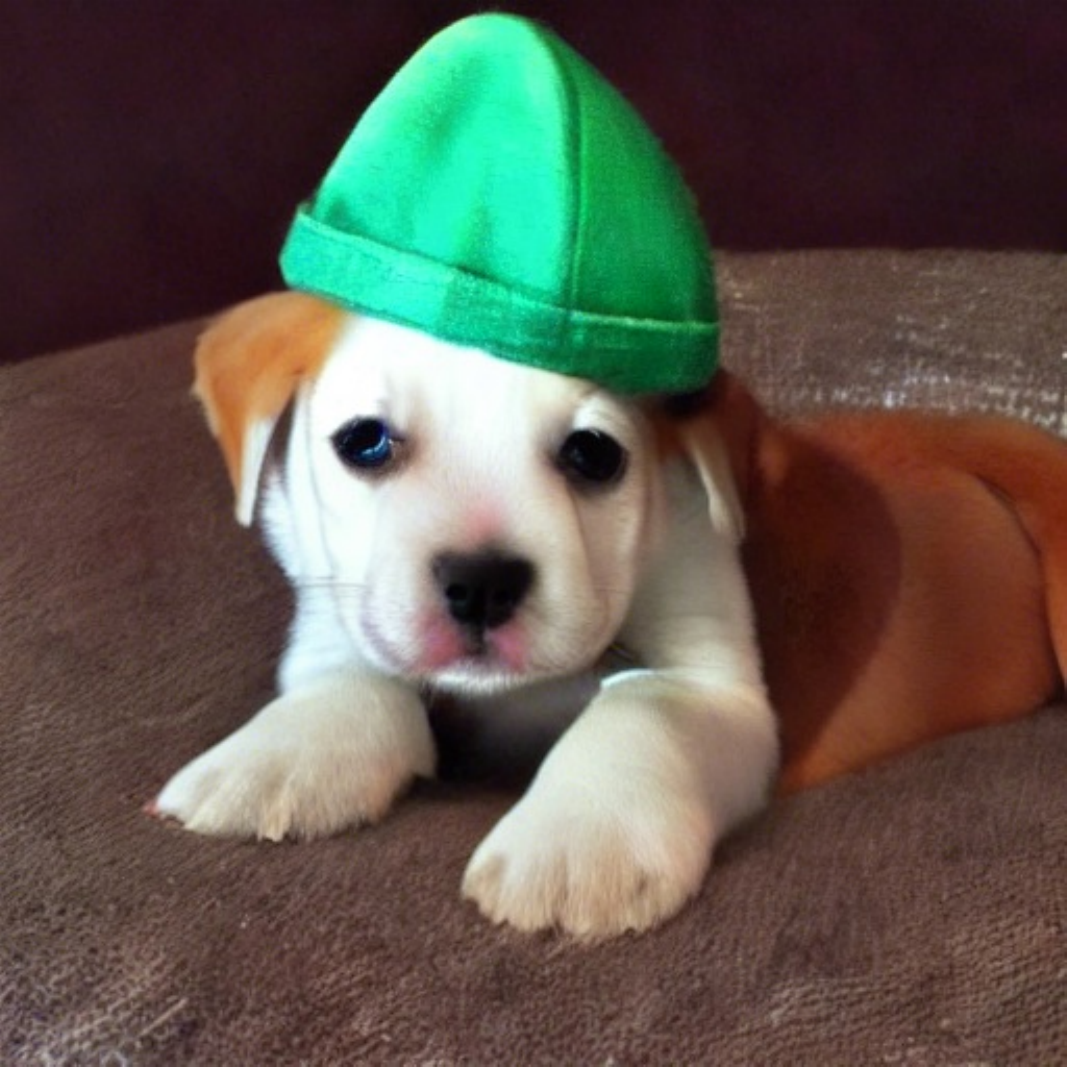} &
        \includegraphics[width=0.48\columnwidth]{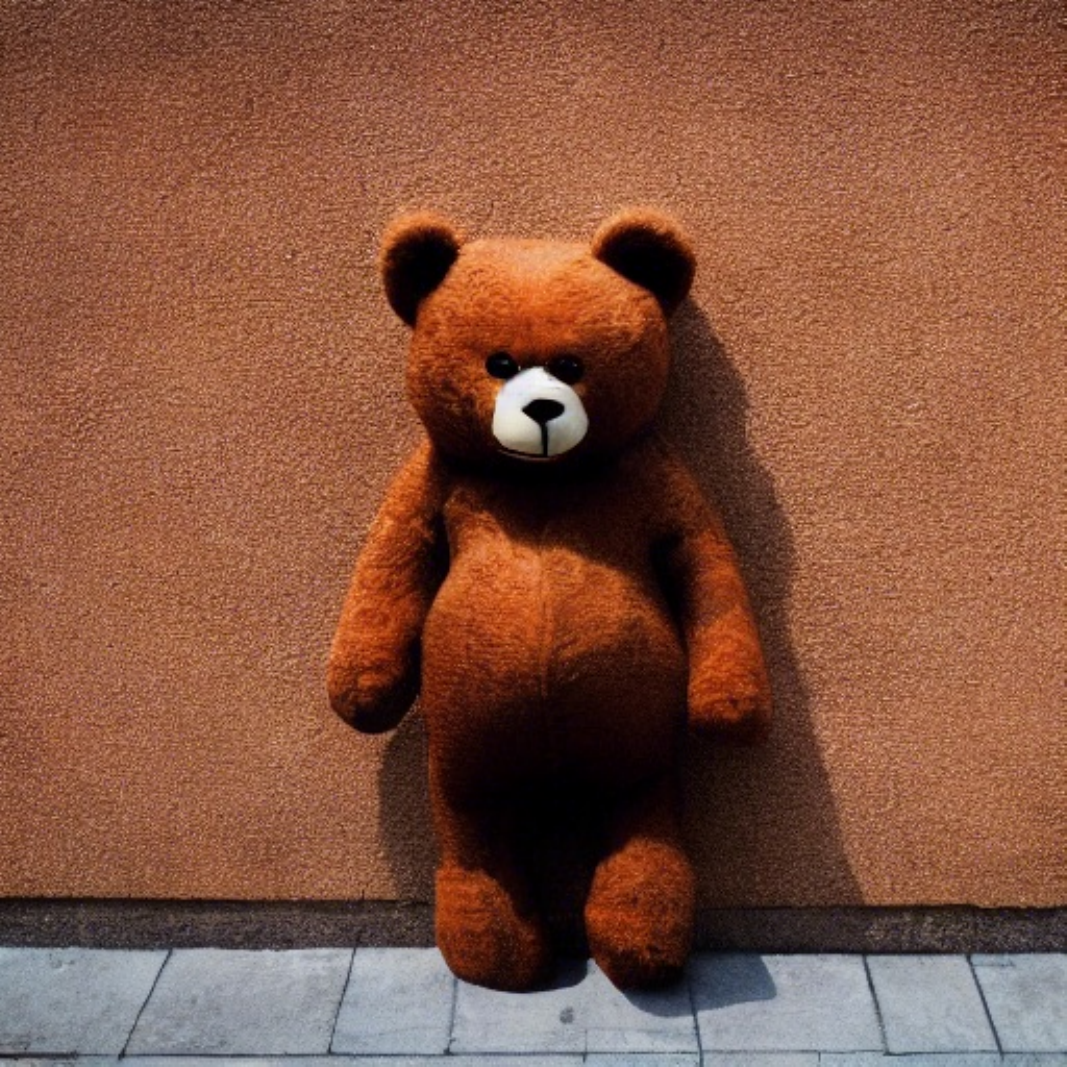} \\
      \end{tabular}
      \label{fig:ours}
      \vspace{-3mm}
      \caption{Ours}
     \end{subfigure}
   \end{center}
   \vspace{-3mm}
   \caption{Visual comparisons of generated images conditioned on text prompts. We generate images by SD v1.4~\cite{rombach2022high} and its quantized versions using PCR~\cite{tang2025post} under a 4/8.4-bit setting and Q-Diffusion~\cite{li2023q}, AccuQuant under a 4/8-bit setting. $\tau=0.20$ indicates setting 20\% of entire denoising steps into 10 bits. The prompts corresponding to each column are: \textit{``A puppy wearing a hat"} and \textit{``A cute teddy bear in front of a plain wall, warm and brown fur."}}
   \vspace{-3mm}
   
   \label{fig:sd_results_updated}
 \end{figure*}

{\textbf{Qualitative results.}}    We provide in Fig.~\ref{fig:lsun_results} visual comparisons of generated images by full-precision model, Q-Diffusion~\cite{li2023q}, TFMQ-DM~\cite{huang2024tfmq} and AccuQuant on LSUN-Bedrooms and LSUN-Churches~\cite{yu2015lsun} for unconditional image generation. 
We can see that AccuQuant provides more realistic images, and they are more close to the results from full-precision models, verifying once more that AccuQuant minimizes the discrepancies between full-precision and quantized models effectively. We also show Fig.~\ref{fig:sd_results_updated} generated images, conditioned on text prompts, by SD v1.4~\cite{rombach2022high} and its quantized versions using Q-Diffusion~\cite{li2023q}, PCR~\cite{tang2025post} and AccuQuant on MS-COCO~\cite{lin2014microsoft}. We can observe that AccuQuant generates high-quality images that closely resemble those obtained by the full-precision model, compared to Q-Diffusion~\cite{li2023q} and PCR~\cite{tang2025post}. Note that PCR~\cite{tang2025post} exploits more bit-widths than ours, demonstrating the effectiveness of our approach to minimizing accumulated quantization errors. For example, the detailed shape of the hat, and the pose of the dog and teddy bear are preserved well for our method, while Q-Diffusion~\cite{li2023q} and PCR~\cite{tang2025post} do not. More qualitative results can be found in the appendix.

\vspace{-2mm}

\subsection{Discussion}
\label{sec:discussion}

\paragraph{Analysis of a group size.}

\begin{wrapfigure}{r}{0.45\textwidth}
    \vspace{-4mm}
    \begin{minipage}{\linewidth}
        \centering
        
        \captionof{table}{Quantitative comparisons of AccuQuant under a 6/6-bit setting with varying group size.}
        \label{tab:group_size}
        \resizebox{0.97\linewidth}{!}{
        \begin{tabular}{cccc}
            \toprule
            \textbf{Group size} & \textbf{IS~$\uparrow$} & \textbf{FID~$\downarrow$} & \textbf{FID2FP32~$\downarrow$} \\
            \midrule
            ~~~~1   & 9.17 & ~~6.96 & ~~5.62 \\
            ~~~~2   & 9.02 & ~~6.82 & ~~4.48 \\
            ~~~~5   & \textbf{9.18} & ~~\textbf{5.79} & ~~\textbf{3.30} \\
            ~~10  & 9.13 & ~~6.18 & ~~4.30 \\
            ~~50  & 8.69 & 14.52 & 14.19 \\
            100 & 7.86 & 50.11 & 57.96 \\
            \bottomrule
        \end{tabular}}
        \vspace{-3mm}
    \end{minipage}
\end{wrapfigure}
\newcommand{\smallmath}[1]{{\footnotesize$#1$}}

\newcommand{\decpm}[3]{%
  \shortstack[c]{#1.#2 \\ {\footnotesize$\pm$#3}}%
}

\newcommand{\deconly}[2]{%
  #1.#2%
}

\begin{table}[t]
\centering
\begin{minipage}{0.58\linewidth}
  \centering
  \caption{Comparison of the full gradient ($\frac{\partial \tilde{x}_{t-1}}{\partial \tilde{x}_t}$), Jacobian term ($c_t\frac{\partial \tilde{\epsilon}_\theta(\tilde{x}_t,t)}{\partial \tilde{x}_{t}}$), and our approximation. We quantize DDIM~\cite{song2020denoising} with CIFAR-10~\cite{krizhevsky2009learning} under W4A8 settings and report the mean and the entire range of the gradient at every 20 timestep.}
  \label{tab:grad_minmax}
  \vspace{1mm}
  \small
  \setlength{\tabcolsep}{5pt}
  \resizebox{\linewidth}{!}{
    \begin{tabular}{cccccc}
      \toprule
      \textbf{Timestep} & \textbf{100} & \textbf{80} & \textbf{60} & \textbf{40} & \textbf{20} \\
      \midrule
      \multirow{2}{*}{\large$\frac{\partial \tilde{x}_{t-1}}{\partial \tilde{x}_t}$} 
      & 1.0221 
      & 1.0586 
      & 1.0273 
      & 1.0082 
      & 1.0008 \\
      
      & \textbf{$\pm$0.112} 
      & \textbf{$\pm$0.059} 
      & \textbf{$\pm$0.034} 
      & \textbf{$\pm$0.034} 
      & \textbf{$\pm$0.071} \\
      \addlinespace[3pt]
      \multirow{2}{*}{\large$c_t\frac{\partial \tilde{\epsilon}_\theta(\tilde{x}_t,t)}{\partial \tilde{x}_{t}}$} 
      &
      -0.1236 
      & -0.0097 
      & -0.0015 
      & -0.0006 
      & -0.0002 \\
      & \textbf{$\pm$0.112} 
      & \textbf{$\pm$0.059} 
      & \textbf{$\pm$0.034} 
      & \textbf{$\pm$0.034} 
      & \textbf{$\pm$0.071} \\
      \midrule
      $\sqrt{\frac{\alpha_{t-1}}{\alpha_t}}$ 
      & ~1.1457 
      & ~1.0683 
      & ~1.0288 
      & ~1.0088 
      & ~1.0010 \\
      \bottomrule
      \end{tabular}}
\end{minipage}
\hfill
\begin{minipage}{0.37\linewidth}
  \centering
  \caption{Quantitative comparisons of AccuQuant with and without gradient approximation. We quantize DDIM~\cite{song2020denoising} on CIFAR-10~\cite{krizhevsky2009learning} with group size 5.}
  \label{tab:grad_approx}
  \vspace{3.4mm}
  \renewcommand{\arraystretch}{1.15}
  \small
  \resizebox{\linewidth}{!}{
  \begin{tabular}{ccc}
    \toprule
    \multirow{2}{*}{\textbf{Bits (W/A)}} & \multicolumn{2}{c}{\textbf{FID~$\downarrow$ / FID2FP32~$\downarrow$}} \\ \cline{2-3}
    & w/o Approx. & w/ Approx. \\
    \midrule
    6/6 & 6.13 / 4.02 & \textbf{5.79} / \textbf{3.30} \\
    4/8 & 5.26 / 2.32 & \textbf{4.75} / \textbf{1.15} \\
    4/6 & 7.21 / 5.18 & \textbf{7.07} / \textbf{3.94} \\
    3/8 & 9.72 / 5.34 & \textbf{9.03} / \textbf{5.15} \\
    3/6 & 10.26 / 7.80 & \textbf{9.89} / \textbf{6.69} \\
    \bottomrule
  \end{tabular}}
\end{minipage}
\vspace{-1mm}
\end{table}

\label{sec:discussion_group}

We compare in Table~\ref{tab:group_size} quantization results of AccuQuant, with different group sizes, for DDIM~\cite{song2020denoising} on CIFAR-10~\cite{krizhevsky2009learning}. We can see that AccuQuant performs better accordingly with an increase of a group size from 1 to 5, confirming once again that it is effective for quantizing diffusion models to account for the behaviors of full-precision and quantized models within multiple denoising steps. This also suggests that considering accumulated errors is not enough with few denoising steps. On the other hand, the quantization performance degrades, when the number of denoising steps is too large. This indicates that it is hard to estimate appropriate quantization parameters reducing the accumulated error across many denoising steps. To this end, the group size is treated as a hyperparameter that balances two objectives: (1) sufficiently capturing accumulated errors across multiple timesteps, and (2) ensuring stable optimization of quantization parameters. As shown in Table~\ref{tab:group_size}, if the group size is too small, it may fail to capture the long term error accumulation; if it is too large, the optimization becomes unstable as a single set of parameters must account for the diverse behaviors of many timesteps. The optimal group size can vary depending on the model and dataset, but we empirically found that dividing the timesteps into 10 to 20 groups yields consistently strong results across our experiments.

\paragraph{Analysis of gradient approximation.}
\label{sec:discussion_gradient}
We show in Table~\ref{tab:grad_approx} quantitative comparisons of AccuQuant with and without the gradient approximation technique. It shows that 1)~AccuQuant with our gradient approximation technique even provides better results quantitatively, with much less memory, compared with those obtained without using the approximation, and 2)~the performance gains are more significant for FID2FP32. This suggests that gradients of Eq.~\eqref{eq:real_gradient} are corrupted by substantial noise before the quantization parameters~$s_l$ of Eq.~\eqref{eq:s_l} converge, which makes the calibration process unstable. Specifically, we show in Table~\ref{tab:grad_minmax} the same statistics as in Fig.~\ref{fig:gradient_approximation} with the entire range to better illustrate the variability of the gradients. As shown in the Table~\ref{tab:grad_minmax}, Jacobian component~(\ie, Row 2 in Table~\ref{tab:grad_minmax}) is significantly smaller in average magnitude compared to the dominant scalar coefficient~(\ie, Row 3 in Table~\ref{tab:grad_minmax}), but it exhibits a high dynamic range that introduce substantial noise into the full gradient~(\ie, Row 1 in Table~\ref{tab:grad_minmax}). This high-variance noise causes the quantization parameters to update inconsistently at each step, leading to an unstable and unreliable calibration process. In contrast, our gradient approximation exploit only the dominant scalar coefficient~(\ie, Row 3 in Table~\ref{tab:grad_minmax}) by omitting the highly dynamic Jacobian term. This leads to more stable convergence during calibration and ultimately yields more optimal results, as demonstrated in Table~\ref{tab:grad_approx}.

\section{Limitation}
\label{sec:limitation}
AccuQuant requires multiple denoising steps during the calibration phase. While AccuQuant effectively mitigates accumulated errors and achieves strong performance even in few-step settings such as 20 steps on ImageNet~\cite{deng2009imagenet}, it may have limited efficacy when applied to diffusion models employing only 1–2 denoising steps~\cite{meng2023distillation,yin2024one}. In addition, AccuQuant has a hyperparameter, the group size, which is currently fixed across all groups. Exploring adaptive strategies to dynamically determine the optimal group size for each group would be an interesting direction for future work.

\vspace{-2mm}
\section{Conclusion}
\label{sec:conclusion}
We have shown a detailed analysis on quantization errors of diffusion models that the errors are accumulated over denoising steps. Based on this, we have introduced a novel PTQ method, dubbed AccuQuant, that alleviates the error accumulation problem by simulating multiple denoising steps in a sampling process of a diffusion model. We have also presented a gradient approximation technique to reduce the computational overhead of storing gradients for intermediate activations along the denoising steps. We have demonstrated that AccuQuant outperforms state-of-the-art PTQ methods across various bit-widths on standard benchmarks.
\clearpage      

\begin{ack}
This work was supported by Institute of Information \& Communications Technology Planning \& Evaluation (IITP) grants funded by the Korea government (MSIT) (No.RS-2022-00143524, Development of Fundamental Technology and Integrated Solution for Next-Generation Automatic Artificial Intelligence System, 
No.RS-2025-09942968, AI Semiconductor Innovation Lab(Yonsei University)), 
the National Research Foundation of Korea(NRF) grants funded by the Korea government(MSIT) (No. 2023R1A2C2004306, 
RS-2025-02216328),
Samsung Electronics Co., Ltd (IO240520-10013-01), and 
the Yonsei Signature Research Cluster Program of 2025 (2025-22-0013).
\end{ack}

\bibliographystyle{plain}
{\small \bibliography{main-neurips.bib}}

\begin{thebibliography}{10}

\bibitem{blattmann2023stable}
Andreas Blattmann, Tim Dockhorn, Sumith Kulal, Daniel Mendelevitch, Maciej Kilian, Dominik Lorenz, Yam Levi, Zion English, Vikram Voleti, Adam Letts, et~al.
\newblock Stable video diffusion: Scaling latent video diffusion models to large datasets.
\newblock {\em arXiv preprint arXiv:2311.15127}, 2023.

\bibitem{blattmann2023align}
Andreas Blattmann, Robin Rombach, Huan Ling, Tim Dockhorn, Seung~Wook Kim, Sanja Fidler, and Karsten Kreis.
\newblock Align your latents: High-resolution video synthesis with latent diffusion models.
\newblock In {\em CVPR}, 2023.

\bibitem{cai2017deep}
Zhaowei Cai, Xiaodong He, Jian Sun, and Nuno Vasconcelos.
\newblock Deep learning with low precision by half-wave gaussian quantization.
\newblock In {\em CVPR}, 2017.

\bibitem{castells2024ld}
Thibault Castells, Hyoung-Kyu Song, Bo-Kyeong Kim, and Shinkook Choi.
\newblock {LD-Pruner}: Efficient pruning of latent diffusion models using task-agnostic insights.
\newblock In {\em CVPR}, 2024.

\bibitem{choi2018pact}
Jungwook Choi, Zhuo Wang, Swagath Venkataramani, I~Pierce, Jen Chuang, Vijayalakshmi Srinivasan, and Kailash Gopalakrishnan.
\newblock {PACT}: Parameterized clipping activation for quantized neural networks.
\newblock {\em arXiv preprint arXiv:1805.06085}, 2018.

\bibitem{deng2009imagenet}
Jia Deng, Wei Dong, Richard Socher, Li-Jia Li, Kai Li, and Li~Fei-Fei.
\newblock {I}mage{N}et: A large-scale hierarchical image database.
\newblock In {\em CVPR}, 2009.

\bibitem{dhariwal2021diffusion}
Prafulla Dhariwal and Alexander Nichol.
\newblock Diffusion models beat {GAN}s on image synthesis.
\newblock In {\em NeurIPS}, 2021.

\bibitem{dosovitskiy2020image}
Alexey Dosovitskiy.
\newblock An image is worth 16x16 words: Transformers for image recognition at scale.
\newblock In {\em ICLR}, 2021.

\bibitem{fang2020post}
Jun Fang, Ali Shafiee, Hamzah Abdel-Aziz, David Thorsley, Georgios Georgiadis, and Joseph~H Hassoun.
\newblock Post-training piecewise linear quantization for deep neural networks.
\newblock In {\em ECCV}, 2020.

\bibitem{Garber_2024_CVPR}
Tomer Garber and Tom Tirer.
\newblock Image restoration by denoising diffusion models with iteratively preconditioned guidance.
\newblock In {\em CVPR}, 2024.

\bibitem{Gatys_2016_CVPR}
Leon~A. Gatys, Alexander~S. Ecker, and Matthias Bethge.
\newblock Image style transfer using convolutional neural networks.
\newblock In {\em CVPR}, 2016.

\bibitem{guo2023animatediff}
Yuwei Guo, Ceyuan Yang, Anyi Rao, Zhengyang Liang, Yaohui Wang, Yu~Qiao, Maneesh Agrawala, Dahua Lin, and Bo~Dai.
\newblock Animate{D}iff: Animate your personalized text-to-image diffusion models without specific tuning.
\newblock In {\em ICLR}, 2023.

\bibitem{He_2024_ICLR}
Yefei He, Jing Liu, Weijia Wu, Hong Zhou, and Bohan Zhuang.
\newblock Efficientdm: Efficient quantization-aware fine-tuning of low-bit diffusion models.
\newblock In {\em ICLR}, 2024.

\bibitem{he2023ptqd}
Yefei He, Luping Liu, Jing Liu, Weijia Wu, Hong Zhou, and Bohan Zhuang.
\newblock Ptqd: Accurate post-training quantization for diffusion models.
\newblock In {\em NeurIPS}, 2023.

\bibitem{hessel2022clipscore}
Jack Hessel, Ari Holtzman, Maxwell Forbes, Ronan Le~Bras, and Yejin Choi.
\newblock {CLIPS}core: A reference-free evaluation metric for image captioning.
\newblock In {\em EMNLP}, 2021.

\bibitem{heusel2017gans}
Martin Heusel, Hubert Ramsauer, Thomas Unterthiner, Bernhard Nessler, and Sepp Hochreiter.
\newblock {GAN}s trained by a two time-scale update rule converge to a local nash equilibrium.
\newblock In {\em NeurIPS}, 2017.

\bibitem{ho2020denoising}
Jonathan Ho, Ajay Jain, and Pieter Abbeel.
\newblock Denoising diffusion probabilistic models.
\newblock In {\em NeurIPS}, 2020.

\bibitem{Ho_2021_NeurIPSW}
Jonathan Ho and Tim Salimans.
\newblock Classifier-free diffusion guidance.
\newblock In {\em NeurIPSW}, 2021.

\bibitem{huang2024tfmq}
Yushi Huang, Ruihao Gong, Jing Liu, Tianlong Chen, and Xianglong Liu.
\newblock {TFMQ-DM}: Temporal feature maintenance quantization for diffusion models.
\newblock In {\em CVPR}, 2024.

\bibitem{jeon2022mr}
Yongkweon Jeon, Chungman Lee, Eulrang Cho, and Yeonju Ro.
\newblock {Mr.BiQ}: Post-training non-uniform quantization based on minimizing the reconstruction error.
\newblock In {\em CVPR}, 2022.

\bibitem{kim2021distance}
Dohyung Kim, Junghyup Lee, and Bumsub Ham.
\newblock Distance-aware quantization.
\newblock In {\em ICCV}, 2021.

\bibitem{kingma2014adam}
Diederik~P Kingma.
\newblock Adam: A method for stochastic optimization.
\newblock In {\em ICLR}, 2015.

\bibitem{krizhevsky2009learning}
Alex Krizhevsky, Geoffrey Hinton, et~al.
\newblock Learning multiple layers of features from tiny images.
\newblock {\em Technical report}, 2009.

\bibitem{lee2021network}
Junghyup Lee, Dohyung Kim, and Bumsub Ham.
\newblock Network quantization with element-wise gradient scaling.
\newblock In {\em CVPR}, 2021.

\bibitem{li2023q}
Xiuyu Li, Yijiang Liu, Long Lian, Huanrui Yang, Zhen Dong, Daniel Kang, Shanghang Zhang, and Kurt Keutzer.
\newblock Q-{D}iffusion: Quantizing diffusion models.
\newblock In {\em ICCV}, 2023.

\bibitem{li2021brecq}
Yuhang Li, Ruihao Gong, Xu~Tan, Yang Yang, Peng Hu, Qi~Zhang, Fengwei Yu, Wei Wang, and Shi Gu.
\newblock {BRECQ}: Pushing the limit of post-training quantization by block reconstruction.
\newblock In {\em ICLR}, 2021.

\bibitem{li2023repq}
Zhikai Li, Junrui Xiao, Lianwei Yang, and Qingyi Gu.
\newblock {RepQ-ViT}: Scale reparameterization for post-training quantization of vision transformers.
\newblock In {\em ICCV}, 2023.

\bibitem{lin2014microsoft}
Tsung-Yi Lin, Michael Maire, Serge Belongie, James Hays, Pietro Perona, Deva Ramanan, Piotr Doll{\'a}r, and C~Lawrence Zitnick.
\newblock Microsoft {COCO}: Common objects in context.
\newblock In {\em ECCV}, 2014.

\bibitem{liu2023audioldm}
Haohe Liu, Zehua Chen, Yi~Yuan, Xinhao Mei, Xubo Liu, Danilo Mandic, Wenwu Wang, and Mark~D Plumbley.
\newblock Audio{LDM}: Text-to-audio generation with latent diffusion models.
\newblock In {\em ICML}, 2023.

\bibitem{liu2023pd}
Jiawei Liu, Lin Niu, Zhihang Yuan, Dawei Yang, Xinggang Wang, and Wenyu Liu.
\newblock {PD-Quant}: Post-training quantization based on prediction difference metric.
\newblock In {\em CVPR}, 2023.

\bibitem{liu2022pseudo}
Luping Liu, Yi~Ren, Zhijie Lin, and Zhou Zhao.
\newblock Pseudo numerical methods for diffusion models on manifolds.
\newblock In {\em ICLR}, 2022.

\bibitem{Liu_2015_ICCV}
Ziwei Liu, Ping Luo, Xiaogang Wang, and Xiaoou Tang.
\newblock Deep learning face attributes in the wild.
\newblock In {\em ICCV}, 2015.

\bibitem{lu2022dpmsolver}
Cheng Lu, Zhou Yuhao, Fan Bao, Jianfei Chen, Chongxuan Li, and Jun Zhu.
\newblock Dpm-solver: A fast ode solver for diffusion probabilistic model sampling in around 10 steps.
\newblock In {\em NeurIPS}, 2022.

\bibitem{lv2024ptq4sam}
Chengtao Lv, Hong Chen, Jinyang Guo, Yifu Ding, and Xianglong Liu.
\newblock {PTQ4SAM}: Post-training quantization for segment anything.
\newblock In {\em CVPR}, 2024.

\bibitem{meng2023distillation}
Chenlin Meng, Robin Rombach, Ruiqi Gao, Diederik Kingma, Stefano Ermon, Jonathan Ho, and Tim Salimans.
\newblock On distillation of guided diffusion models.
\newblock In {\em CVPR}, 2023.

\bibitem{moon2024instance}
Jaehyeon Moon, Dohyung Kim, Junyong Cheon, and Bumsub Ham.
\newblock Instance-aware group quantization for vision transformers.
\newblock In {\em CVPR}, 2024.

\bibitem{nagel2020up}
Markus Nagel, Rana~Ali Amjad, Mart Van~Baalen, Christos Louizos, and Tijmen Blankevoort.
\newblock Up or down? adaptive rounding for post-training quantization.
\newblock In {\em ICML}, 2020.

\bibitem{nash2021generating}
Charlie Nash, Jacob Menick, Sander Dieleman, and Peter~W Battaglia.
\newblock Generating images with sparse representations.
\newblock In {\em ICML}, 2021.

\bibitem{Peebles_2023_ICCV}
William Peebles and Saining Xie.
\newblock Scalable diffusion models with transformers.
\newblock In {\em ICCV}, 2023.

\bibitem{poole2022dreamfusion}
Ben Poole, Ajay Jain, Jonathan T.~Barron, and Ben Mildenhall.
\newblock Dreamfusion: Text-to-3d using 2d diffusion.
\newblock In {\em ICLR}, 2023.

\bibitem{rombach2022high}
Robin Rombach, Andreas Blattmann, Dominik Lorenz, Patrick Esser, and Bj{\"o}rn Ommer.
\newblock High-resolution image synthesis with latent diffusion models.
\newblock In {\em CVPR}, 2022.

\bibitem{ronneberger2015u}
Olaf Ronneberger, Philipp Fischer, and Thomas Brox.
\newblock U-{N}et: Convolutional networks for biomedical image segmentation.
\newblock In {\em MICCAI}, 2015.

\bibitem{ruiz2023dreambooth}
Nataniel Ruiz, Yuanzhen Li, Varun Jampani, Yael Pritch, Michael Rubinstein, and Kfir Aberman.
\newblock Dream{B}ooth: Fine tuning text-to-image diffusion models for subject-driven generation.
\newblock In {\em CVPR}, 2023.

\bibitem{salimans2016improved}
Tim Salimans, Ian Goodfellow, Wojciech Zaremba, Vicki Cheung, Alec Radford, and Xi~Chen.
\newblock Improved techniques for training {GAN}s.
\newblock In {\em NeurIPS}, 2016.

\bibitem{shang2023post}
Yuzhang Shang, Zhihang Yuan, Bin Xie, Bingzhe Wu, and Yan Yan.
\newblock Post-training quantization on diffusion models.
\newblock In {\em CVPR}, 2023.

\bibitem{shomron2021post}
Gil Shomron, Freddy Gabbay, Samer Kurzum, and Uri Weiser.
\newblock Post-training sparsity-aware quantization.
\newblock In {\em NeurIPS}, 2021.

\bibitem{Simonyan_2015_ICLR}
Karen Simonyan and Andrew Zisserman.
\newblock Very deep convolutional networks for large-scale image recognition.
\newblock In {\em ICLR}, 2015.

\bibitem{so2024temporal}
Junhyuk So, Jungwon Lee, Daehyun Ahn, Hyungjun Kim, and Eunhyeok Park.
\newblock Temporal dynamic quantization for diffusion models.
\newblock In {\em NeurIPS}, 2024.

\bibitem{song2020denoising}
Jiaming Song, Chenlin Meng, and Stefano Ermon.
\newblock Denoising diffusion implicit models.
\newblock In {\em ICLR}, 2020.

\bibitem{tang2025post}
Siao Tang, Xin Wang, Hong Chen, Chaoyu Guan, Zewen Wu, Yansong Tang, and Wenwu Zhu.
\newblock Post-training quantization with progressive calibration and activation relaxing for text-to-image diffusion models.
\newblock In {\em ECCV}, 2024.

\bibitem{wei2022qdrop}
Xiuying Wei, Ruihao Gong, Yuhang Li, Xianglong Liu, and Fengwei Yu.
\newblock Q{D}rop: Randomly dropping quantization for extremely low-bit post-training quantization.
\newblock In {\em ICLR}, 2022.

\bibitem{Wu_2024_NeurIPS}
Junyi Wu, Haoxuan Wang, Yuzhang Shang, Mubarak Shah, and Yan Yan.
\newblock Ptq4dit: Post-training quantization for diffusion transformers.
\newblock In {\em NeurIPS}, 2024.

\bibitem{yang2019quantization}
Jiwei Yang, Xu~Shen, Jun Xing, Xinmei Tian, Houqiang Li, Bing Deng, Jianqiang Huang, and Xian-sheng Hua.
\newblock Quantization networks.
\newblock In {\em CVPR}, 2019.

\bibitem{yao2024timestep}
Yuzhe Yao, Feng Tian, Jun Chen, Haonan Lin, Guang Dai, Yong Liu, and Jingdong Wang.
\newblock Timestep-aware correction for quantized diffusion models.
\newblock In {\em ECCV}, 2024.

\bibitem{yin2024one}
Tianwei Yin, Micha{\"e}l Gharbi, Richard Zhang, Eli Shechtman, Fredo Durand, William~T Freeman, and Taesung Park.
\newblock One-step diffusion with distribution matching distillation.
\newblock In {\em CVPR}, 2024.

\bibitem{yu2015lsun}
Fisher Yu, Ari Seff, Yinda Zhang, Shuran Song, Thomas Funkhouser, and Jianxiong Xiao.
\newblock {LSUN}: Construction of a large-scale image dataset using deep learning with humans in the loop.
\newblock {\em arXiv preprint arXiv:1506.03365}, 2015.

\bibitem{yuan2021ptq4vit}
Zhihang Yuan, Chenhao Xue, Yiqi Chen, Qiang Wu, and Guangyu Sun.
\newblock {PTQ4ViT}: Post-training quantization framework for vision transformers with twin uniform quantization.
\newblock In {\em ECCV}, 2020.

\bibitem{zhang2023redi}
Kexun Zhang, Xianjun Yang, William Yang~Wang, and Lei Li.
\newblock Redi: Efficient learning-free diffusion inference via trajectory retrieval.
\newblock In {\em ICML}, 2023.

\bibitem{zhang2018unreasonable}
Richard Zhang, Phillip Isola, Alexei~A. Efros, Eli Shechtman, and Oliver Wang.
\newblock The unreasonable effectiveness of deep features as a perceptual metric.
\newblock In {\em CVPR}, 2018.

\bibitem{zhao2019improving}
Ritchie Zhao, Yuwei Hu, Jordan Dotzel, Chris De~Sa, and Zhiru Zhang.
\newblock Improving neural network quantization without retraining using outlier channel splitting.
\newblock In {\em ICML}, 2019.

\end{thebibliography}


\newpage
\appendix
\section{Detailed algorithm for AccuQuant}
\label{appendix:algorithm}
\subsection{Algorithm table for AccuQuant}
\vspace{-10mm}
\begin{algorithm}
\caption{Pseudo code of AccuQuant.}
\label{alg:accuquant}
\begin{lstlisting}[language=Python]
# M : group size
# x_T : random gaussian noise
# fp_model : full-precision model
# quant_model : quantized model
# gather_output() : one iteration of diffusion denoising step
# sg() : stop gradient operation

# Initialize
x_t = x_T
t = T
group_index = 0

for i in range(number_of_groups):
    # Gather x_(t-M) from Full Precision Model
    with torch.no_grad():
        fp_x_M = gather_output(fp_model,x_t,t)

    # Reconstruction stage
    quant_model.set_group(group_index)
    optimizer=torch.optim.Adam(quant_params,learning_rate)

    for epoch in range(epochs):
        # Gather \tilde x_(t-M) with stop gradient
        with torch.no_grad():
            sg_quant_x_M = gather_output(quant_model,x_m,m,quant_params)
        
        optimizer.zero_grad()

        # Accumulate the gradient
        for m in range(M):
            # Calculate gradient scaling factor
            g_m =sqrt_alpha_M / sqrt_alpha_m

            # Gather \tilde x_(t-m)
            quant_x_m = gather_output(quant_model,x_m,m,quant_params)

            # Compute \tilde x_(t-M) with stop gradient
            quant_x_M = sg_quant_x_M - sg(quant_x_m) + quant_x_m

            # Compute Loss for current step
            loss_accuquant = torch.mse(fp_x_M - quant_x_M) * g_m

            # Update quantization parameters with accumulated gradients
            loss_accuquant.backward()
        
        optimizer.step()

    # update indices
    x_t = fp_x_M
    t = t-M
    group_index += 1
\end{lstlisting}
\end{algorithm}
    
\subsection{Detailed derivations of Eq.~\eqref{eq:gradient_approximation}}
\label{appendix:gm_proof}

We provide a detailed derivation of Eq.~\eqref{eq:gradient_approximation}. We can represent the gradient between consecutive denoising steps for the quantized diffusion model as follows:
\begin{equation}
    \frac{\partial \tilde{x}_{t-1}}{\partial \tilde{x}_{t}} = \frac{\sqrt{\alpha_{t-1}}}{\sqrt{\alpha_{t}}}I+\left( -\frac{\sqrt{\alpha_{t-1}}\sqrt{1-\alpha_{t}}}{\sqrt{\alpha_{t}}} + \sqrt{1-\alpha_{t-1}} \right) \frac{\partial \tilde{\epsilon}_\theta(\tilde{x}_t,t)}{\partial \tilde{x}_{t}}.
    \label{eq:real_gradient_supple}
\end{equation}
Here, $I$ is the identity matrix, and $\tilde{\epsilon}_\theta(\tilde{x}_t,t)$ is the output of the quantized model. We have empirically observed in Fig.~\ref{fig:gradient_approximation} that the first term in~Eq.~\eqref{eq:real_gradient} dominates the entire gradient, whereas the second one is negligible. Based on this observation, we approximate the gradient in Eq.~\eqref{eq:real_gradient} by omitting the second term as follows:
\begin{equation} 
    \frac{\partial \tilde{x}_{t-1}}{\partial \tilde{x}_{t}} \approx \frac{\sqrt{\alpha_{t-1}}}{\sqrt{\alpha_{t}}}I. \label{eq:appendix_grad} 
\end{equation}
Using Eq.~\eqref{eq:appendix_grad} in Eq.~\eqref{eq:g_m_approximation}, we obtain the approximated $g_m$ as follows:
\begin{align} 
    g_m &= 
    \begin{cases} 
        1 & m = M \\ 
        \prod_{j=1}^{M-m}\frac{\partial \tilde{x}_{{t}-M-1+j}}{\partial \tilde{x}_{{t}-M+j}}, & m \neq M, 
    \end{cases}\\
    &\approx
    \begin{cases} 
        1 & m = M \\ 
        \prod_{j=1}^{M-m}\frac{\partial x_{{t}-M-1+j}}{\partial x_{{t}-M+j}}, & m \neq M, 
    \end{cases}\\
    &\approx    
    \begin{cases} 
        1 & m = M \\
        \prod_{j=1}^{M-m}\frac{\sqrt{\alpha_{t-M-1+j}}}{\sqrt{\alpha_{t-M+j}}}I, & m \neq M, 
    \end{cases}\\
    &= \frac{\sqrt{\alpha_{t-M}}}{\sqrt{\alpha_{t-m}}}I, \quad \text{for } m = 1,2,\dots,M. \label{eq:appendix_g_m} 
\end{align}

\subsection{Detailed derivation of the objective in Eq.~\eqref{eq:AccuQuantloss} with the gradient approximation}
\label{appendix:loss_proof}
We derive a loss function in Eq.~\eqref{eq:AccuQuantloss} incorporating our gradient approximation of Eq.~\eqref{eq:appendix_g_m}. First, by substituting Eq.\eqref{eq:appendix_g_m} into Eq.~\eqref{eq:group_gradient}, we obtain:
\begin{align} 
    \frac{\partial \mathcal{L}_{\text{MSE}}}{\partial s_l} 
    &= \frac{\partial\mathcal{L}_{\text{MSE}}}{\partial \tilde{x}_{{t}-M}} \left[\sum_{m=1}^{M} g_m \frac{\partial \tilde{x}_{{t}-m}}{\partial s_l} \right] \\
    &\approx \frac{\partial\mathcal{L}_{\text{MSE}}}{\partial \tilde{x}_{{t}-M}} \left[\sum_{m=1}^{M} \frac{\sqrt{\alpha_{t-M}}}{\sqrt{\alpha_{t-m}}} \frac{\partial \tilde{x}_{{t}-m}}{\partial s_l} \right]\\
    &= 2 (\tilde{x}_{t-M} - x_{t-M})^\top \left[\sum_{m=1}^{M} \frac{\sqrt{\alpha_{t-M}}}{\sqrt{\alpha_{t-m}}} \frac{\partial \tilde{x}_{{t}-m}}{\partial s_l} \right] \\
    &= \sum_{m=1}^{M} \frac{\sqrt{\alpha_{t-M}}}{\sqrt{\alpha_{t-m}}} \left[ 2(\tilde{x}_{t-M} - x_{t-M})^\top \frac{\partial \tilde{x}_{{t}-m}}{\partial s_l} \right]. 
    \label{eq:appendix_group_gradient} 
\end{align}
Note that we omit identity matrix in Eq.~\eqref{eq:appendix_g_m} for simple notation. From Eq.~\eqref{eq:appendix_group_gradient}, we derive a loss function whose gradient matches the derived expression by adding $x_{t-m}$ and subtracting a detached version of $x_{t-m}$ within the squared term:
\begin{equation}
    \mathcal{L}_{\text{AccuQuant}} = \sum_{m=1}^{M} \frac{\sqrt{\alpha_{t-M}}}{\sqrt{\alpha_{t-m}}} \lVert \texttt{sg}(x_{t-M} - \tilde{x}_{t-M}) + \tilde{x}_{t-m} - \texttt{sg}(\tilde{x}_{t-m})\rVert_2^2 .
    \label{eq:appendix_loss}
\end{equation}
In this formulation, we compute $x_{t-M}$ and $\tilde{x}_{t-M}$ before updating gradients without saving intermediate feature maps and treat it as a constant during optimization. We then compute loss term inside the summation and accumulate its gradients progressively over multiple denoising steps. After completing $M$ denoising steps, we update the step size using the cumulative gradient. The detailed pipeline of our method is provided in Algorithm~\ref{alg:accuquant}.

\section{Implementation details}
\label{sec:details_appendix}
For weight quantization, we adopt a uniform quantizer identical for all denoising steps, following Q-Diffusion~\cite{li2023q}, since the weights remain invariant across different denoising steps. 
For the CIFAR-10~\cite{krizhevsky2009learning}, we employ the DDIM~\cite{song2020denoising} with 100 total denoising steps, grouping them into 20 sets of 5 denoising steps each and setting learning rate in $\{1\times10^{-3},\,4\times10^{-4}\}$. For the LSUN-Bedroom~\cite{yu2015lsun}, we use LDM-4~\cite{rombach2022high} and the DDIM sampler~\cite{song2020denoising} with 200 total denoising steps, grouping them into 20 sets of 10 denoising steps each and setting learning rate into $4\times10^{-5}$. For the LSUN-Church~\cite{yu2015lsun}, we use LDM-8~\cite{rombach2022high} and the DDIM sampler~\cite{song2020denoising} with 500 total denoising steps, grouping them into 20 sets of 25 denoising steps each and setting learning rate into $4\times10^{-5}$. For the ImageNet~\cite{deng2009imagenet}, we employ LDM-4~\cite{rombach2022high} and the DDIM sampler~\cite{song2020denoising} with 20 total denoising steps and forming 10 groups and setting learning rate into $1\times10^{-3}$. For text-to-image generation, we employ Stable-Diffusion v1.4~\footnote{\url{https://huggingface.co/CompVis/stable-diffusion-v-1-4-original}\label{fn:sd}} with the PNDM sampler~\cite{liu2022pseudo}, using 50 total denoising steps and forming 25 groups and setting learning rate into $1\times10^{-5}$. We set the calibration batch size to 8 for DDIM~\cite{song2020denoising} and LDMs~\cite{rombach2022high}, and to 1 for Stable Diffusion~\cite{rombach2022high}. All other experimental settings are identical to those of Q-Diffusion~\cite{li2023q}. We also retain the default settings~\cite{rombach2022high} during the sampling phase, ensuring that we can conduct all experiments on a single A100~(80 GB) GPU. Also, to ensure a fair comparison, we quantize every layer of TFMQ-DM~\cite{huang2024tfmq}, such as attention-modules and skip-connections for unconditional image generation. Finally, we evaluate the FID~\cite{heusel2017gans} and IS~\cite{salimans2016improved} scores, including the FID2FP32 metric~\cite{tang2025post}, using the torch-fidelity library~\footnote{\url{https://github.com/toshas/torch-fidelity}} and the official Guided Diffusion~\cite{dhariwal2021diffusion} codebase.

\section{Detailed analysis of quantitative results}
In this section, we analyze the quantitative results presented in Tables 1–3.
Firstly, Table 1 shows that AccuQuant outperforms other methods in most settings. In particular, AccuQuant achieves overwhelmingly superior performance on the FID2FP32~\cite{tang2025post} metric across all bit-widths, and this effect becomes more pronounced at lower bit-widths~(\textit{\eg, 3/6 bits}). Furthermore, AccuQuant achieves higher scores on the FID~\cite{heusel2017gans}, sFID~\cite{nash2021generating}, and IS~\cite{salimans2016improved} metrics compared to previous methods~\cite{li2023q,huang2024tfmq,yao2024timestep}, which are not designed for reducing the accumulated quantization error. These findings indicate that AccuQuant effectively reduces accumulated quantization error allowing better performance and confirms that minimizing the accumulated quantization error constitutes a crucial factor in diffusion model quantization.
Secondly, for the class-conditional image generation task reported in Table 2, we draw on previous work~\cite{yao2024timestep} showing that FID~\cite{heusel2017gans} does not perform reliably on ImageNet~\cite{deng2009imagenet} and therefore we report LPIPS~\cite{zhang2018unreasonable}, PSNR, and SSIM metrics to quantify differences between the full precision and quantized models. Likewise to FID2FP32~\cite{tang2025post}, these metrics measure how closely the quantized model’s outputs match those of the full precision model. We find that AccuQuant outperforms previous methods in terms of FID2FP32~\cite{tang2025post}, LPIPS~\cite{zhang2018unreasonable}, PSNR, and SSIM across all bit-widths, demonstrating that AccuQuant generates images that most closely resemble the full precision model’s outputs while maintaining high visual quality. These results imply that accumulated quantization error remains critical even when employing few denoising steps, such as 20. Furthermore, even compared to PTQD~\cite{he2023ptqd}, which is designed to reduce accumulating quantization error, AccuQuant outperforms it in FID2FP32~\cite{tang2025post}, LPIPS~\cite{zhang2018unreasonable}, PSNR, and SSIM across all bit widths, demonstrating the superiority of our framework in explicitly reducing accumulating quantization error.
Finally, the text-to-image generation results in Table 3 reveal that, despite PCR~\cite{tang2025post} use of more bit-widths~(\textit{\ie,} 8.4 bits for activations) and adopting independent quantizer at every denoising steps, AccuQuant achieves a better FID2FP32~\cite{tang2025post} score. Additionally, AccuQuant attains the highest CLIP-Score~\cite{hessel2022clipscore}, indicating that it not only reproduces images similar to the full precision model but also aligns them effectively with the text prompts.
To this end, our findings demonstrate that the AccuQuant framework outperforms existing methods for reducing accumulated quantization error, and that AccuQuant not only achieves results comparable to those of the full-precision model but also confirms the fidelity term by resulting superior performance on FID~\cite{heusel2017gans}, sFID~\cite{nash2021generating}, and CLIP score~\cite{hessel2022clipscore}.

\begin{figure*}[!t]
    \centering
    \captionsetup[subfigure]{font=small, labelformat=empty}
    
    \begin{subfigure}[c]{1.0\linewidth}
       \centering
       \begin{tabular}{c @{} c @{} c @{} c @{} c @{} c @{} c @{} c @{} c}
          \includegraphics[width=0.1\linewidth]{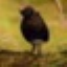} &
          \includegraphics[width=0.1\linewidth]{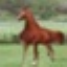} &
          \includegraphics[width=0.1\linewidth]{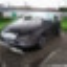} &
          \includegraphics[width=0.1\linewidth]{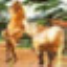} &
          \includegraphics[width=0.1\linewidth]{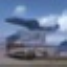} &
          \includegraphics[width=0.1\linewidth]{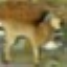} &
          \includegraphics[width=0.1\linewidth]{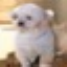} &
          \includegraphics[width=0.1\linewidth]{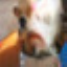} &
          \includegraphics[width=0.1\linewidth]{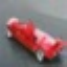} \\
       \end{tabular}
    \end{subfigure}
    \vspace{-1.5mm}

    \begin{subfigure}[c]{1.0\linewidth}
       \centering
       \begin{tabular}{c @{} c @{} c @{} c @{} c @{} c @{} c @{} c @{} c}
          \includegraphics[width=0.1\linewidth]{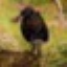} &
          \includegraphics[width=0.1\linewidth]{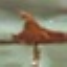} &
          \includegraphics[width=0.1\linewidth]{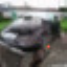} &
          \includegraphics[width=0.1\linewidth]{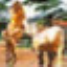} &
          \includegraphics[width=0.1\linewidth]{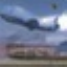} &
          \includegraphics[width=0.1\linewidth]{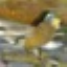} &
          \includegraphics[width=0.1\linewidth]{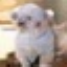} &
          \includegraphics[width=0.1\linewidth]{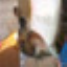} &
          \includegraphics[width=0.1\linewidth]{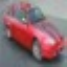} \\
       \end{tabular}
    \end{subfigure}
    \vspace{-1.5mm}

    \begin{subfigure}[c]{1.0\linewidth}
       \centering
       \begin{tabular}{c @{} c @{} c @{} c @{} c @{} c @{} c @{} c @{} c}
          \includegraphics[width=0.1\linewidth]{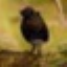} &
          \includegraphics[width=0.1\linewidth]{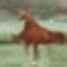} &
          \includegraphics[width=0.1\linewidth]{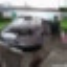} &
          \includegraphics[width=0.1\linewidth]{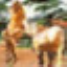} &
          \includegraphics[width=0.1\linewidth]{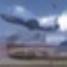} &
          \includegraphics[width=0.1\linewidth]{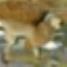} &
          \includegraphics[width=0.1\linewidth]{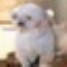} &
          \includegraphics[width=0.1\linewidth]{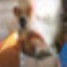} &
          \includegraphics[width=0.1\linewidth]{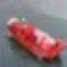} \\
       \end{tabular}
    \end{subfigure}
    \vspace{-1.5mm}

    \begin{subfigure}[c]{1.0\linewidth}
       \centering
       \begin{tabular}{c @{} c @{} c @{} c @{} c @{} c @{} c @{} c @{} c}
          \includegraphics[width=0.1\linewidth]{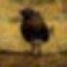} &
          \includegraphics[width=0.1\linewidth]{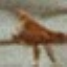} &
          \includegraphics[width=0.1\linewidth]{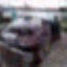} &
          \includegraphics[width=0.1\linewidth]{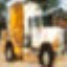} &
          \includegraphics[width=0.1\linewidth]{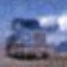} &
          \includegraphics[width=0.1\linewidth]{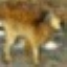} &
          \includegraphics[width=0.1\linewidth]{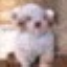} &
          \includegraphics[width=0.1\linewidth]{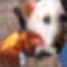} &
          \includegraphics[width=0.1\linewidth]{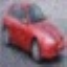} \\
       \end{tabular}
    \end{subfigure}
    \vspace{-1.5mm}
    
    \caption{Visual comparisons of DDIM~\cite{song2020denoising} for unconditional image generation on CIFAR-10~\cite{krizhevsky2009learning} under various group sizes. Both weights and activations were quantized to 6 bits for a visually distinct and clear comparison. From top to bottom, the presented sequences correspond to the full-precision model, our method with group sizes of 1, 5, and 100.}
    \label{fig:quali_groupsize}
    \vspace{-5mm}
\end{figure*}

\section{Qualitative results for group sizes}
In this section, we present the qualitative results for various group sizes, as reported in Table~\ref{tab:group_size}. We observe from Fig.~\ref{fig:quali_groupsize} that selecting an appropriate group size (e.g., group size of 5) generate images that closely resembling those of the full-precision model, under low bit-widhts.
 For example, in the first row of generated horse images, we can see that when the group size is set to 1, the quantized model fails to adequately account for accumulated errors, leading to the generation of distorted and unrealistic shapes. Similarly, when the group size is excessively large (e.g., group size of 100), the quantized model is unable to effectively manage the substantial accumulation of errors, resulting in unrealistic image outputs. In contrast, with an appropriately chosen group size (e.g., group size of 5), the accumulated errors are effectively mitigated, allowing the generated images to maintain a structure and appearance comparable to those of the full-precision model. In the dog images of column 6, using a group size of 5 yields images that most closely match the full-precision model in both texture and shape, outperforming group sizes 1 and 100. This indicates that an appropriate group sizes is a crucial factor to mitigate the accumulating quantization error.

 \begin{table}[t]
    \caption{Quantitative comparisons of AccuQuant with and without the gradient approximation via various group sizes. We quantize DDIM~\cite{song2020denoising} on CIFAR-10~\cite{krizhevsky2009learning} under a 4/8-bit setting.}
    \vspace{2mm}
    \centering
    \resizebox{0.6\textwidth}{!}{
    \begin{tabular}{ccc}
    \toprule
    \multirow{2}{*}{Group size} & \multicolumn{2}{c}{FID~$\downarrow$ / FID2FP32~$\downarrow$} \\ \cline{2-3}
    & \multicolumn{1}{c}{w/o Approximation} & \multicolumn{1}{c}{w/ Approximation} \\
    \midrule
    ~~1 & 6.88 / 3.11 & 6.88 / 3.11 \\
    ~~5 & 5.26 / 2.32 & \textbf{4.75} / \textbf{1.15} \\
    10 & 5.27 / \textbf{1.62} & \textbf{5.07} / 1.88 \\
    25 & 5.73 / 1.49     & \textbf{5.24} / 1.49 \\
    \bottomrule
    \end{tabular}}
    \label{tab:ablation_groupgrad}
    \vspace{-3mm}
\end{table}

 \begin{figure}[!t]
    \vspace{-1mm}        
    \centering
    \includegraphics[width=0.6\linewidth]{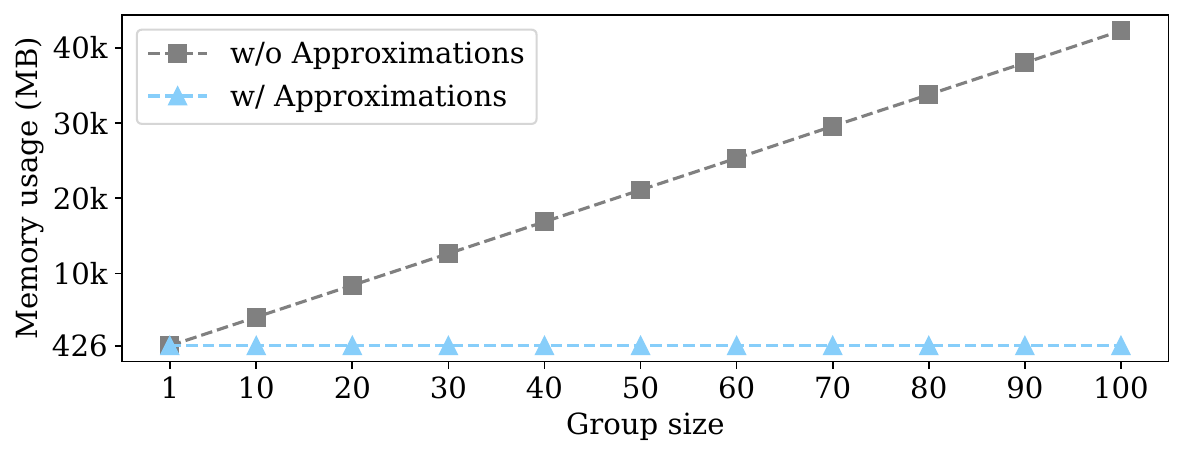}
    \vspace{-1mm}        
    \caption{Comparisons of a total amount of memory of AccuQuant with and without the gradient approximation. We measure the memory usage during the calibration process of AccuQuant under a 6/6-bit setting for DDIM~\cite{song2020denoising} on CIFAR-10~\cite{krizhevsky2009learning}.}
    \label{fig:grad_memory}
    \vspace{-2mm}        
\end{figure}
 
\section{Quantitative results for group sizes and gradient approximation}
In this section, we show in Table.~\ref{tab:ablation_groupgrad} quantitative ablation results for varying group sizes, both with and without gradient approximation, for 4/8-bit quantized DDIM~\cite{song2020denoising} on CIFAR-10~\cite{krizhevsky2009learning}. We observe that both FID~\cite{heusel2017gans} and FID2FP32~\cite{tang2025post} degrade at group sizes of 1 and 25, since a group size of 1 fails to capture accumulated quantization error, whereas 25 is too large to find quantization parameters minimizing the error. This result highlights the importance of selecting appropriate group sizes and aligns with the findings discussed in the main paper. We also investigate the impact of gradient approximation. When the group size is 1, we calculate the gradient at every denoising steps and therefore the approximation has no effect on performance. We show that using gradient approximation usually yields better performance, since the gradients of Eq.~\eqref{eq:real_gradient} are corrupted before the quantization parameters $s_l$ converge, leading unstable calibration process. We also show in Fig.~\ref{fig:grad_memory} a total amount of memory of AccuQuant with and without the gradient approximation, \wrt the group size. The results show that the memory requirement grows linearly with group size in the absence of the gradient approximation, whereas it remains constant across all group sizes when the approximation is applied. Consequently, omitting the noisy gradient term effectively reduce memory requirement and ensure calibration process stable.

\begin{figure*}[!t]
    \centering
    \captionsetup[subfigure]{font=small, labelformat=empty, justification=centering}
    \captionsetup{justification=raggedright, singlelinecheck=false}

    \begin{subfigure}[t]{0.48\textwidth}
      \centering
      \makebox[0pt][l]{\hspace*{-36mm}\includegraphics[width=\linewidth]{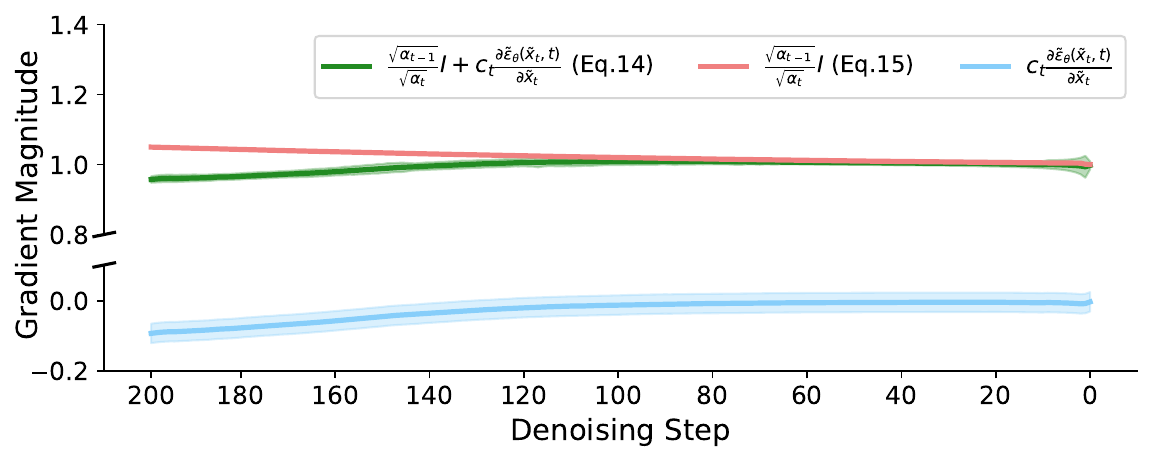}}
      \caption{LSUN-Bedroom}
      \label{fig:grad_bedroom}
    \end{subfigure}
    \hfill
    \begin{subfigure}[t]{0.48\textwidth}
      \centering
      \makebox[0pt][l]{\hspace*{-37mm}\includegraphics[width=\linewidth]{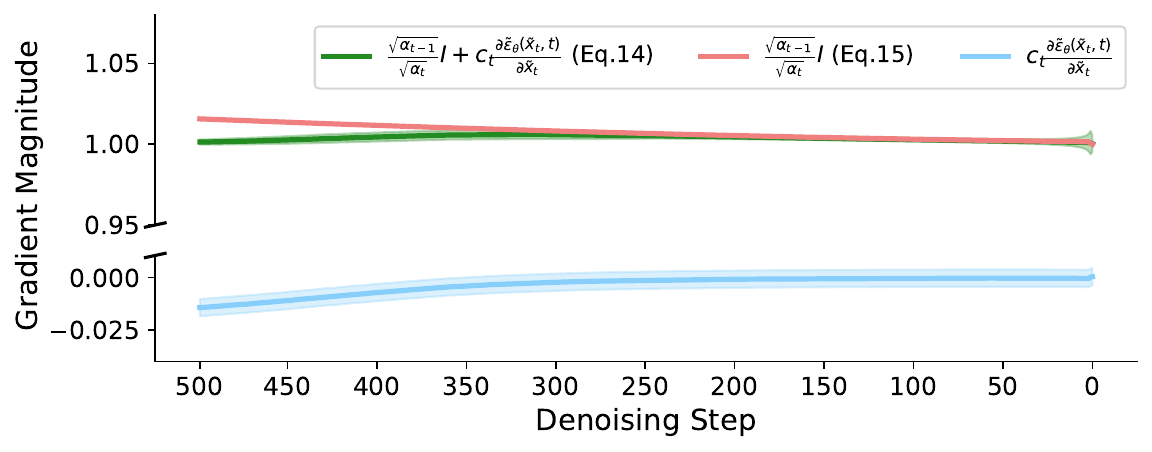}}

      \caption{LSUN-Church}
      \label{fig:grad_church}
    \end{subfigure}

    \caption{Plots of the gradient in Eq.~\eqref{eq:real_gradient_supple} and its components over denoising steps. The gradients are calculated during a calibration process of AccuQuant for LSUN-Bedroom~\cite{yu2015lsun} on the left and LSUN-Church~\cite{yu2015lsun} on the right.}
    \label{fig:gradients_bedroom_church}
\end{figure*}

\section{Gradient approximation across different dataset}
In this section, we visualize the gradient approximation on LDM-4 and LDM-8~\cite{rombach2022high} across diverse datasets, including LSUN-Bedroom and LSUN-Church~\cite{yu2015lsun}, as illustrated in Fig.~\ref{fig:gradients_bedroom_church}. We plot the full gradients of Eq.~\eqref{eq:real_gradient_supple}, the gradient approximation of Eq.~\eqref{eq:appendix_grad}, and the residual term~(\textit{\ie}, $c_t\frac{\partial \tilde{\epsilon}_\theta(\tilde{x}_t,t)}{\partial \tilde{x}_{t}}$) at 200 and 500 denoising steps. The results show that, consistent with the behavior observed in Fig.~\ref{fig:gradients_bedroom_church}, the gradient approximation remains closely aligned with the full gradient throughout multiple denoising steps on both datasets, thereby corroborating the validity of our approximation. Moreover, as discussed in Table.~\ref{tab:grad_approx} and Table.~\ref{tab:ablation_groupgrad}, by omitting the residual term we avoid noisy updates during the calibration phase, resulting in a more stable calibration process.

\begin{table}[!t]
        \sisetup{
        detect-weight   = true,  
        detect-family   = true,   
        mode            = text,  
        }
        \centering
        \scriptsize
        \setlength{\tabcolsep}{3pt}  
        \caption{Quantization results for unconditional image generation with DPM-Solver++ sampler~\cite{lu2022dpmsolver} on CIFAR-10 ($32\times32$)~\cite{krizhevsky2009learning}.}
        \vspace{2mm}
        \label{tab:dpm_solver}
           \centering
            \resizebox{0.7\linewidth}{!}{
            \begin{tabular}{@{}c l c S[table-format=1.2] S[table-format=2.2] S[table-format=1.2]@{}}
            \toprule
            \textbf{Model} & \textbf{Method} & \textbf{Bits(W/A)} & \textbf{IS}$\uparrow$ & \textbf{FID}$\downarrow$ & \textbf{FID2FP32}$\downarrow$ \\
            \midrule
            \multirow{10}{*}{\parbox{1.8cm}{\centering DPM-Solver++\\CIFAR-10\\(steps=50)}} 
                & Full-Precision                    & 32/32 &  9.42 &  3.59 &  0.00 \\
            \cmidrule(lr){2-6}
                & Q-Diffusion~\cite{li2023q}        &  4/8  &  \bfseries\tablenum{9.40} & 10.25 &  7.65 \\
                & \cellcolor[HTML]{ECF9FF}Ours                  &  \cellcolor[HTML]{ECF9FF}4/8  & \cellcolor[HTML]{ECF9FF}9.31 &\cellcolor[HTML]{ECF9FF}\bfseries\tablenum{5.75} &\cellcolor[HTML]{ECF9FF}\bfseries\tablenum{2.57} \\
            \cmidrule(lr){2-6}
                & Q-Diffusion~\cite{li2023q}        &  4/6  & 7.88 & 52.65 & 49.21 \\
                & \cellcolor[HTML]{ECF9FF}Ours                  &  \cellcolor[HTML]{ECF9FF}4/6  & \cellcolor[HTML]{ECF9FF}\bfseries\tablenum{9.33} & \cellcolor[HTML]{ECF9FF}\bfseries\tablenum{7.34} &\cellcolor[HTML]{ECF9FF}\bfseries\tablenum{4.28} \\
            \cmidrule(lr){2-6}
                & Q-Diffusion~\cite{li2023q}        &  3/8  & \bfseries\tablenum{9.57} & 40.41 & 39.04 \\
                & \cellcolor[HTML]{ECF9FF}Ours                  &  \cellcolor[HTML]{ECF9FF}3/8  & \cellcolor[HTML]{ECF9FF}{9.11} & \cellcolor[HTML]{ECF9FF}\bfseries\tablenum{22.36} &\cellcolor[HTML]{ECF9FF}\bfseries\tablenum{20.90} \\
            \bottomrule
            \end{tabular}
            }
        \vspace{-2mm}
\end{table}
    
\section{Evaluation with advanced sampler}
\begin{figure*}[!t]
    \centering
    \captionsetup[subfigure]{font=small, labelformat=empty}
    
    \begin{subfigure}[c]{1.0\linewidth}
       \centering
       \begin{tabular}{c @{} c @{} c @{} c @{} c @{} c @{} c @{} c @{} c @{} c}
          \includegraphics[width=0.097\linewidth]{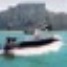} &
          \includegraphics[width=0.097\linewidth]{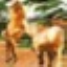} &
          \includegraphics[width=0.097\linewidth]{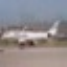} &
          \includegraphics[width=0.097\linewidth]{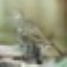} &
          \includegraphics[width=0.097\linewidth]{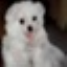} &
          \includegraphics[width=0.097\linewidth]{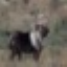} &
          \includegraphics[width=0.097\linewidth]{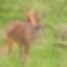} &
          \includegraphics[width=0.097\linewidth]{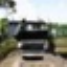} &
          \includegraphics[width=0.097\linewidth]{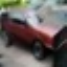} &
          \includegraphics[width=0.097\linewidth]{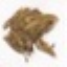} \\
       \end{tabular}
       \vspace{-1.5mm}
    \end{subfigure}
 
    \begin{subfigure}[c]{1.0\linewidth}
       \centering
       \begin{tabular}{c @{} c @{} c @{} c @{} c @{} c @{} c @{} c @{} c @{} c}
          \includegraphics[width=0.097\linewidth]{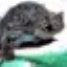} &
          \includegraphics[width=0.097\linewidth]{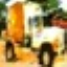} &
          \includegraphics[width=0.097\linewidth]{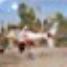} &
          \includegraphics[width=0.097\linewidth]{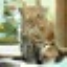} &
          \includegraphics[width=0.097\linewidth]{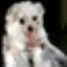} &
          \includegraphics[width=0.097\linewidth]{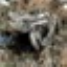} &
          \includegraphics[width=0.097\linewidth]{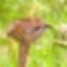} &
          \includegraphics[width=0.097\linewidth]{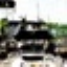} &
          \includegraphics[width=0.097\linewidth]{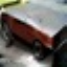} &
          \includegraphics[width=0.097\linewidth]{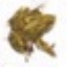} \\
       \end{tabular}
       \vspace{-1.5mm}
    \end{subfigure}

    \begin{subfigure}[c]{1.0\linewidth}
       \centering
       \begin{tabular}{c @{} c @{} c @{} c @{} c @{} c @{} c @{} c @{} c @{} c}
          \includegraphics[width=0.097\linewidth]{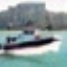} &
          \includegraphics[width=0.097\linewidth]{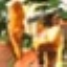} &
          \includegraphics[width=0.097\linewidth]{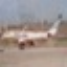} &
          \includegraphics[width=0.097\linewidth]{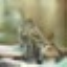} &
          \includegraphics[width=0.097\linewidth]{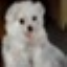} &
          \includegraphics[width=0.097\linewidth]{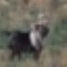} &
          \includegraphics[width=0.097\linewidth]{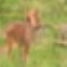} &
          \includegraphics[width=0.097\linewidth]{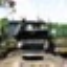} &
          \includegraphics[width=0.097\linewidth]{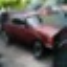} &
          \includegraphics[width=0.097\linewidth]{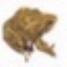} \\
       \end{tabular}
       \vspace{-2mm}
    \end{subfigure}
 
    \caption{Visual comparisons of generated images on CIFAR-10~\cite{krizhevsky2009learning} (32$\times$32) for unconditional image generation with 3rd order DPM-Solver++~\cite{lu2022dpmsolver} under a 4/8-bit setting. From top to bottom, the presented sequences correspond to the Full precision, Q-Diffusion~\cite{li2023q}, and Ours.}
    \label{fig:quali_dpm_suppl}
 \end{figure*}

In this section, we report additional results for the advanced sampler~(\ie, 3rd-order DPM-Solver++~\cite{lu2022dpmsolver}). We have already reported performance for Stable Diffusion~\cite{rombach2022high} with the PNDM sampler~\cite{liu2022pseudo} in Table.~\ref{tab:text_cond}. Here, we show in Table.~\ref{tab:dpm_solver} a quantitative comparisons between our method and Q-Diffusion~\cite{li2023q} for unconditional CIFAR-10 generation~\cite{krizhevsky2009learning} over 50 denoising steps. For quantization, we devide the 50-step denoising process into 17 groups: the first 16 groups contain three consecutive steps each, and the final group contains two steps. 
We observe that our method consistently outperforms Q-Diffusion~\cite{li2023q} in both FID~\cite{heusel2017gans} and FID2FP32~\cite{tang2025post}, demonstrating robustness across different samplers. Notably, in the 4/6-bit setting, our approach yields substantial gains, indicating enhanced tolerance to accumulated quantization error under low-bit activations.
In Fig.~\ref{fig:quali_dpm_suppl}, we present qualitative comparisons between our method and Q-Diffusion~\cite{li2023q} under 4/8-bit quantization. It demonstrates that our approach recovers visual fidelity close to the full-precision model, whereas Q-Diffusion~\cite{li2023q} produces visibly degraded outputs. This qualitative results further validates the ability of our method to mitigate accumulated quantization error.

\section{Additional visualization on various benchmarks}
\begin{figure*}[!t]
   \centering
   \captionsetup[subfigure]{font=small, labelformat=empty}
   
   \begin{subfigure}[c]{1.0\linewidth}
      \centering
      \begin{tabular}{c @{} c @{} c @{} c @{} c @{} c}
         \includegraphics[width=0.16\linewidth]{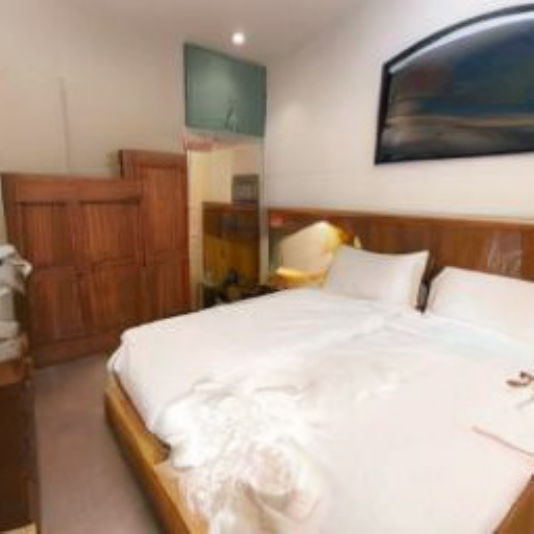} &
         \includegraphics[width=0.16\linewidth]{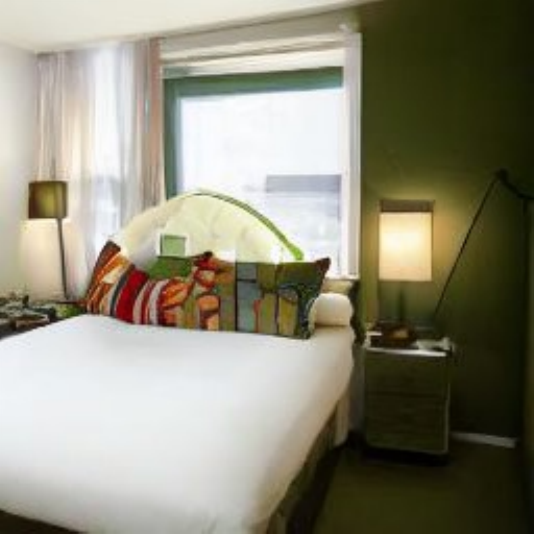} &
         \includegraphics[width=0.16\linewidth]{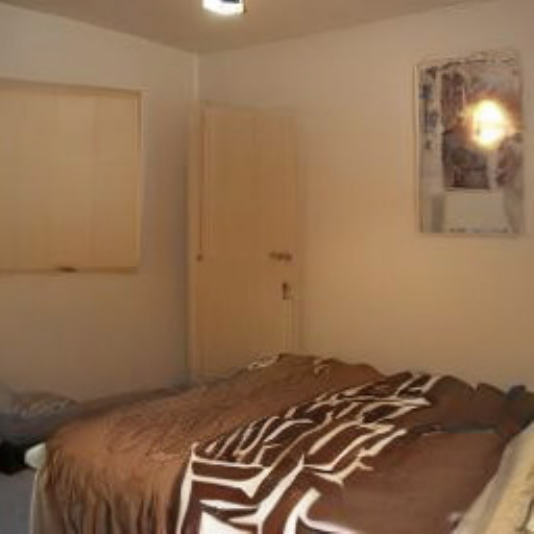} &
         \includegraphics[width=0.16\linewidth]{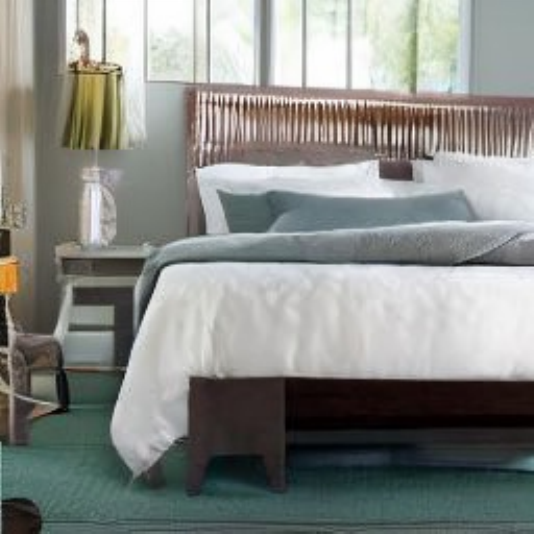} &
         \includegraphics[width=0.16\linewidth]{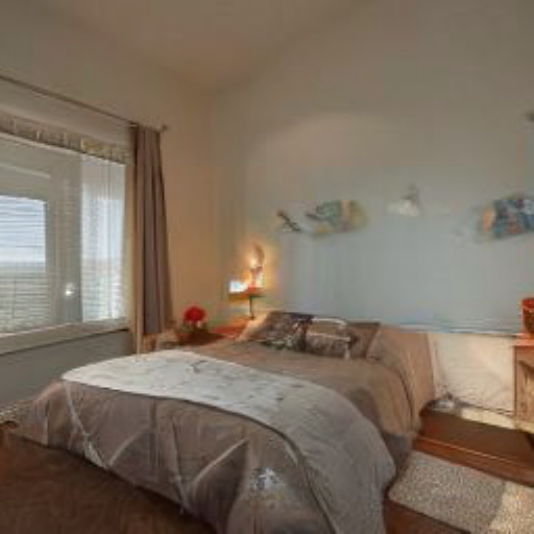} &
         \includegraphics[width=0.16\linewidth]{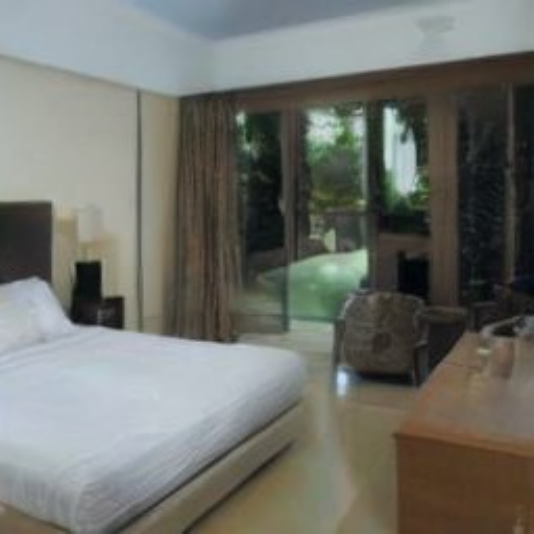} \\

         \includegraphics[width=0.16\linewidth]{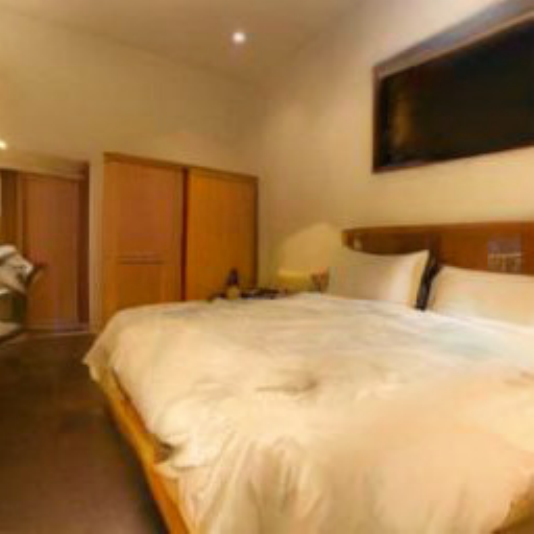} &
         \includegraphics[width=0.16\linewidth]{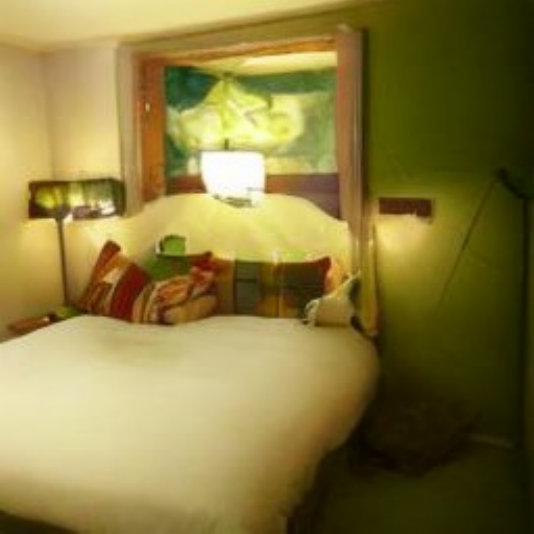} &
         \includegraphics[width=0.16\linewidth]{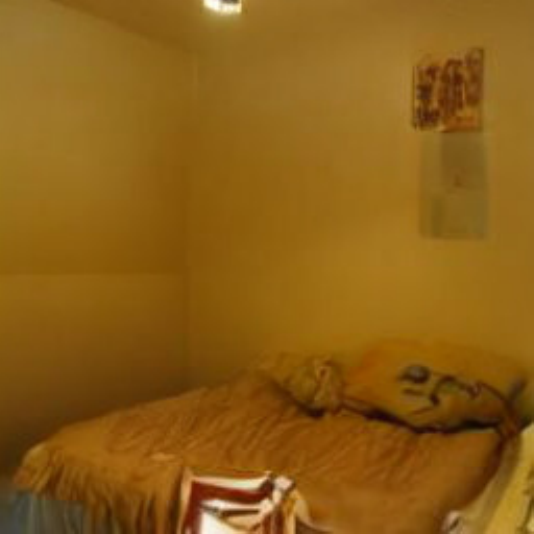} &
         \includegraphics[width=0.16\linewidth]{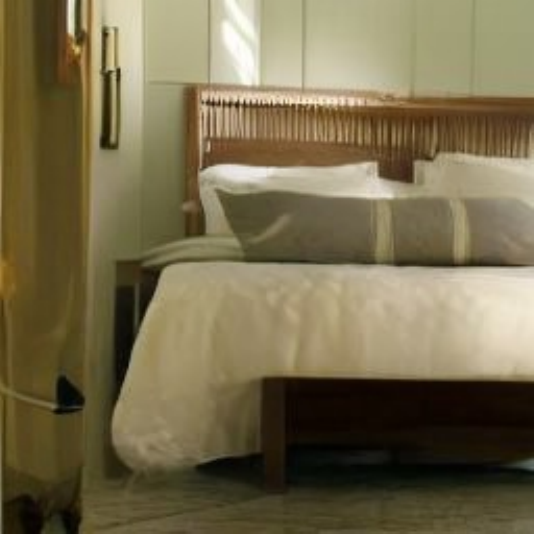} &
         \includegraphics[width=0.16\linewidth]{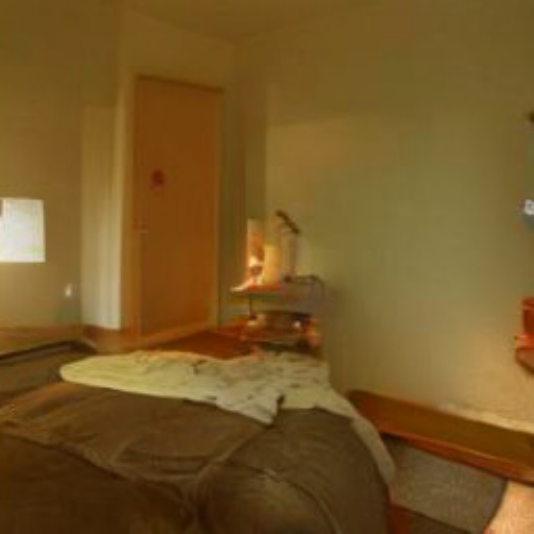} &
         \includegraphics[width=0.16\linewidth]{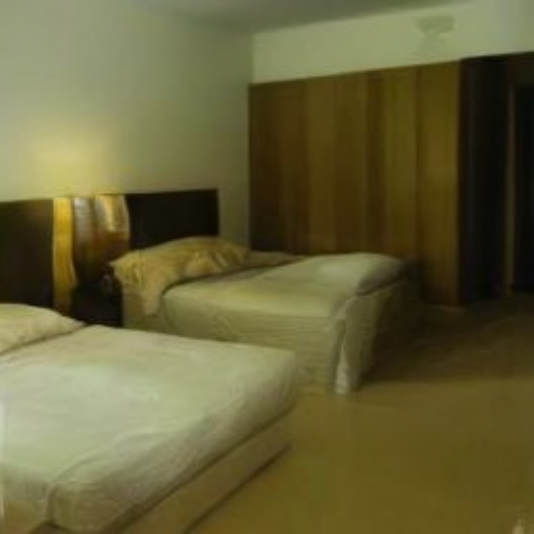} \\

         \includegraphics[width=0.16\linewidth]{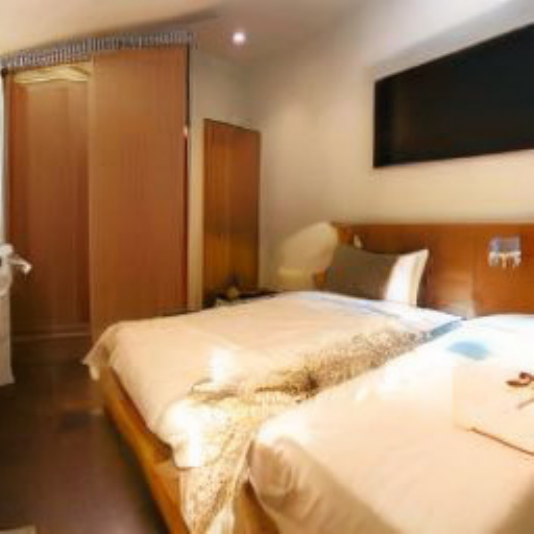} &
         \includegraphics[width=0.16\linewidth]{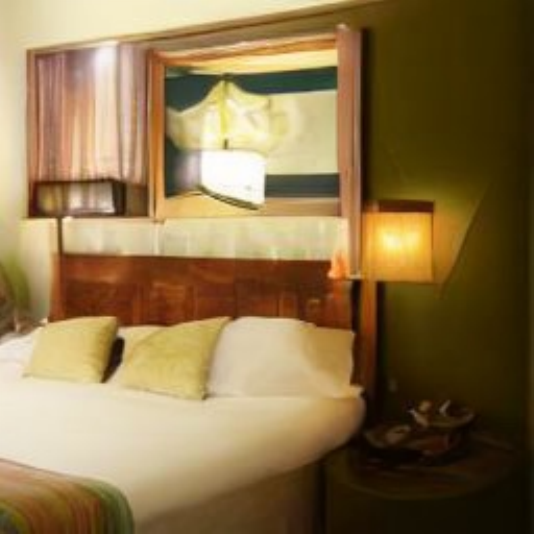} &
         \includegraphics[width=0.16\linewidth]{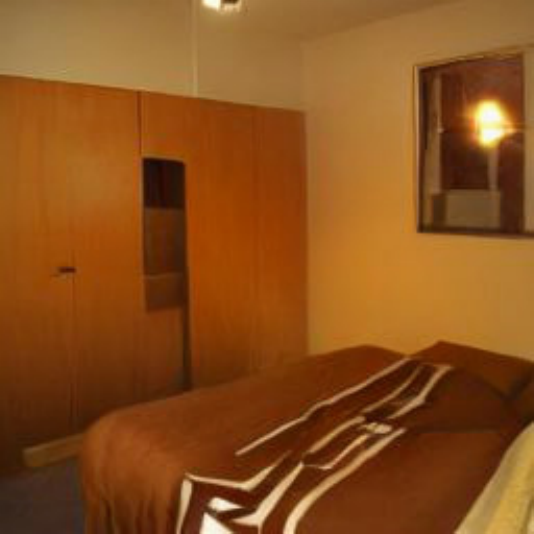} &
         \includegraphics[width=0.16\linewidth]{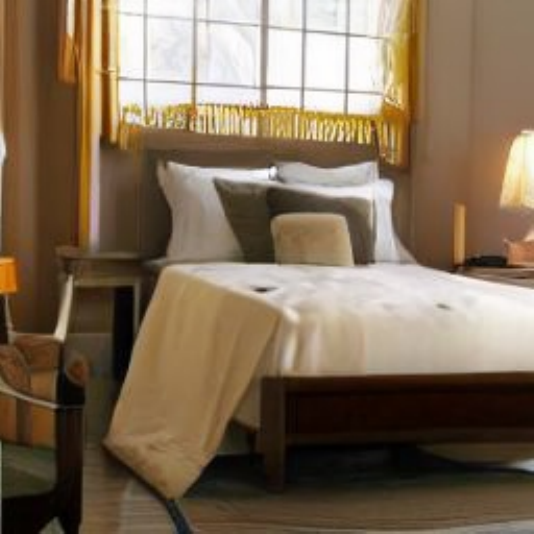} &
         \includegraphics[width=0.16\linewidth]{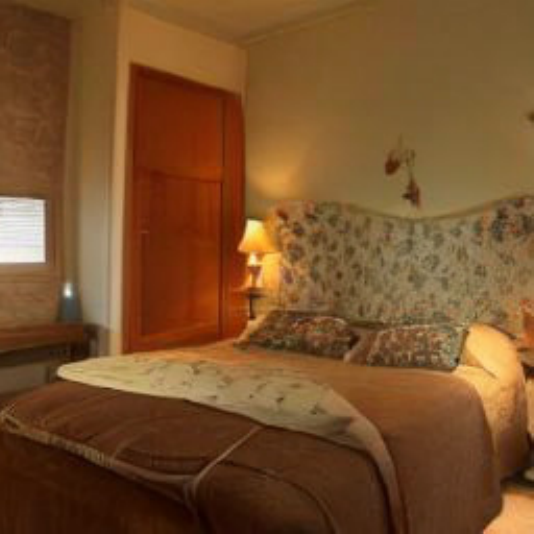} &
         \includegraphics[width=0.16\linewidth]{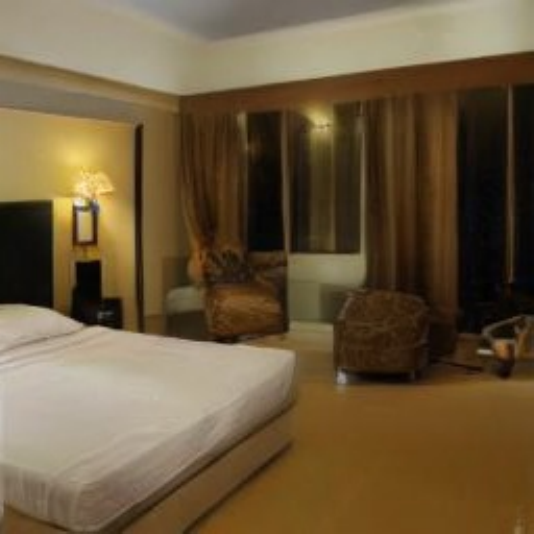} \\

         \includegraphics[width=0.16\linewidth]{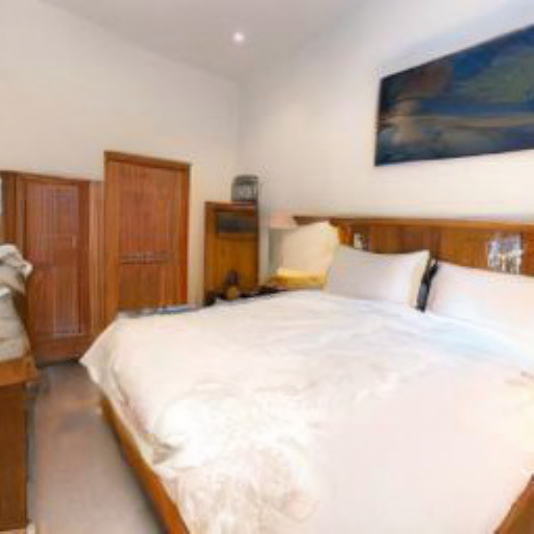} &
         \includegraphics[width=0.16\linewidth]{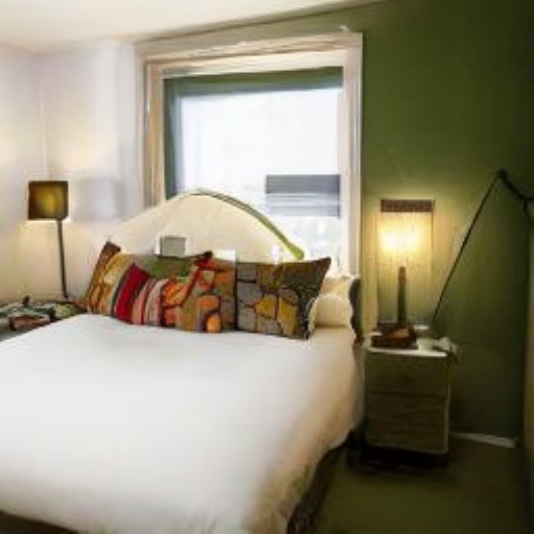} &
         \includegraphics[width=0.16\linewidth]{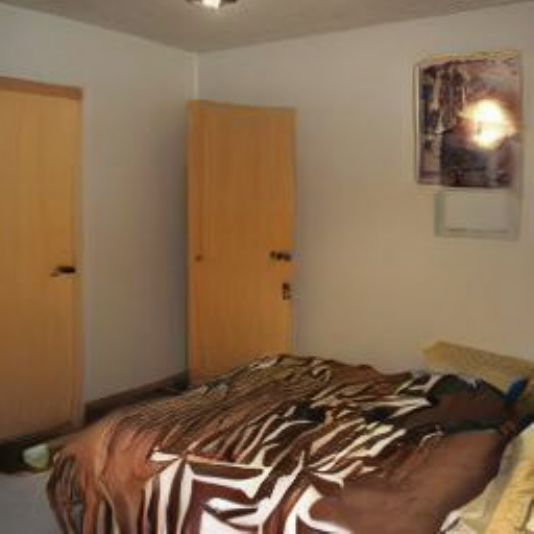} &
         \includegraphics[width=0.16\linewidth]{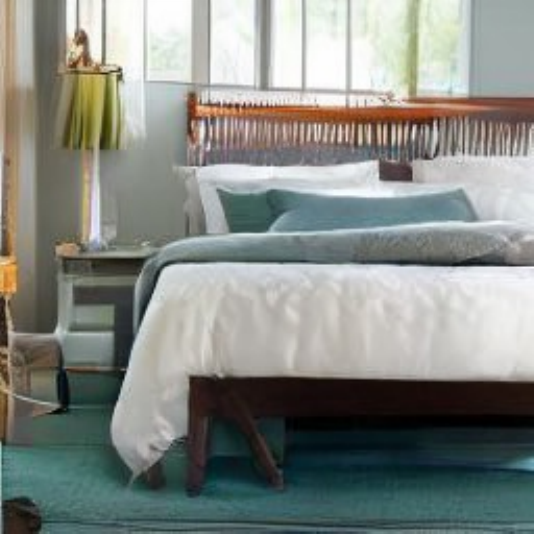} &
         \includegraphics[width=0.16\linewidth]{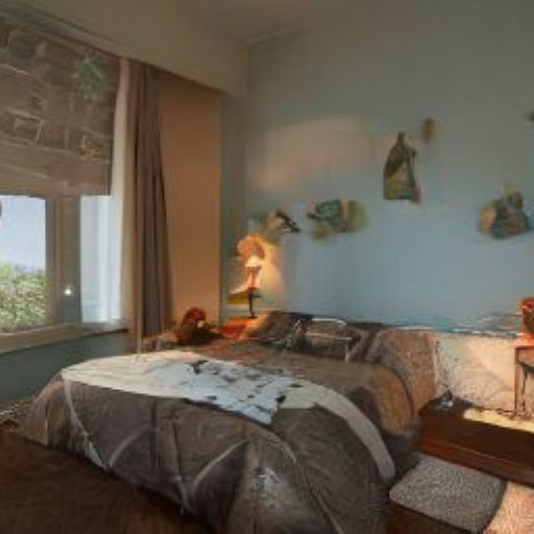} &
         \includegraphics[width=0.16\linewidth]{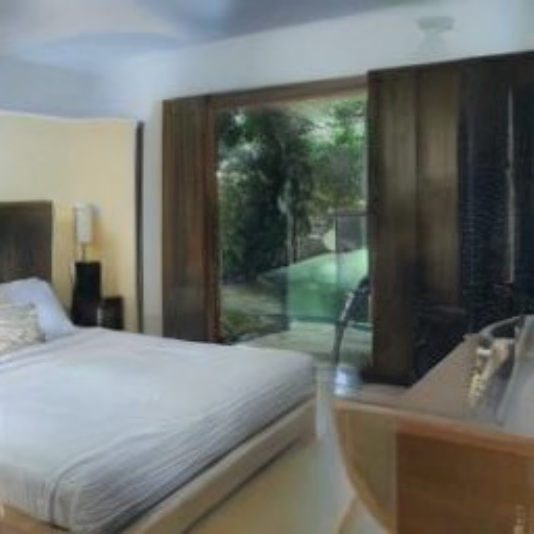} \\

      \end{tabular}
   \end{subfigure}
   \vspace{3mm}

   \begin{subfigure}[c]{1.0\linewidth}
      \centering
      \begin{tabular}{c @{} c @{} c @{} c @{} c @{} c}
         \includegraphics[width=0.16\linewidth]{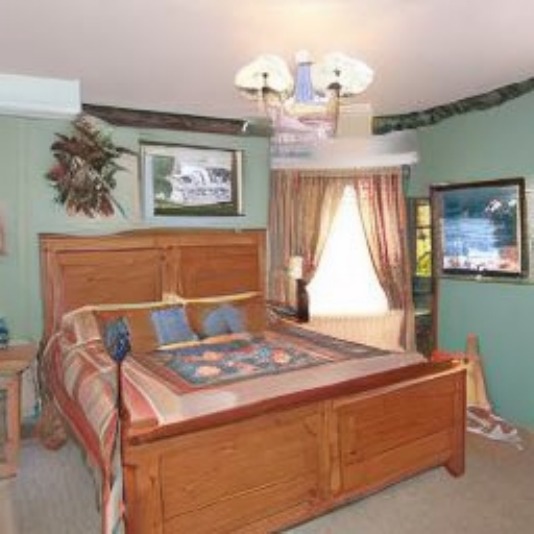} &
         \includegraphics[width=0.16\linewidth]{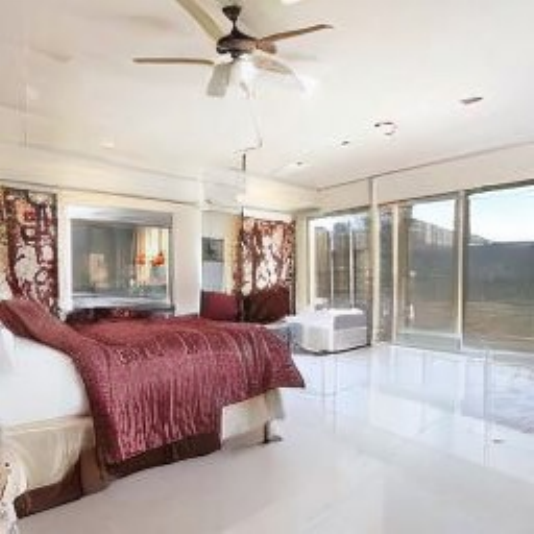} &
         \includegraphics[width=0.16\linewidth]{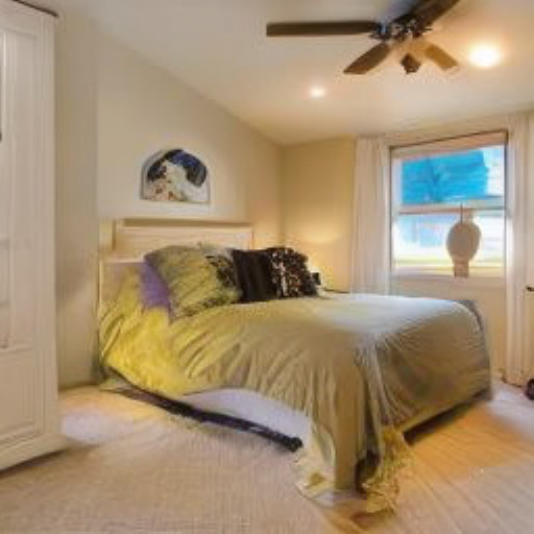} &
         \includegraphics[width=0.16\linewidth]{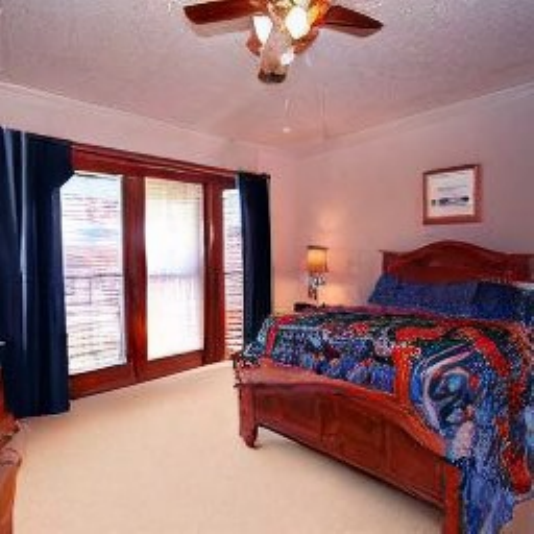} &
         \includegraphics[width=0.16\linewidth]{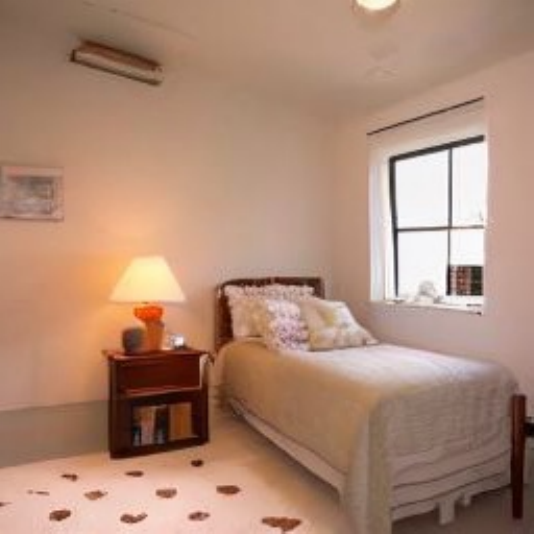} &
         \includegraphics[width=0.16\linewidth]{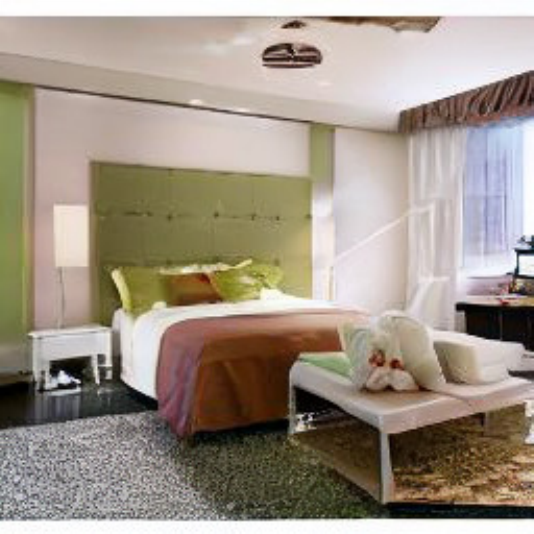} \\

         \includegraphics[width=0.16\linewidth]{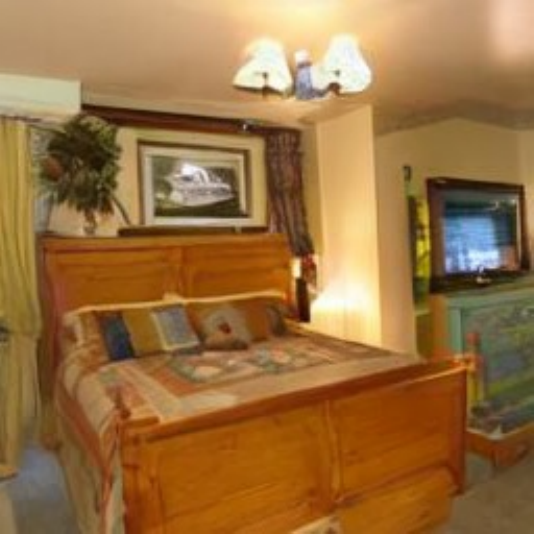} &
         \includegraphics[width=0.16\linewidth]{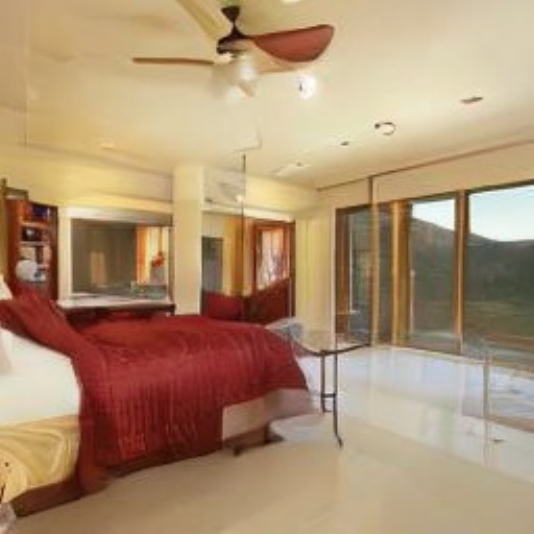} &
         \includegraphics[width=0.16\linewidth]{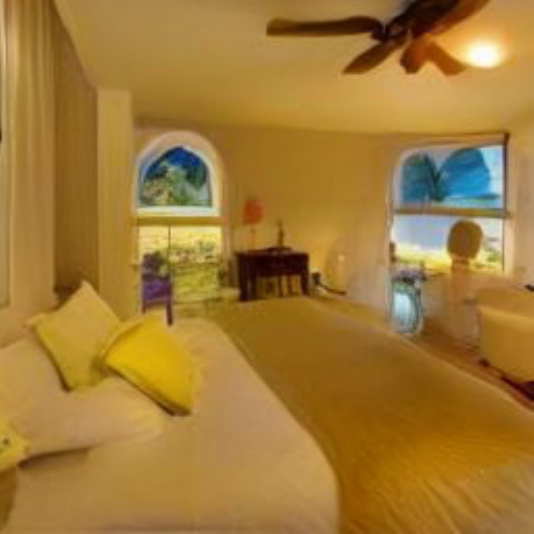} &
         \includegraphics[width=0.16\linewidth]{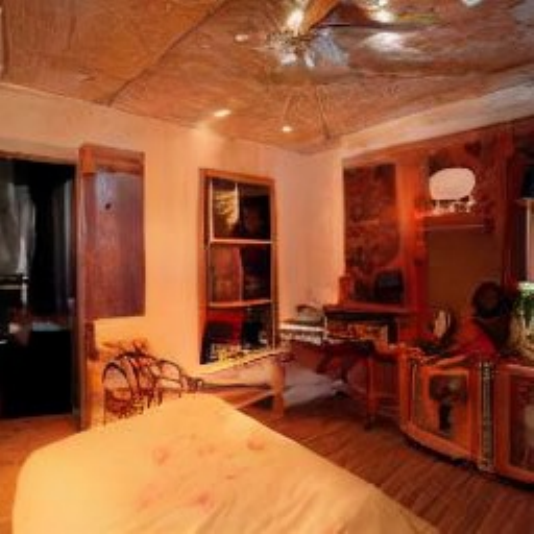} &
         \includegraphics[width=0.16\linewidth]{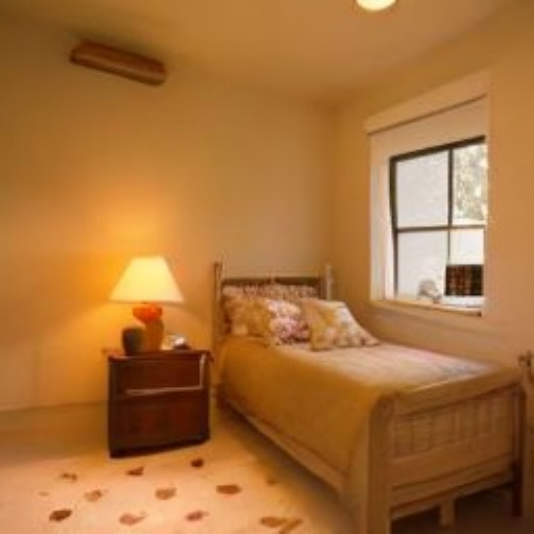} &
         \includegraphics[width=0.16\linewidth]{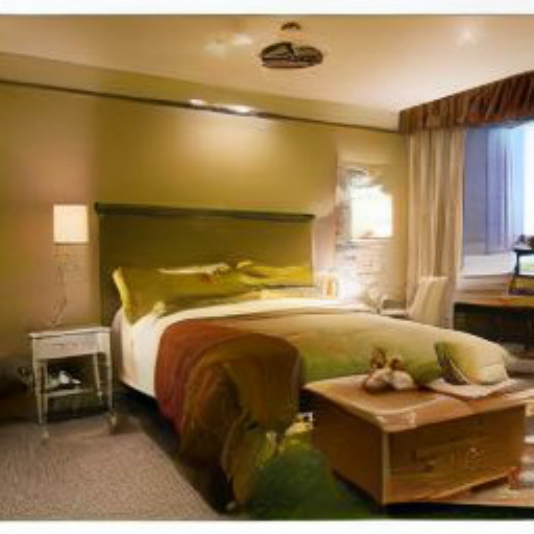} \\

         \includegraphics[width=0.16\linewidth]{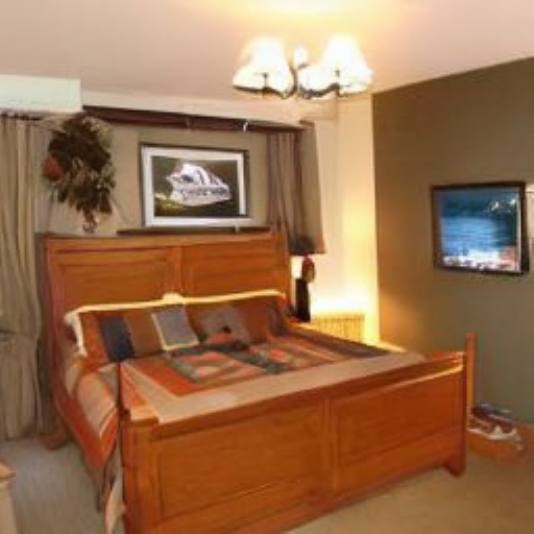} &
         \includegraphics[width=0.16\linewidth]{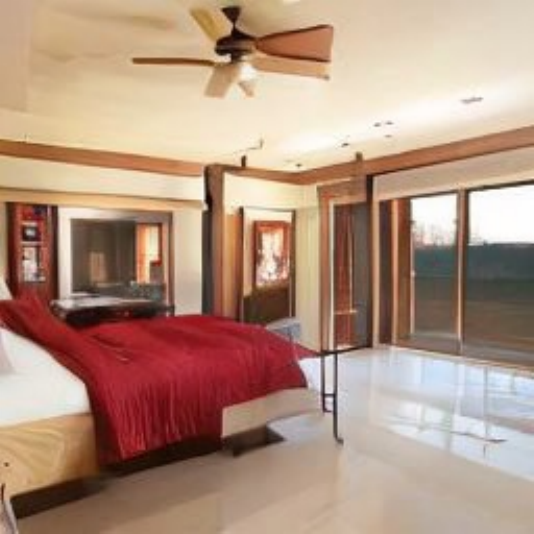} &
         \includegraphics[width=0.16\linewidth]{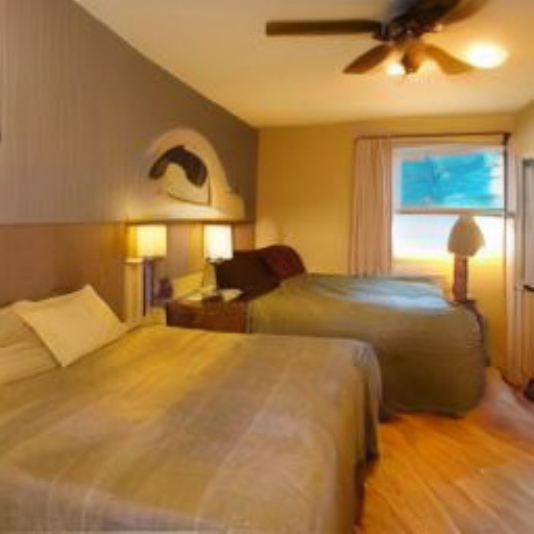} &
         \includegraphics[width=0.16\linewidth]{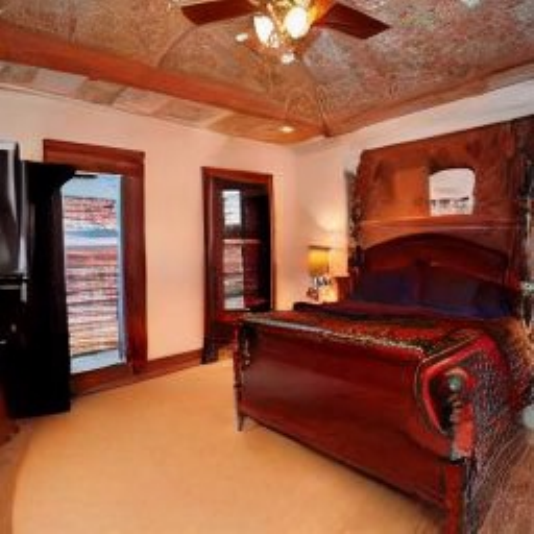} &
         \includegraphics[width=0.16\linewidth]{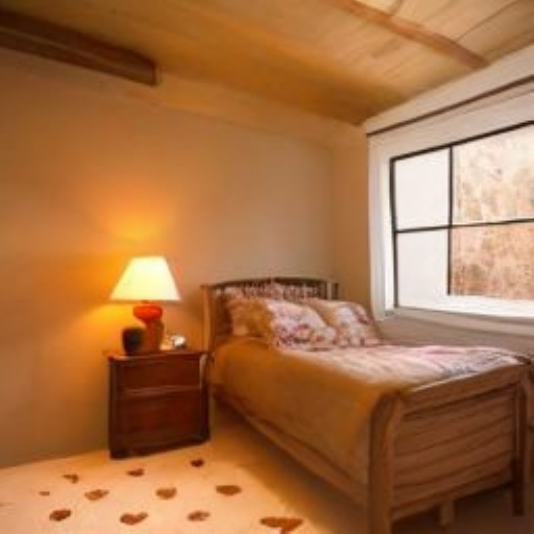} &
         \includegraphics[width=0.16\linewidth]{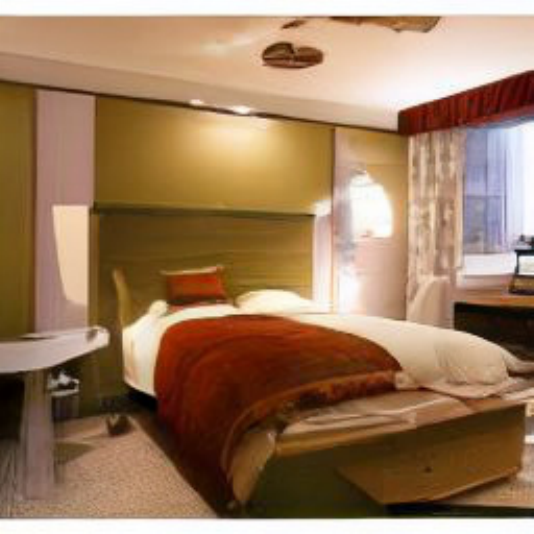} \\

         \includegraphics[width=0.16\linewidth]{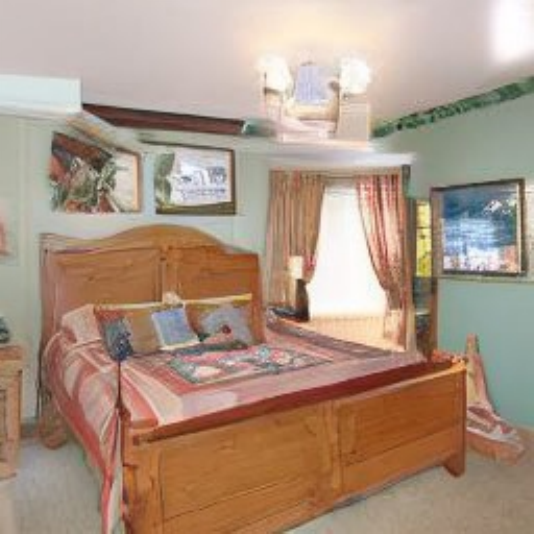} &
         \includegraphics[width=0.16\linewidth]{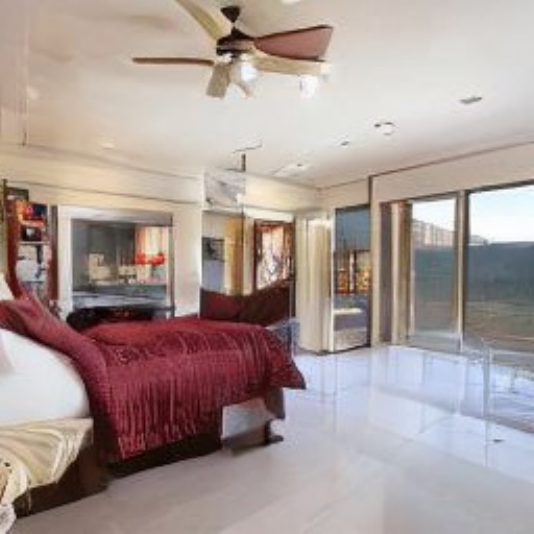} &
         \includegraphics[width=0.16\linewidth]{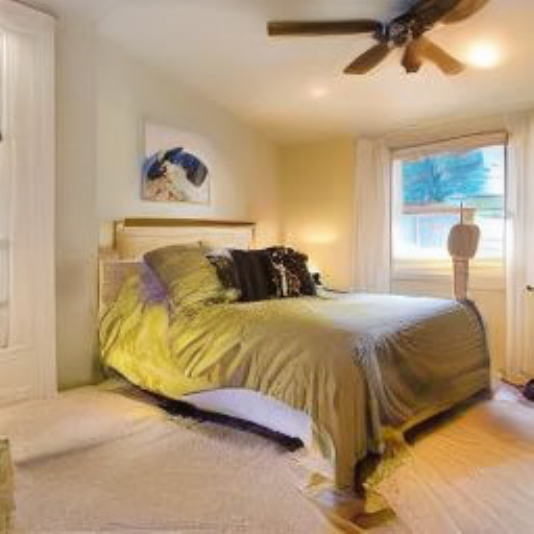} &
         \includegraphics[width=0.16\linewidth]{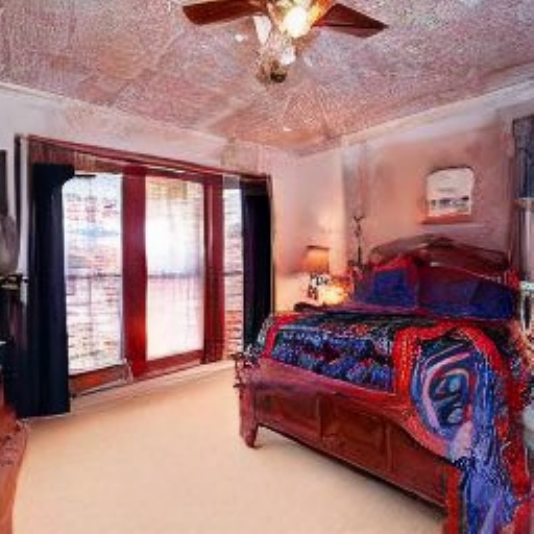} &
         \includegraphics[width=0.16\linewidth]{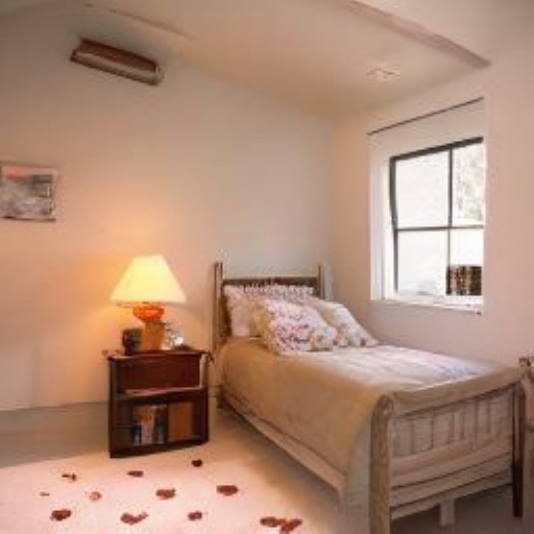} &
         \includegraphics[width=0.16\linewidth]{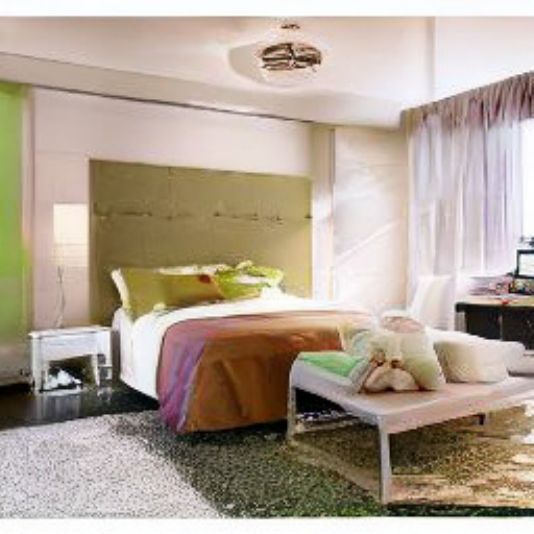} \\
      \end{tabular}
   \end{subfigure}

   \caption{Visual comparisons of generated images on LSUN-Bedrooms~\cite{yu2015lsun} (256$\times$256) for unconditional image generation with LDM-4~\cite{rombach2022high} under a 3/8-bit setting. Each row corresponds to Full Precision, Q-Diffusion~\cite{li2023q}, TFMQ-DM~\cite{huang2024tfmq}, and Ours.}
   \label{fig:quali_ldm_bedroom_suppl}
\end{figure*}

\begin{figure*}[!t]
   \centering
   \captionsetup[subfigure]{font=small, labelformat=empty}
   
   \begin{subfigure}[c]{1.0\linewidth}
      \centering
      \begin{tabular}{c @{} c @{} c @{} c @{} c @{} c}
         \includegraphics[width=0.16\linewidth]{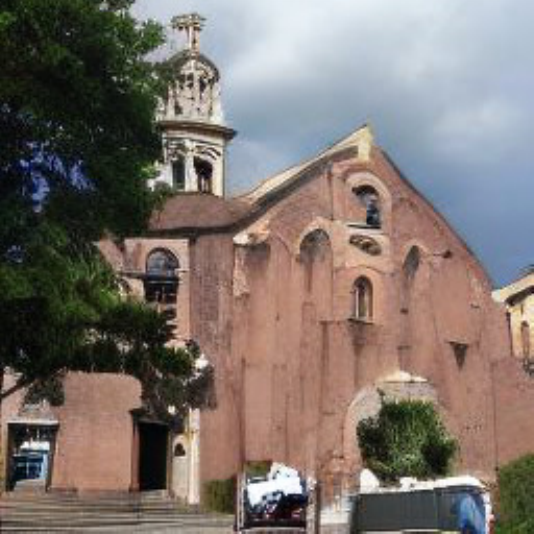} &
         \includegraphics[width=0.16\linewidth]{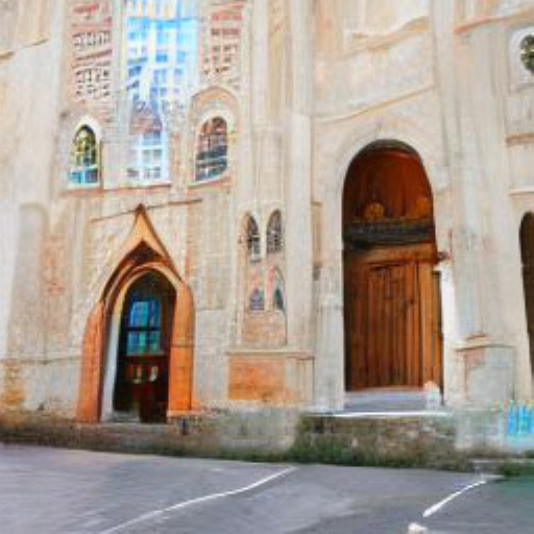} &
         \includegraphics[width=0.16\linewidth]{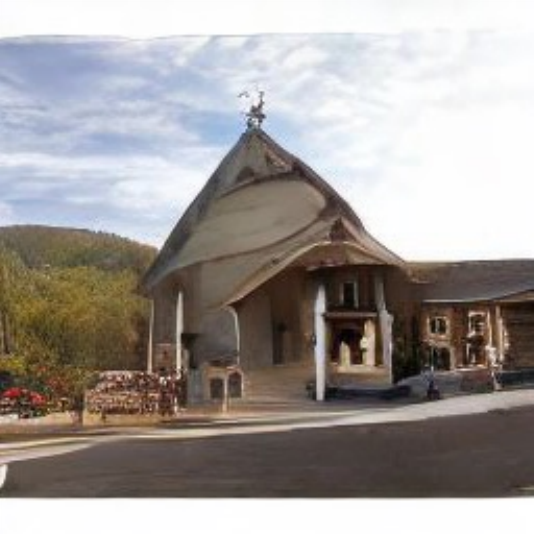} &
         \includegraphics[width=0.16\linewidth]{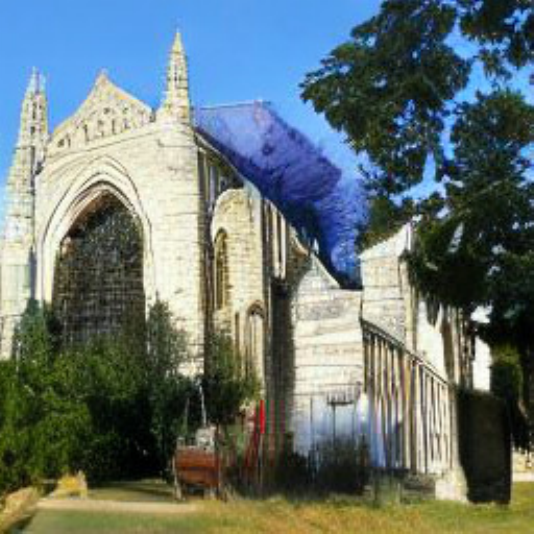} &
         \includegraphics[width=0.16\linewidth]{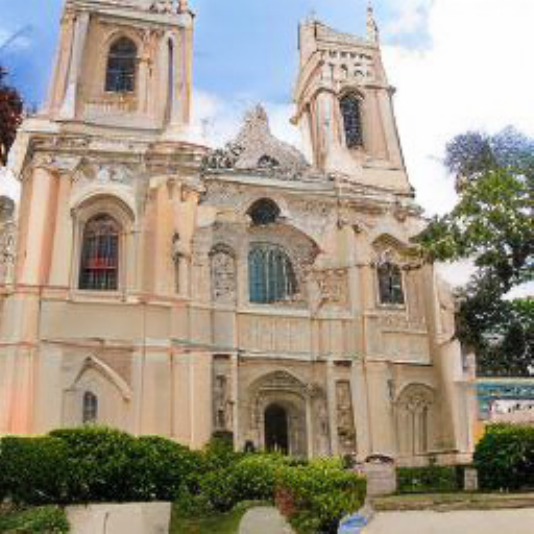} &
         \includegraphics[width=0.16\linewidth]{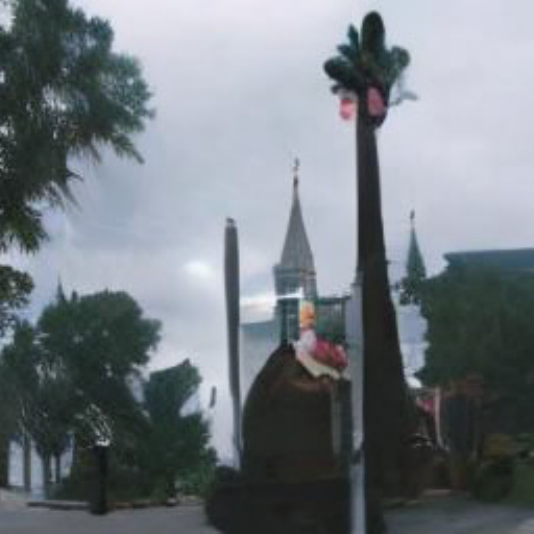} \\

         \includegraphics[width=0.16\linewidth]{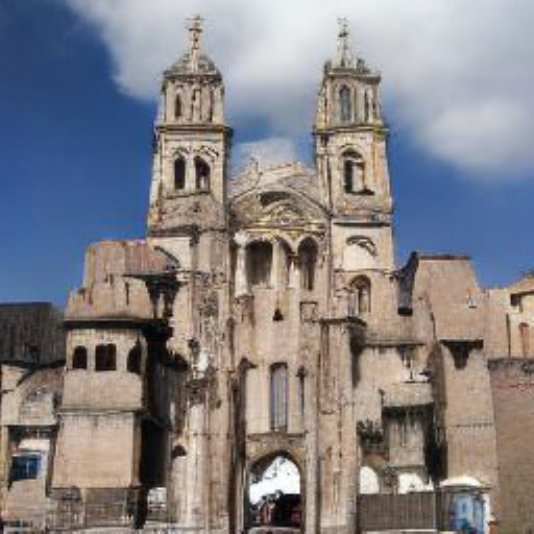} &
         \includegraphics[width=0.16\linewidth]{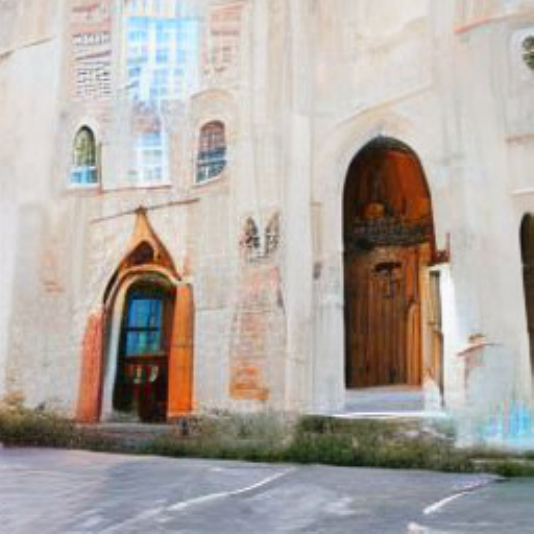} &
         \includegraphics[width=0.16\linewidth]{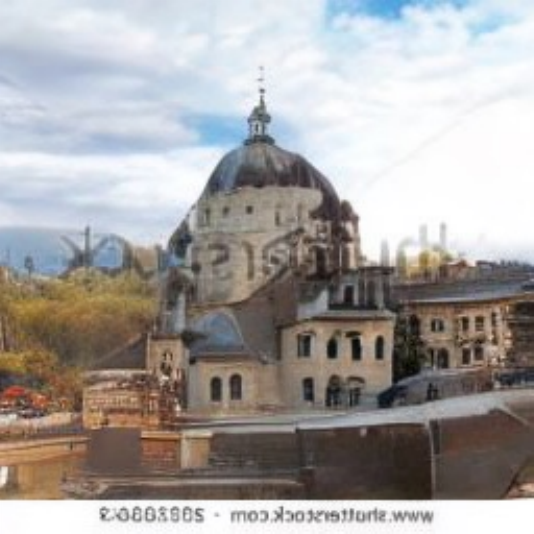} &
         \includegraphics[width=0.16\linewidth]{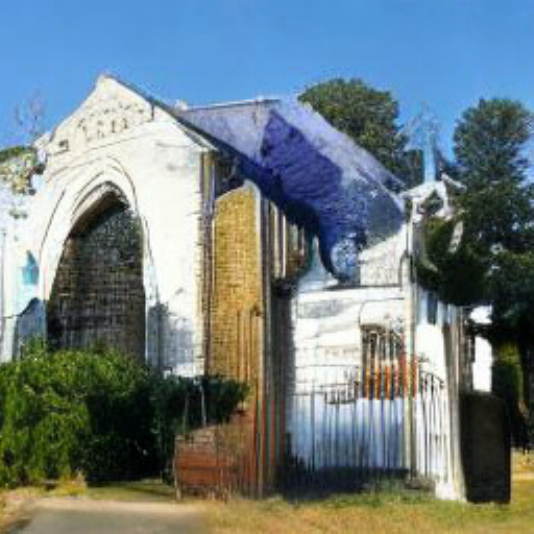} &
         \includegraphics[width=0.16\linewidth]{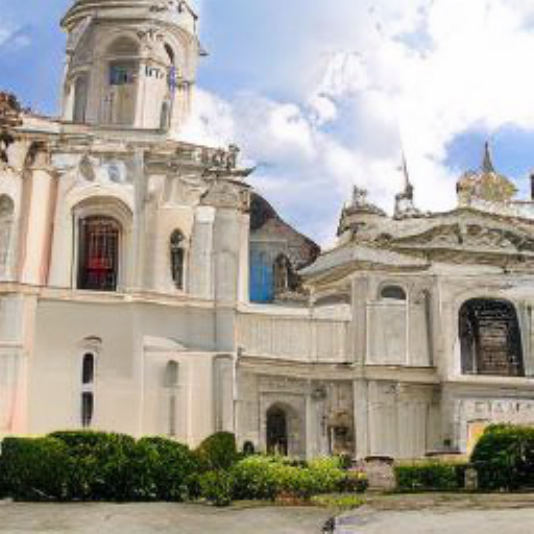} &
         \includegraphics[width=0.16\linewidth]{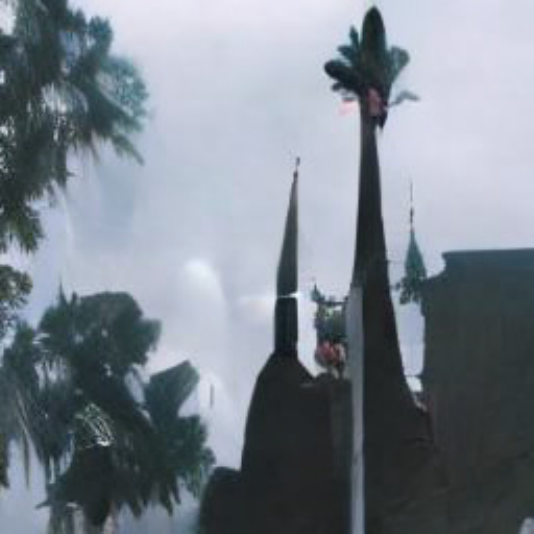} \\

         \includegraphics[width=0.16\linewidth]{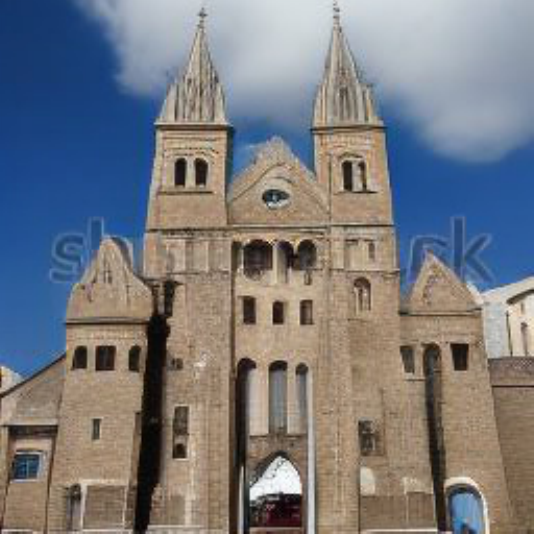} &
         \includegraphics[width=0.16\linewidth]{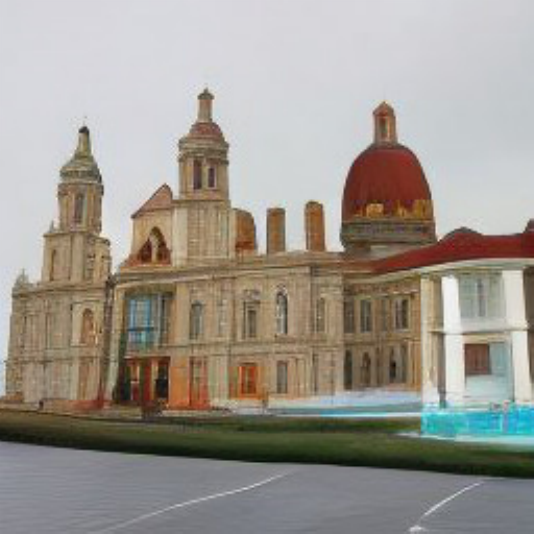} &
         \includegraphics[width=0.16\linewidth]{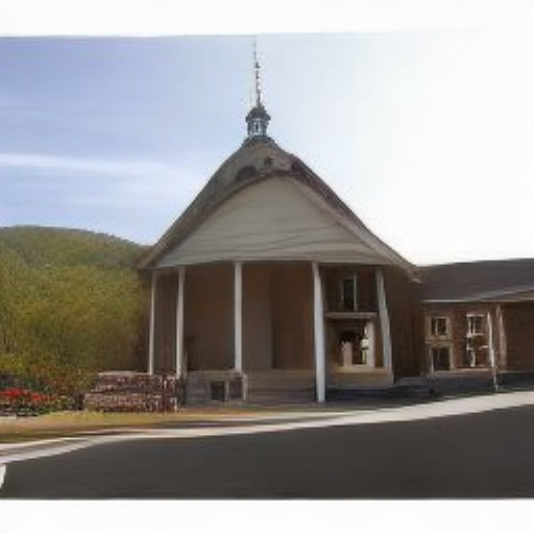} &
         \includegraphics[width=0.16\linewidth]{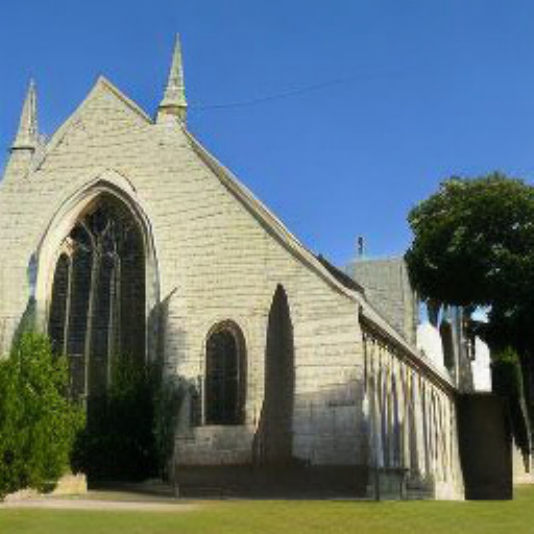} &
         \includegraphics[width=0.16\linewidth]{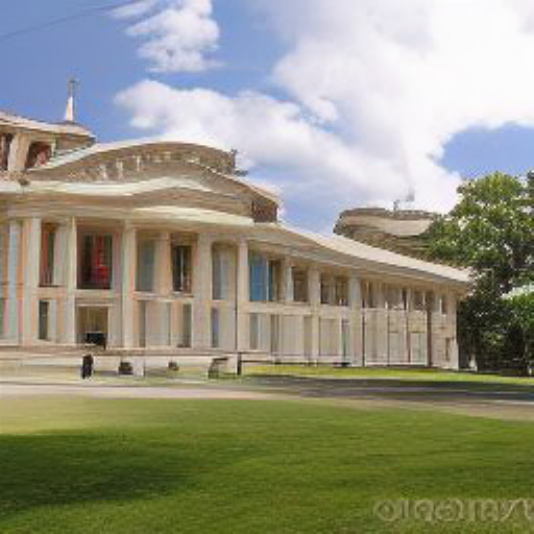} &
         \includegraphics[width=0.16\linewidth]{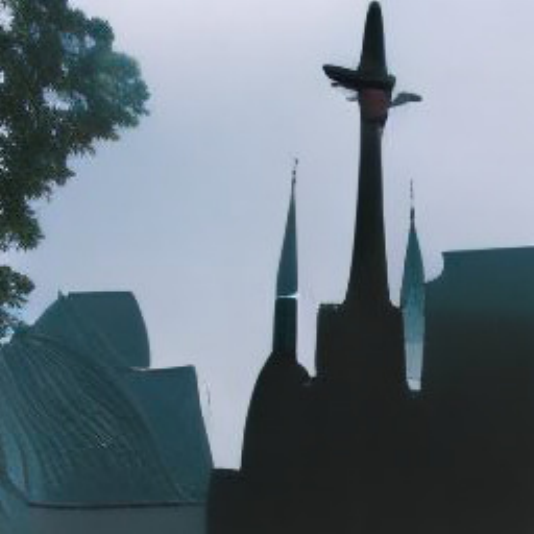} \\

         \includegraphics[width=0.16\linewidth]{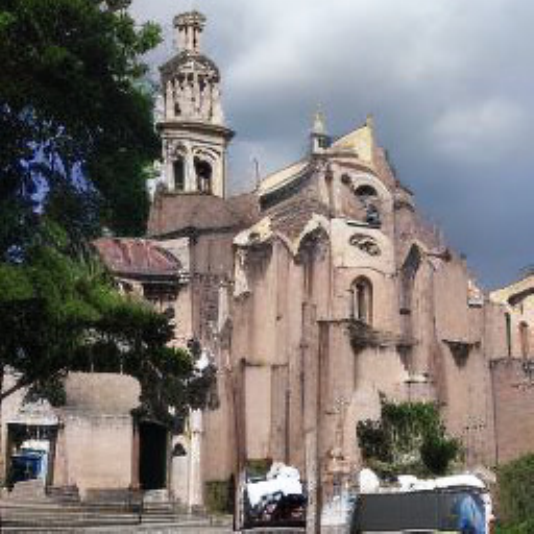} &
         \includegraphics[width=0.16\linewidth]{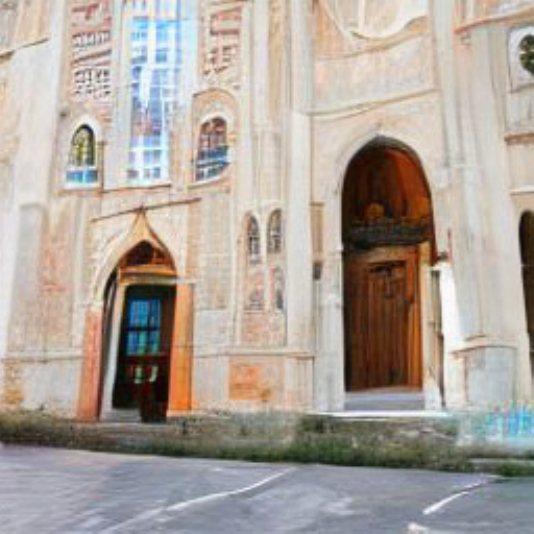} &
         \includegraphics[width=0.16\linewidth]{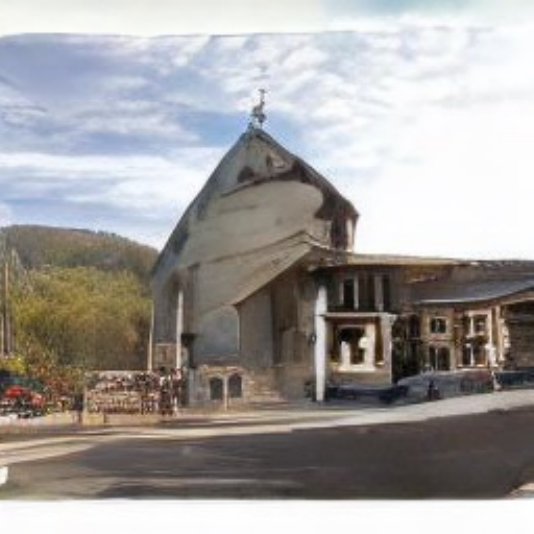} &
         \includegraphics[width=0.16\linewidth]{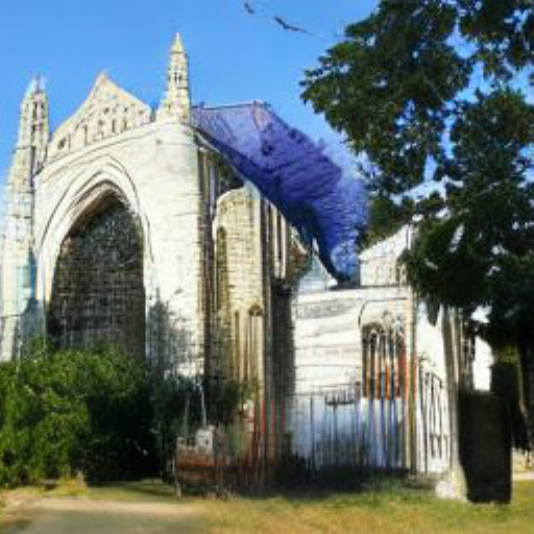} &
         \includegraphics[width=0.16\linewidth]{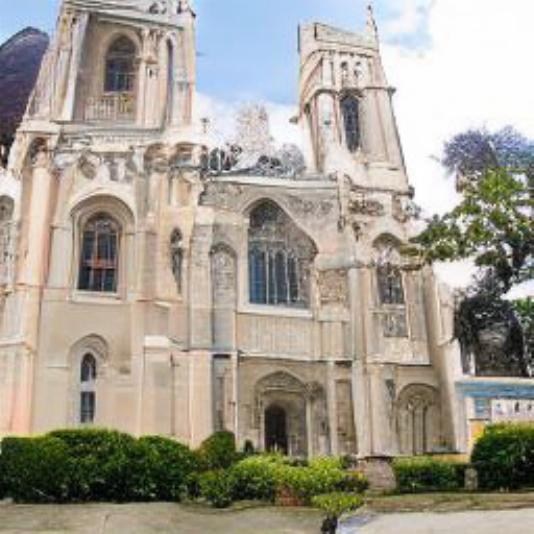} &
         \includegraphics[width=0.16\linewidth]{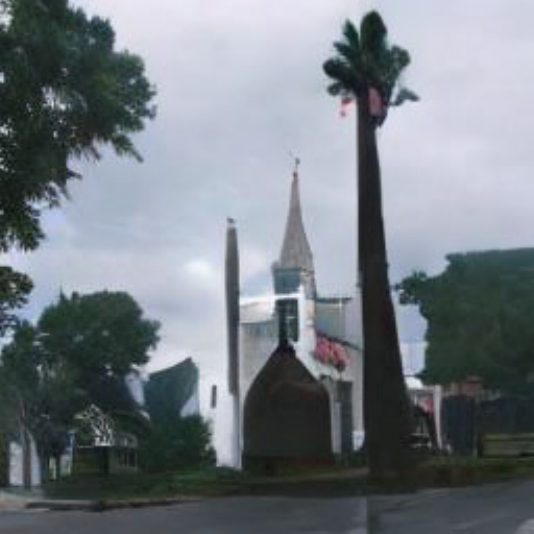} \\

      \end{tabular}
   \end{subfigure}
   \vspace{3mm}

   \begin{subfigure}[c]{1.0\linewidth}
      \centering
      \begin{tabular}{c @{} c @{} c @{} c @{} c @{} c}
         \includegraphics[width=0.16\linewidth]{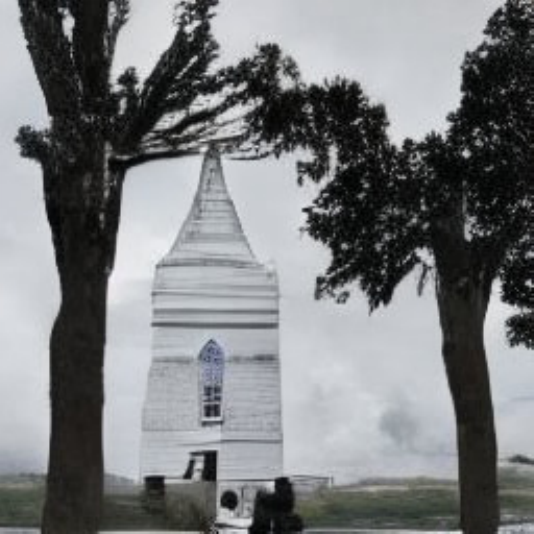} &
         \includegraphics[width=0.16\linewidth]{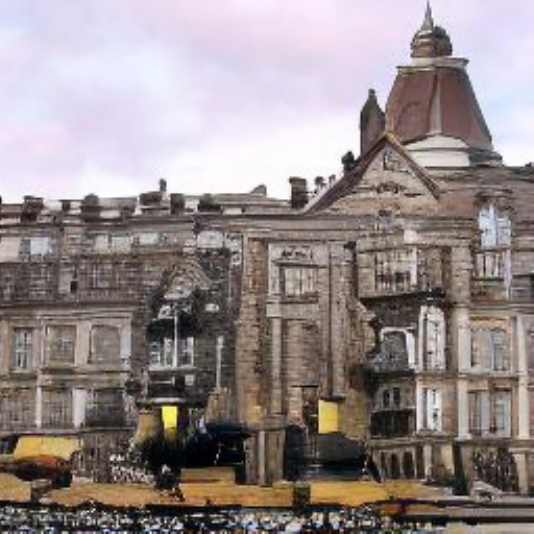} &
         \includegraphics[width=0.16\linewidth]{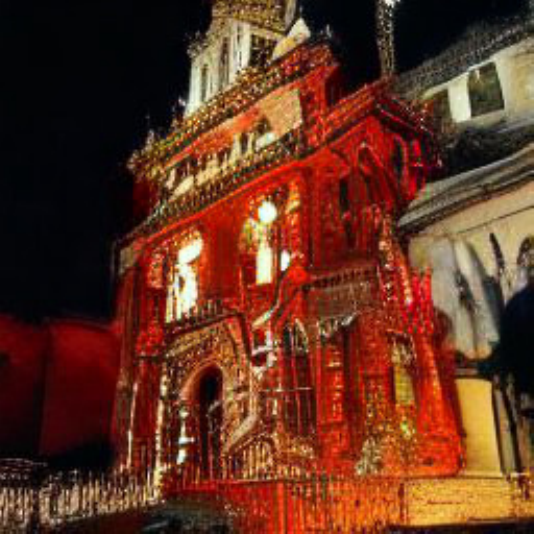} &
         \includegraphics[width=0.16\linewidth]{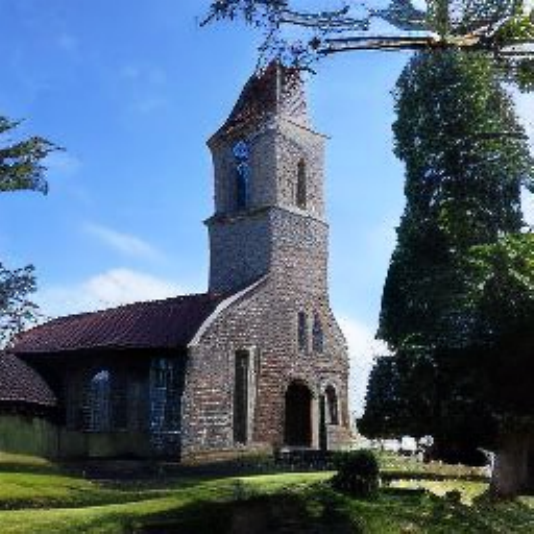} &
         \includegraphics[width=0.16\linewidth]{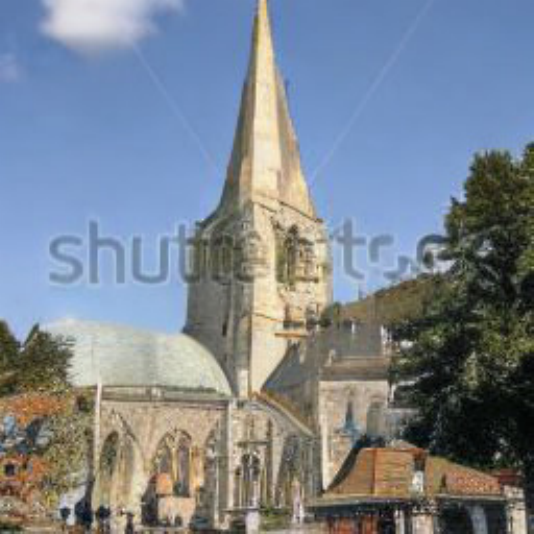} &
         \includegraphics[width=0.16\linewidth]{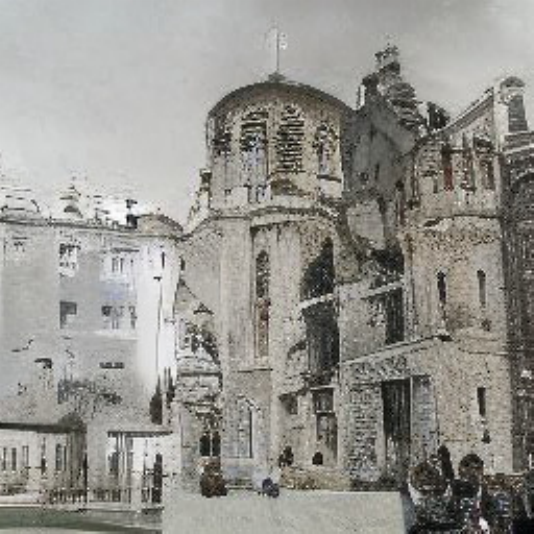} \\

         \includegraphics[width=0.16\linewidth]{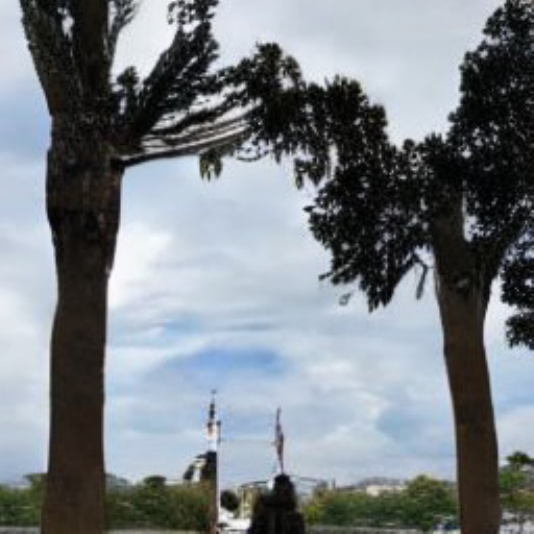} &
         \includegraphics[width=0.16\linewidth]{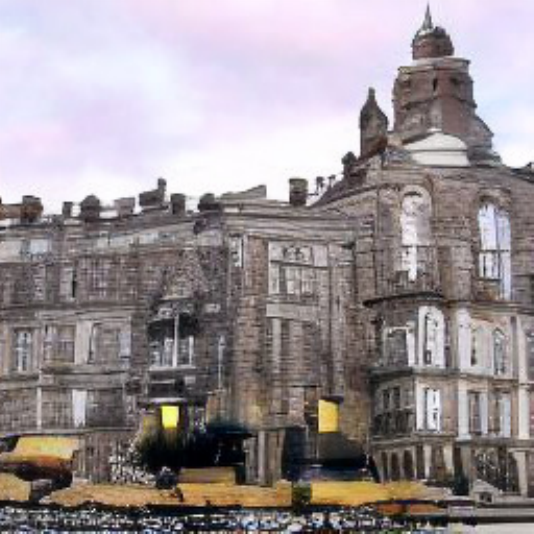} &
         \includegraphics[width=0.16\linewidth]{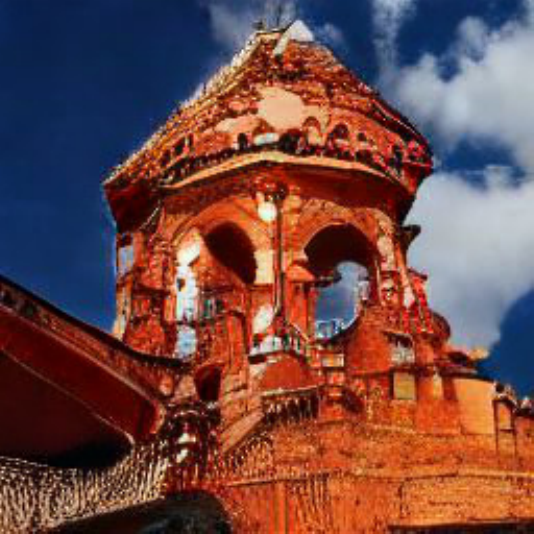} &
         \includegraphics[width=0.16\linewidth]{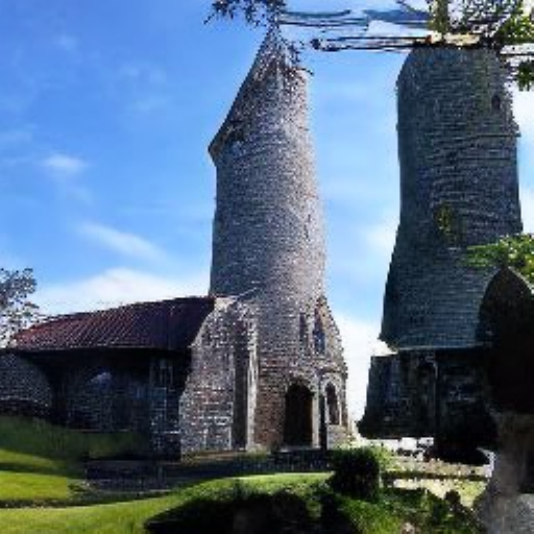} &
         \includegraphics[width=0.16\linewidth]{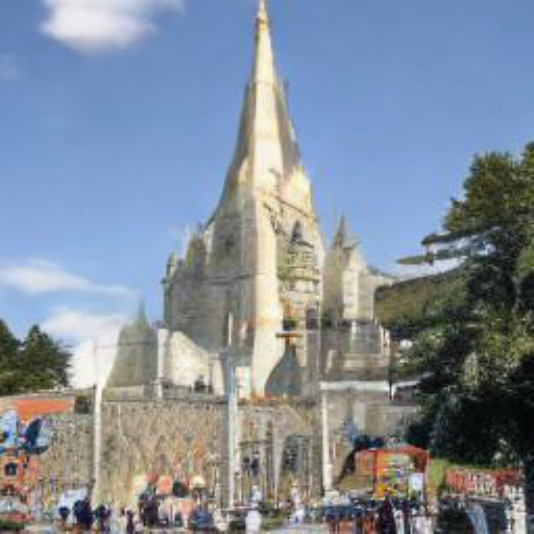} &
         \includegraphics[width=0.16\linewidth]{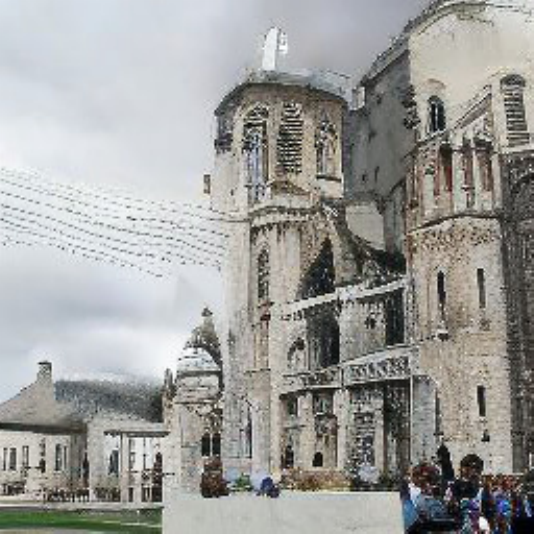} \\

         \includegraphics[width=0.16\linewidth]{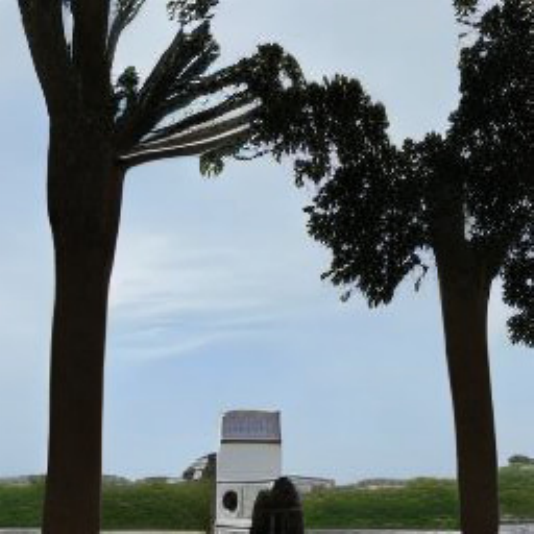} &
         \includegraphics[width=0.16\linewidth]{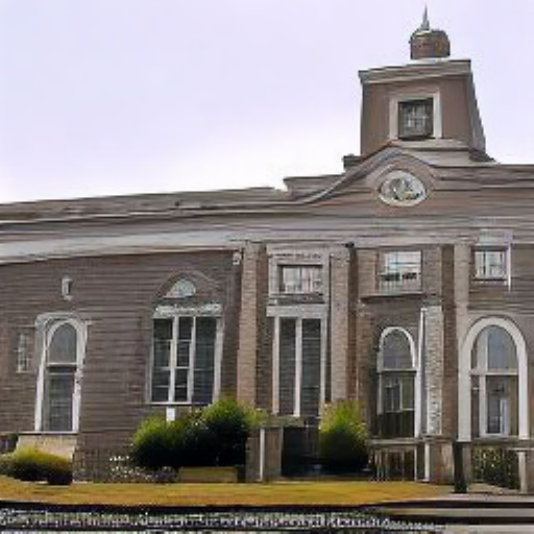} &
         \includegraphics[width=0.16\linewidth]{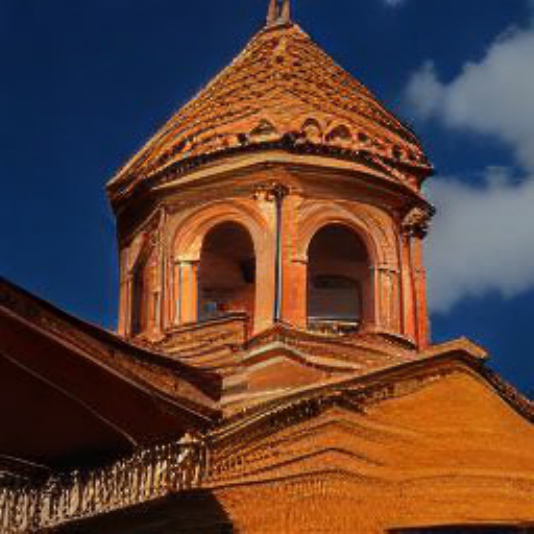} &
         \includegraphics[width=0.16\linewidth]{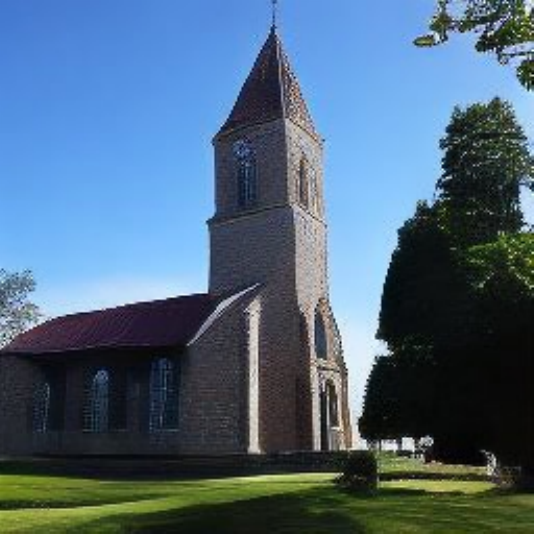} &
         \includegraphics[width=0.16\linewidth]{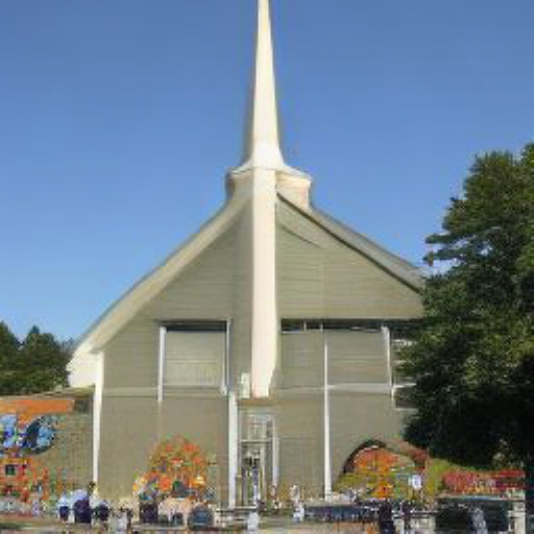} &
         \includegraphics[width=0.16\linewidth]{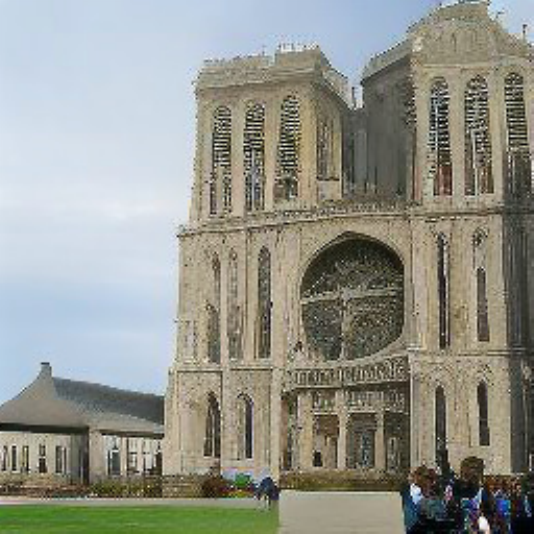} \\

         \includegraphics[width=0.16\linewidth]{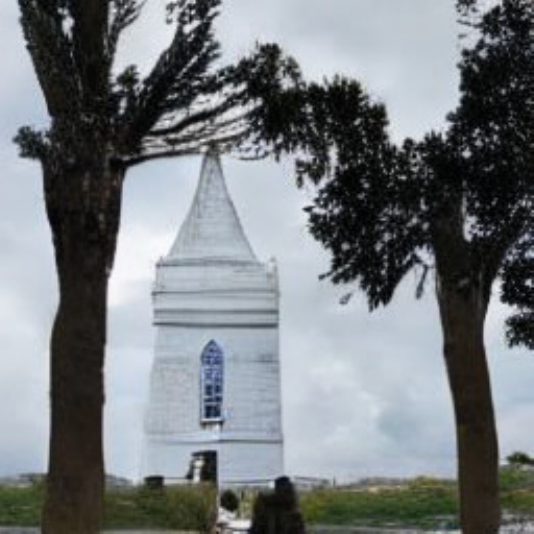} &
         \includegraphics[width=0.16\linewidth]{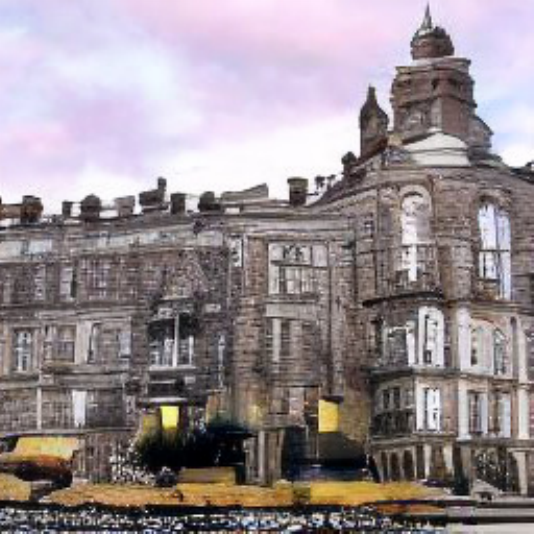} &
         \includegraphics[width=0.16\linewidth]{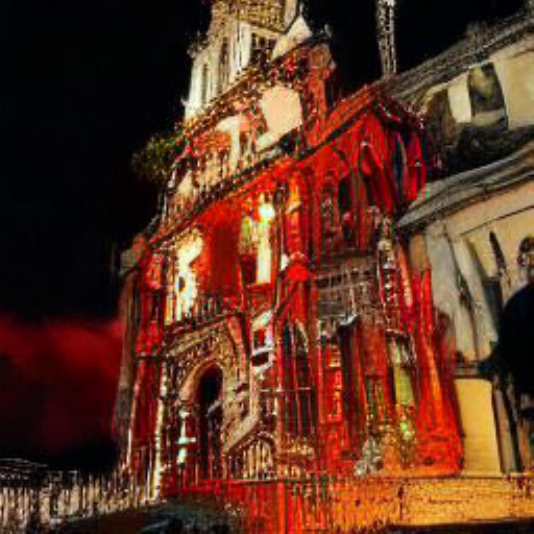} &
         \includegraphics[width=0.16\linewidth]{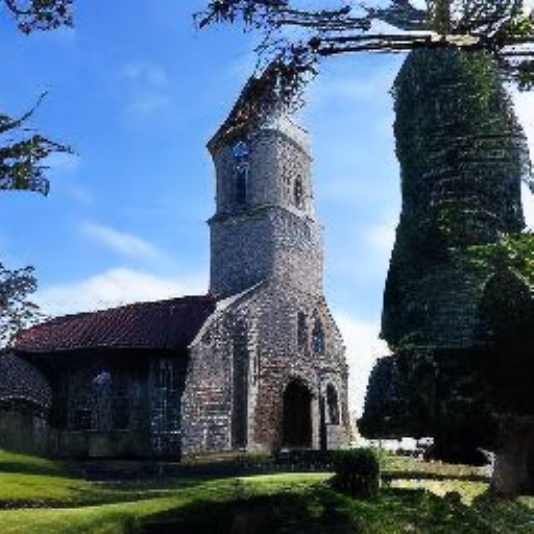} &
         \includegraphics[width=0.16\linewidth]{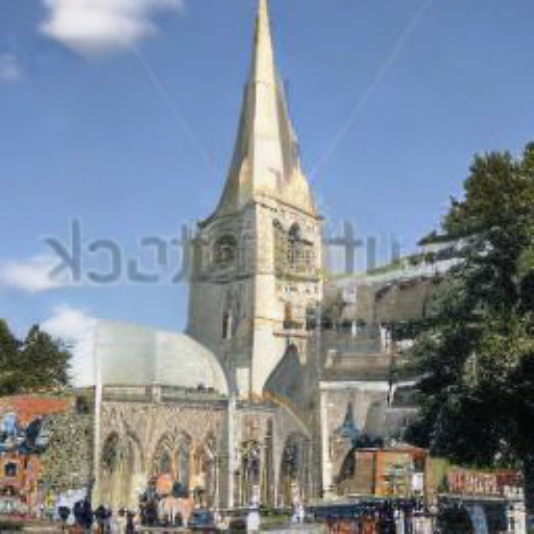} &
         \includegraphics[width=0.16\linewidth]{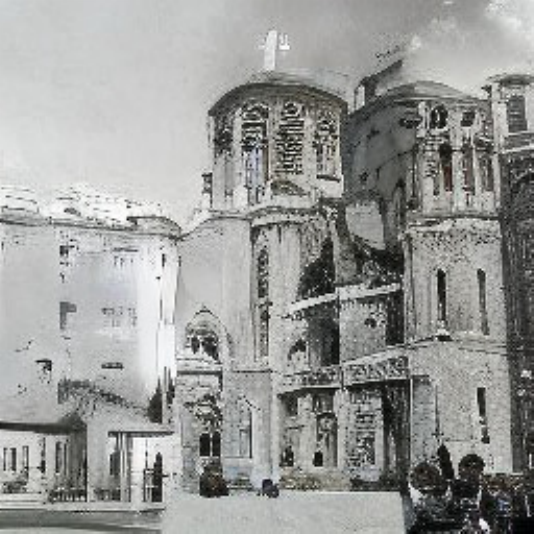} \\

      \end{tabular}
   \end{subfigure}

   \caption{Visual comparisons of generated images on LSUN-Church~\cite{yu2015lsun} (256$\times$256) for unconditional image generation with LDM-8~\cite{rombach2022high} under a 3/8-bit setting. Each row corresponds to Full Precision, Q-Diffusion~\cite{li2023q}, TFMQ-DM~\cite{huang2024tfmq}, and Ours.}
   \label{fig:quali_ldm_church_suppl}
\end{figure*}

\begin{figure*}[!t]
  \vspace{0.1mm}
  \captionsetup[subfigure]{font=small, labelformat=empty, justification=centering}
  \begin{center}
    \begin{subfigure}[c]{\linewidth}
      \centering
      \begin{tabular}{c@{}c@{}c@{}c@{}c@{}c@{}c@{}c}
        \includegraphics[width=0.12\columnwidth]{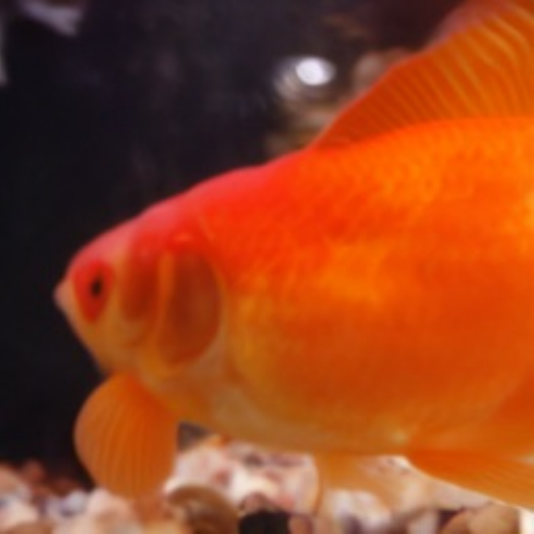}&
        \includegraphics[width=0.12\columnwidth]{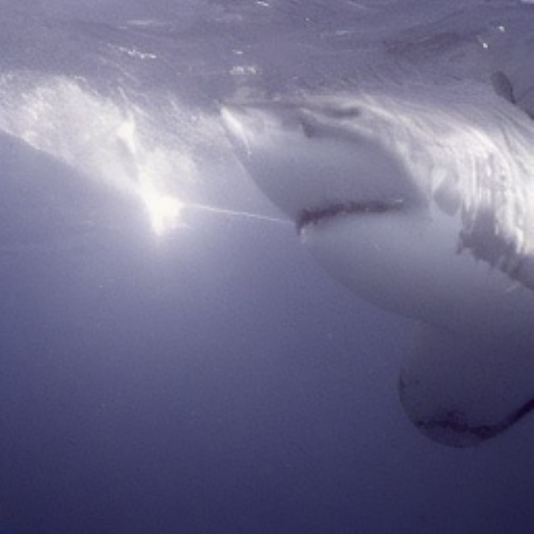}&
        \includegraphics[width=0.12\columnwidth]{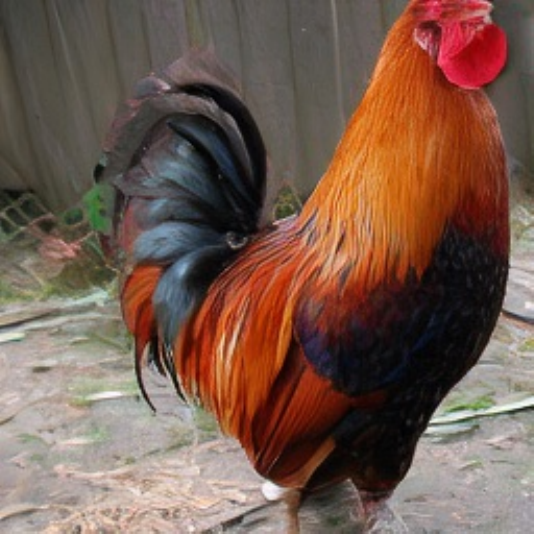}&
        \includegraphics[width=0.12\columnwidth]{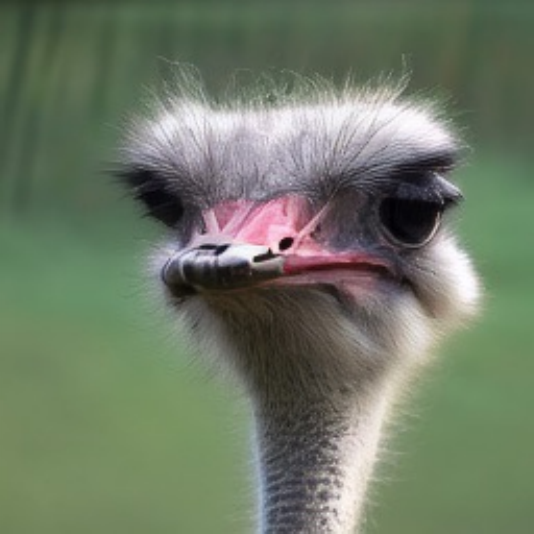}&
        \includegraphics[width=0.12\columnwidth]{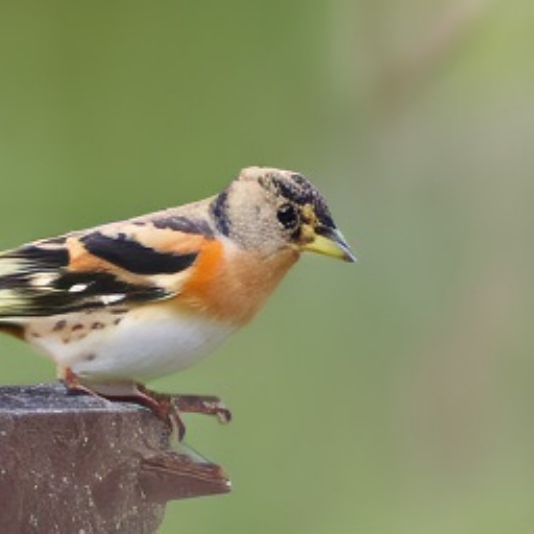}&
        \includegraphics[width=0.12\columnwidth]{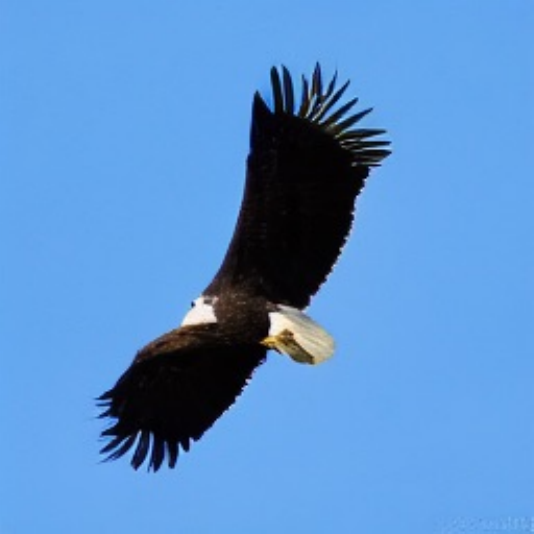}&
        \includegraphics[width=0.12\columnwidth]{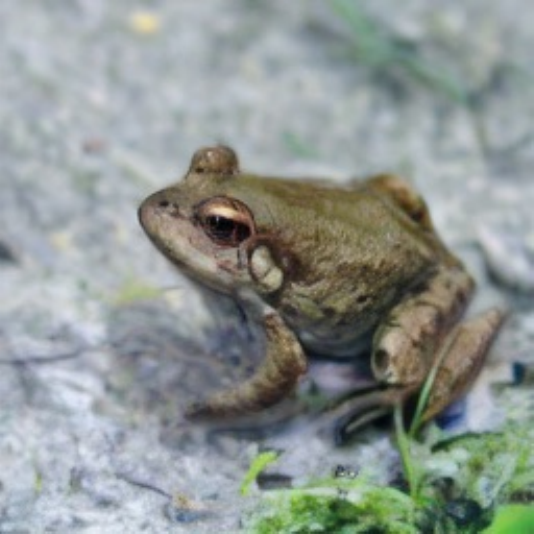}&
        \includegraphics[width=0.12\columnwidth]{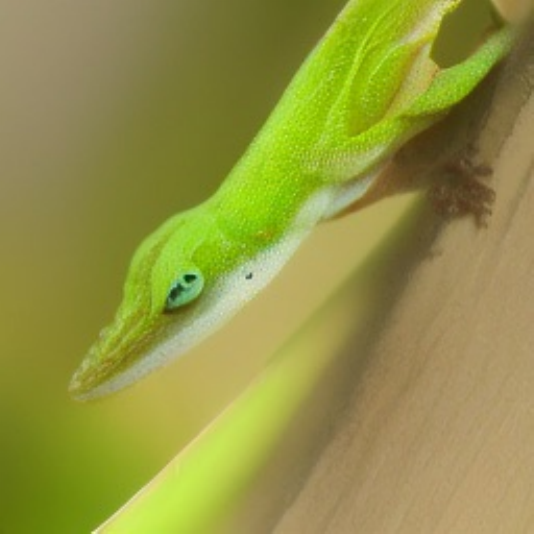}\\[-1pt]

        \includegraphics[width=0.12\columnwidth]{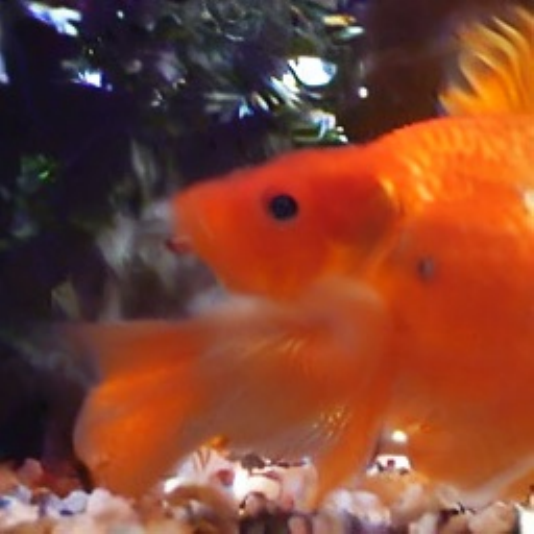}&
        \includegraphics[width=0.12\columnwidth]{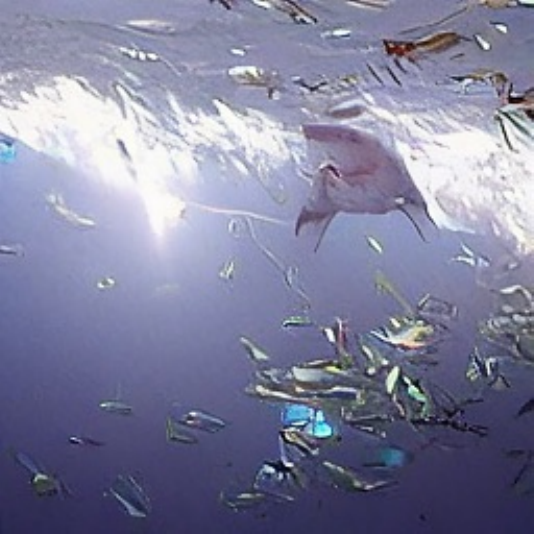}&
        \includegraphics[width=0.12\columnwidth]{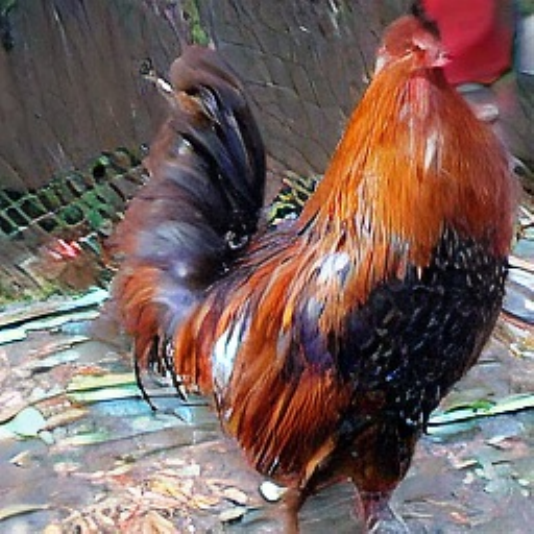}&
        \includegraphics[width=0.12\columnwidth]{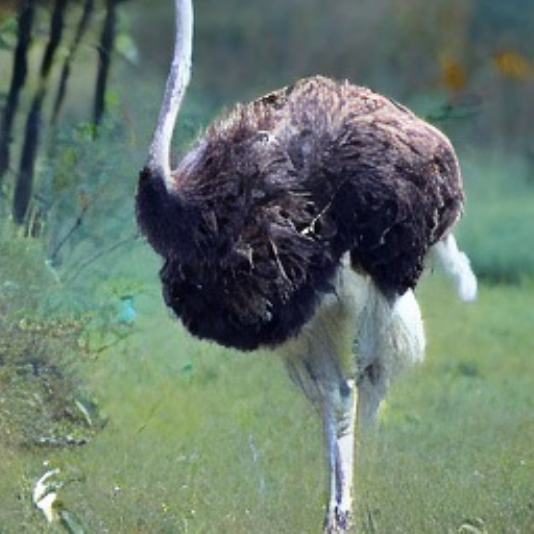}&
        \includegraphics[width=0.12\columnwidth]{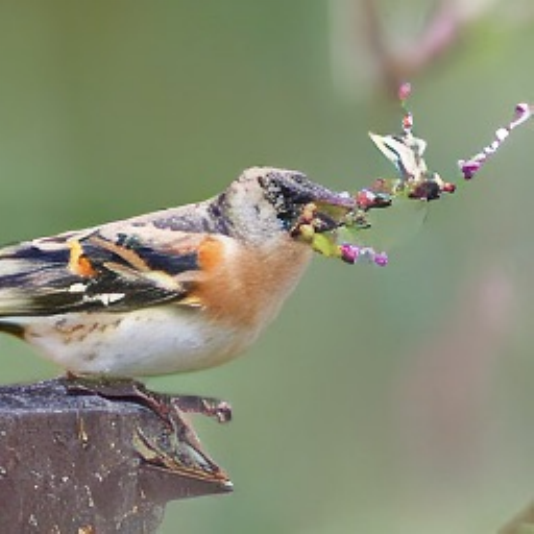}&
        \includegraphics[width=0.12\columnwidth]{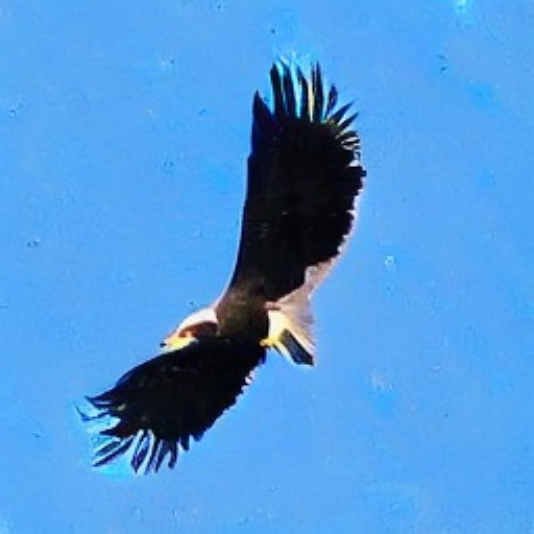}&
        \includegraphics[width=0.12\columnwidth]{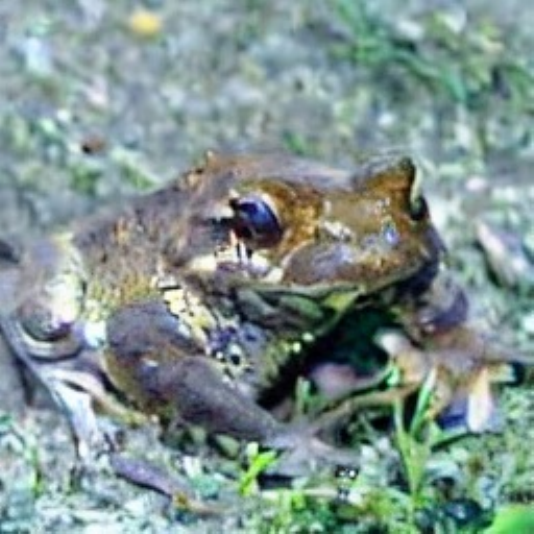}&
        \includegraphics[width=0.12\columnwidth]{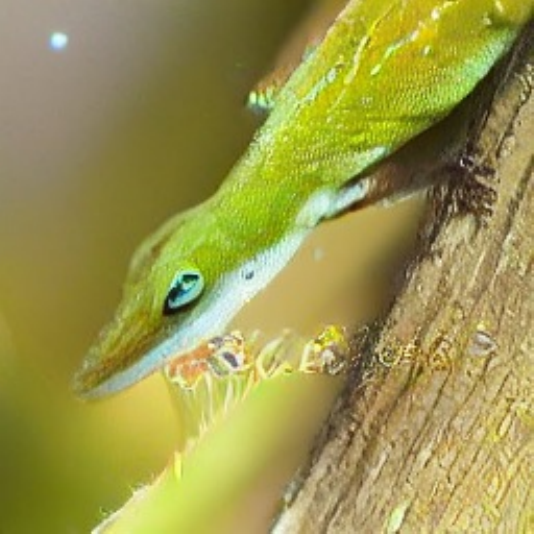}\\[-1pt]

        \includegraphics[width=0.12\columnwidth]{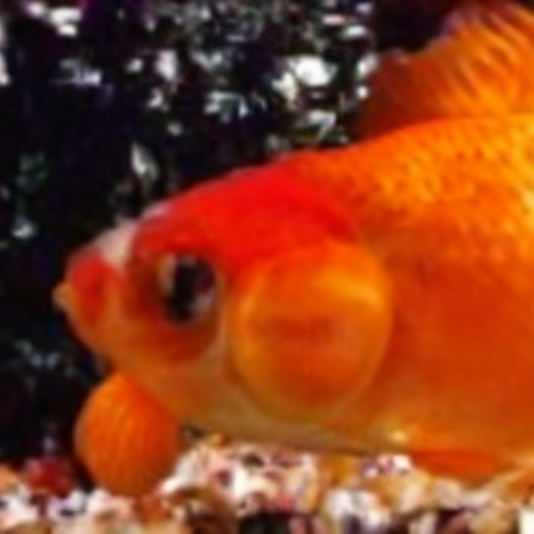}&
        \includegraphics[width=0.12\columnwidth]{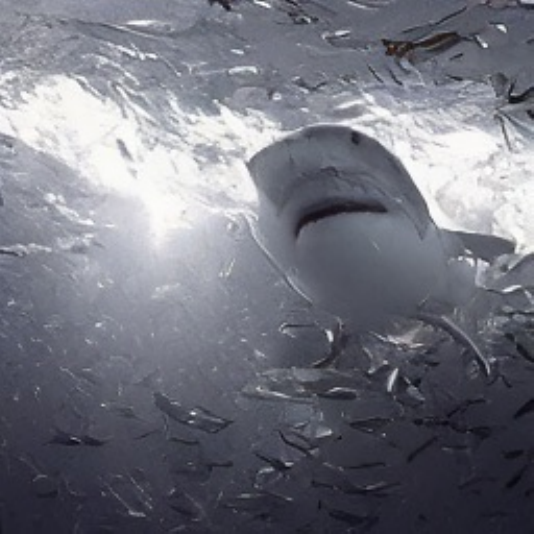}&
        \includegraphics[width=0.12\columnwidth]{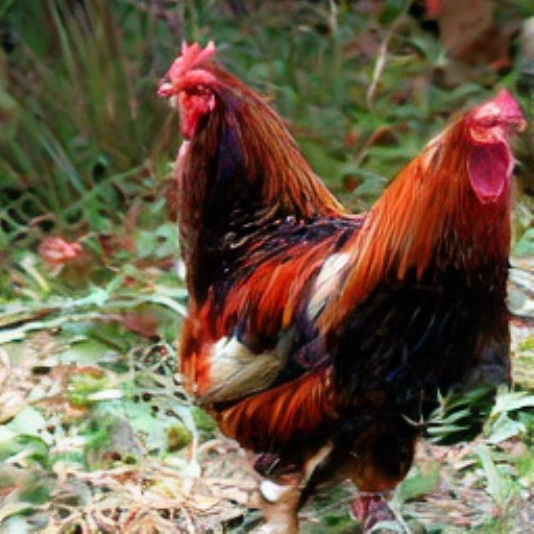}&
        \includegraphics[width=0.12\columnwidth]{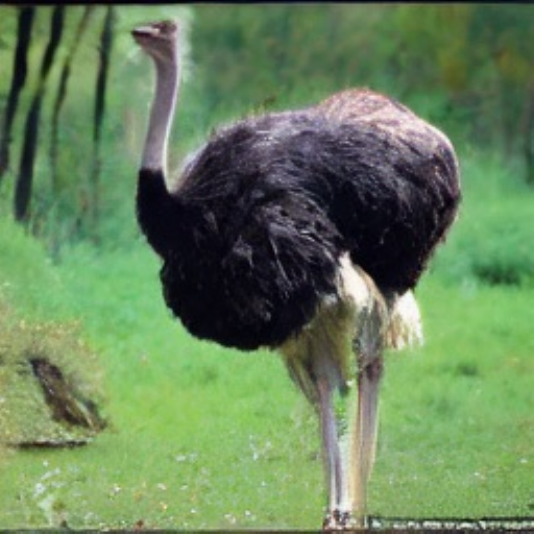}&
        \includegraphics[width=0.12\columnwidth]{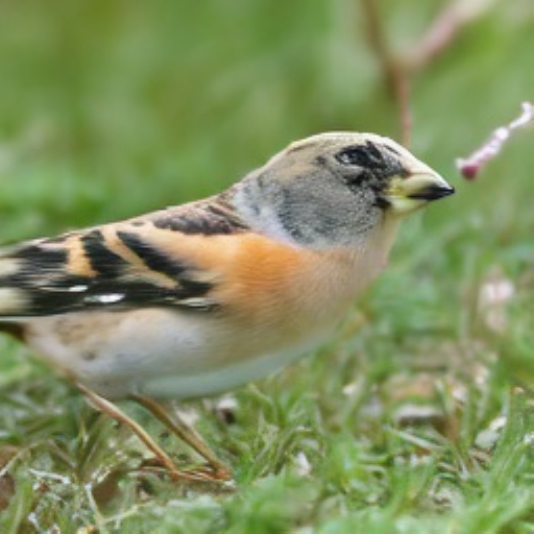}&
        \includegraphics[width=0.12\columnwidth]{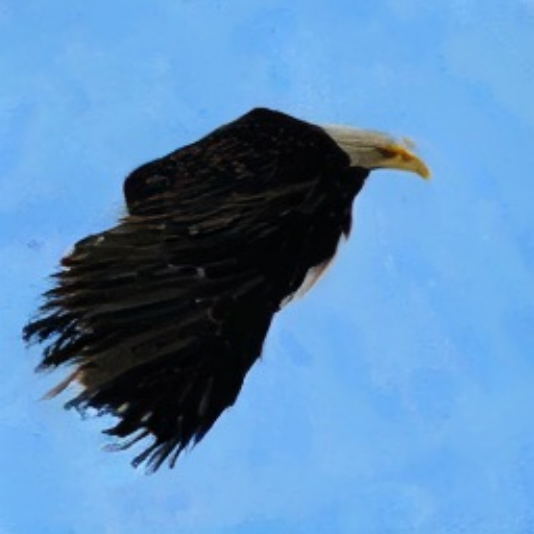}&
        \includegraphics[width=0.12\columnwidth]{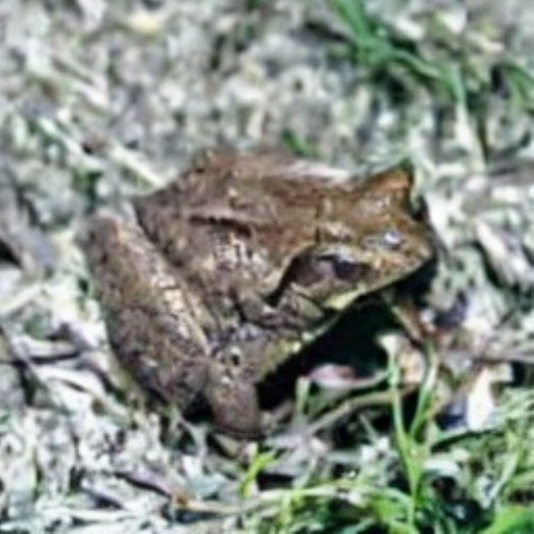}&
        \includegraphics[width=0.12\columnwidth]{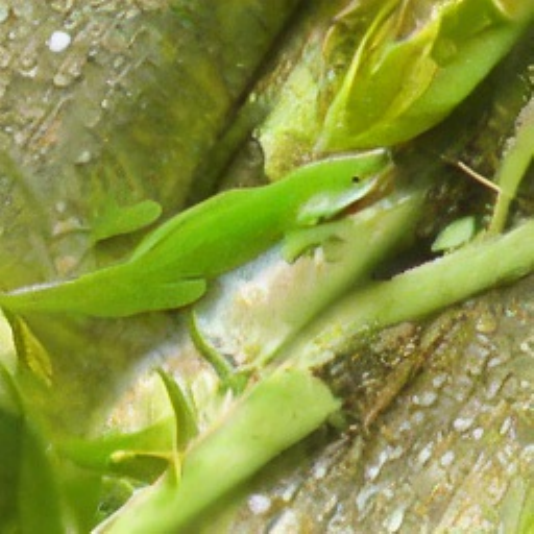}\\[-1pt]

        \includegraphics[width=0.12\columnwidth]{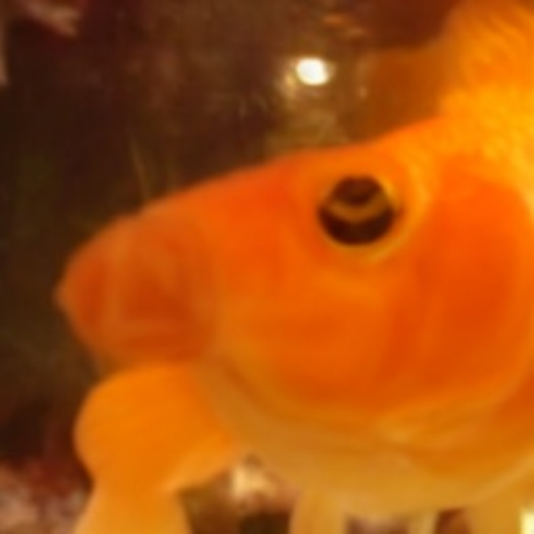}&
        \includegraphics[width=0.12\columnwidth]{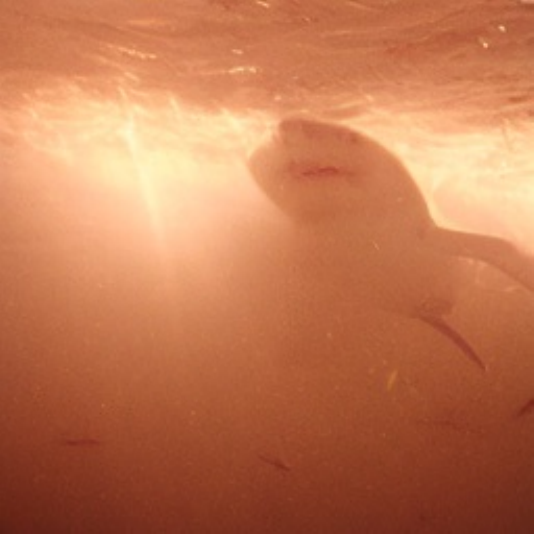}&
        \includegraphics[width=0.12\columnwidth]{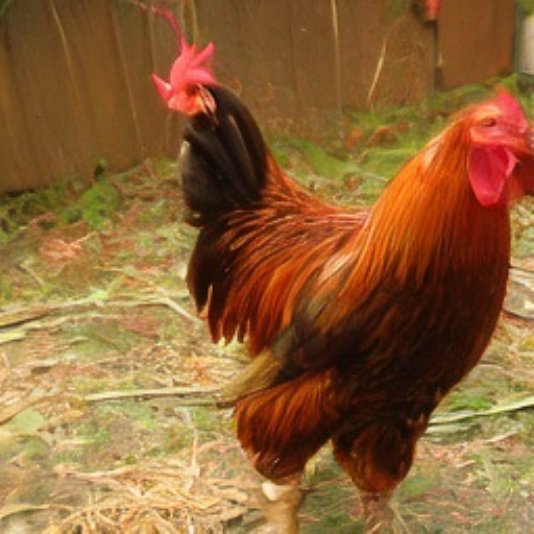}&
        \includegraphics[width=0.12\columnwidth]{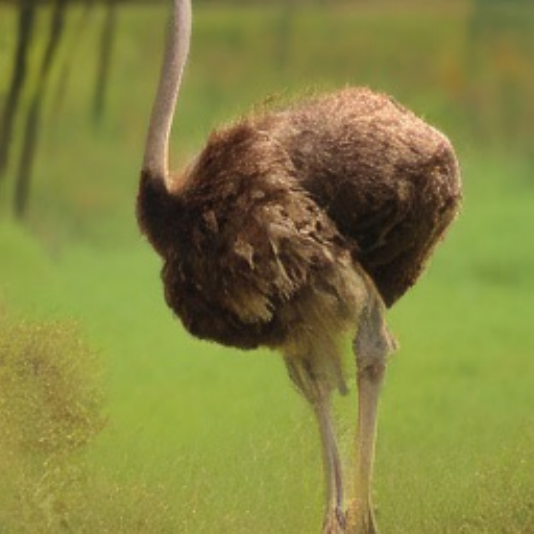}&
        \includegraphics[width=0.12\columnwidth]{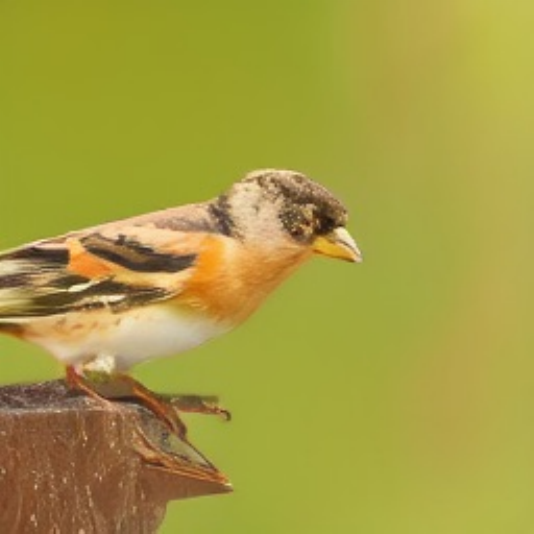}&
        \includegraphics[width=0.12\columnwidth]{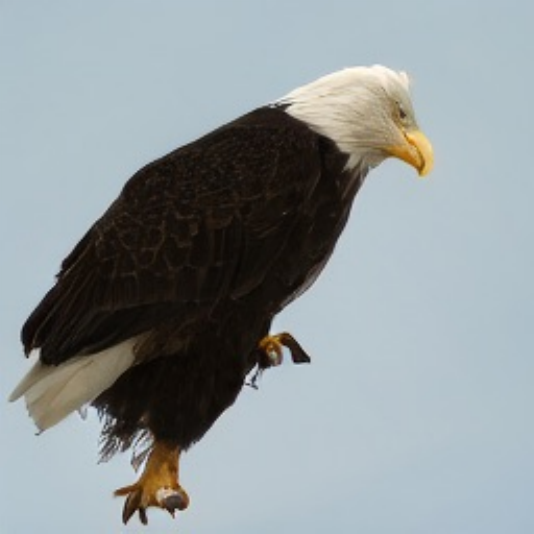}&
        \includegraphics[width=0.12\columnwidth]{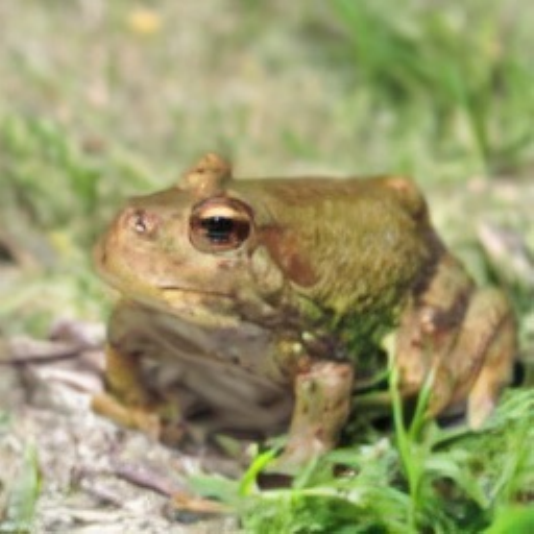}&
        \includegraphics[width=0.12\columnwidth]{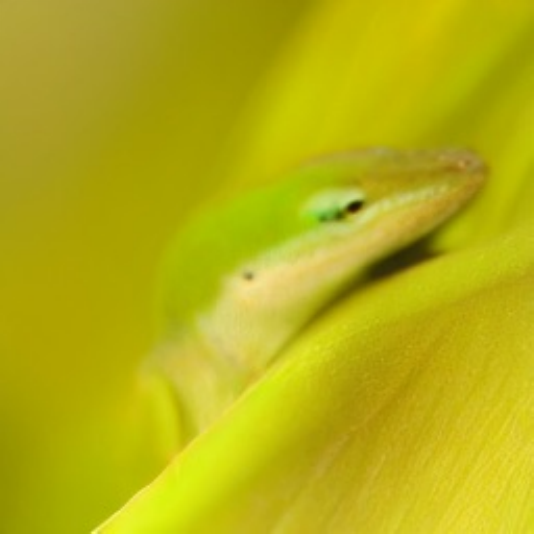}\\[-1pt]

        \includegraphics[width=0.12\columnwidth]{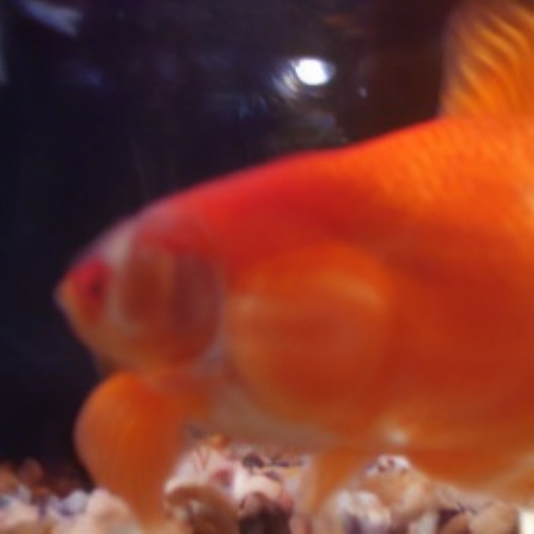}&
        \includegraphics[width=0.12\columnwidth]{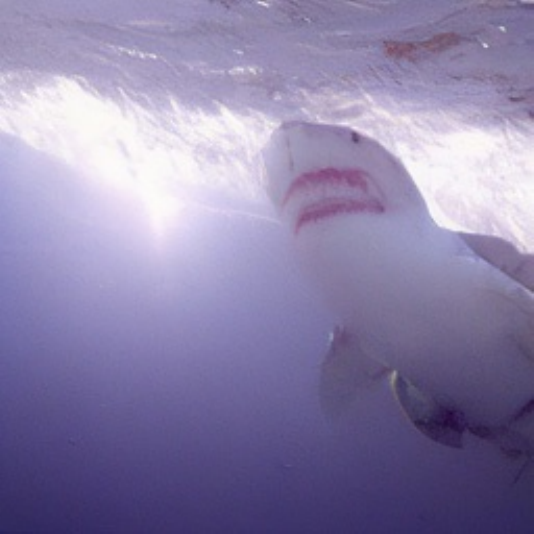}&
        \includegraphics[width=0.12\columnwidth]{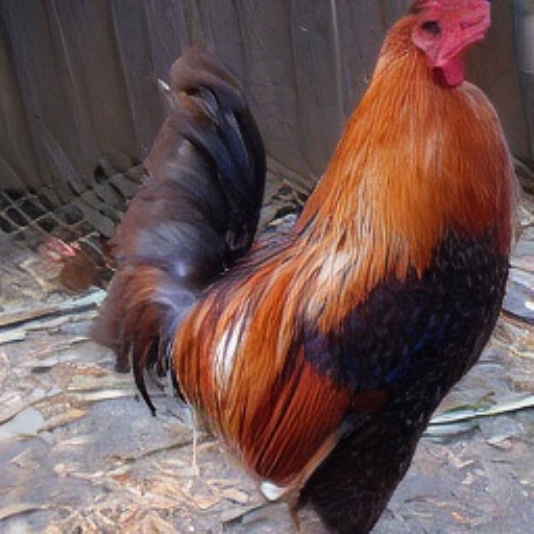}&
        \includegraphics[width=0.12\columnwidth]{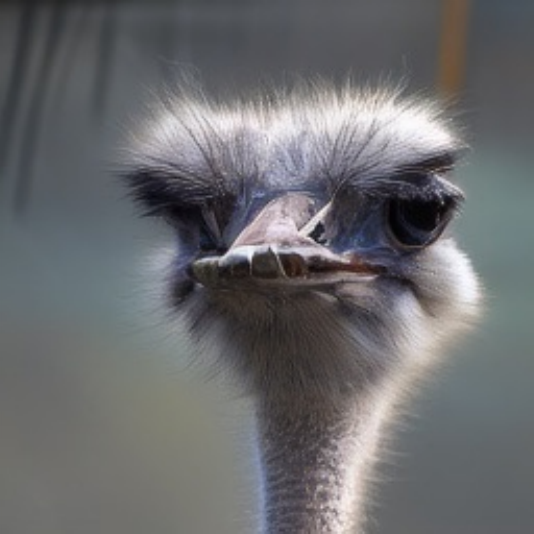}&
        \includegraphics[width=0.12\columnwidth]{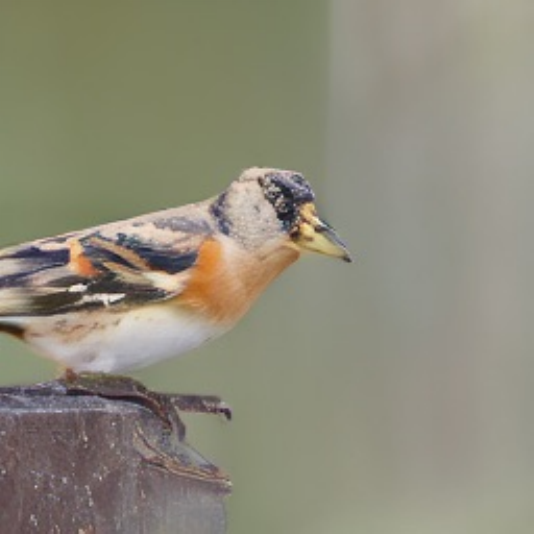}&
        \includegraphics[width=0.12\columnwidth]{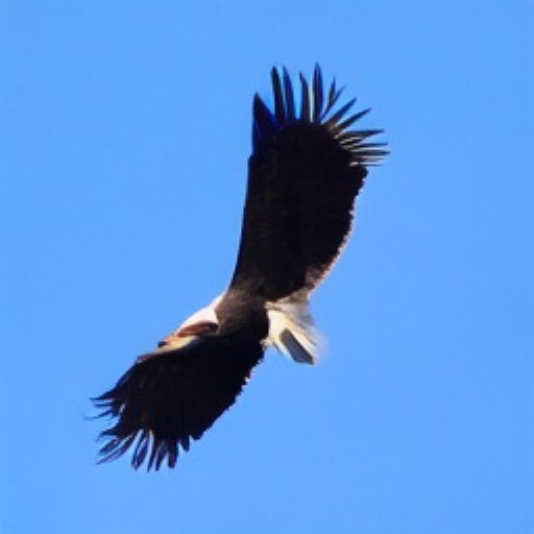}&
        \includegraphics[width=0.12\columnwidth]{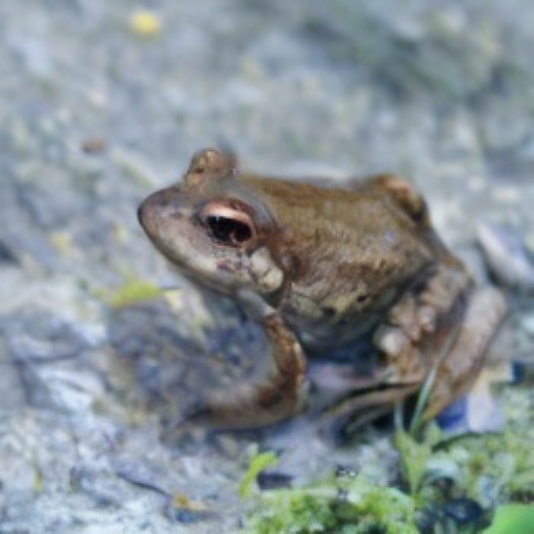}&
        \includegraphics[width=0.12\columnwidth]{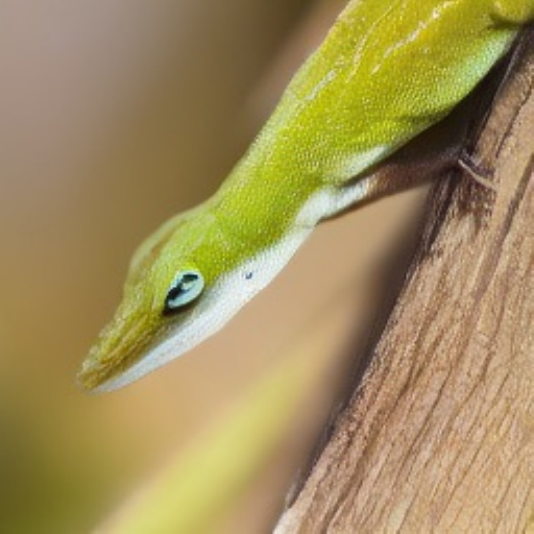}\\[-1pt]

      \end{tabular}
      \label{fig:imagenet_fp}
    \end{subfigure}
    \vspace{3mm}

    \begin{subfigure}[c]{\linewidth}
      \centering
      \begin{tabular}{c@{}c@{}c@{}c@{}c@{}c@{}c@{}c}

        \includegraphics[width=0.12\columnwidth]{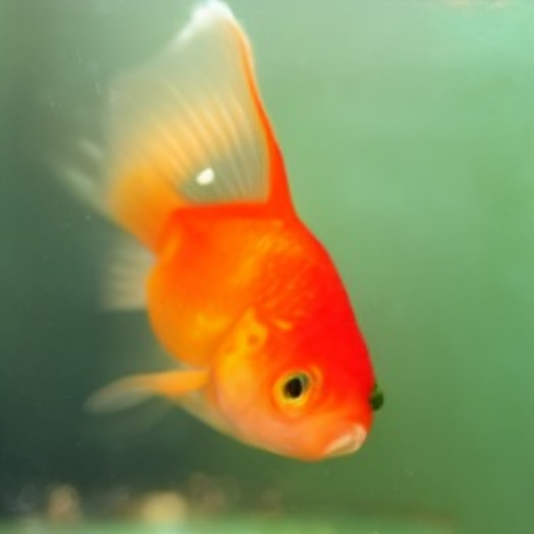}&
        \includegraphics[width=0.12\columnwidth]{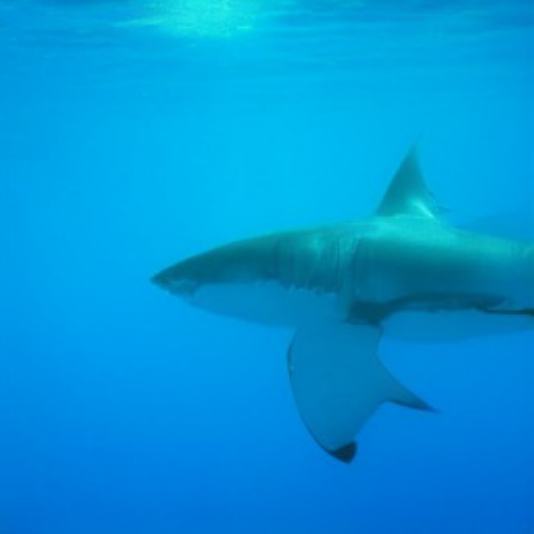}&
        \includegraphics[width=0.12\columnwidth]{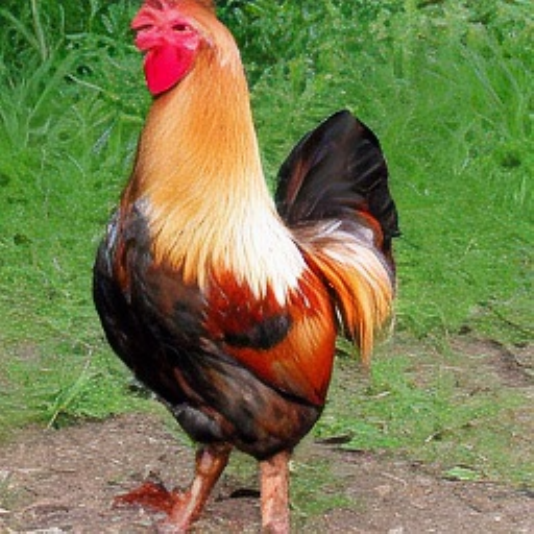}&
        \includegraphics[width=0.12\columnwidth]{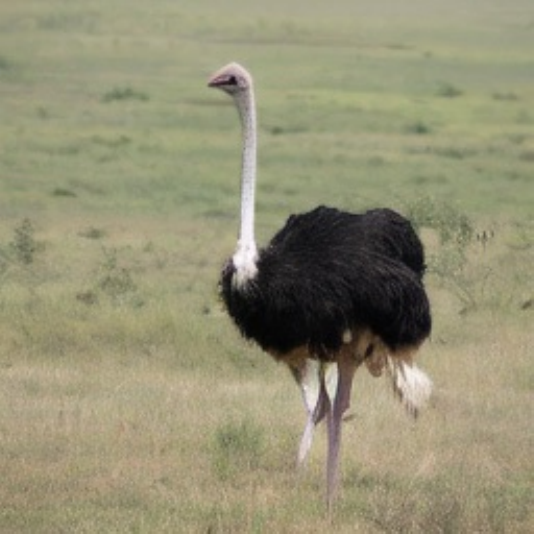}&
        \includegraphics[width=0.12\columnwidth]{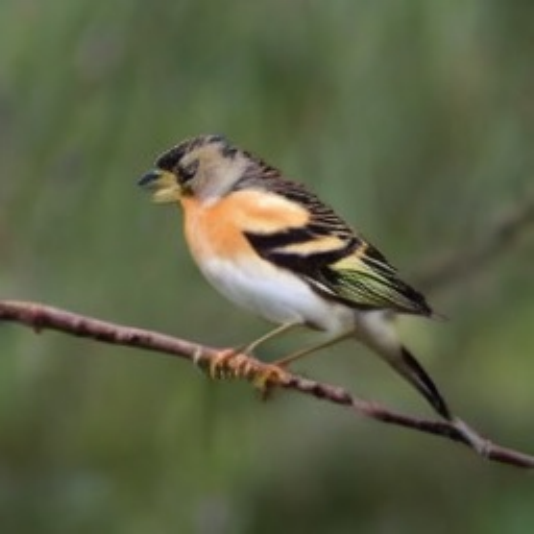}&
        \includegraphics[width=0.12\columnwidth]{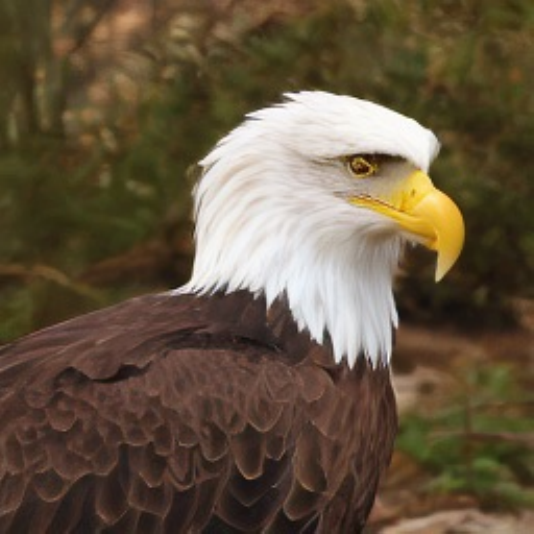}&
        \includegraphics[width=0.12\columnwidth]{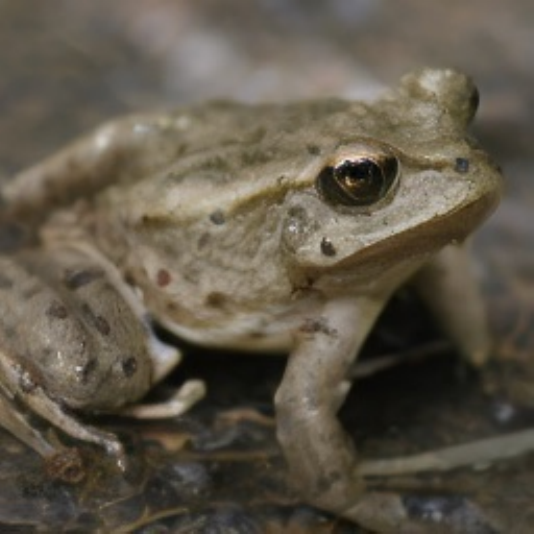}&
        \includegraphics[width=0.12\columnwidth]{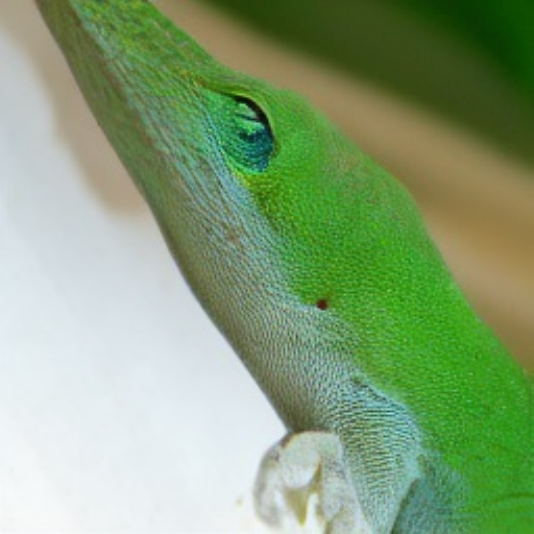}\\

        \includegraphics[width=0.12\columnwidth]{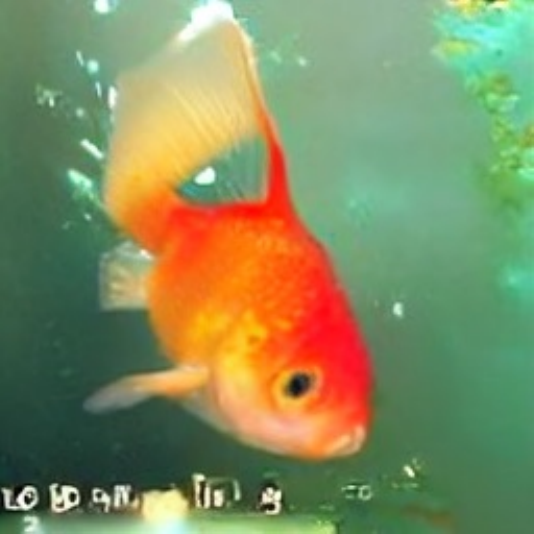}&
        \includegraphics[width=0.12\columnwidth]{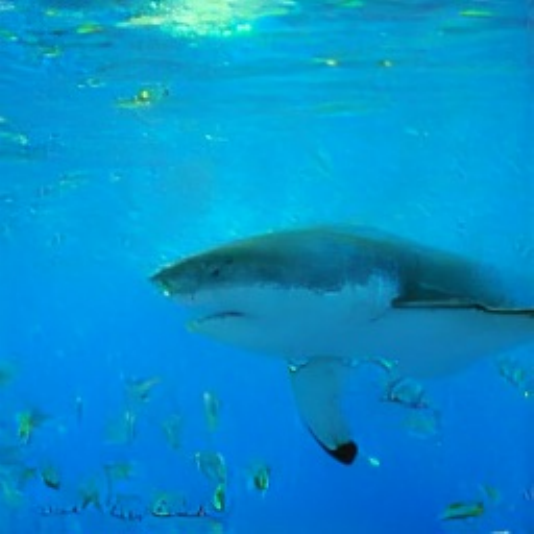}&
        \includegraphics[width=0.12\columnwidth]{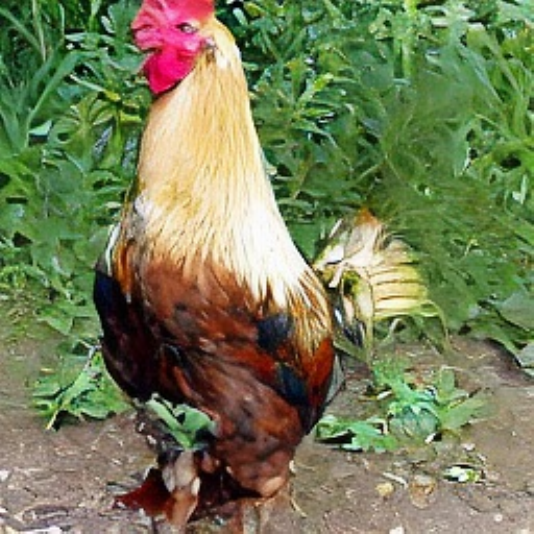}&
        \includegraphics[width=0.12\columnwidth]{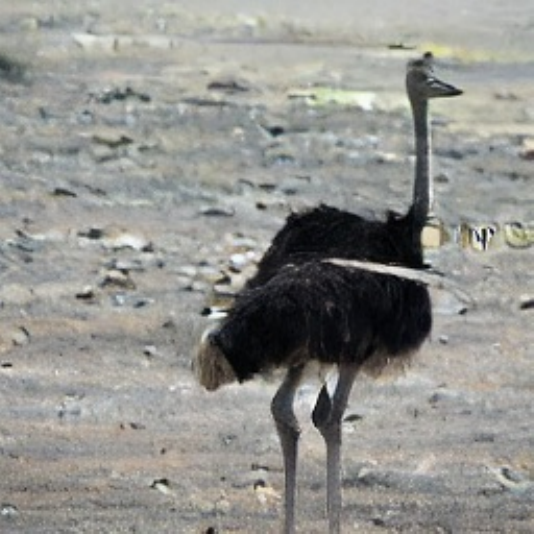}&
        \includegraphics[width=0.12\columnwidth]{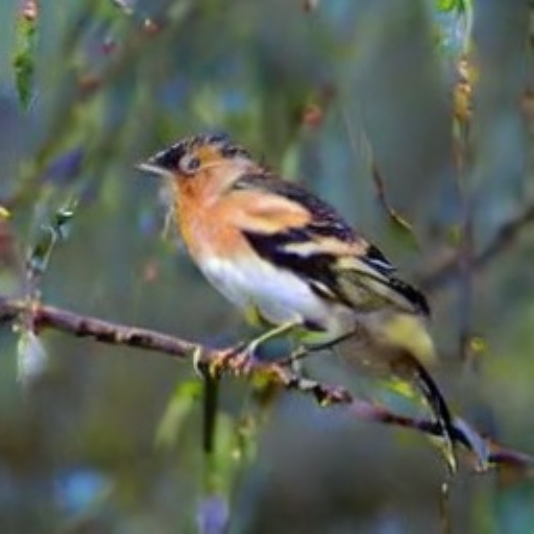}&
        \includegraphics[width=0.12\columnwidth]{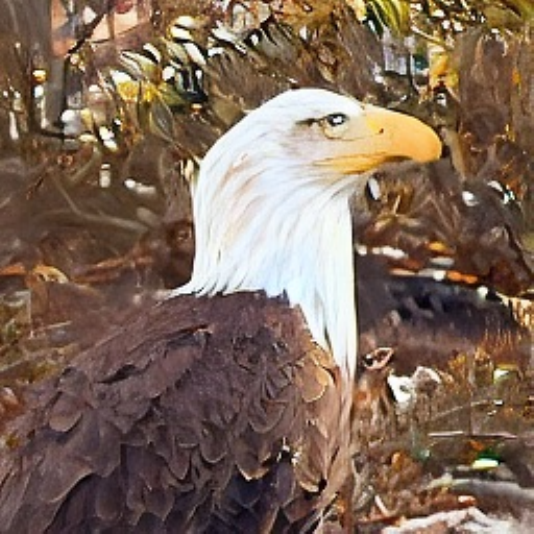}&
        \includegraphics[width=0.12\columnwidth]{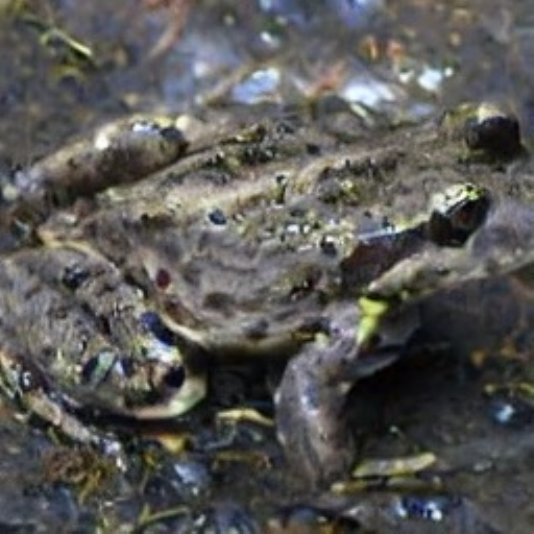}&
        \includegraphics[width=0.12\columnwidth]{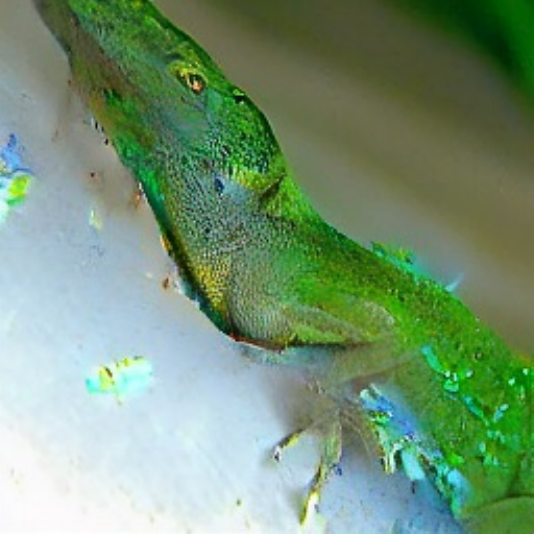}\\

        \includegraphics[width=0.12\columnwidth]{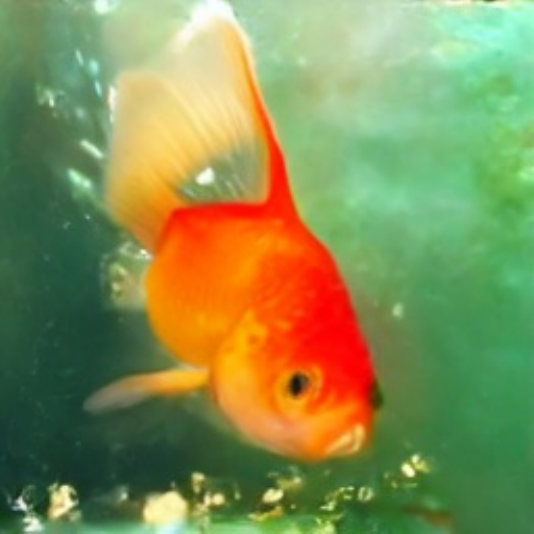}&
        \includegraphics[width=0.12\columnwidth]{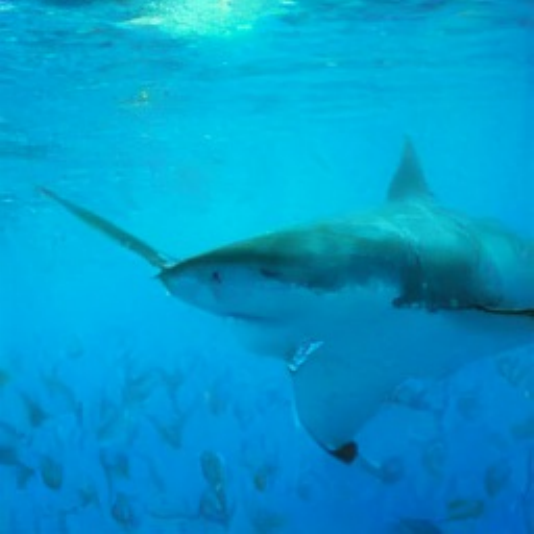}&
        \includegraphics[width=0.12\columnwidth]{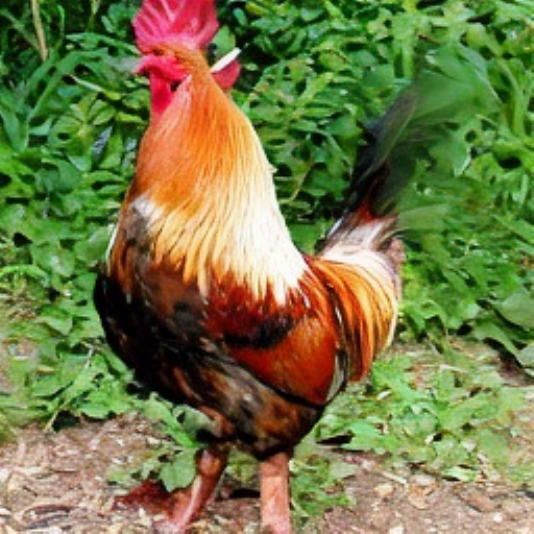}&
        \includegraphics[width=0.12\columnwidth]{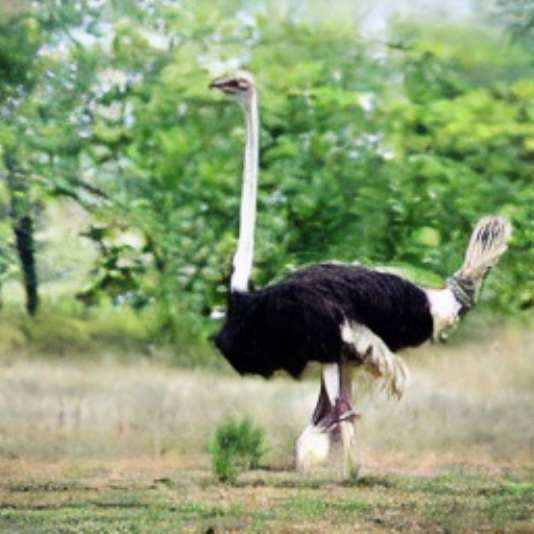}&
        \includegraphics[width=0.12\columnwidth]{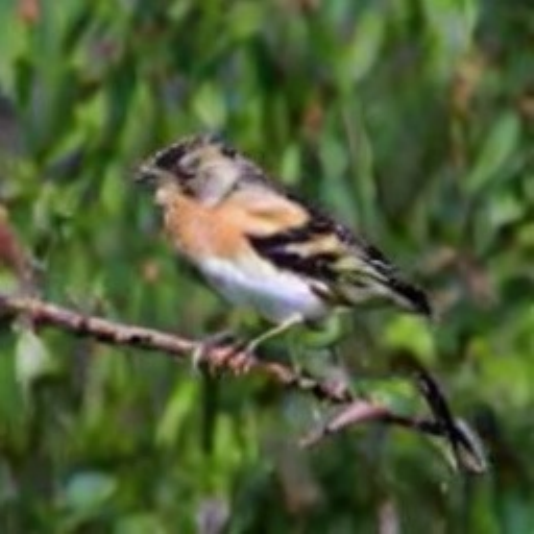}&
        \includegraphics[width=0.12\columnwidth]{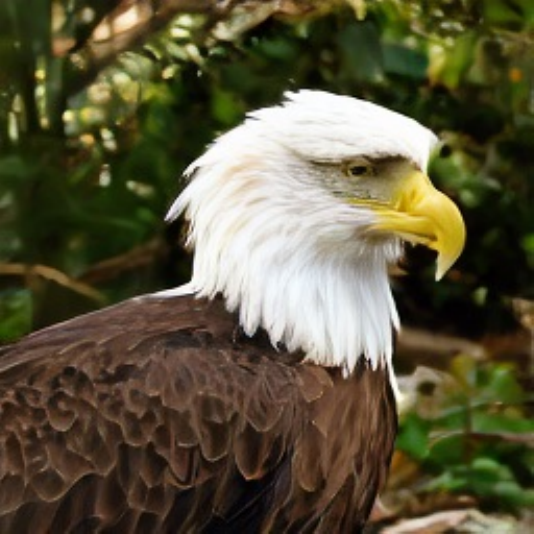}&
        \includegraphics[width=0.12\columnwidth]{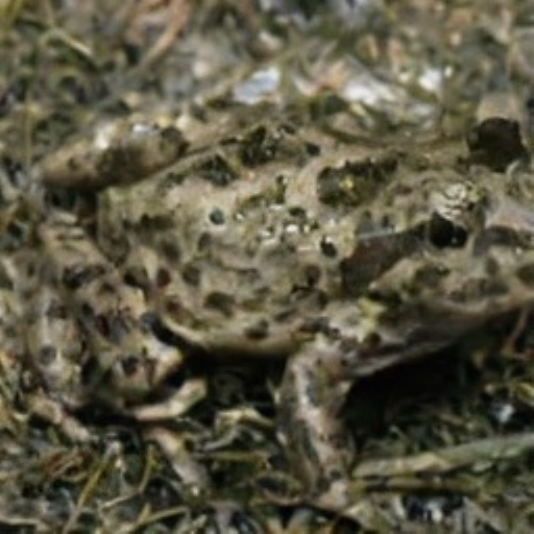}&
        \includegraphics[width=0.12\columnwidth]{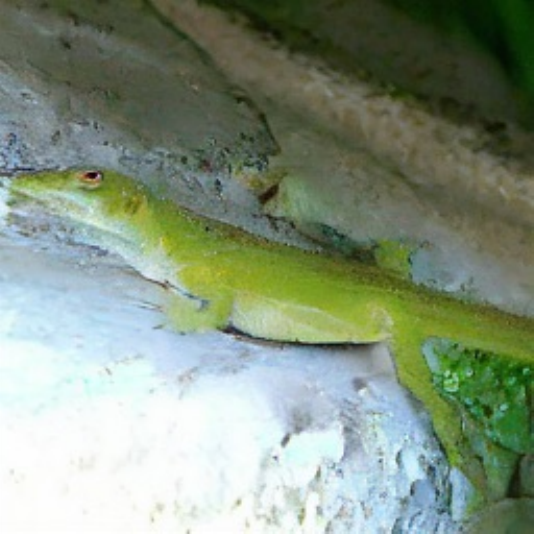}\\

        \includegraphics[width=0.12\columnwidth]{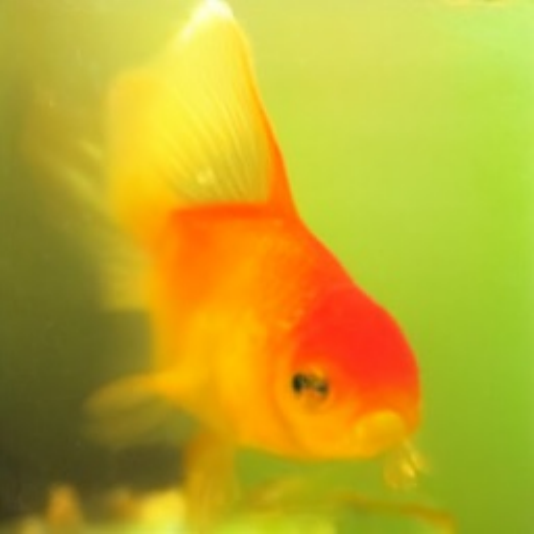}&
        \includegraphics[width=0.12\columnwidth]{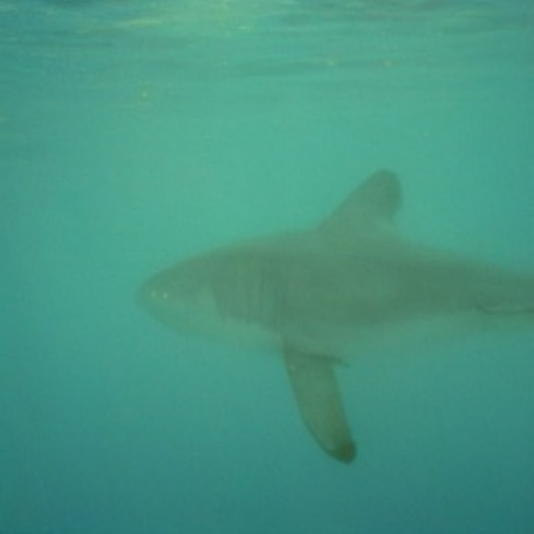}&
        \includegraphics[width=0.12\columnwidth]{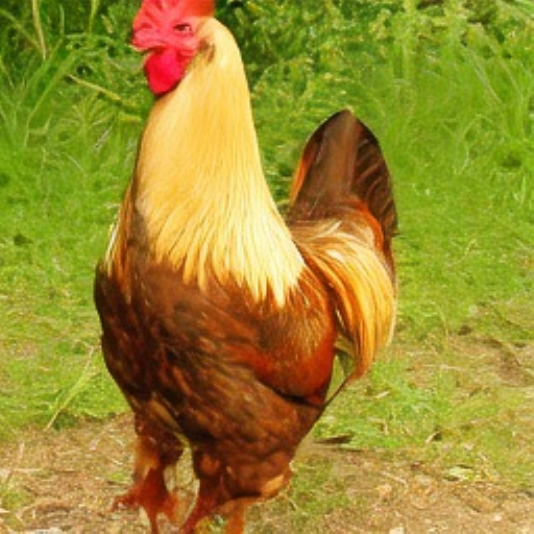}&
        \includegraphics[width=0.12\columnwidth]{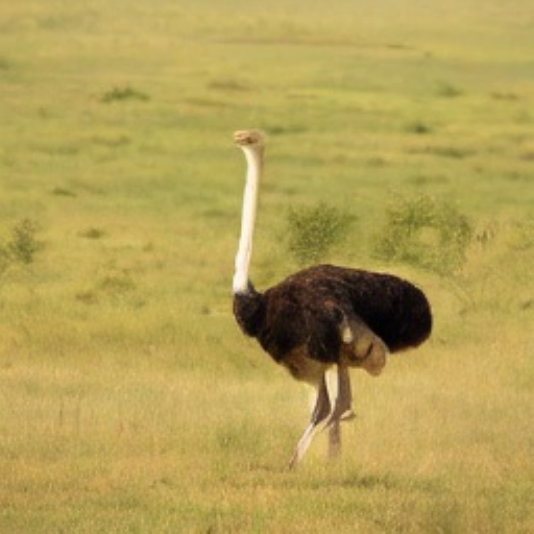}&
        \includegraphics[width=0.12\columnwidth]{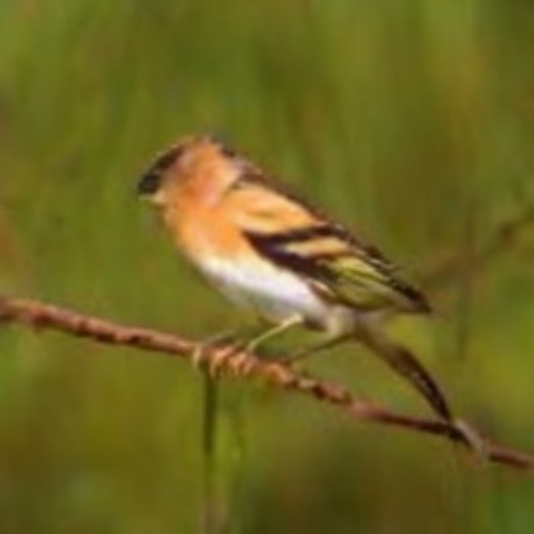}&
        \includegraphics[width=0.12\columnwidth]{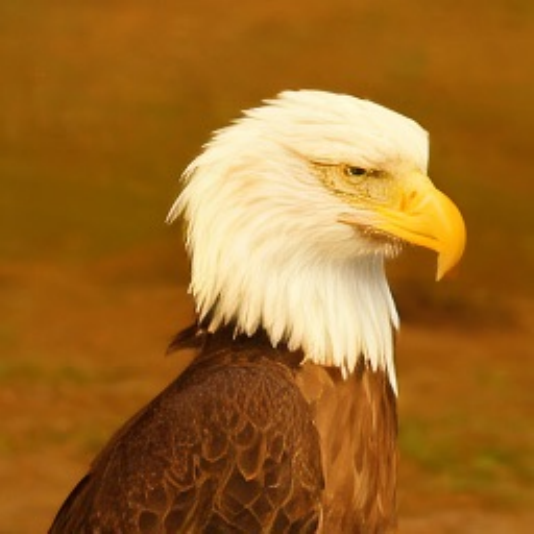}&
        \includegraphics[width=0.12\columnwidth]{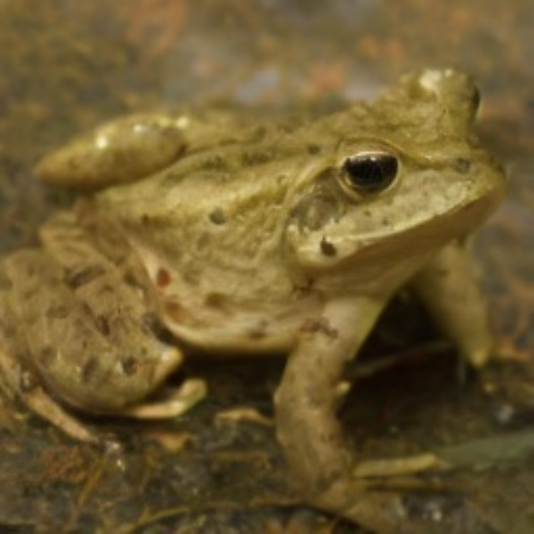}&
        \includegraphics[width=0.12\columnwidth]{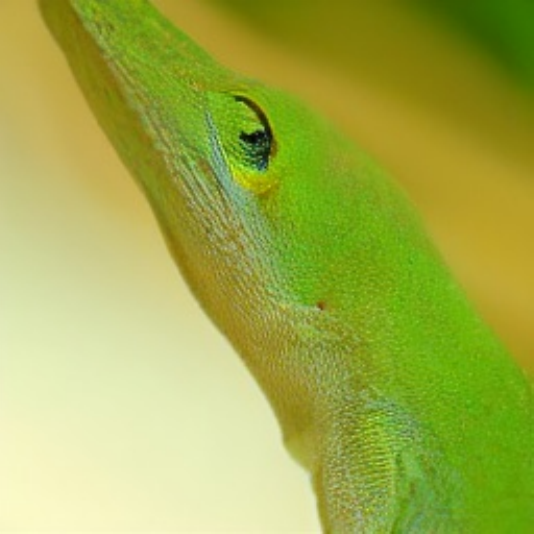}\\

        \includegraphics[width=0.12\columnwidth]{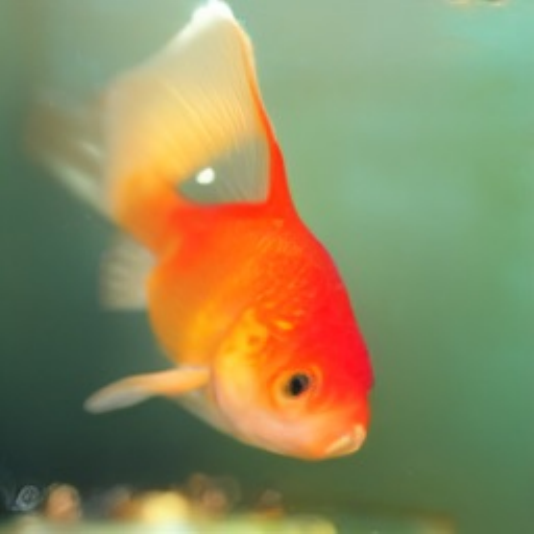}&
        \includegraphics[width=0.12\columnwidth]{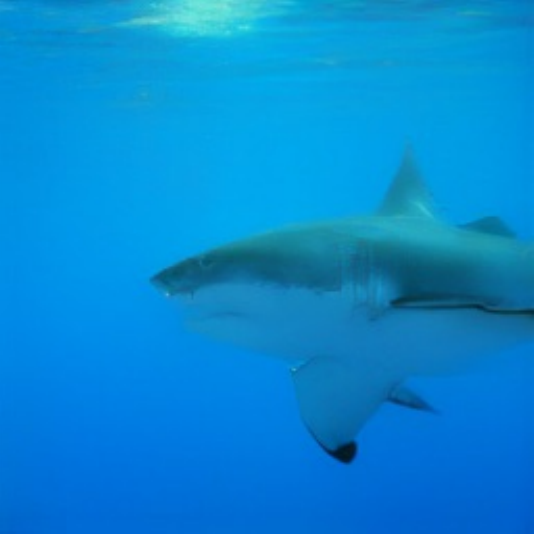}&
        \includegraphics[width=0.12\columnwidth]{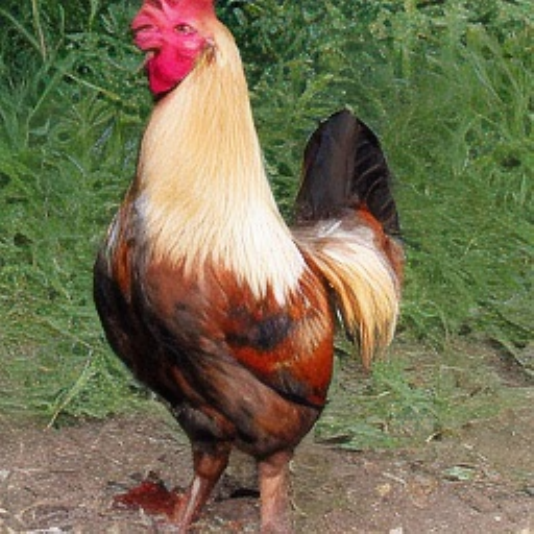}&
        \includegraphics[width=0.12\columnwidth]{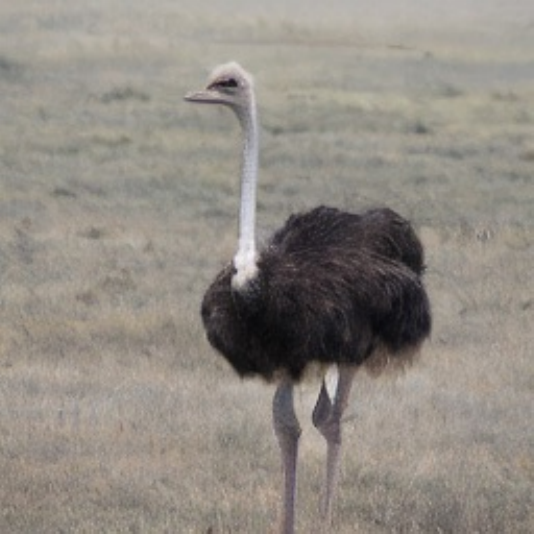}&
        \includegraphics[width=0.12\columnwidth]{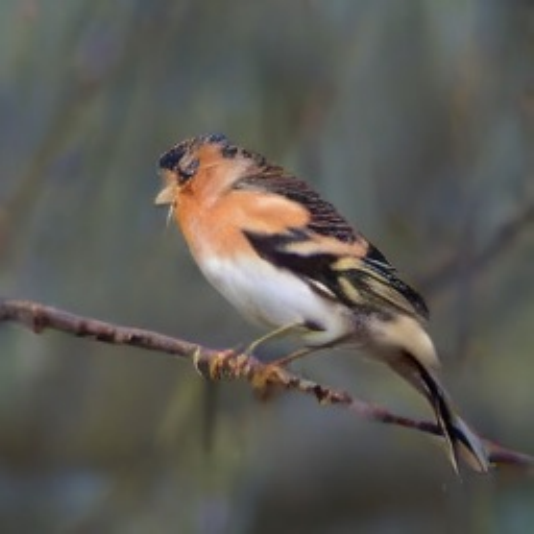}&
        \includegraphics[width=0.12\columnwidth]{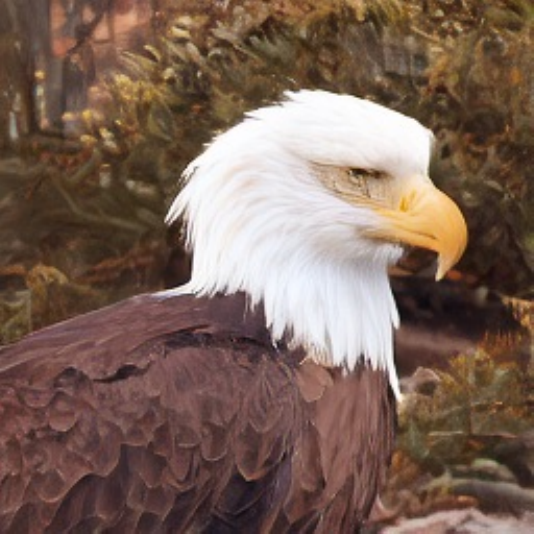}&
        \includegraphics[width=0.12\columnwidth]{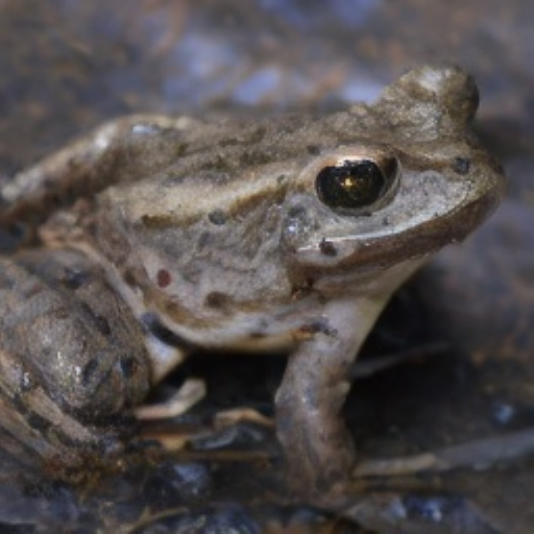}&
        \includegraphics[width=0.12\columnwidth]{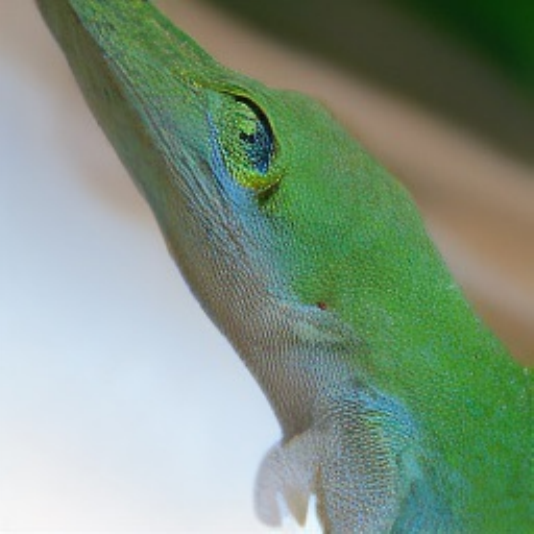}\\
      \end{tabular}
      \label{fig:imagenet_ptqd}
    \end{subfigure}

    \caption{Visual comparisons of generated images on ImageNet~\cite{deng2009imagenet}~(256$\times$256) for class-conditional image generation with LDM-4~\cite{rombach2022high} under a 3/8-bit setting. Each row corresponds to Full Precision, Q-Diffusion~\cite{li2023q}, PTQD~\cite{he2023ptqd}, TFMQ-DM~\cite{huang2024tfmq}, and Ours.}
    \label{fig:imagenet_comparison}
  \end{center}
\end{figure*}

\vspace{-0.5mm}
In this section, we show additional qualitative results for unconditional image generation, class-conditional image generation and text-to-image generation. For unconditional image generation, we conduct the LSUN-Bedroom~\cite{yu2015lsun} using LDM-4~\cite{rombach2022high} with 200 denoising steps, and LSUN-Church~\cite{yu2015lsun} using LDM-8~\cite{rombach2022high} with 500 denoising steps. The model weights and activations are quantized to 3/8 bits respectively.  
As illustrated in Fig.~\ref{fig:quali_ldm_bedroom_suppl} and Fig.~\ref{fig:quali_ldm_church_suppl}, AccuQuant successfully prevents the accumulation of quantization errors, ensuring that generated images closely match their full precision counterparts with images containing rich semantics even in low-bit configurations.
For example, in Fig.~\ref{fig:quali_ldm_bedroom_suppl}, we observe that Q-Diffusion~\cite{li2023q} and TFMQ-DM~\cite{huang2024tfmq} present an overall monotonous coloration compared to the full precision results and generate different content of the bed due to the accumulated quantization error. In contrast, AccuQuant preserves a coloration and the content similar to that of the full precision model.
For class conditional image generation, we conduct the ImageNet~\cite{deng2009imagenet} using LDM-4~\cite{rombach2022high} with 20 denoising steps and quantized weights and activations to 3/8 bits respectively.
As illustrated in Fig.~\ref{fig:imagenet_comparison}, Q-Diffusion~\cite{li2023q} and TFMQ-DM~\cite{huang2024tfmq}, which do not account for accumulating quantization error, respectively exhibit chroma noise or produce monotone coloration. PTQD~\cite{he2023ptqd}, which does account for accumulating quantization error, generates vivid colors but alters content compared to full precision and producing unrealistic artifacts such as a chicken with two heads~(\textit{\ie, row 3, column 3 in Fig.~\ref{fig:imagenet_comparison}}). In contrast, AccuQuant generates vibrant colors, realistic imagery, and outputs that closely match those of the full precision model.
For text-to-image generation, we use COCO validation prompts~\cite{lin2014microsoft} in Fig.~\ref{fig:appendix_sd_coco_prompts} and user defined prompts in Fig.~\ref{fig:appendix_sd_user_prompts} with 50 denoising steps of Stable Diffusion v1.4~\footref{fn:sd} quantized into 4/8 bit-widths.
As illustrated in Fig.~\ref{fig:appendix_sd_coco_prompts}, PCR~\cite{tang2025post}, which is designed to mitigate accumulating quantization error, allocates more bit-width to activations yet still produces images with altered style or distortion compared to the full precision model. In contrast, AccuQuant uses fewer quantizers and a lower bit-width than PCR~\cite{tang2025post} but effectively minimizes accumulated quantization error, generating realistic images that closely match those of the full precision model.
We also visualize generated images conditioned on user defined prompts. As illustrated in Fig.~\ref{fig:appendix_sd_user_prompts}, Q-Diffusion~\cite{li2023q} and PCR~\cite{tang2025post} generates output that differ from those of the full precision model and misalign with the text prompt~(\textit{\eg, No astronaut in row 2, column 2 and row 3, column 4 in Fig.~\ref{fig:astronaut}}), while AccuQuant align strongly with the prompts. In summary, AccuQuant not only achieves superior performance on quantitative evaluation metrics but also qualitatively generates images that most closely resemble full precision outputs among state-of-the-art methods, align strongly with the prompts, and exhibit naturalness and semantic richness, demonstrating the superiority of our framework.

\begin{table}[t]
    \centering
    \sisetup{detect-weight=true,detect-family=true,mode=text}
    \caption{Computational cost of calibration process for DDIM~\cite{song2020denoising} on CIFAR-10 ($32\times32$)~\cite{krizhevsky2009learning} with 100 timesteps.}
    \vspace{3mm}
    
    \label{tab:ddim_cifar_calib_overhead}
    \resizebox{1\linewidth}{!}{
    \begin{tabular}{l c c c c S[table-format=2.2] S[table-format=2.2]}
        \toprule
        \textbf{Method} & \textbf{Bits (W/A)} & \textbf{Batch size} & \textbf{Calibration time (h)} & \textbf{Energy (Wh)} & {\textbf{FID}$\downarrow$} & {\textbf{FID2FP32}$\downarrow$} \\
        \midrule
        Full-Precision            & 32/32 & 8 & --   & --     & 4.26 & 0.00 \\
        \midrule
        Q-Diffusion~\cite{li2023q}   & 6/6   & 8 & 5.97 & \textbf{519.39} & 30.46 & 35.24 \\
        \cellcolor[HTML]{ECF9FF}Ours          & \cellcolor[HTML]{ECF9FF}6/6   & \cellcolor[HTML]{ECF9FF}8 & \cellcolor[HTML]{ECF9FF}\textbf{5.56} & \cellcolor[HTML]{ECF9FF}561.56 & \cellcolor[HTML]{ECF9FF}\textbf{5.79}  & \cellcolor[HTML]{ECF9FF}\textbf{3.30} \\
        \bottomrule
    \end{tabular}}
\end{table}
\begin{table}[!t]
    \centering
    \sisetup{detect-weight=true,detect-family=true,mode=text}
    \caption{Computational cost of real-time CPU inference. Experiments are conducted on CIFAR-10 ($32\times32$)~\cite{krizhevsky2009learning} with DDIM~\cite{song2020denoising} over 100 timesteps.}
    \label{tab:r6_cpu_cost}
    \vspace{5mm}
    \renewcommand{\arraystretch}{1.3}
    \resizebox{1\textwidth}{!}{
    \begin{tabular}{l c c
                    S[table-format=4.0]
                    S[table-format=4.1]
                    S[table-format=3.3]
                    S[table-format=3.2]
                    S[table-format=1.2]}
        \toprule
        \textbf{Method} & \textbf{Bits (W/A)} & \textbf{Batch size}
        & {\textbf{GBops}}
        & {\textbf{Memory (MB)}}
        & {\textbf{CPU latency (s)}}
        & {\textbf{Model size (MB)}}
        & {\textbf{Speedup (×)}} \\
        \midrule
        Full-Precision           & 32/32 & 64 & 6597 & 1726.0 & 94.589 & 143.08 & 1.00 \\
        \midrule
        Q-Diffusion~\cite{li2023q}  & 8/8   & 64 &  798 & 1541.7 & 31.322 &  36.57 & 3.02 \\
        \cellcolor[HTML]{ECF9FF}Ours         & \cellcolor[HTML]{ECF9FF}8/8   & \cellcolor[HTML]{ECF9FF}64 &  \cellcolor[HTML]{ECF9FF}798 & \cellcolor[HTML]{ECF9FF}1541.7 & \cellcolor[HTML]{ECF9FF}31.380 &  \cellcolor[HTML]{ECF9FF}36.58 & \cellcolor[HTML]{ECF9FF}3.01 \\
        \bottomrule
    \end{tabular}}
\end{table}
\begin{table}[!t]
    \centering
    \sisetup{detect-weight=true,detect-family=true,mode=text}
    \caption{Comparison against lightweight QAT. Experiments are conducted on ImageNet ($256\times256$)~\cite{deng2009imagenet} with LDM-4~\cite{blattmann2023align} over 20 timesteps. * denotes result under the same resource constraint.}
    \label{tab:r7_qat_compare}
    \vspace{5mm}
    \resizebox{1\linewidth}{!}{
    \begin{tabular}{l c
                    S[table-format=1.2]
                    S[table-format=5.0]
                    S[table-format=2.2]
                    S[table-format=2.2]
                    S[table-format=3.2]
                    S[table-format=2.2]}
        \toprule
        \textbf{Method} & \textbf{W/A} &
        {\textbf{Calibration time (h)}} &
        {\textbf{\# of calibration data}} &
        {\textbf{FID}$\downarrow$} &
        {\textbf{sFID}$\downarrow$} &
        {\textbf{IS}$\uparrow$} &
        {\textbf{FID2FP32}$\downarrow$} \\
        \midrule
        Full-Precision & 32/32 & {{--}} & {{--}} & 11.13 & 7.85 & 368.19 & 0.00 \\
        \midrule
        EfficientDM*      & 4/8 & 4.34 &  5120  & 12.43 & 25.07 & 197.86 & 14.55 \\
        EfficientDM-Full~\cite{He_2024_ICLR}  & 4/8 & 6.50 & 32000  &  9.92 & \bfseries 7.40 & 351.79 &  1.63 \\
        \cellcolor[HTML]{ECF9FF}Ours              & \cellcolor[HTML]{ECF9FF}4/8 & \cellcolor[HTML]{ECF9FF}4.34 &  \cellcolor[HTML]{ECF9FF}5120  &  \cellcolor[HTML]{ECF9FF}\bfseries9.39 &  \cellcolor[HTML]{ECF9FF}7.41 & \cellcolor[HTML]{ECF9FF}\bfseries 356.48 & \cellcolor[HTML]{ECF9FF}\bfseries 0.65 \\
        \bottomrule
    \end{tabular}}
\end{table}
\section{Computational cost and efficiency}
\label{sec:computation}
In this section, we provide a detailed analysis of computational cost and efficiency in terms of calibration and inference. In Tab.~\ref{tab:ddim_cifar_calib_overhead}, we compare AccuQuant with Q-Diffusion~\cite{li2023q} in terms of computational cost for calibration and the generation performance. We can see that AccuQuant achieves superior performance with shorter calibration time compared to Q-Diffusion. We note that although AccuQuant may require more computational cost for one loss calculation, it only requires 50 epochs per group. In contrast, Q-diffusion calibrates each layer and residual block for 5000 epochs. Although AccuQuant incurs a marginal increase in energy consumption, it delivers a favorable trade-off for real-world deployment.
In addition, we evaluate the inference efficiency of our quantized diffusion model against both Q-Diffusion and the full-precision baseline. Since the official PyTorch quantization API does not support bit-widths lower than 8, we quantize both weights and activations to 8 bits and measure the memory usage and runtime latency using ONNX Runtime with the Intel Xeon Gold 6226R CPU. As shown in Tab.~\ref{tab:r6_cpu_cost}, both quantized models achieve over a 3 times speedup compared to the FP model. Our method incurs a marginally higher latency (by less than 0.18\%) and larger model size (by 0.006 MB) compared to Q-Diffusion, which we attribute to the separate quantization parameters per group. We also compare in Tab.~\ref{tab:r7_qat_compare} AccuQuant with a lightweight QAT approach, EfficientDM~\cite{He_2024_ICLR}. We report the results both under the same resource constraints (denoted as EfficientDM*) and using the official training recipe~\footnote{\url{https://github.com/ThisisBillhe/EfficientDM}} with larger training data and longer training time (denoted as EfficientDM-Full). We find that under an identical setting, AccuQuant achieves substantially better performance than EfficientDM*, suggesting that QAT frameworks cannot converge sufficiently within limited resource budgets. Furthermore, although EfficientDM-Full benefits from extended training time and larger datasets, AccuQuant still outperforms it in terms of FID2FP32, demonstrating the effectiveness of our method. In summary, although our method marginally increases latency and model size compared to Q-Diffusion, the benefits of reduced calibration time and improved generation quality offer a favorable trade-off for practical deployment.

\begin{table}[!t]
    \centering
    \sisetup{detect-weight=true,detect-family=true,mode=text}
    \caption{Quantitative results for Gaussian deblurring on CelebA ($256\times256$)~\cite{Liu_2015_ICCV}.}
    \vspace{5mm}
    \label{tab:deblur}
    \resizebox{0.5\textwidth}{!}{
    \begin{tabular}{l c S[table-format=2.2] S[table-format=1.4]}
        \toprule
        \textbf{Method} & \textbf{W/A} & {\textbf{PSNR}$\uparrow$} & {\textbf{LPIPS}$\downarrow$} \\
        \midrule
        Full precision & 32/32 & 31.23 & 0.0584 \\
        \midrule
        Q-Diffusion~\cite{li2023q}    & 4/8   & 27.63 & 0.2131 \\
        \cellcolor[HTML]{ECF9FF}Ours           & \cellcolor[HTML]{ECF9FF}4/8   & \cellcolor[HTML]{ECF9FF}\bfseries 30.57 & \cellcolor[HTML]{ECF9FF}\bfseries 0.0984 \\
        \midrule
        Q-Diffusion~\cite{li2023q}    & 3/8   & 24.70 & 0.4574 \\
        \cellcolor[HTML]{ECF9FF}Ours           & \cellcolor[HTML]{ECF9FF}3/8   & \cellcolor[HTML]{ECF9FF}\bfseries 26.15 & \cellcolor[HTML]{ECF9FF}\bfseries 0.3979 \\
        \bottomrule
    \end{tabular}}
\end{table}
\begin{table}[!t]
    \centering
    \sisetup{detect-weight=true,detect-family=true,mode=text}
    \caption{Quantitative results for style transfer on the COCO validation set (512×512)~\cite{lin2014microsoft}. We report SSIM, PSNR, LPIPS, style loss, content loss, and FID2FP32. All scores are calculated using the output of the full-precision model, since ground truth images are not available for style transfer.}
    \label{tab:style_transfer}
    \vspace{5mm}
    \resizebox{1\textwidth}{!}{
    \begin{tabular}{l c
                    S[table-format=1.4]
                    S[table-format=2.2]
                    S[table-format=1.4]
                    S[table-format=1.4]
                    S[table-format=2.4]
                    S[table-format=2.2]}
        \toprule
        \textbf{Method} & \textbf{W/A} &
        {\textbf{SSIM}$\uparrow$} &
        {\textbf{PSNR}$\uparrow$} &
        {\textbf{LPIPS}$\downarrow$} &
        {\textbf{Style loss}$\downarrow$} &
        {\textbf{Content loss}$\downarrow$} &
        {\textbf{FID2FP32}$\downarrow$} \\
        \midrule
        Q-Diffusion~\cite{li2023q} & 4/8 & 0.6645 & 18.50 & 0.3310 & 0.0012 & 13.2267 & 14.20 \\
        \cellcolor[HTML]{ECF9FF}Ours        & \cellcolor[HTML]{ECF9FF}4/8 & \cellcolor[HTML]{ECF9FF}\bfseries 0.7444 & \cellcolor[HTML]{ECF9FF}\bfseries 20.49 & \cellcolor[HTML]{ECF9FF}\bfseries 0.2594 & \cellcolor[HTML]{ECF9FF}\bfseries 0.0006 & \cellcolor[HTML]{ECF9FF}\bfseries 9.3687 & \cellcolor[HTML]{ECF9FF}\bfseries 10.81 \\
        \bottomrule
    \end{tabular}}
\end{table}
\begin{table}[!t]
    \centering
    \sisetup{detect-weight=true,detect-family=true,mode=text}
    \caption{Quantization results of DIT-XL/2 with 100 denoising steps on ImageNet ($256\times256$)~\cite{krizhevsky2009learning}. W/A denotes bit-widths of weights and activations. (a) without classifier-free guidance. (b) with classifier-free guidance scale of 1.5.}
    \label{tab:dit48_cfg}
    \vspace{5mm}
    \begin{subtable}[t]{0.48\textwidth}
        \centering
        \resizebox{\linewidth}{!}{
        \begin{tabular}{l c S[table-format=3.2] S[table-format=2.2]}
            \toprule
            \textbf{Method} & \textbf{W/A} & {\textbf{FID}$\downarrow$} & {\textbf{sFID}$\downarrow$} \\
            \midrule
            Full-precision  & 32/32 & 12.41 & 19.23 \\
            \midrule
            PTQ4DM~\cite{shang2023post}    & 4/8   & 213.66 & 85.11 \\
            RepQ-ViT~\cite{li2023repq}  & 4/8   & 224.14 & 81.24 \\
            TFMQ-DM~\cite{huang2024tfmq}   & 4/8   & 143.47 & 61.09 \\
            PTQ4DiT~\cite{Wu_2024_NeurIPS}   & 4/8   & 28.90  & 34.56 \\
            \cellcolor[HTML]{ECF9FF}Ours      & \cellcolor[HTML]{ECF9FF}4/8   & \cellcolor[HTML]{ECF9FF}\bfseries18.60  & \cellcolor[HTML]{ECF9FF}\bfseries 18.83 \\
            \bottomrule
        \end{tabular}}
        \caption{\textbf{without} classifier-free guidance}

    \end{subtable}\hfill
    \begin{subtable}[t]{0.48\textwidth}
        \centering
        \resizebox{\linewidth}{!}{
        \begin{tabular}{l c S[table-format=3.2] S[table-format=2.2]}
            \toprule
            \textbf{Method} & \textbf{W/A} & {\textbf{FID}$\downarrow$} & {\textbf{sFID}$\downarrow$} \\
            \midrule
            Full-precision  & 32/32 & 5.31 & 17.61 \\
            \midrule
            PTQ4DM~\cite{shang2023post}     & 4/8   & 215.68 & 86.63 \\
            RepQ-ViT~\cite{li2023repq}  & 4/8   & 226.60 & 77.93 \\
            TFMQ-DM~\cite{huang2024tfmq}   & 4/8   & 141.90 & 56.01 \\
            PTQ4DiT~\cite{Wu_2024_NeurIPS}   & 4/8   & 7.75   & 22.01 \\
            \cellcolor[HTML]{ECF9FF}Ours      & \cellcolor[HTML]{ECF9FF}4/8   & \cellcolor[HTML]{ECF9FF}\bfseries6.80   & \cellcolor[HTML]{ECF9FF}\bfseries17.78 \\
            \bottomrule
        \end{tabular}}
        \caption{\textbf{with} classifier-free guidance}

    \end{subtable}
\end{table}
\section{Expanding to various tasks and advanced models}
In this section, we demonstrate the generalization ability of AccuQuant by evaluating image restoration (\eg, Gaussian deblurring), style transfer, and applying to transformer-based diffusion models~\cite{Peebles_2023_ICCV}. 
For Gaussian deblurring, we leverage the pre-trained DDPG~\cite{Garber_2024_CVPR} model from the official GitHub repository~\footnote{\url{https://github.com/tirer-lab/DDPG}} on the CelebA dataset (256 $\times$ 256 resolution, 1K images)~\cite{Liu_2015_ICCV}, injecting noise with $\sigma_Y$ = 0.05. Also, we use 100 total timesteps with a group size of 5 and calibrate with 64 samples obtained from a full-precision model. In Tab.~\ref{tab:deblur}, AccuQuant consistently outperforms Q-Diffusion~\cite{li2023q} across both PSNR and LPIPS metrics, demonstrating its effectiveness in the restoration setting. For style transfer (Fig.~\ref{tab:style_transfer}), we use Stable Diffusion v1.4~\footnote{\url{https://huggingface.co/CompVis/stable-diffusion-v-1-4-original}} with the prompt \textit{'A cartoon style'} and evaluated on the COCO validation set~\cite{lin2014microsoft}, with the entire timestep of 35 using the DDIM sampler~\cite{song2020denoising}. Since ground-truth images are unavailable for style transfer, we evaluate by comparing to the outputs obtained from the full-precision model. Note that we compute style and content losses following~\cite{Gatys_2016_CVPR}, using feature maps from a VGG network~\cite{Simonyan_2015_ICLR}. In Tab.~\ref{tab:style_transfer}, AccuQuant outperforms all evaluated metrics, highlighting its robustness for diverse generative tasks beyond standard image synthesis. 
For transformer-based diffusion model~\cite{Peebles_2023_ICCV} adaptation, we compare AccuQuant to recent diffusion-focused quantization methods~\cite{shang2023post,huang2024tfmq} including diffusion-transformer based quantization~\cite{Wu_2024_NeurIPS} and advanced transformer quantization method~\cite{li2023repq}. We conduct experiments on the ImageNet dataset~\cite{deng2009imagenet} using the DiT-XL/2~\cite{Peebles_2023_ICCV} with 100 timesteps, under both without and with classifier-free guidance settings~\cite{Ho_2021_NeurIPSW}. As shown in Tab.~\ref{tab:dit48_cfg}, AccuQuant outperforms transformer-based, diffusion-focused, and even DiT-specific quantization methods, demonstrating that reducing the accumulating error across diffusion timesteps is the key factor for quantizing diffusion models. In summary, by designing the quantization methods at the diffusion framework level rather than for a specific diffusion architecture, AccuQuant generalizes to diverse tasks and diffusion models.


\begin{figure*}[p]
  \centering
  \captionsetup[subfigure]{font=small,labelformat=empty,justification=centering}

  \begin{subfigure}[t]{\textwidth}
    \centering
    \begin{tabular}{@{}c@{\hspace{1mm}}c@{\hspace{1mm}}c@{\hspace{1mm}}c@{\hspace{1mm}}c@{}}
      \includegraphics[width=0.18\linewidth]{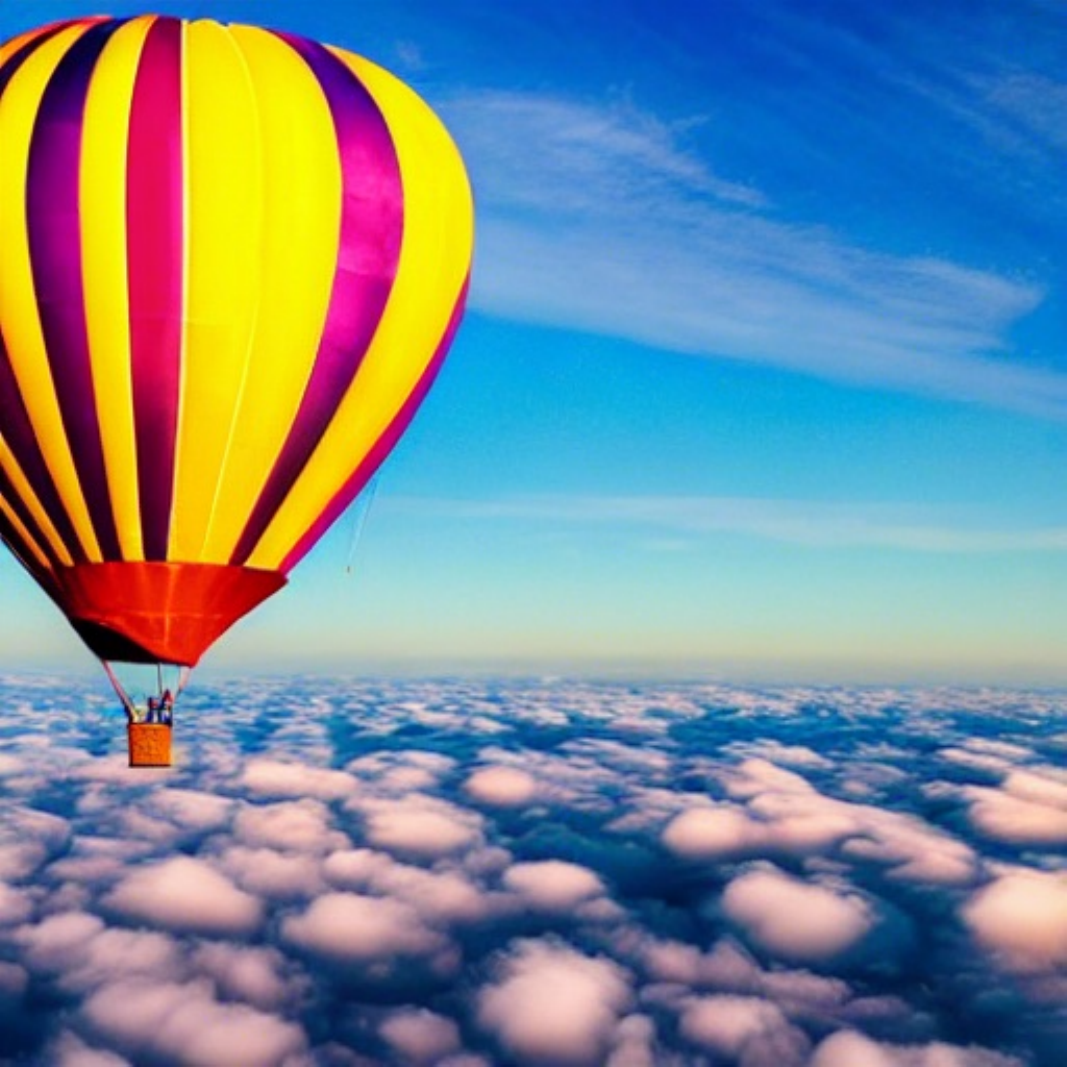} &
      \includegraphics[width=0.18\linewidth]{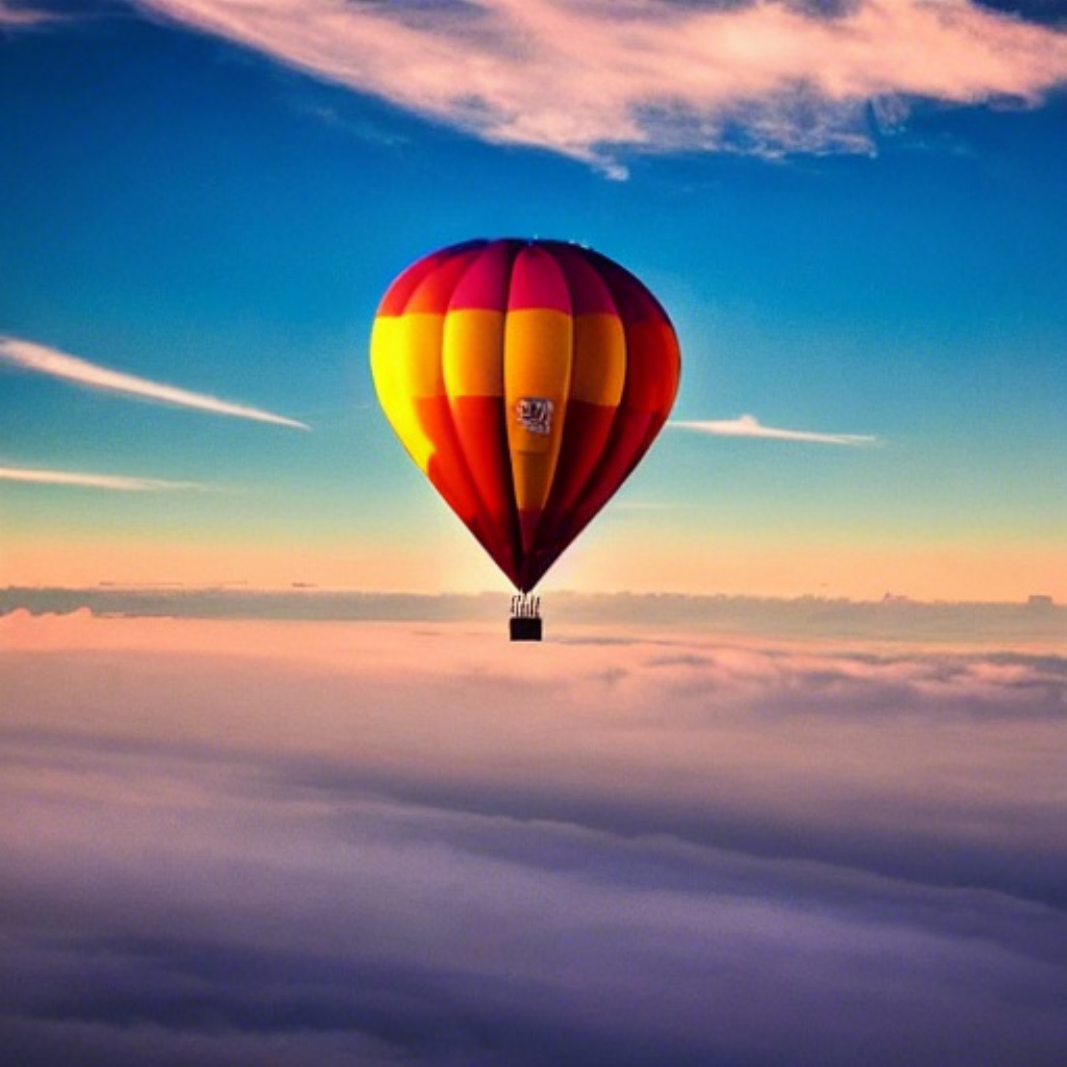} &
      \includegraphics[width=0.18\linewidth]{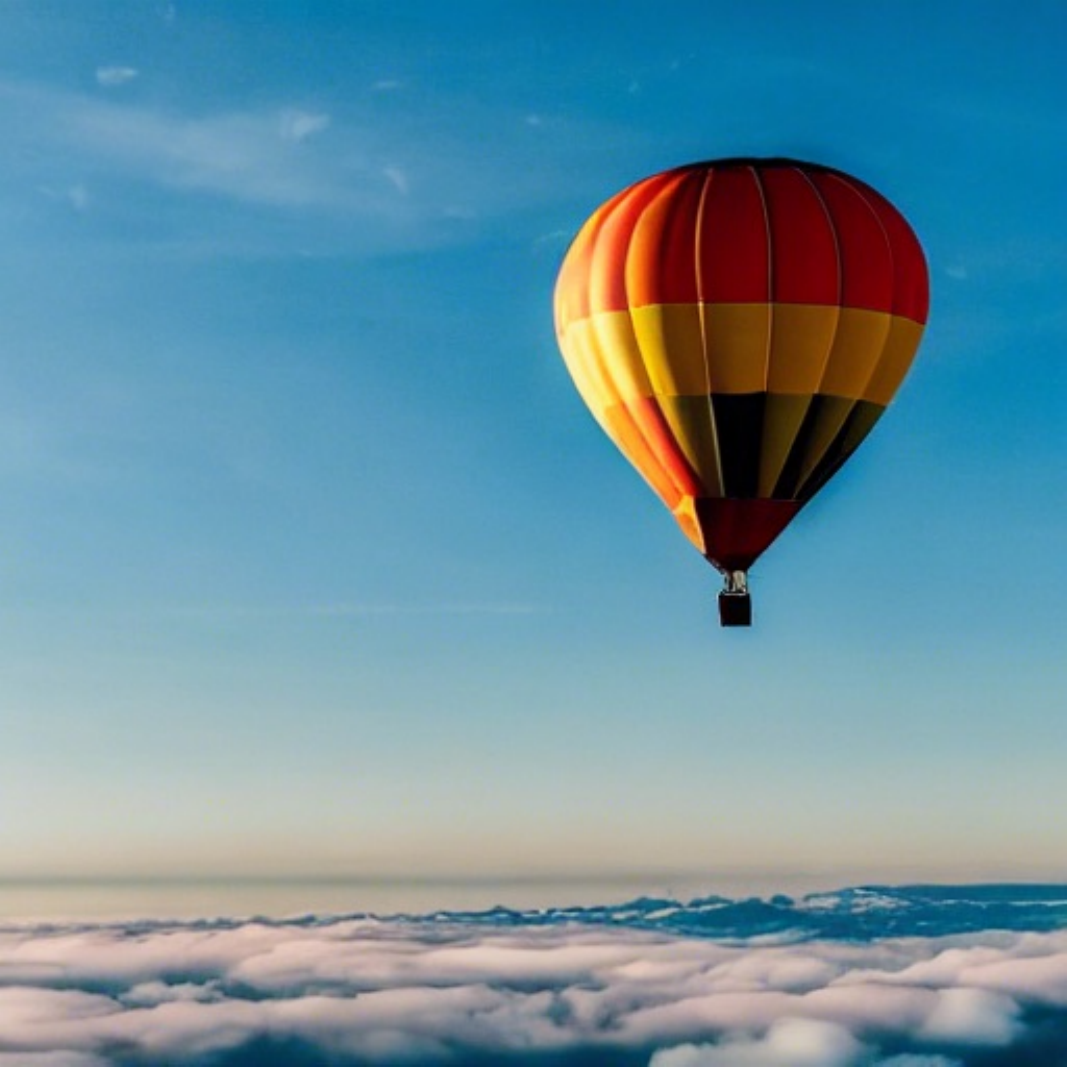} &
      \includegraphics[width=0.18\linewidth]{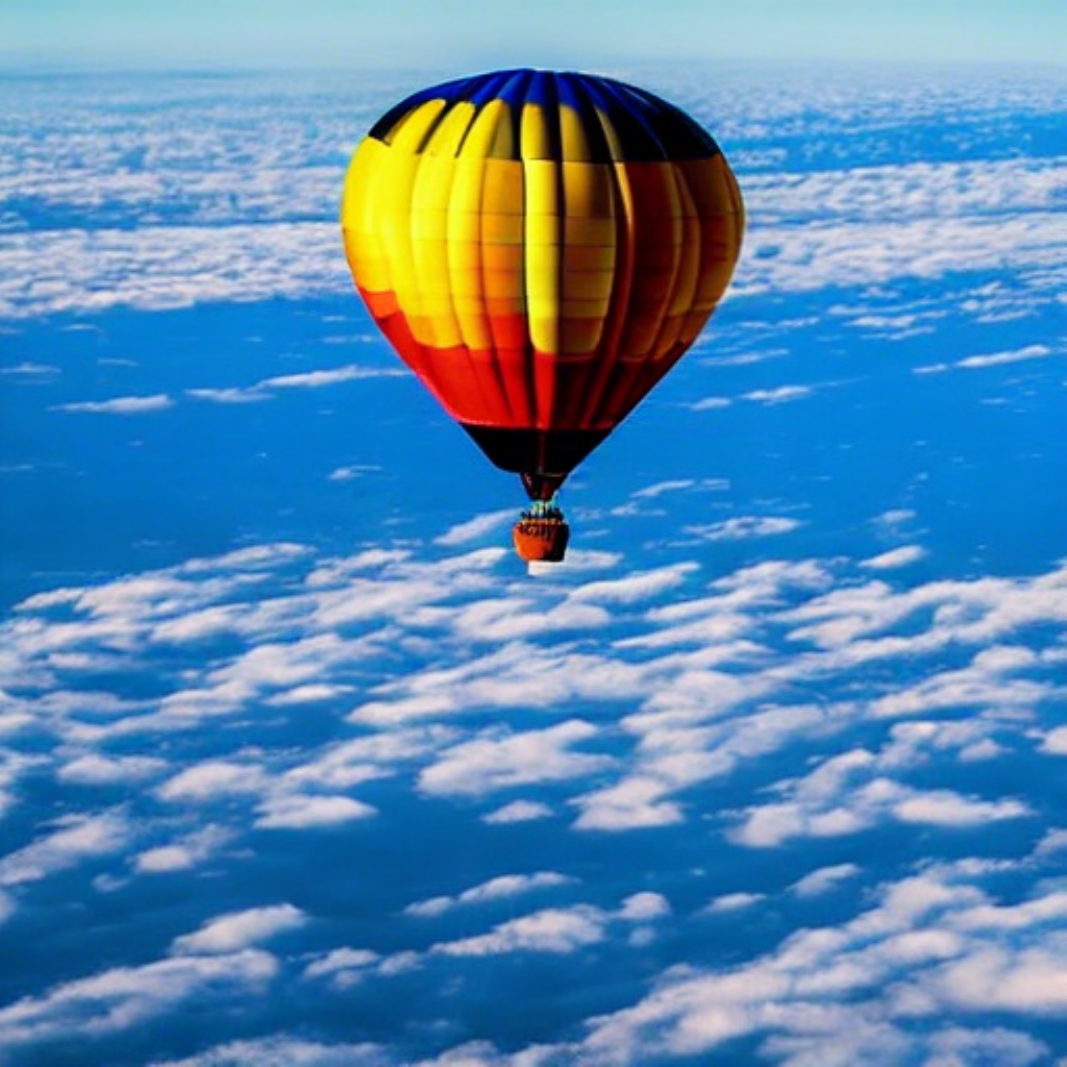} &
      \includegraphics[width=0.18\linewidth]{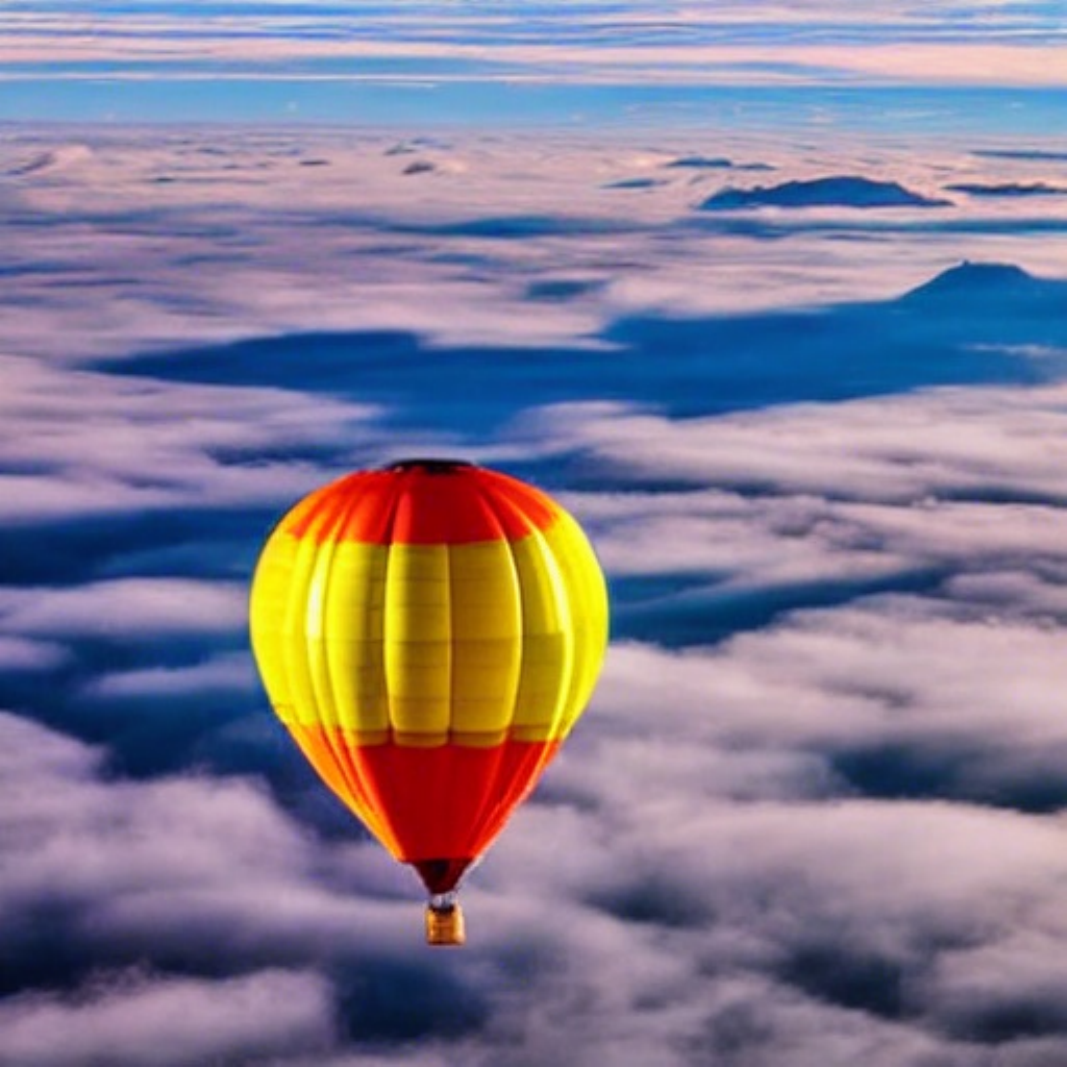} \\[-1pt]
      \includegraphics[width=0.18\linewidth]{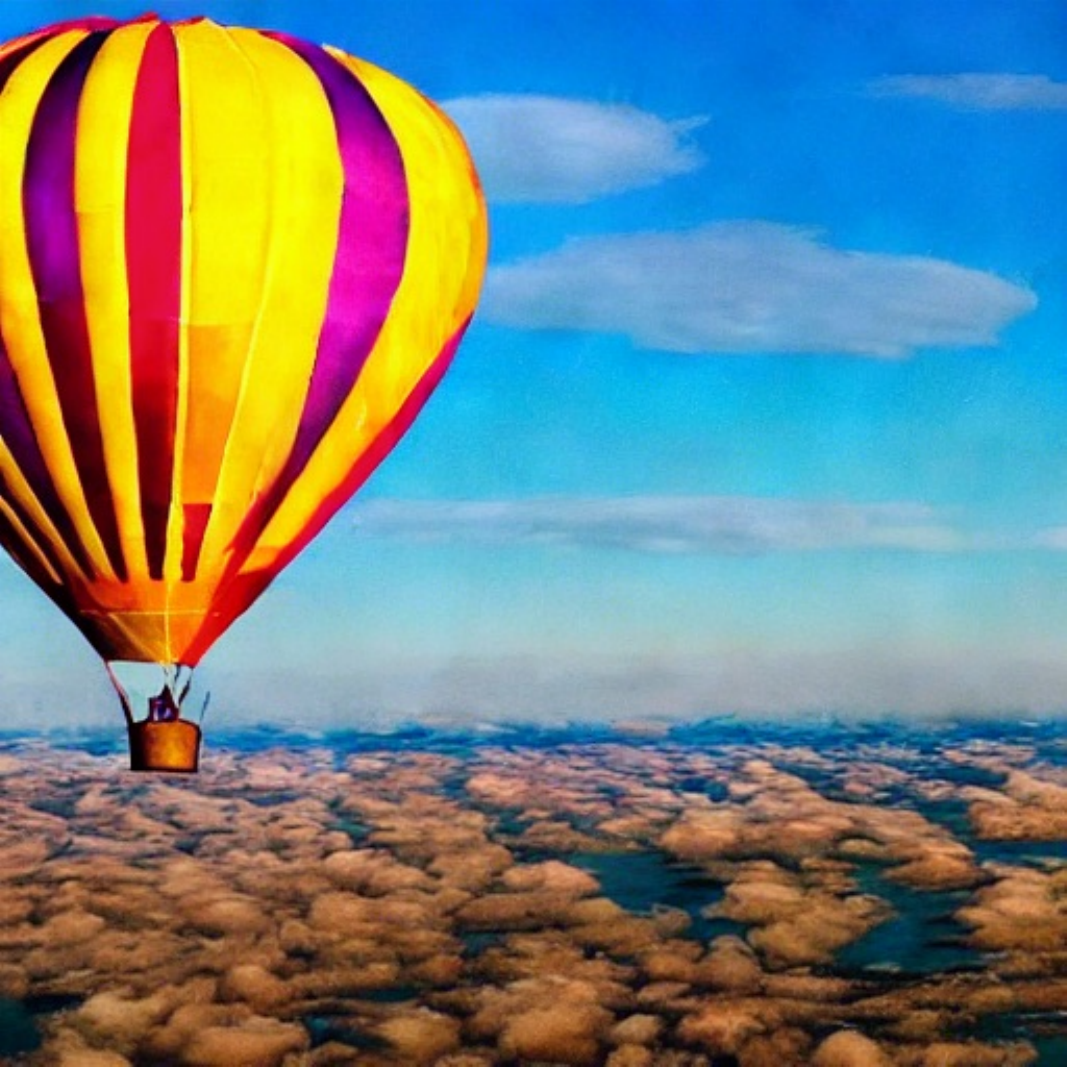} &
      \includegraphics[width=0.18\linewidth]{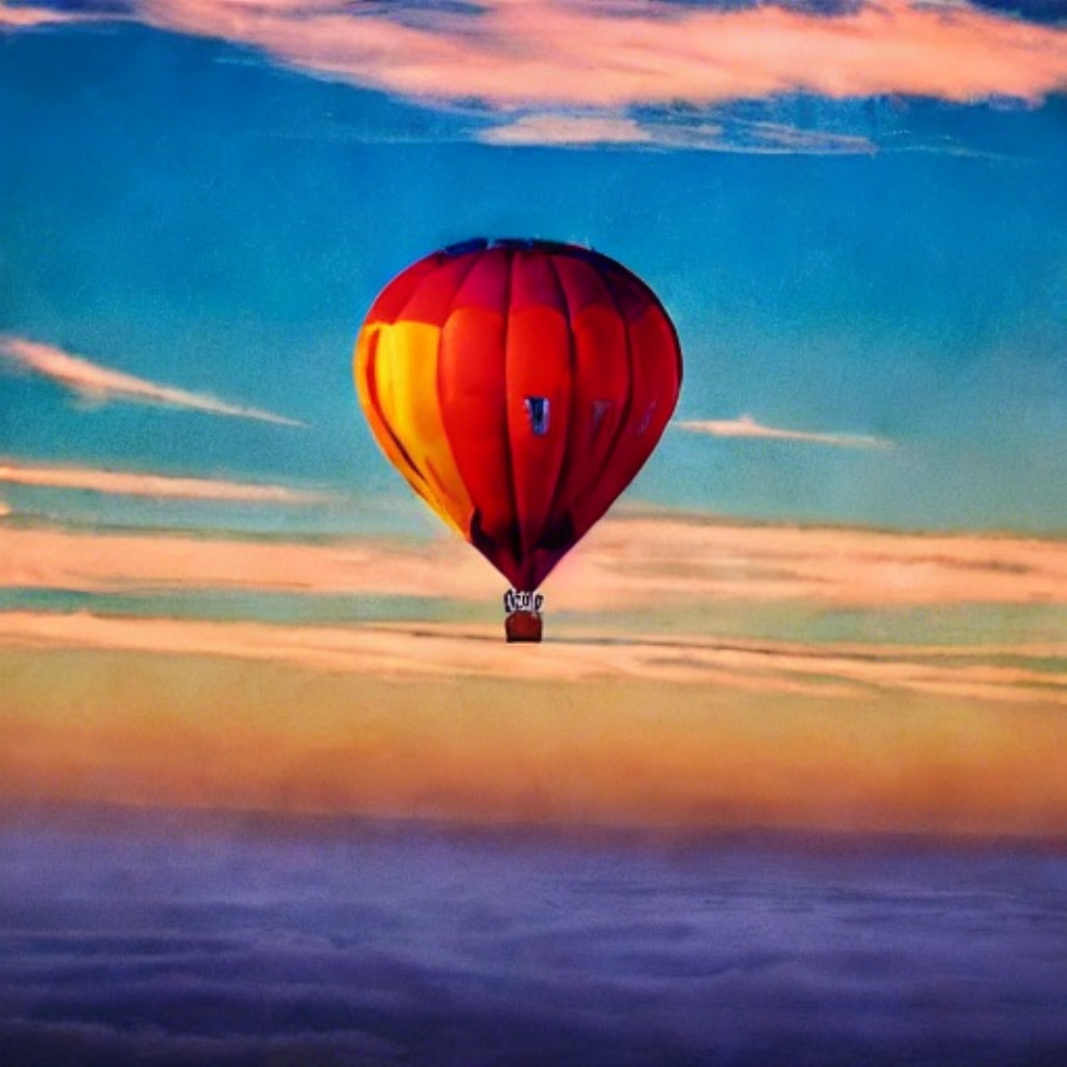} &
      \includegraphics[width=0.18\linewidth]{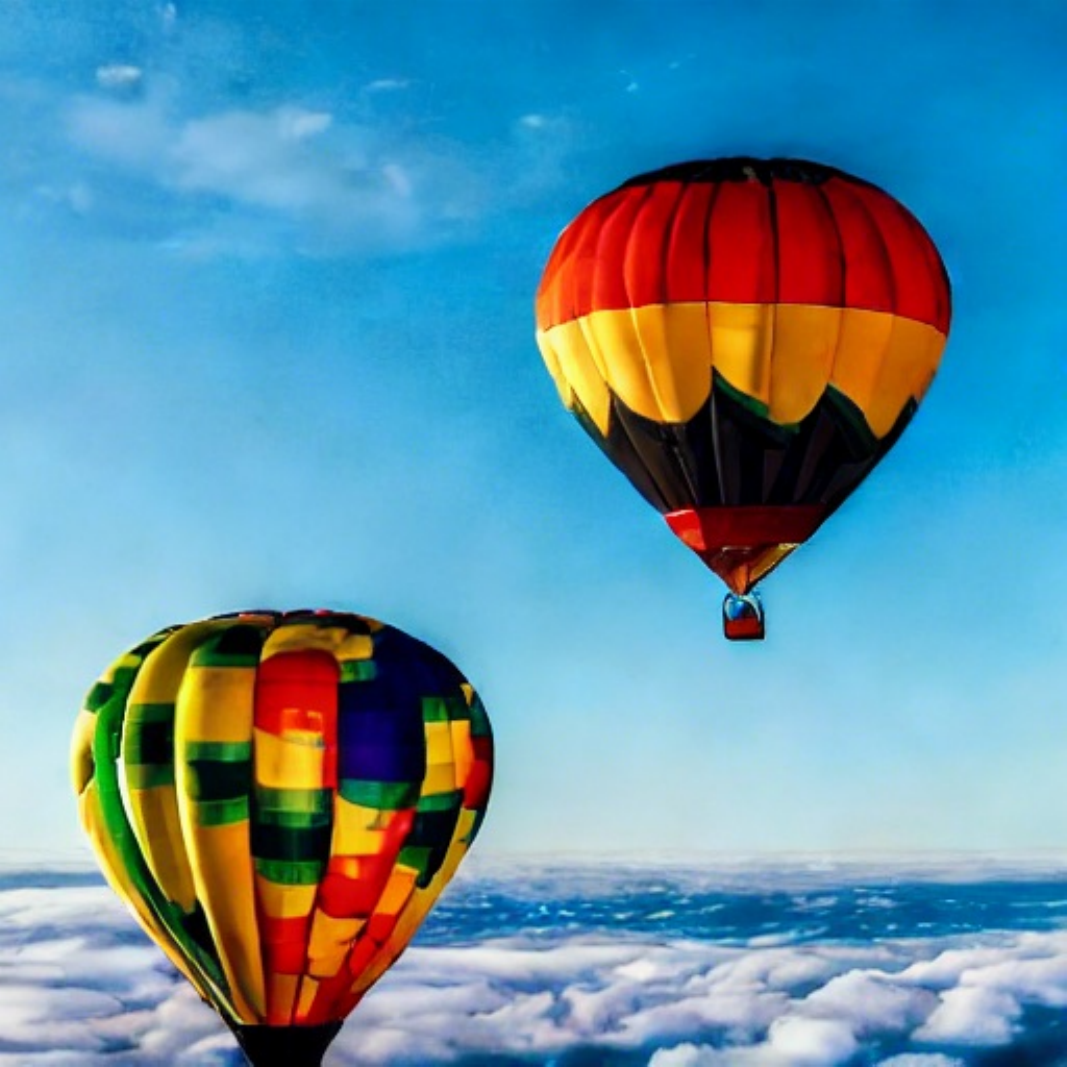} &
      \includegraphics[width=0.18\linewidth]{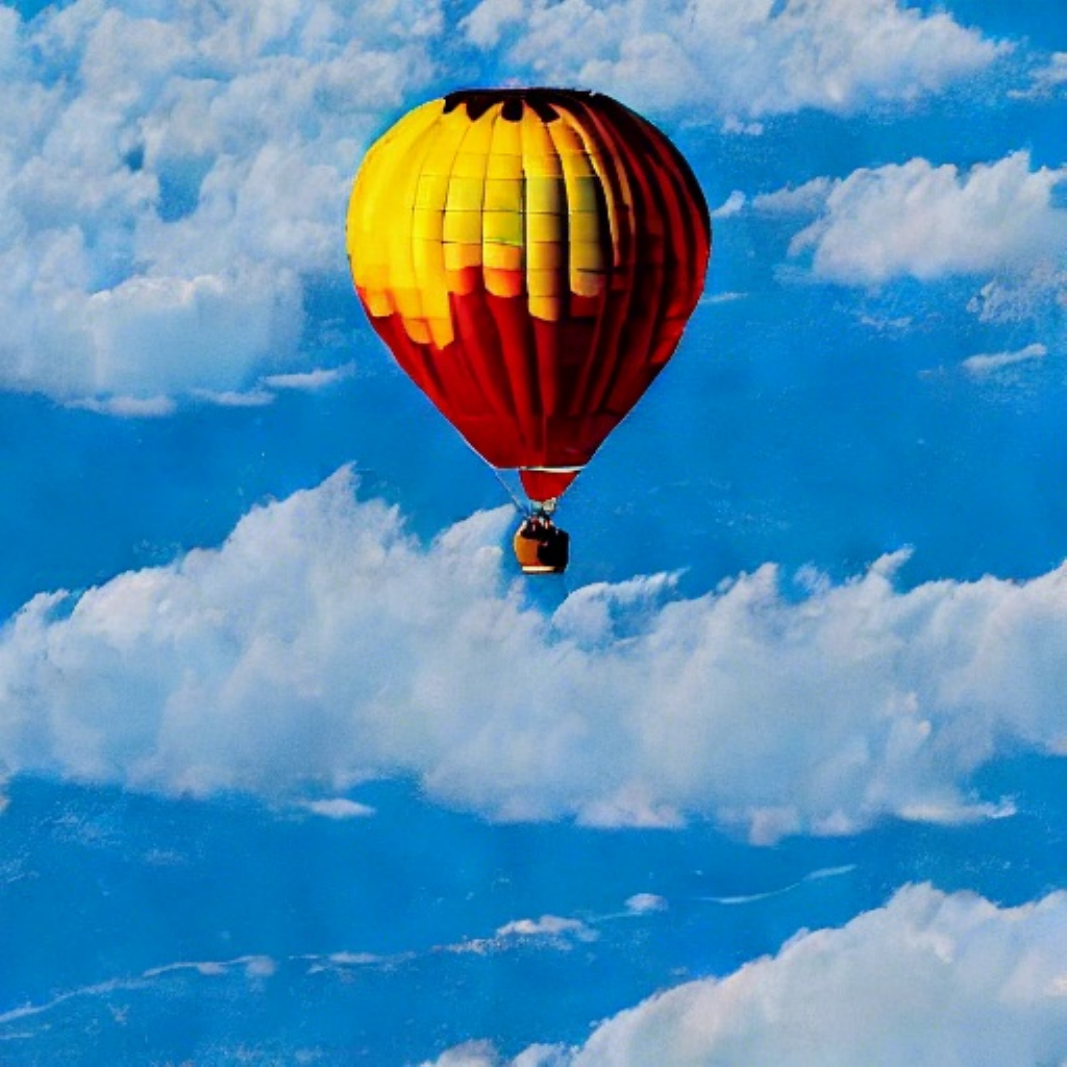} &
      \includegraphics[width=0.18\linewidth]{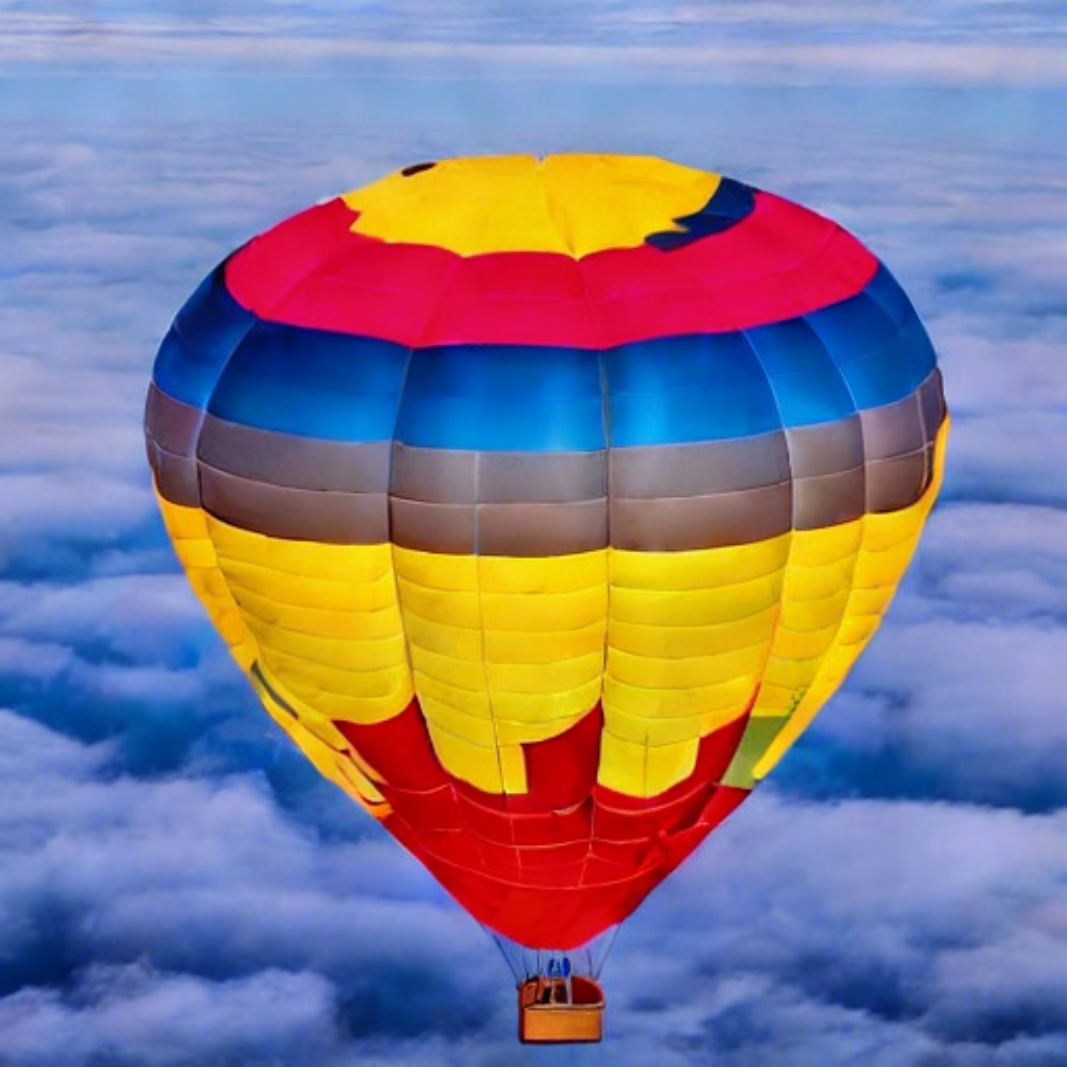} \\[-1pt]
      \includegraphics[width=0.18\linewidth]{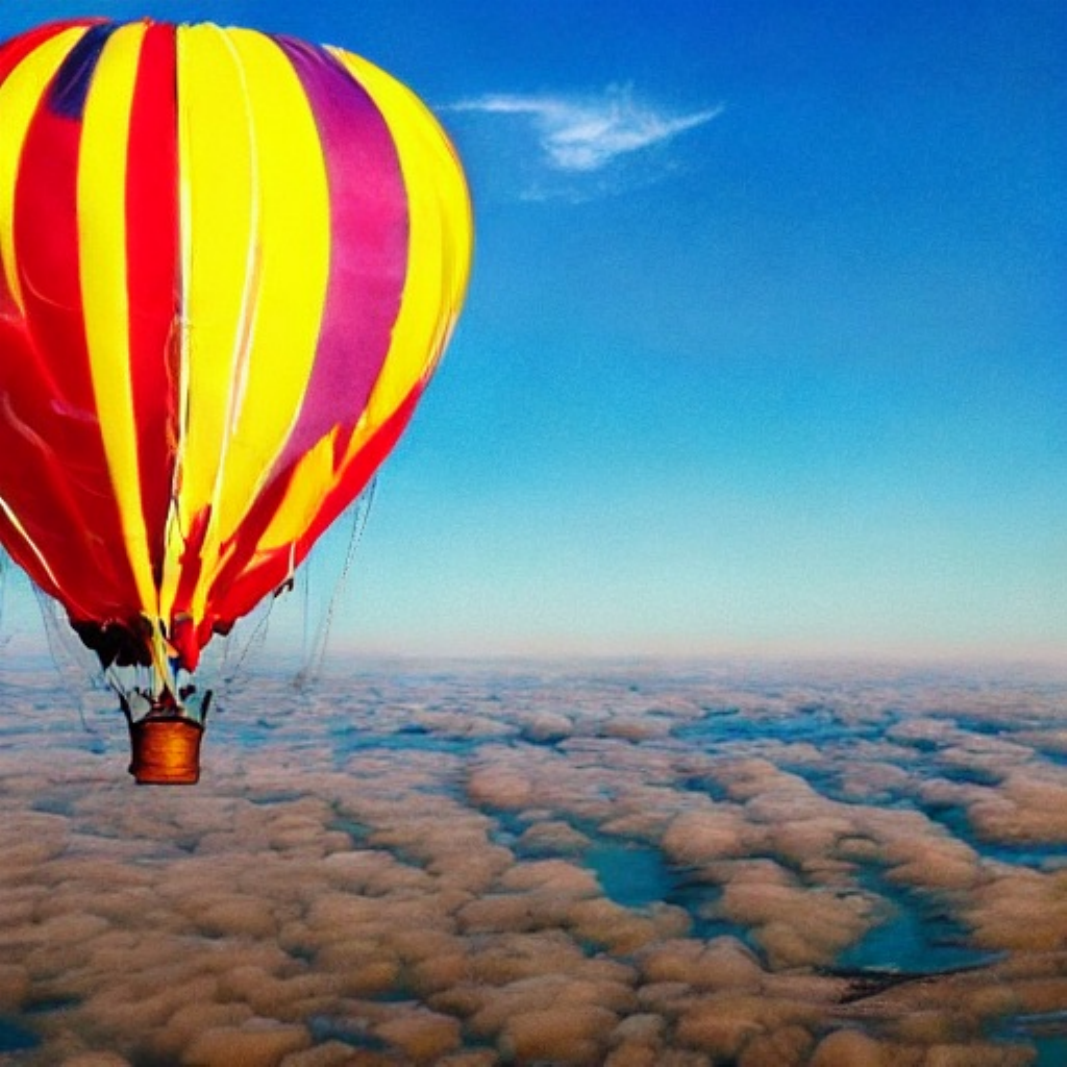} &
      \includegraphics[width=0.18\linewidth]{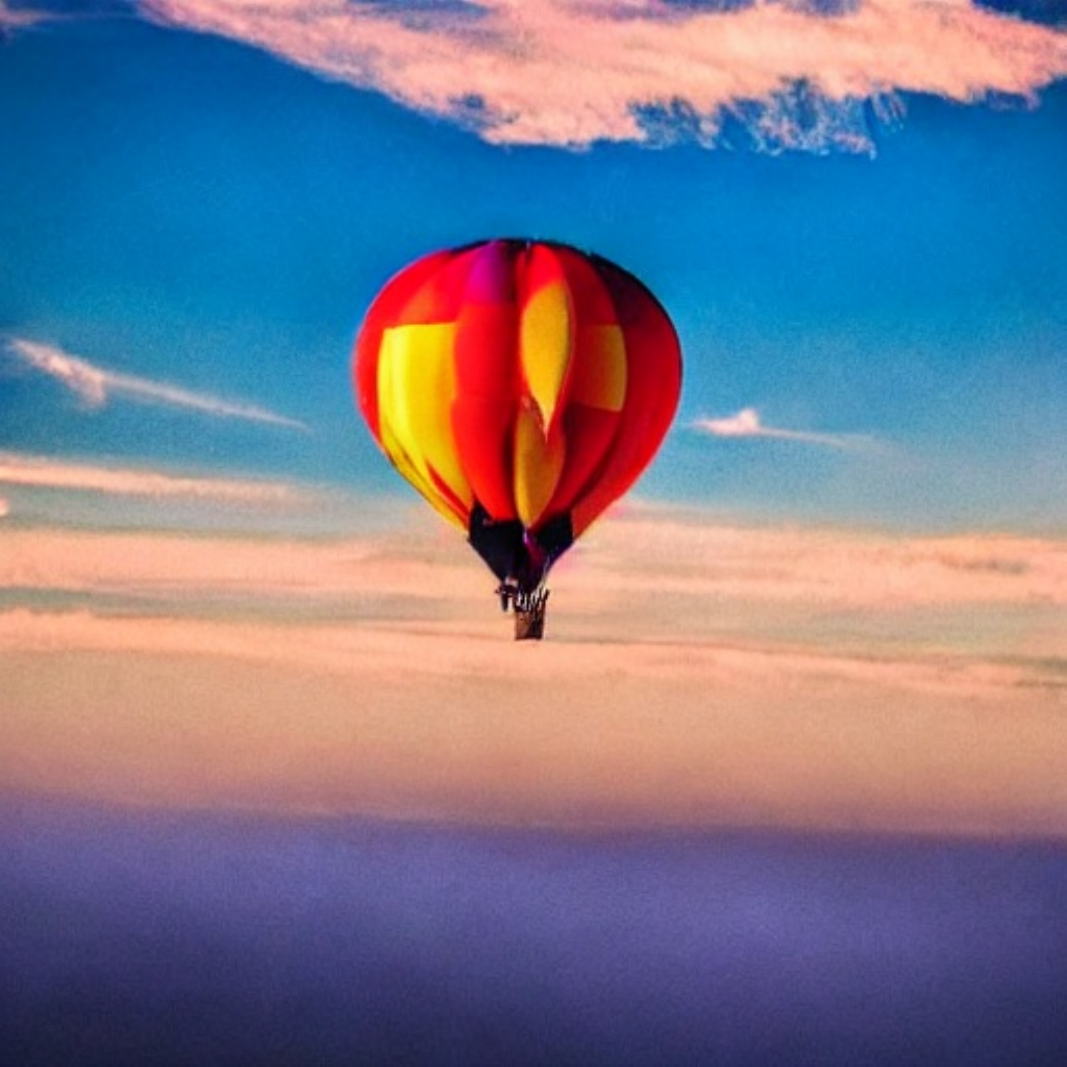} &
      \includegraphics[width=0.18\linewidth]{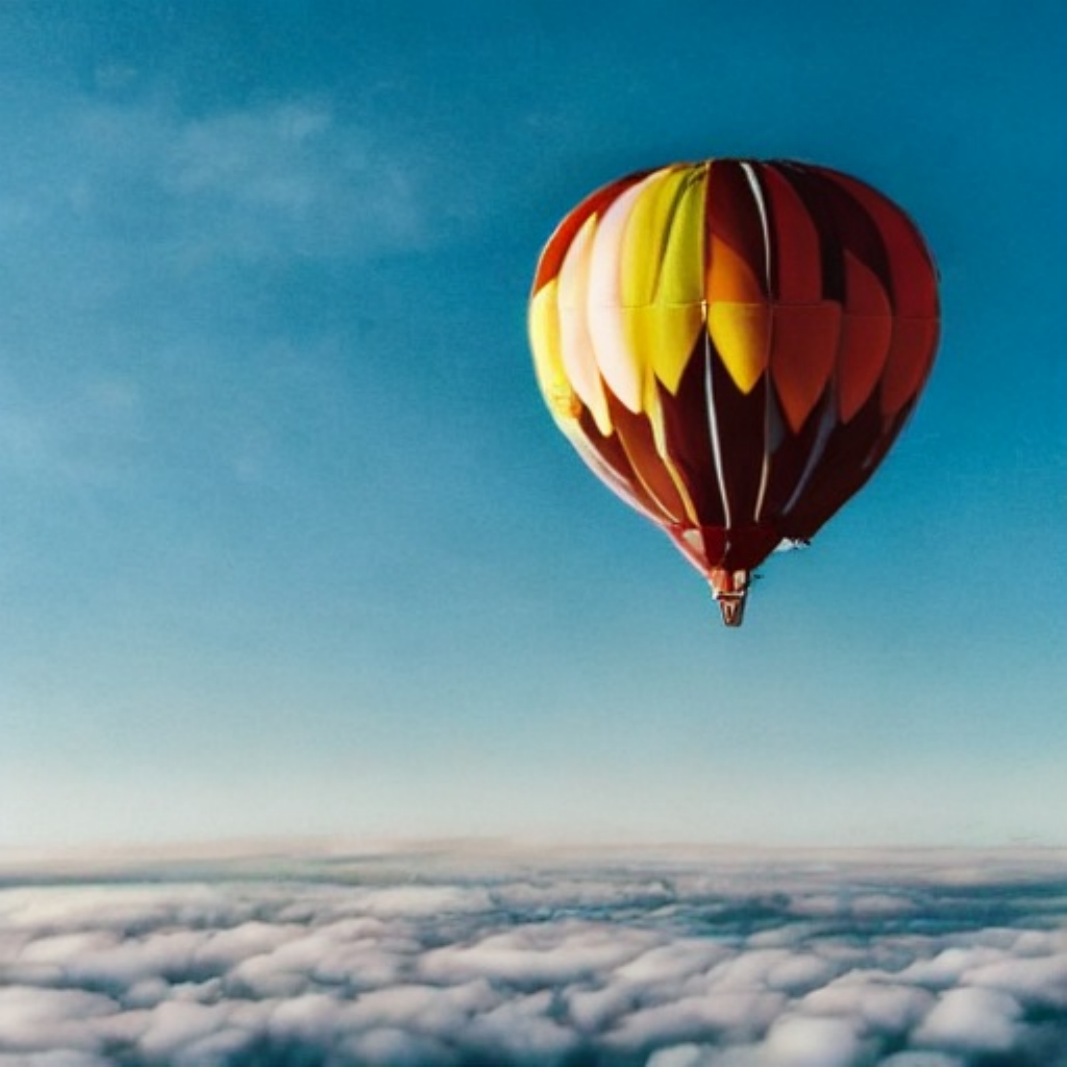} &
      \includegraphics[width=0.18\linewidth]{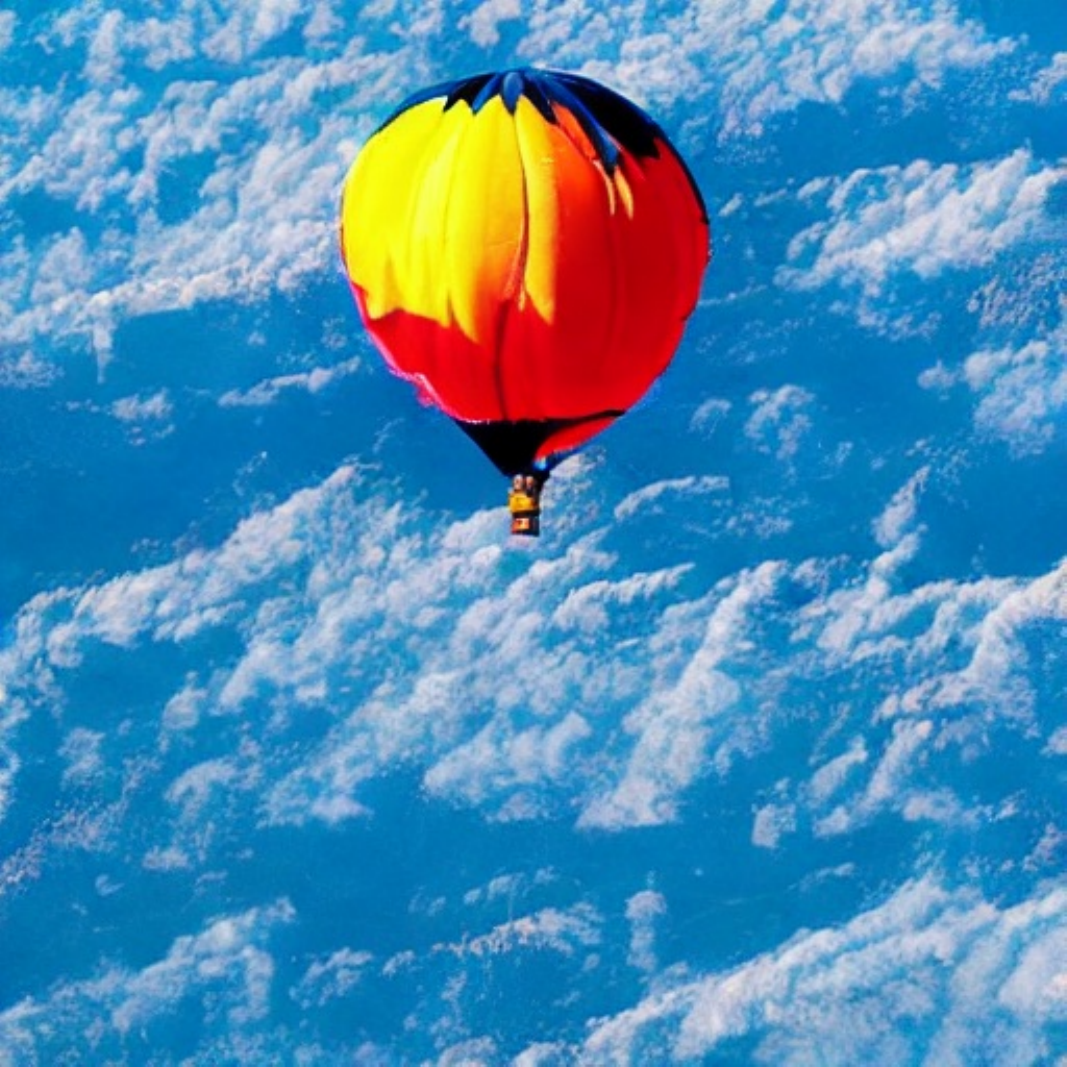} &
      \includegraphics[width=0.18\linewidth]{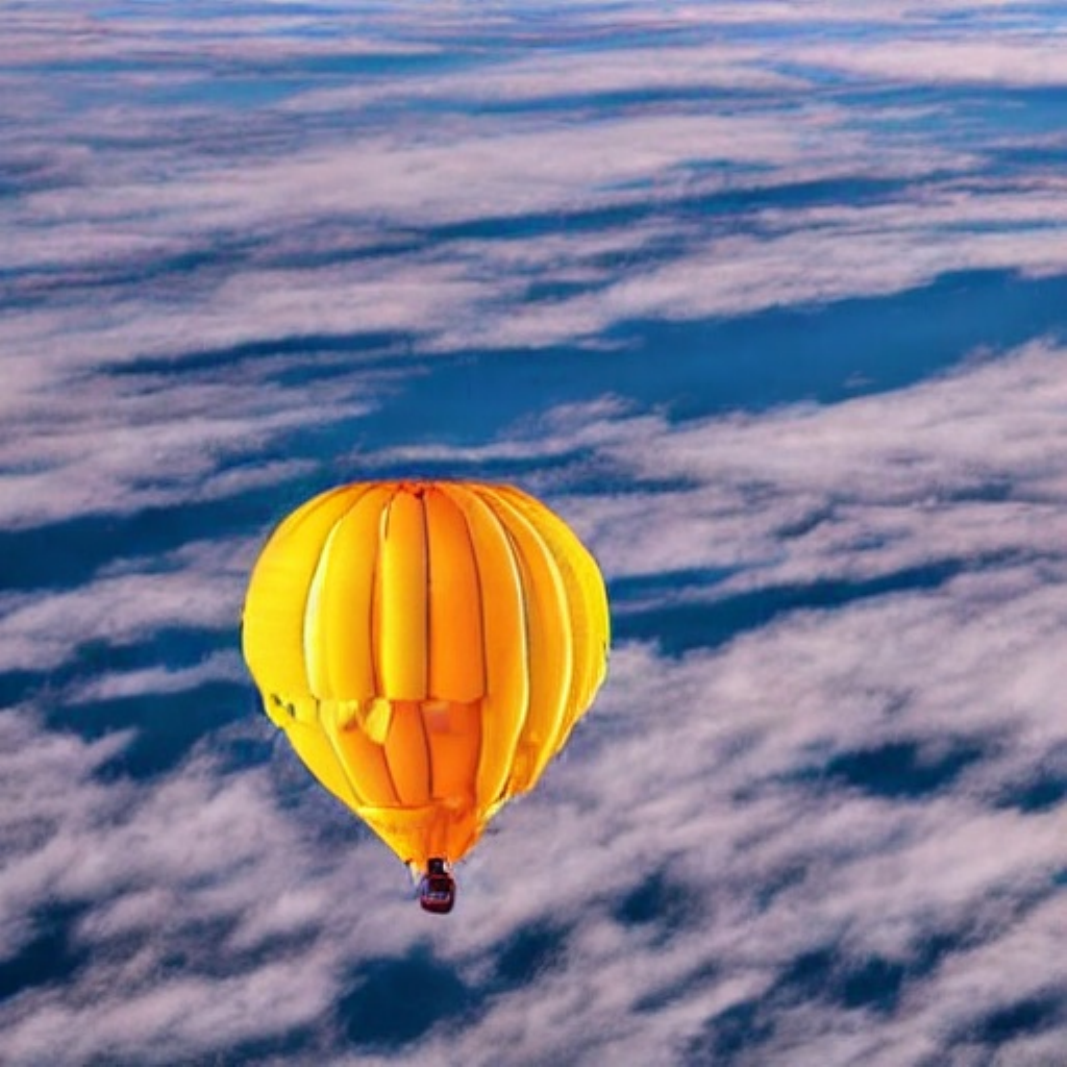} \\[-1pt]
      \includegraphics[width=0.18\linewidth]{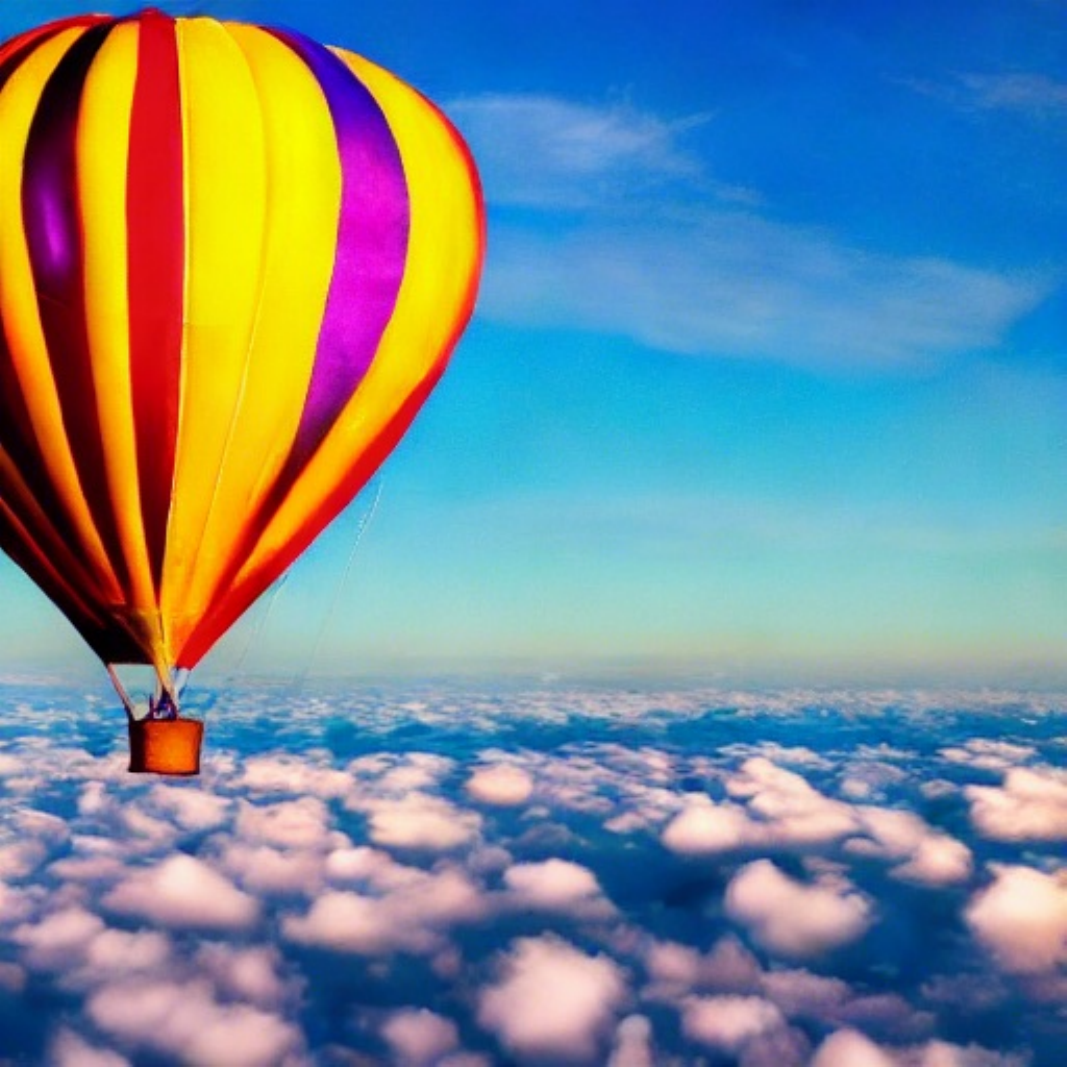} &
      \includegraphics[width=0.18\linewidth]{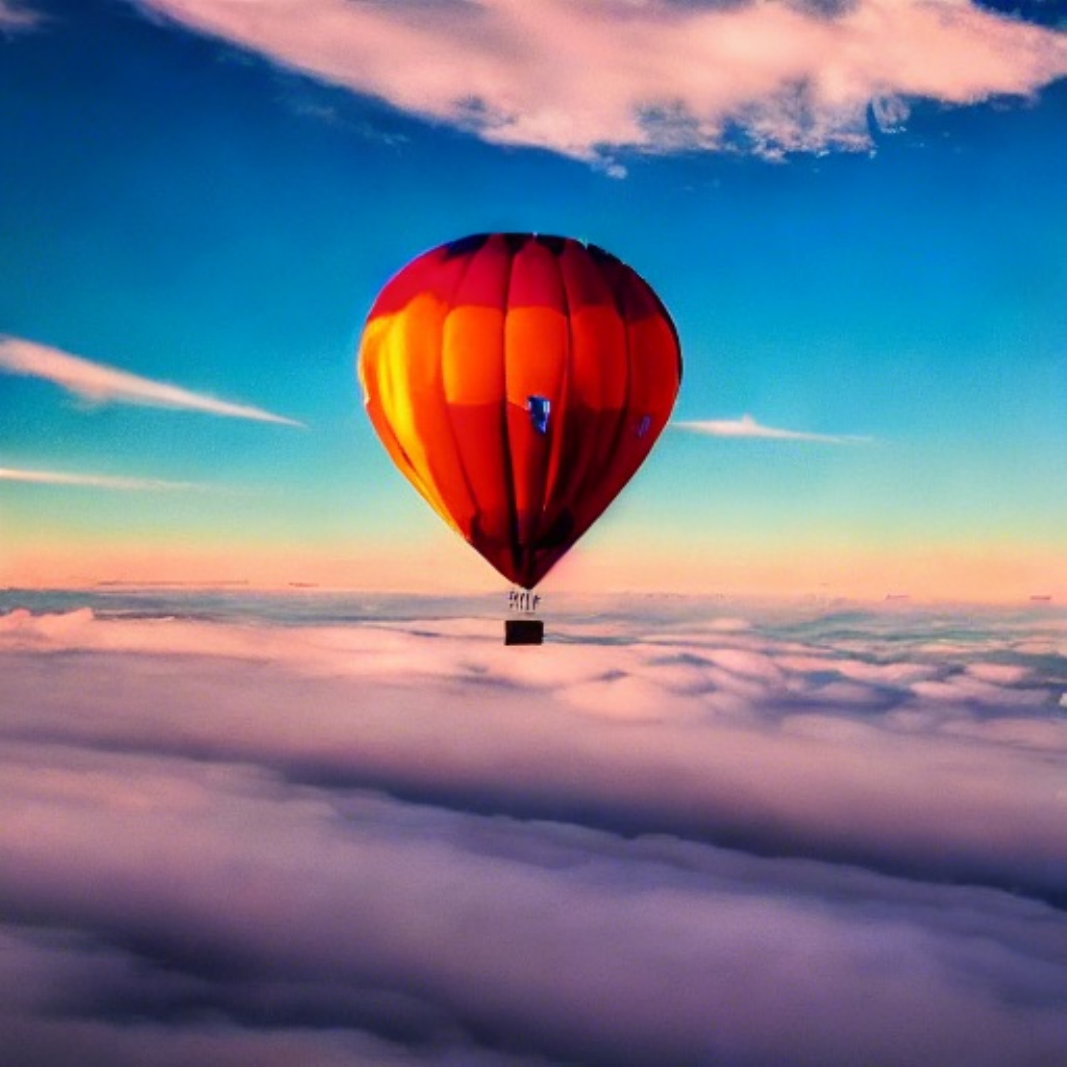} &
      \includegraphics[width=0.18\linewidth]{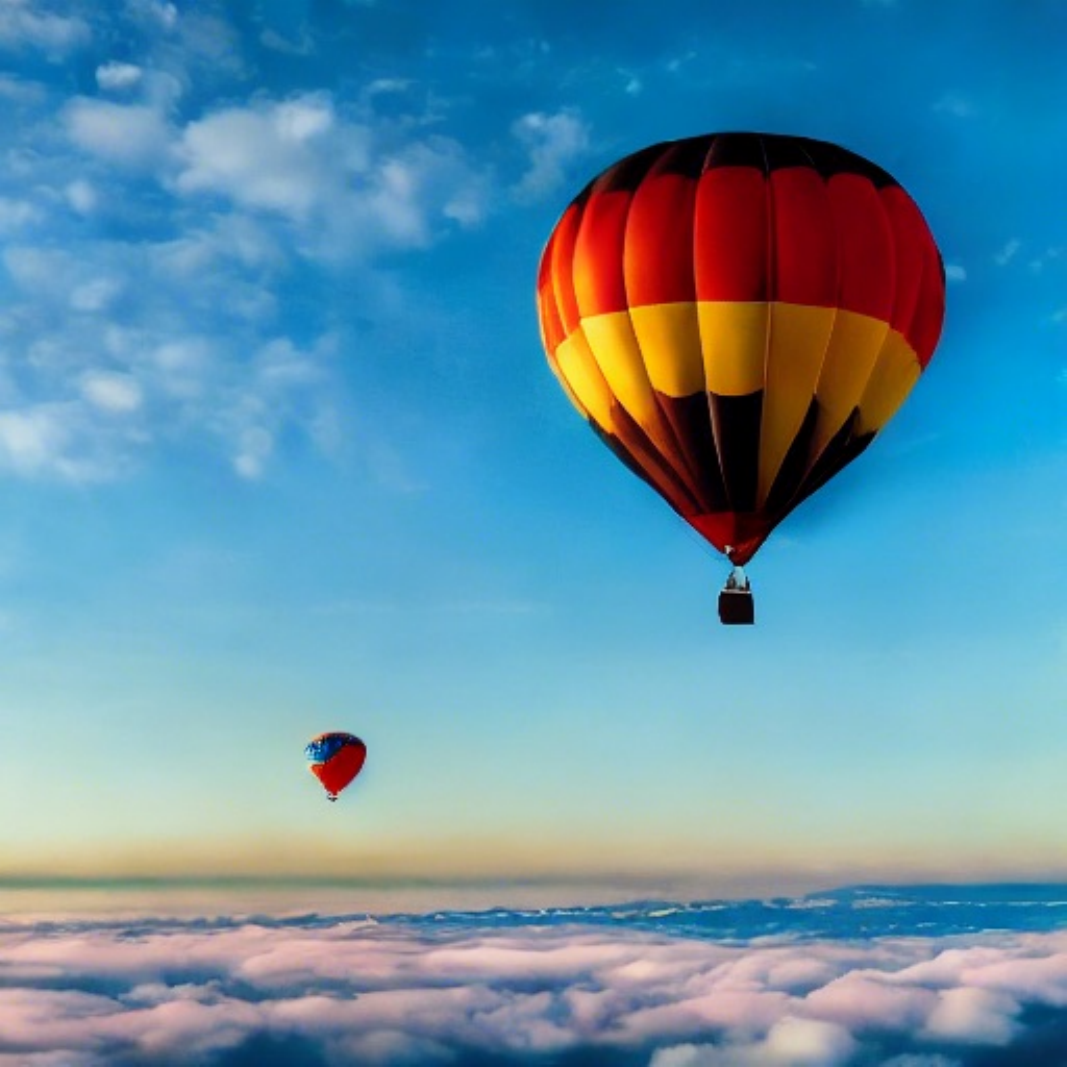} &
      \includegraphics[width=0.18\linewidth]{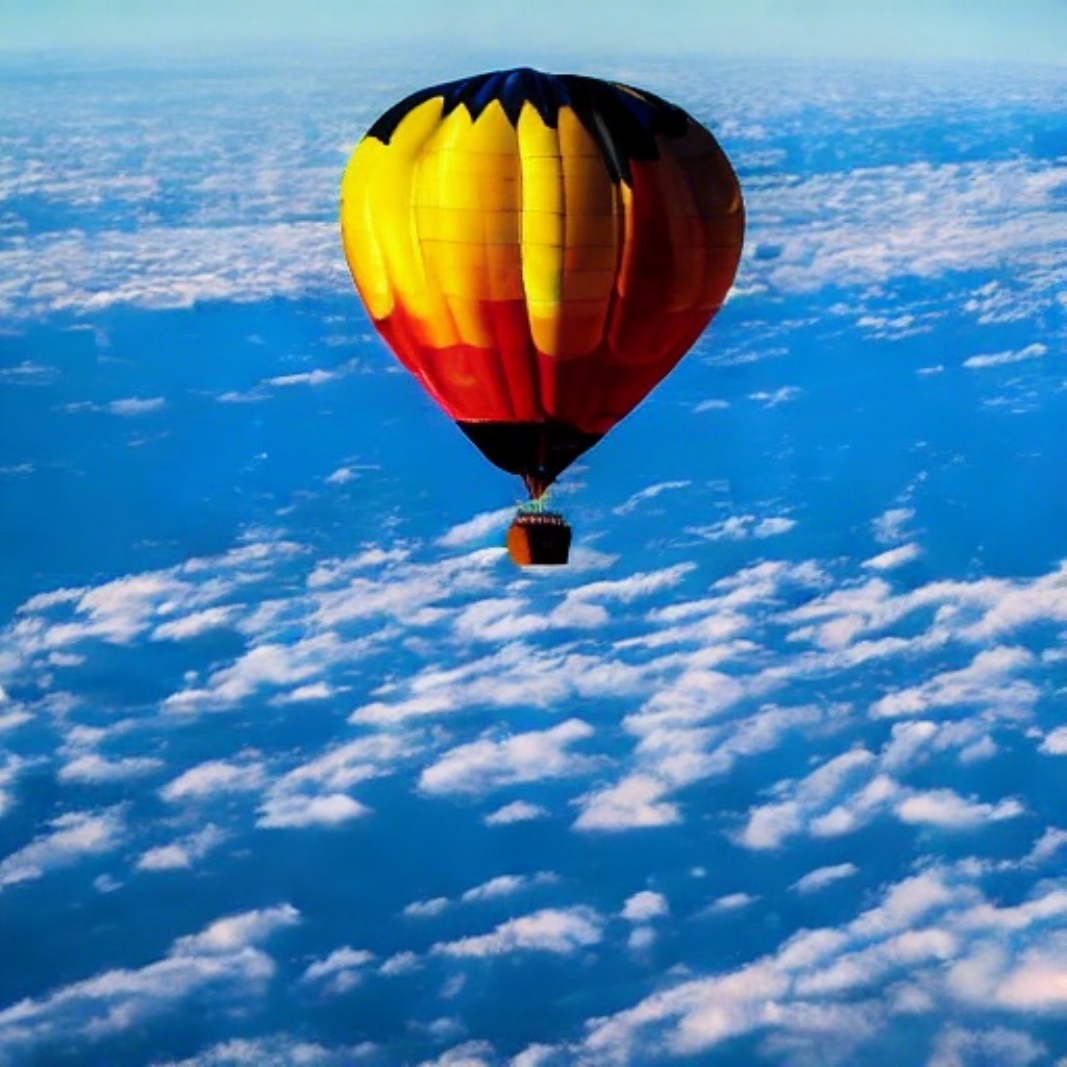} &
      \includegraphics[width=0.18\linewidth]{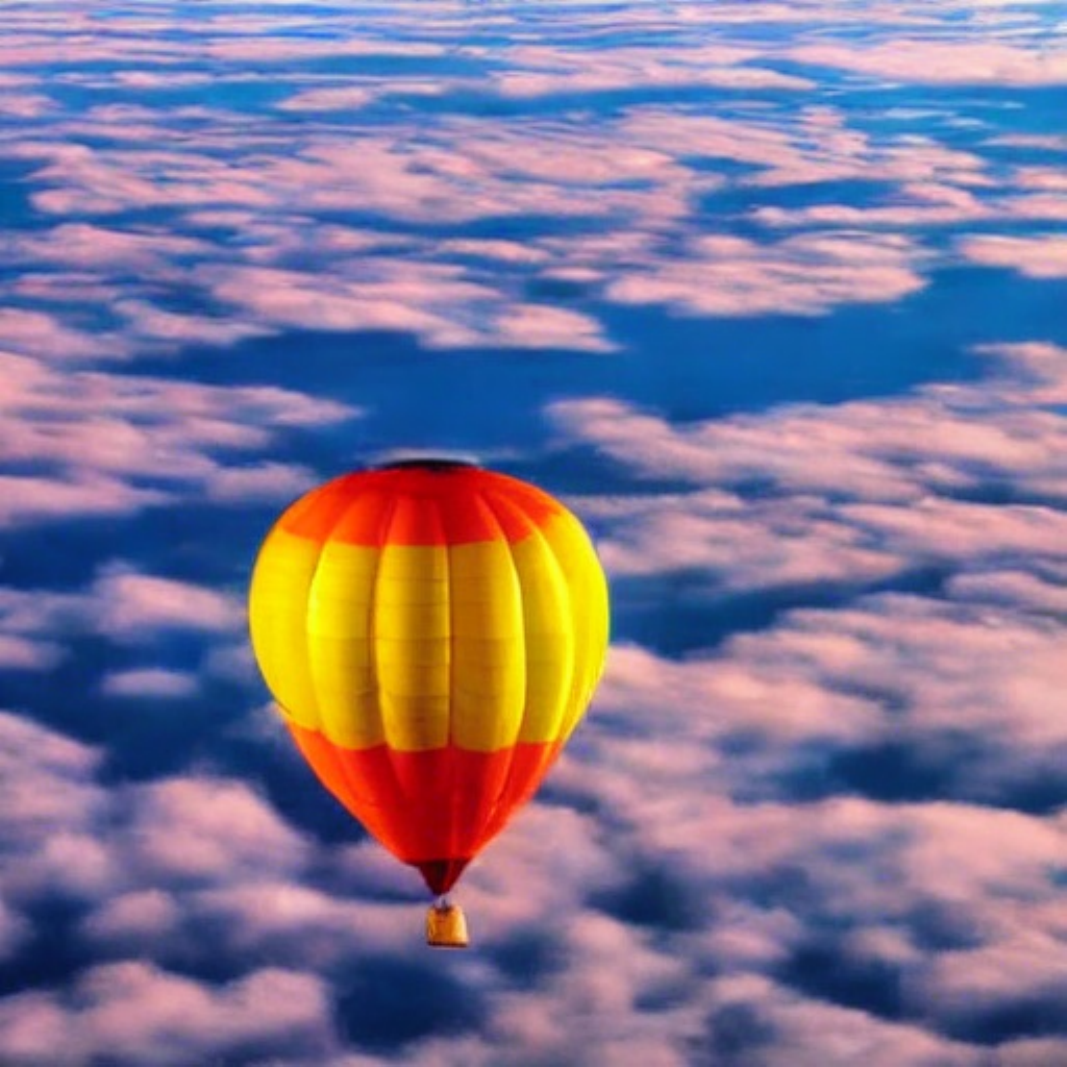}
    \end{tabular}
    \vspace{-2mm}
    \caption{\textit{"A colorful hot air balloon floating in a blue sky above clouds."}}
  \end{subfigure}

  \vspace{5mm}

  \begin{subfigure}[t]{\textwidth}
    \centering
    \begin{tabular}{@{}c@{\hspace{1mm}}c@{\hspace{1mm}}c@{\hspace{1mm}}c@{\hspace{1mm}}c@{}}
      \includegraphics[width=0.18\linewidth]{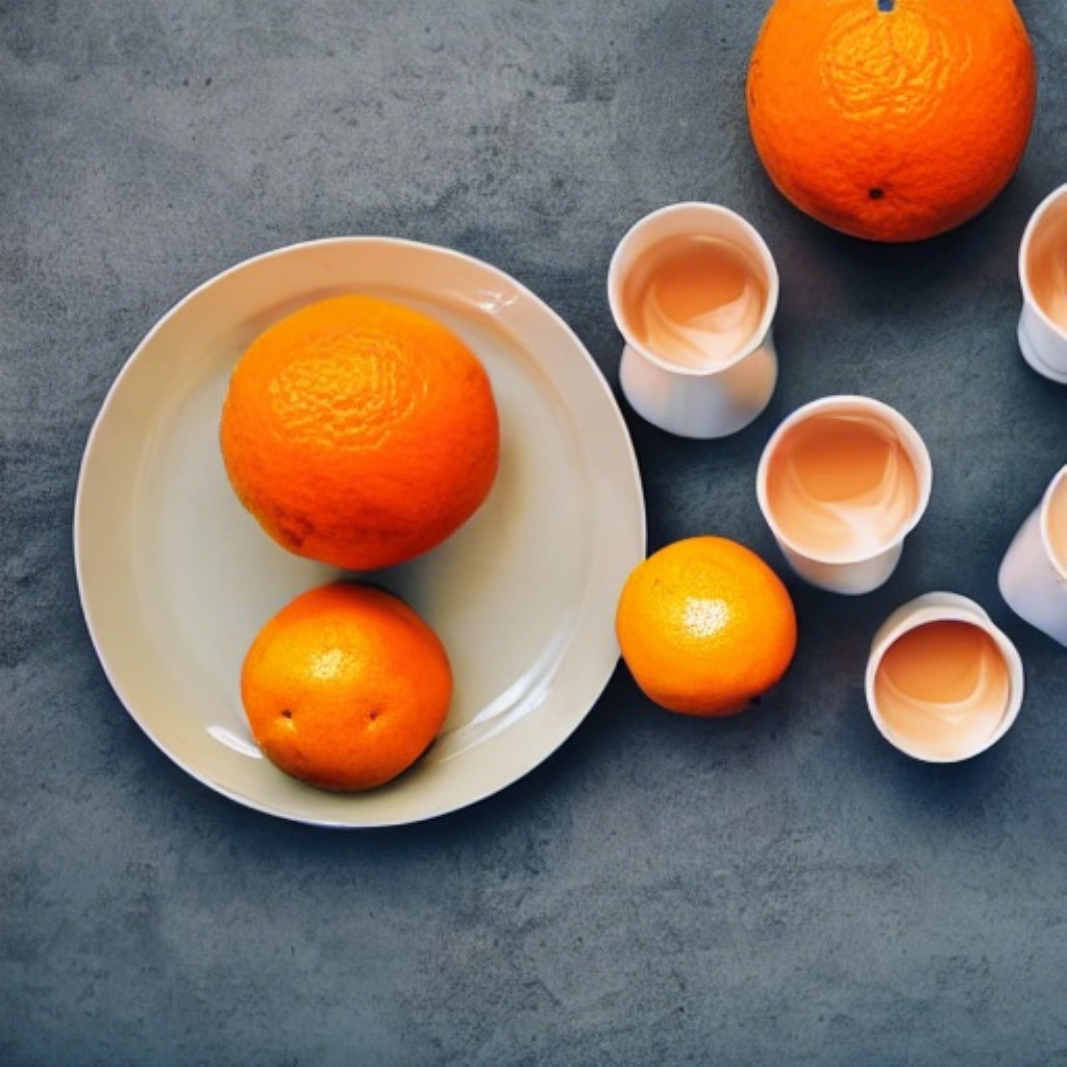} &
      \includegraphics[width=0.18\linewidth]{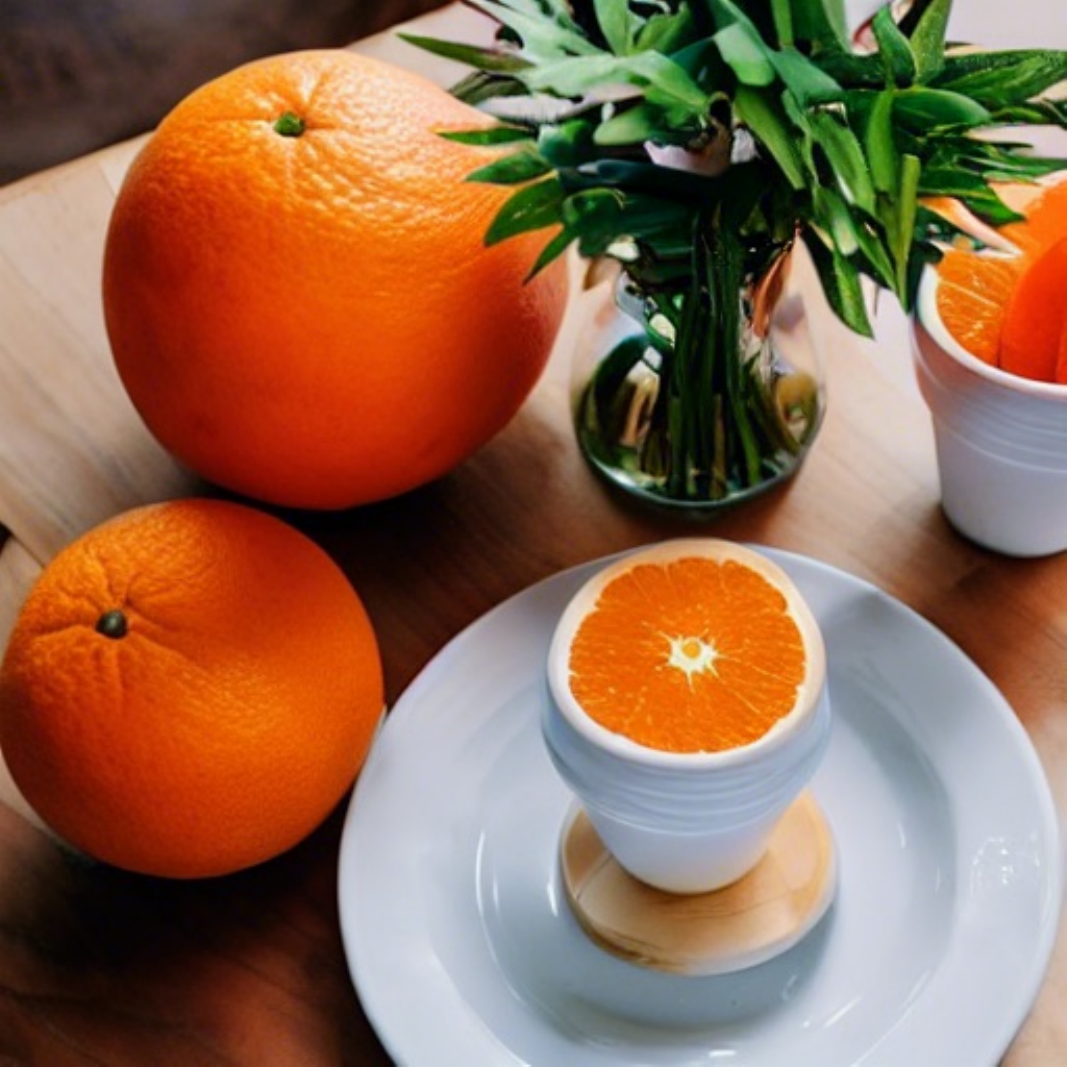} &
      \includegraphics[width=0.18\linewidth]{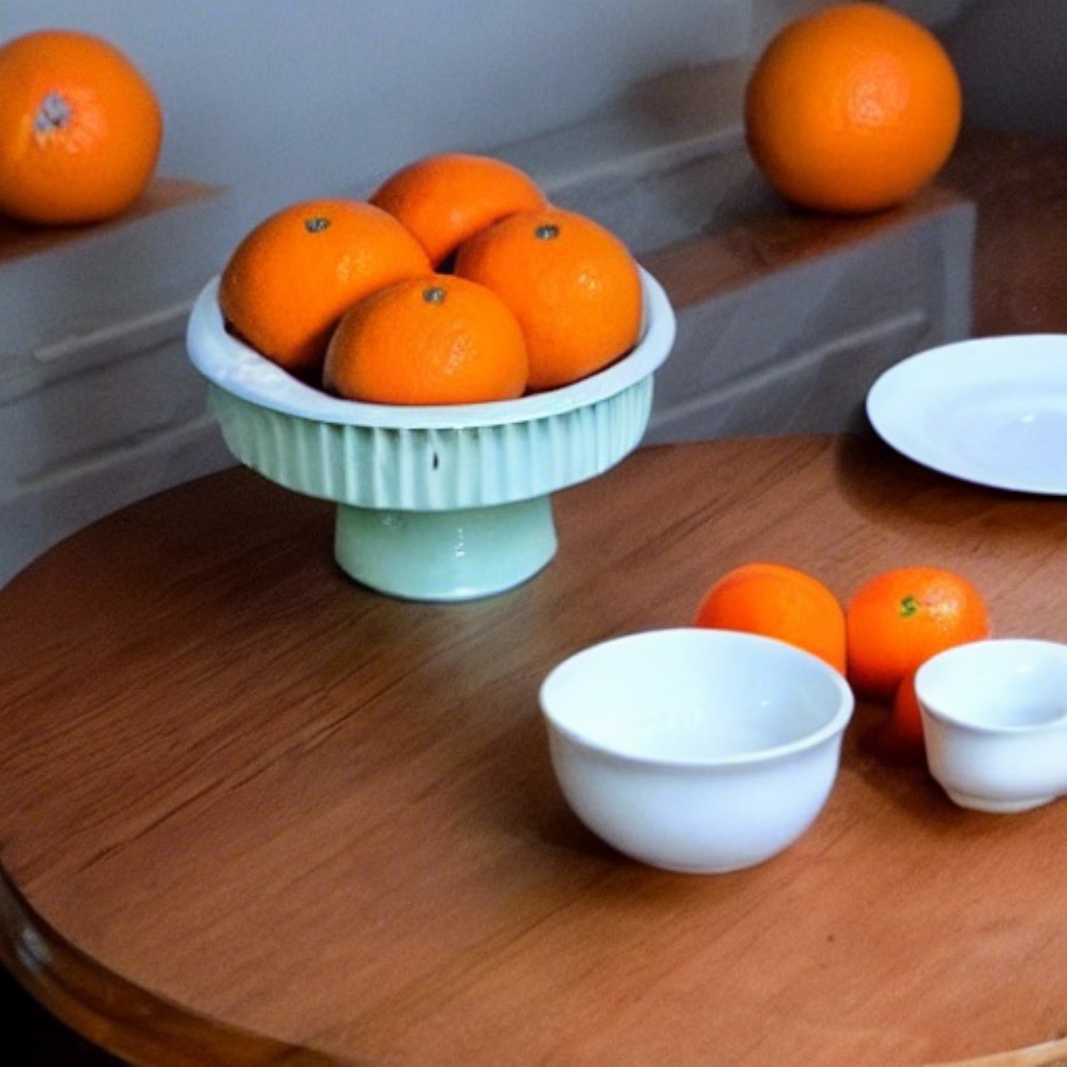} &
      \includegraphics[width=0.18\linewidth]{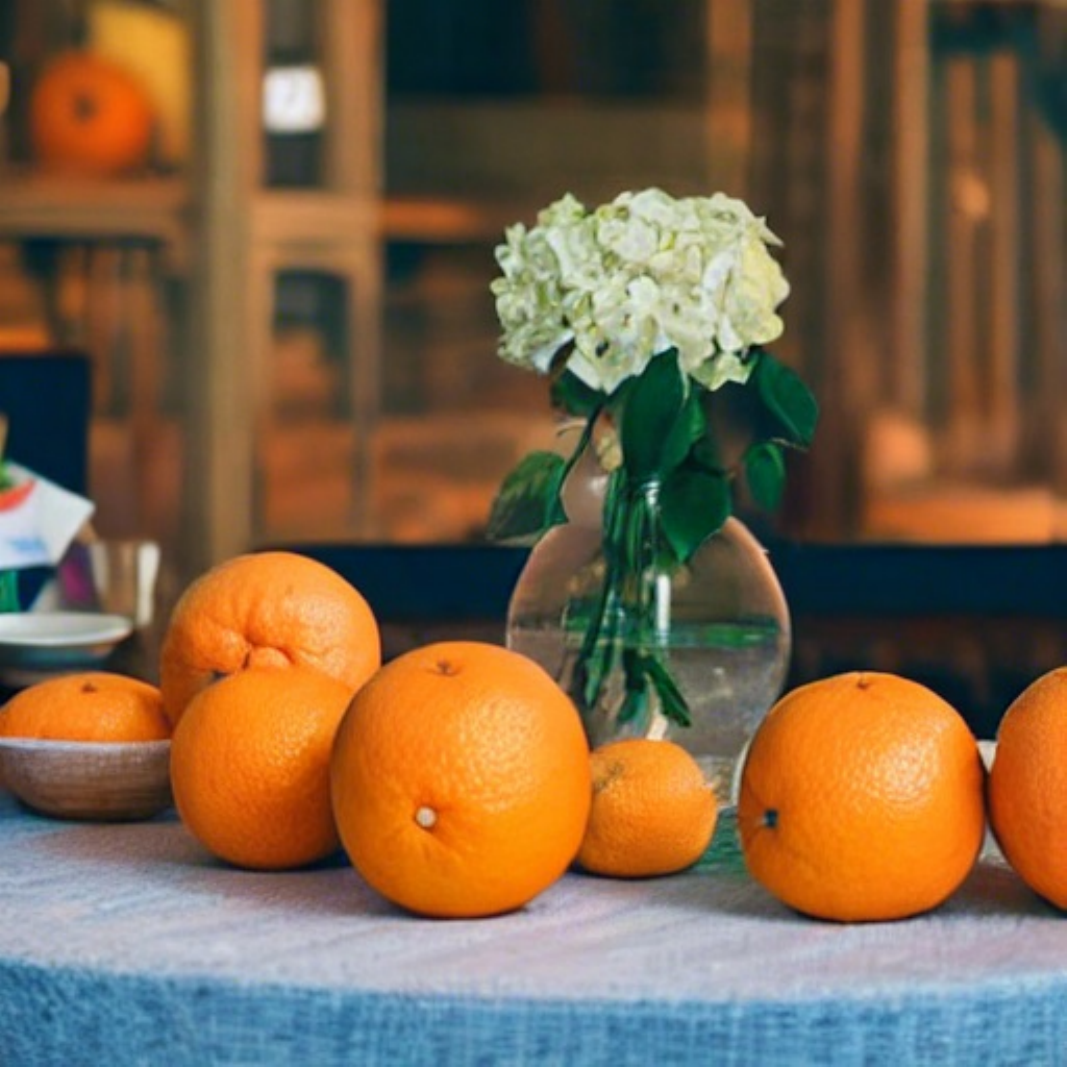} &
      \includegraphics[width=0.18\linewidth]{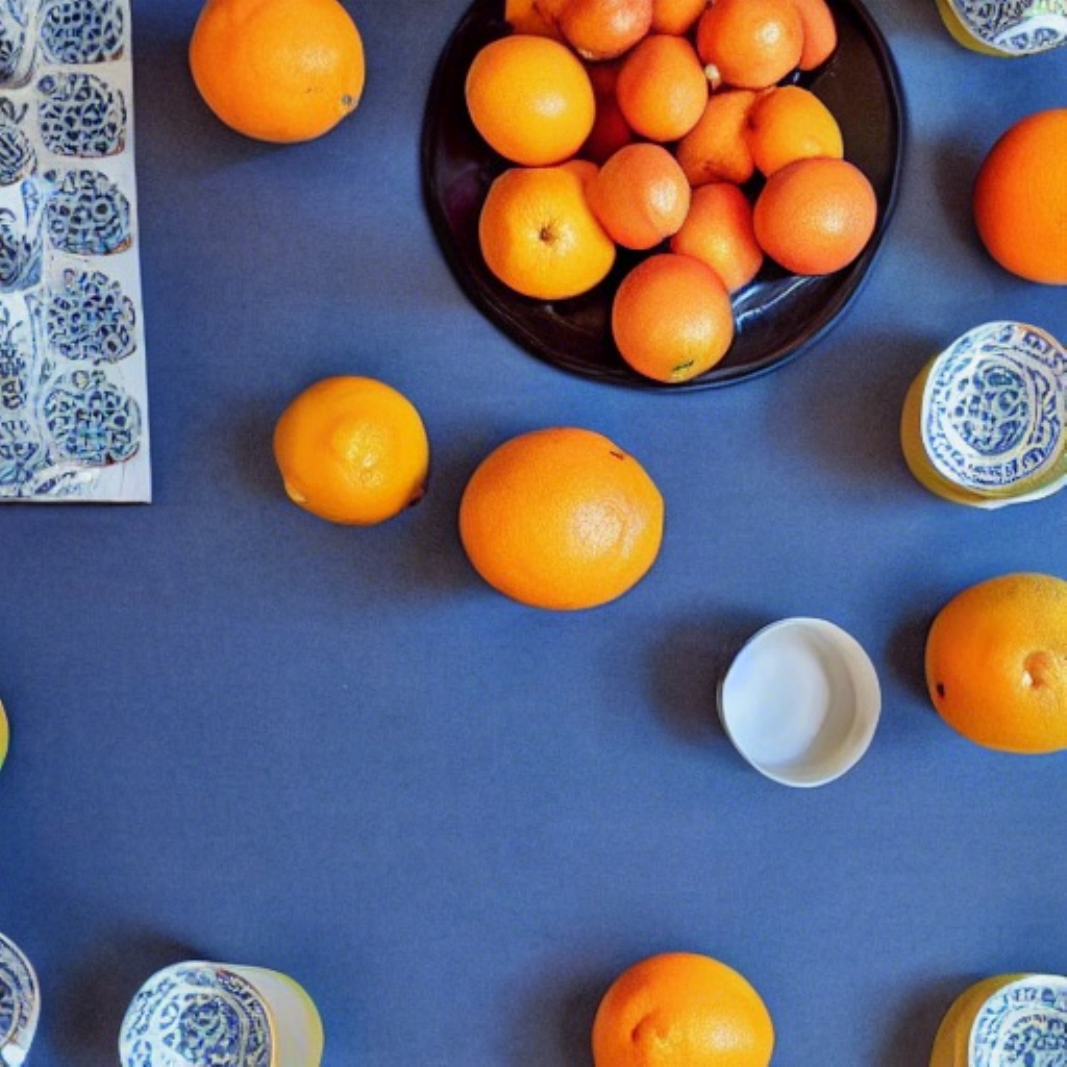} \\[-1pt]
      \includegraphics[width=0.18\linewidth]{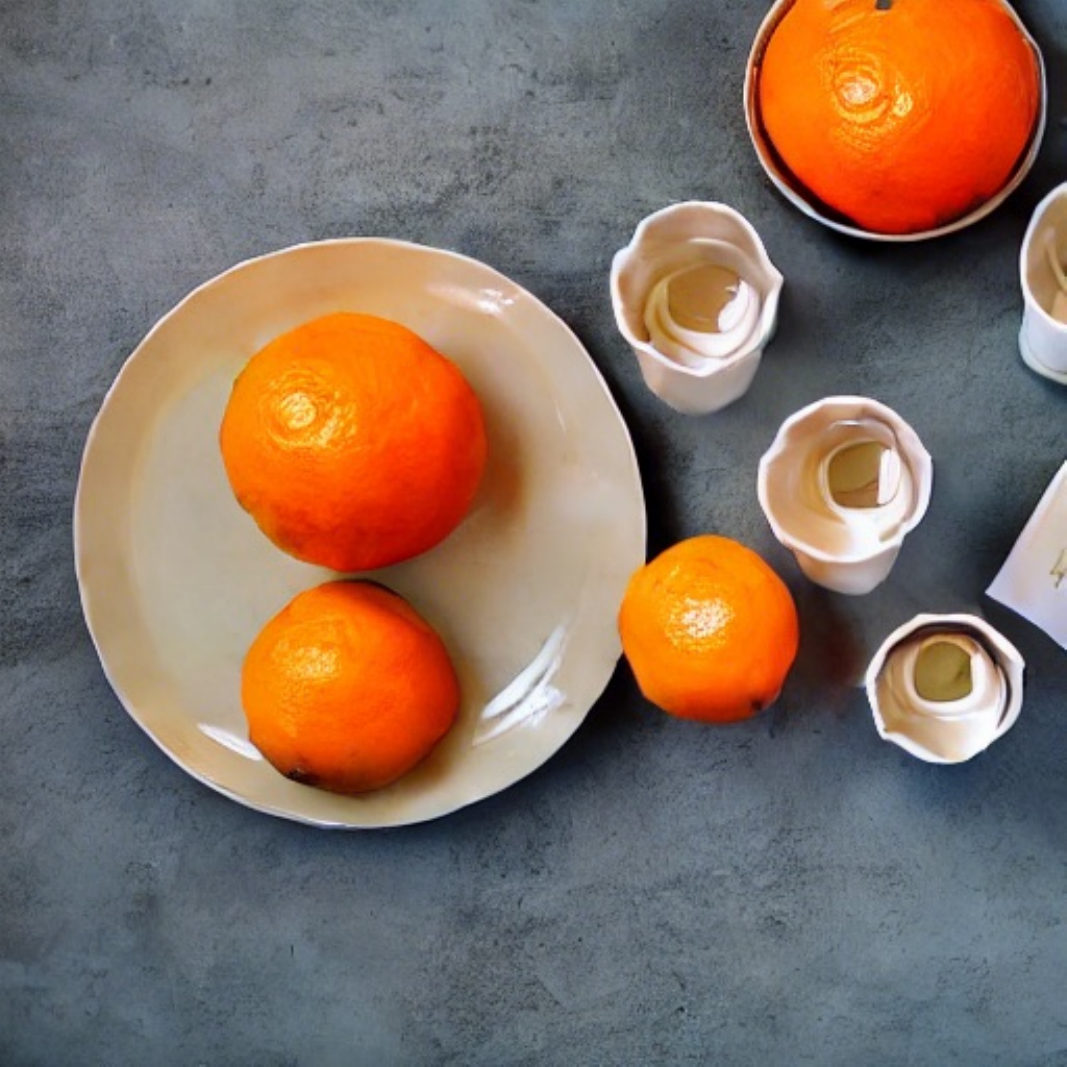} &
      \includegraphics[width=0.18\linewidth]{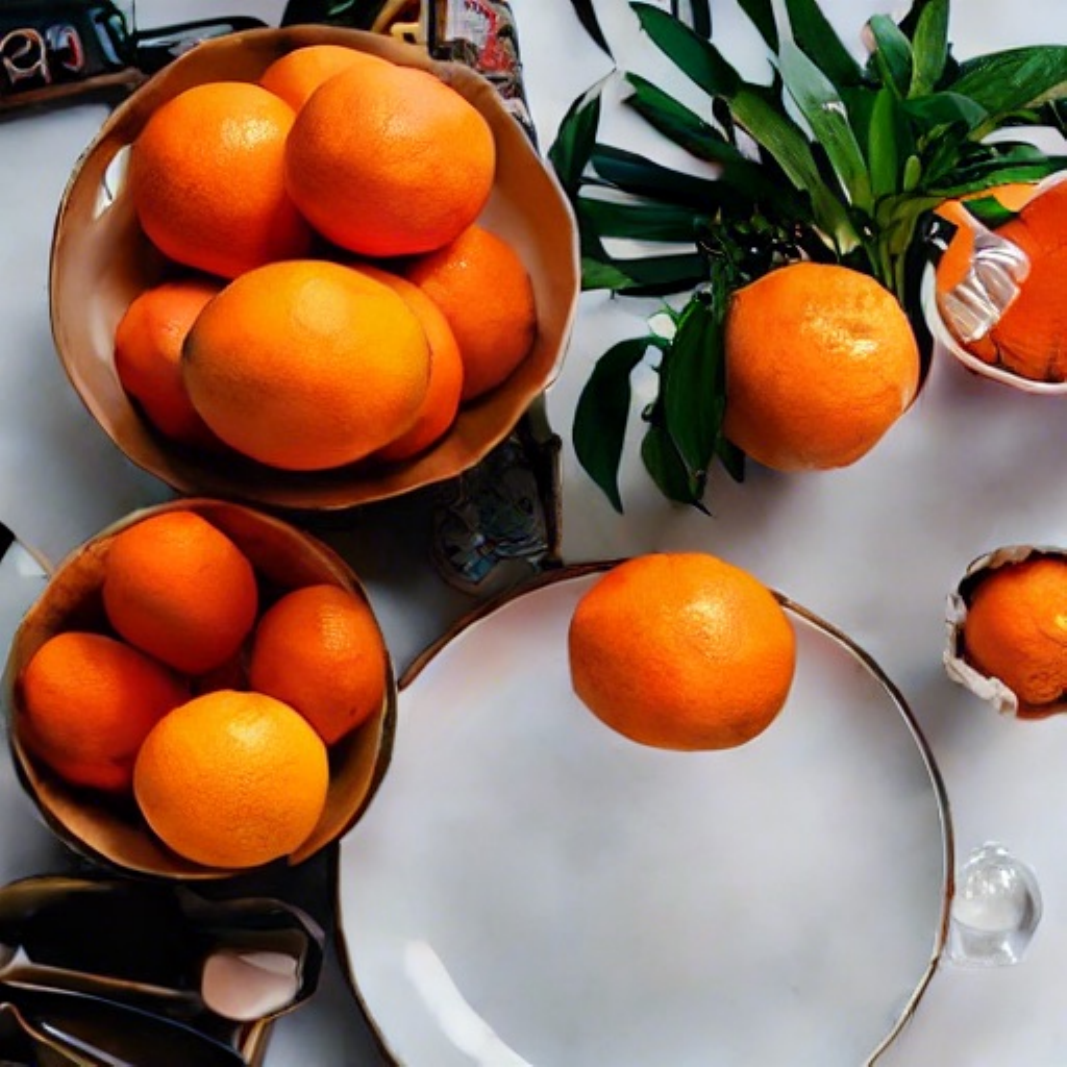} &
      \includegraphics[width=0.18\linewidth]{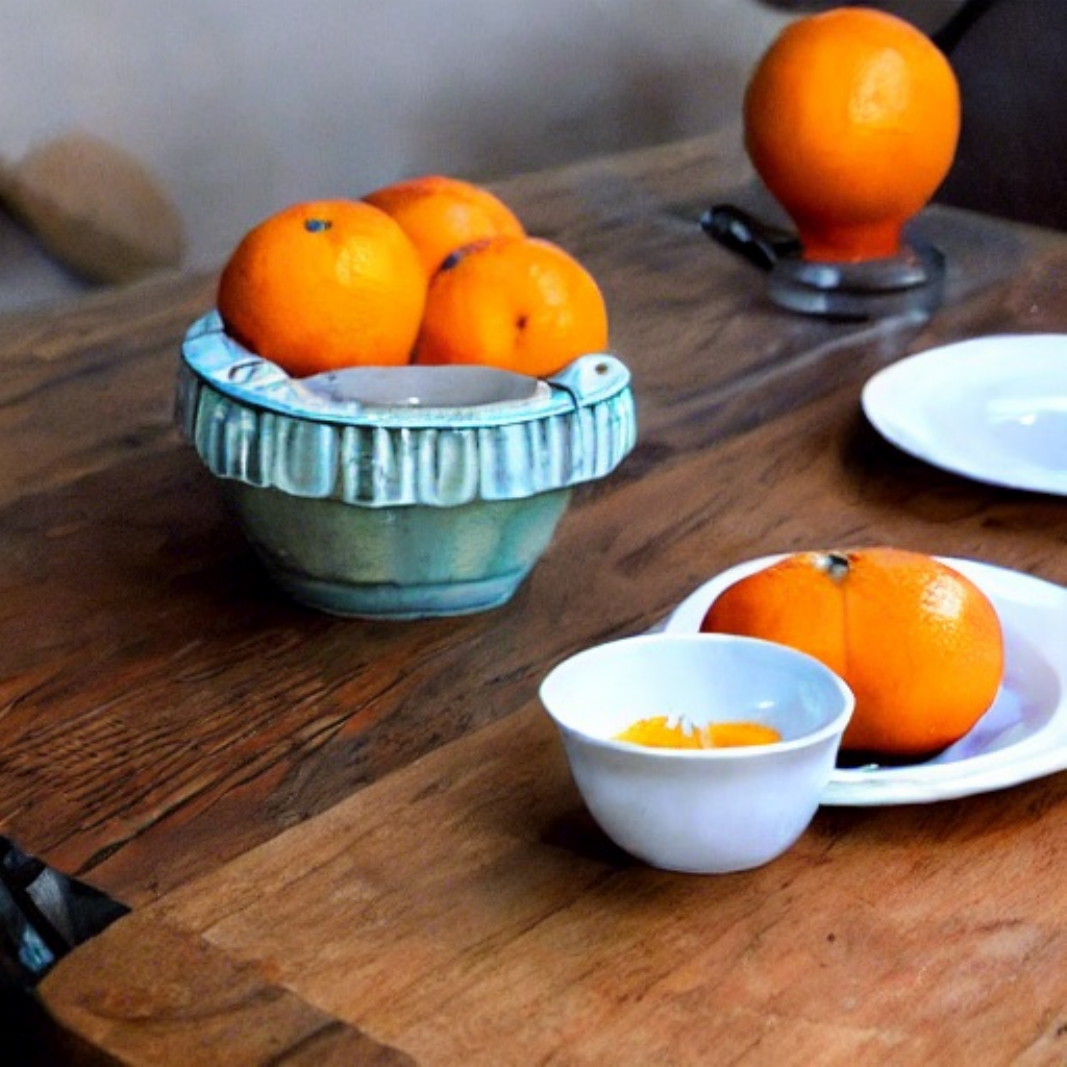} &
      \includegraphics[width=0.18\linewidth]{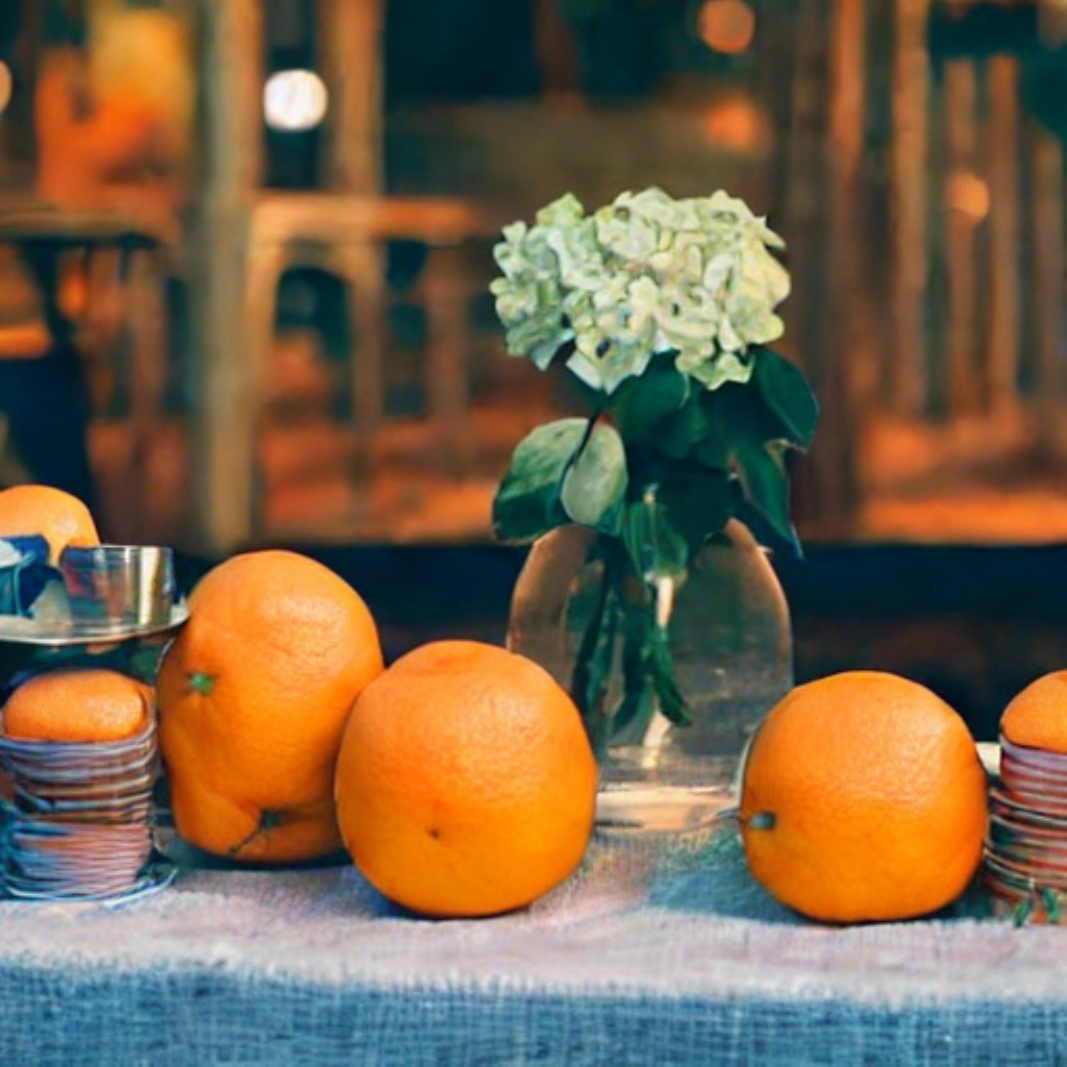} &
      \includegraphics[width=0.18\linewidth]{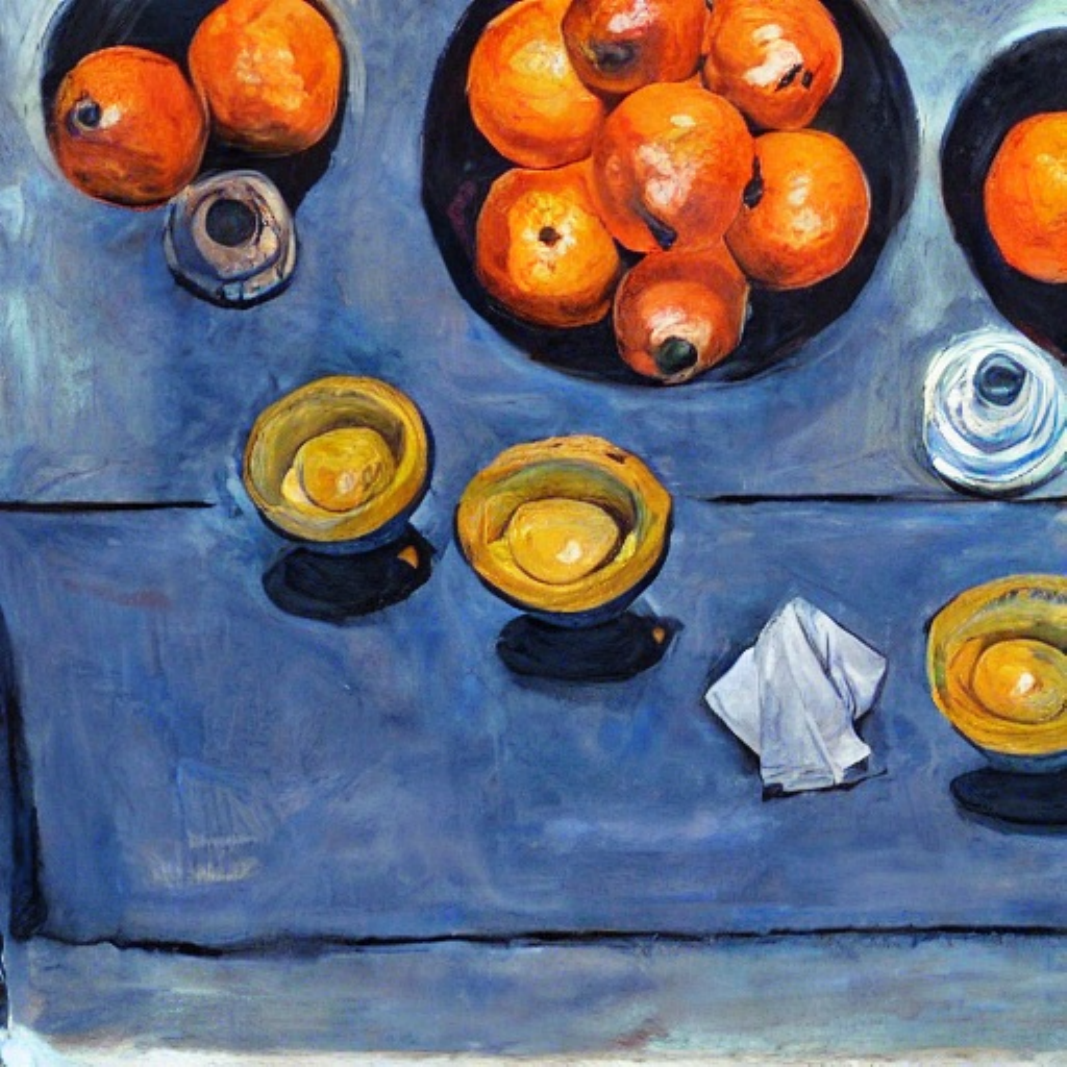} \\[-1pt]
      \includegraphics[width=0.18\linewidth]{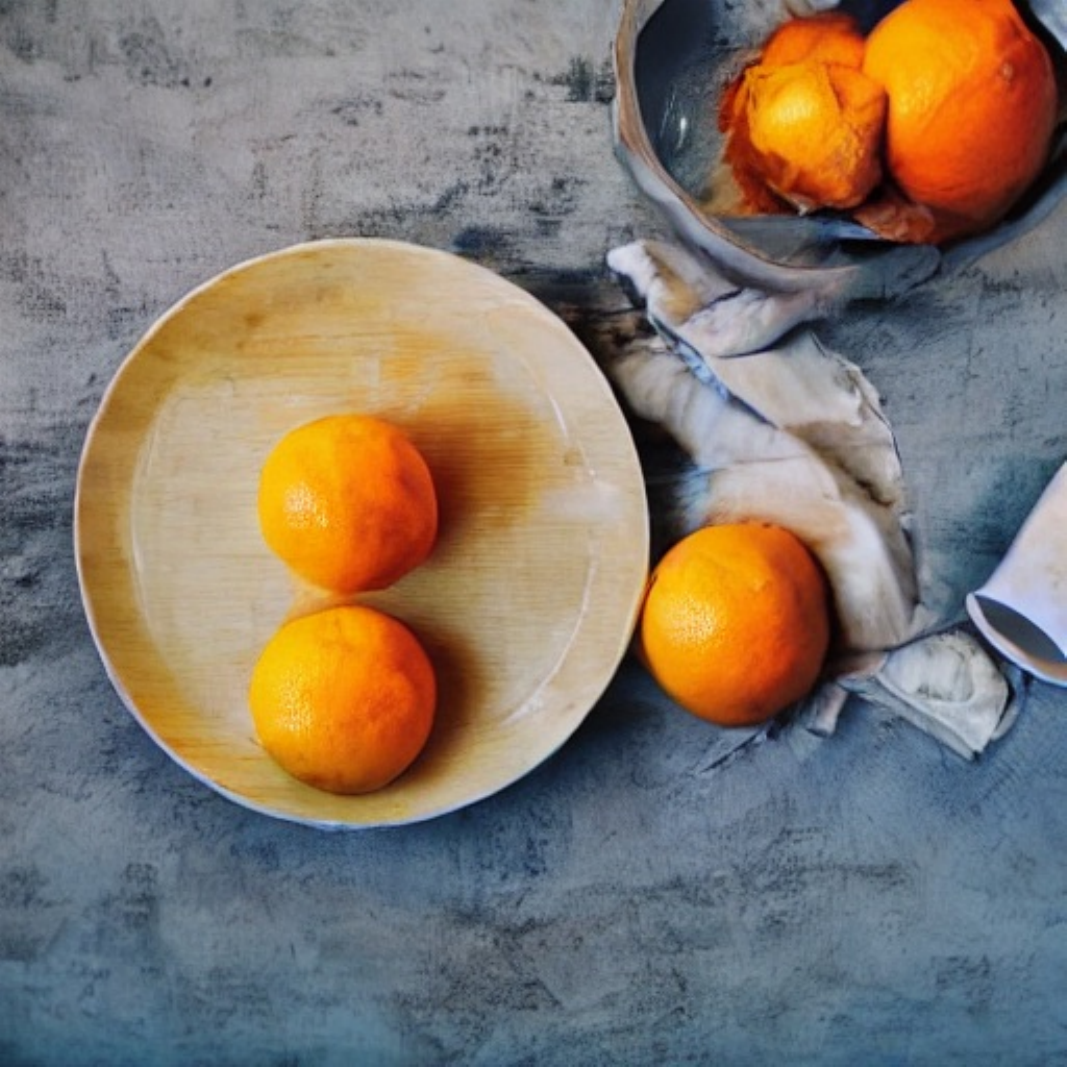} &
      \includegraphics[width=0.18\linewidth]{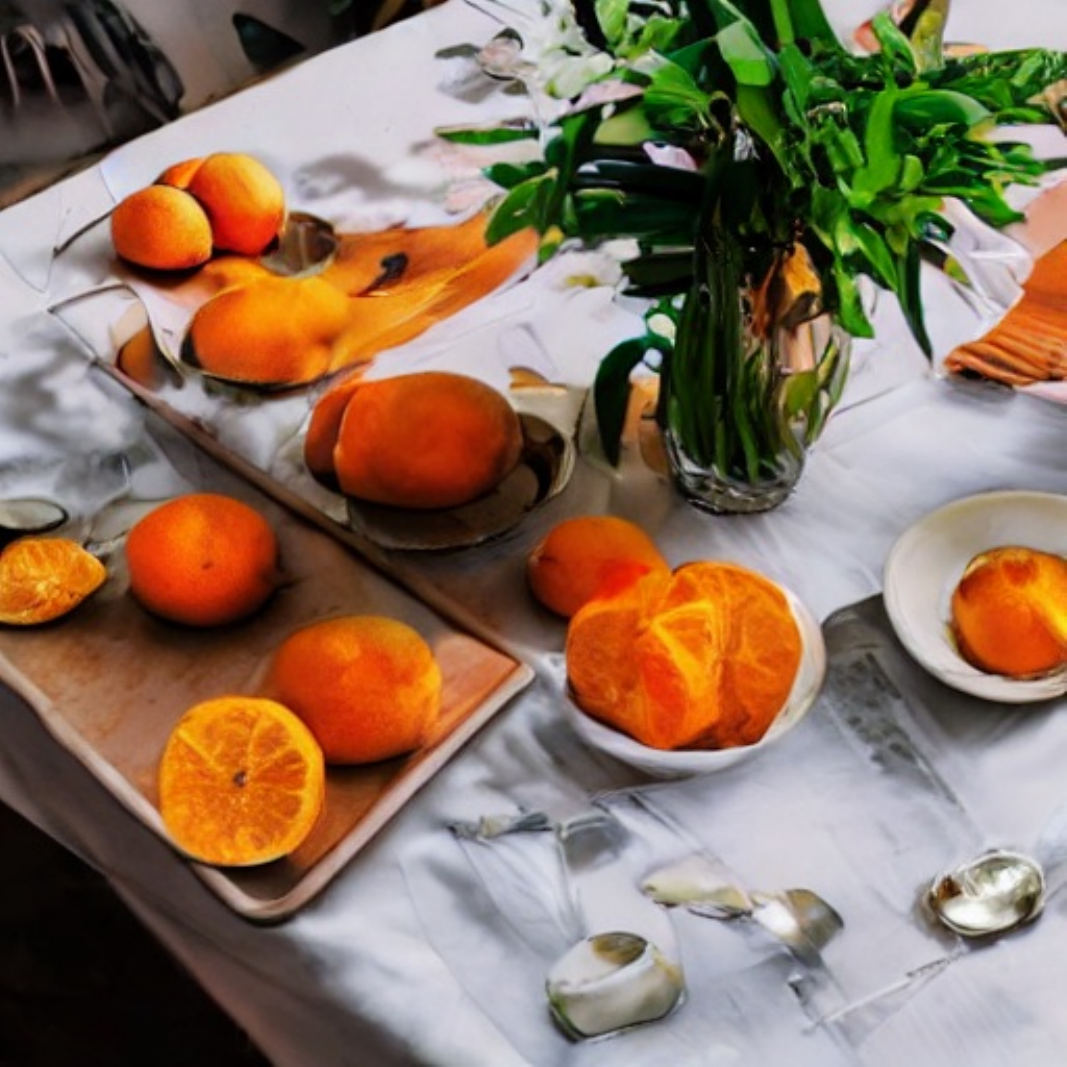} &
      \includegraphics[width=0.18\linewidth]{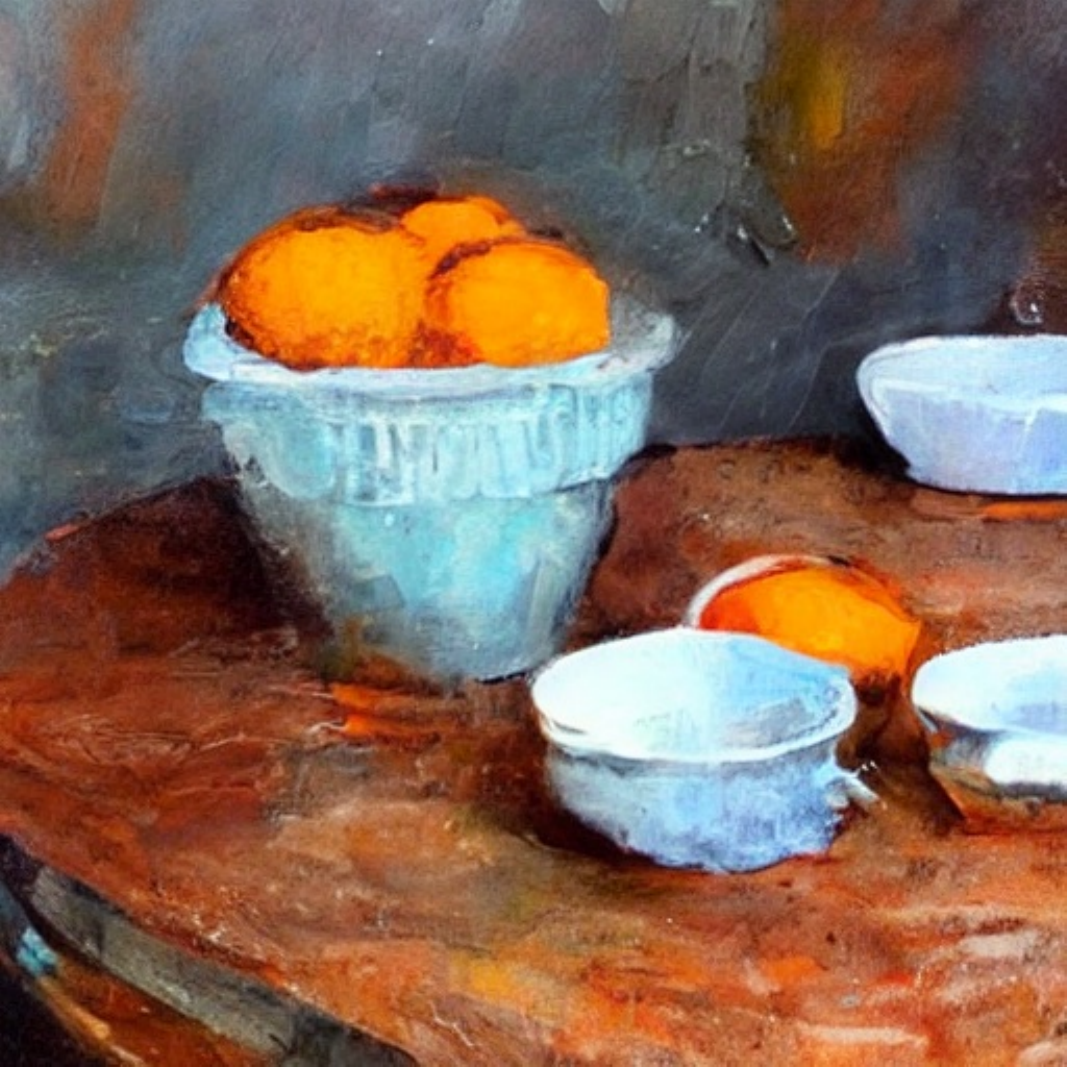} &
      \includegraphics[width=0.18\linewidth]{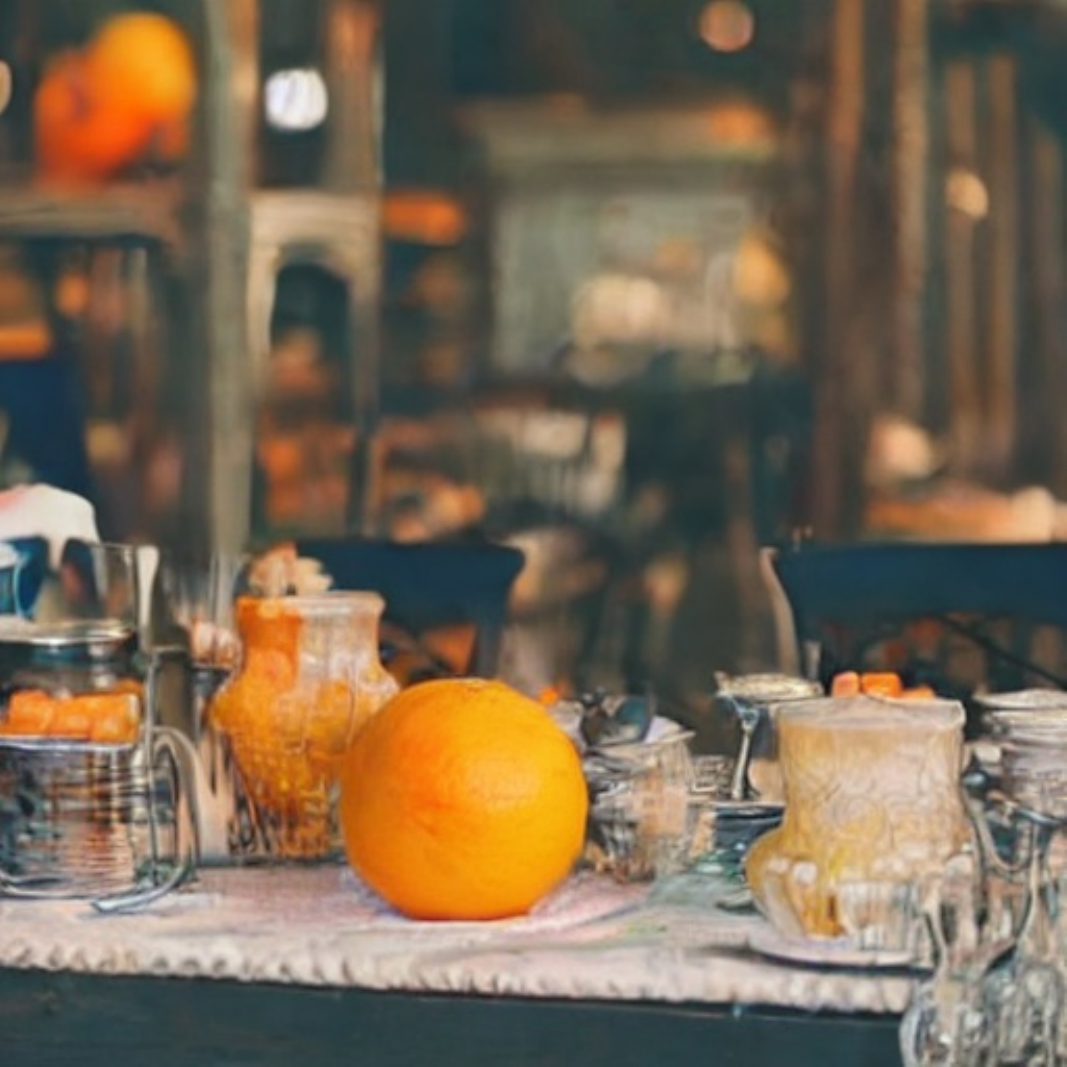} &
      \includegraphics[width=0.18\linewidth]{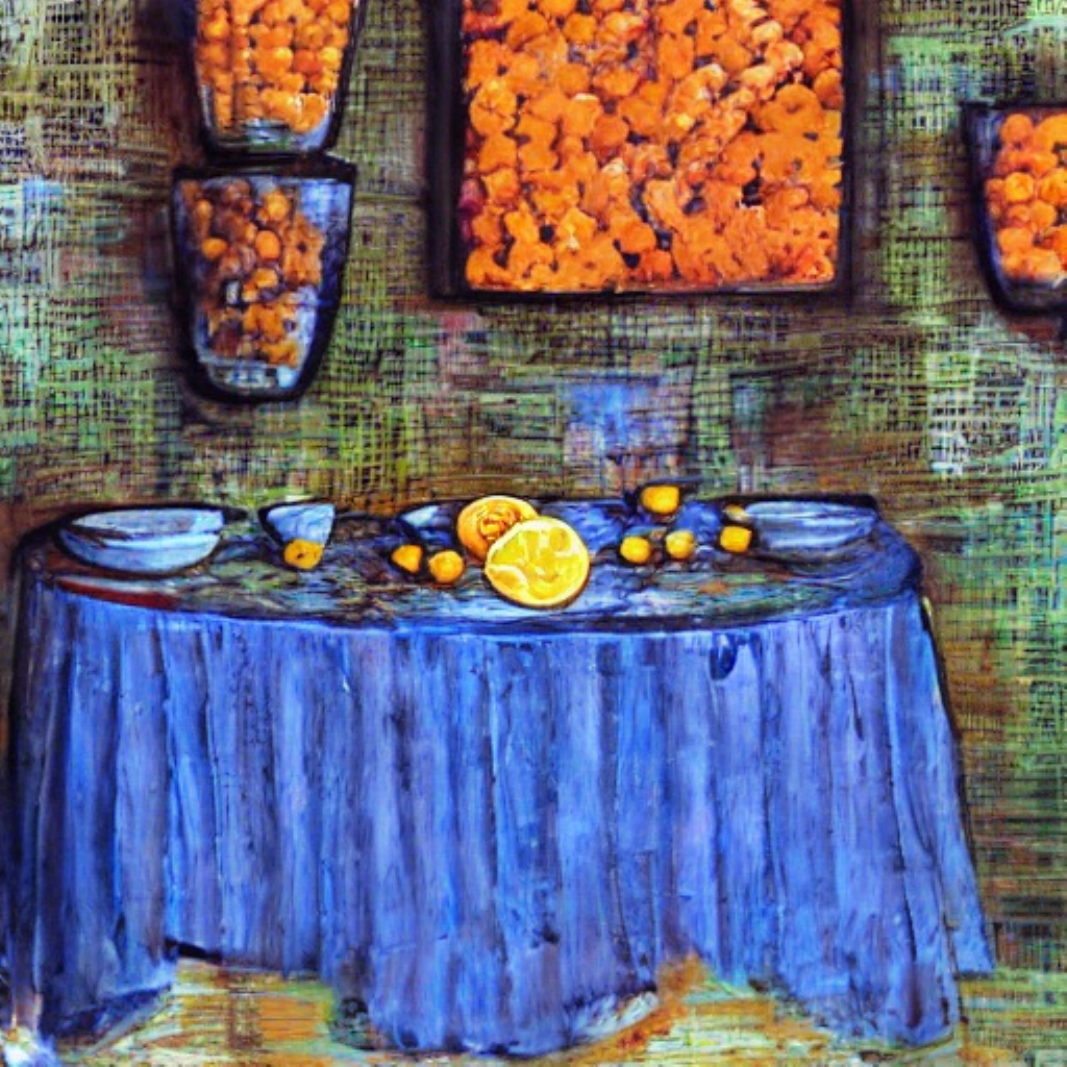} \\[-1pt]
      \includegraphics[width=0.18\linewidth]{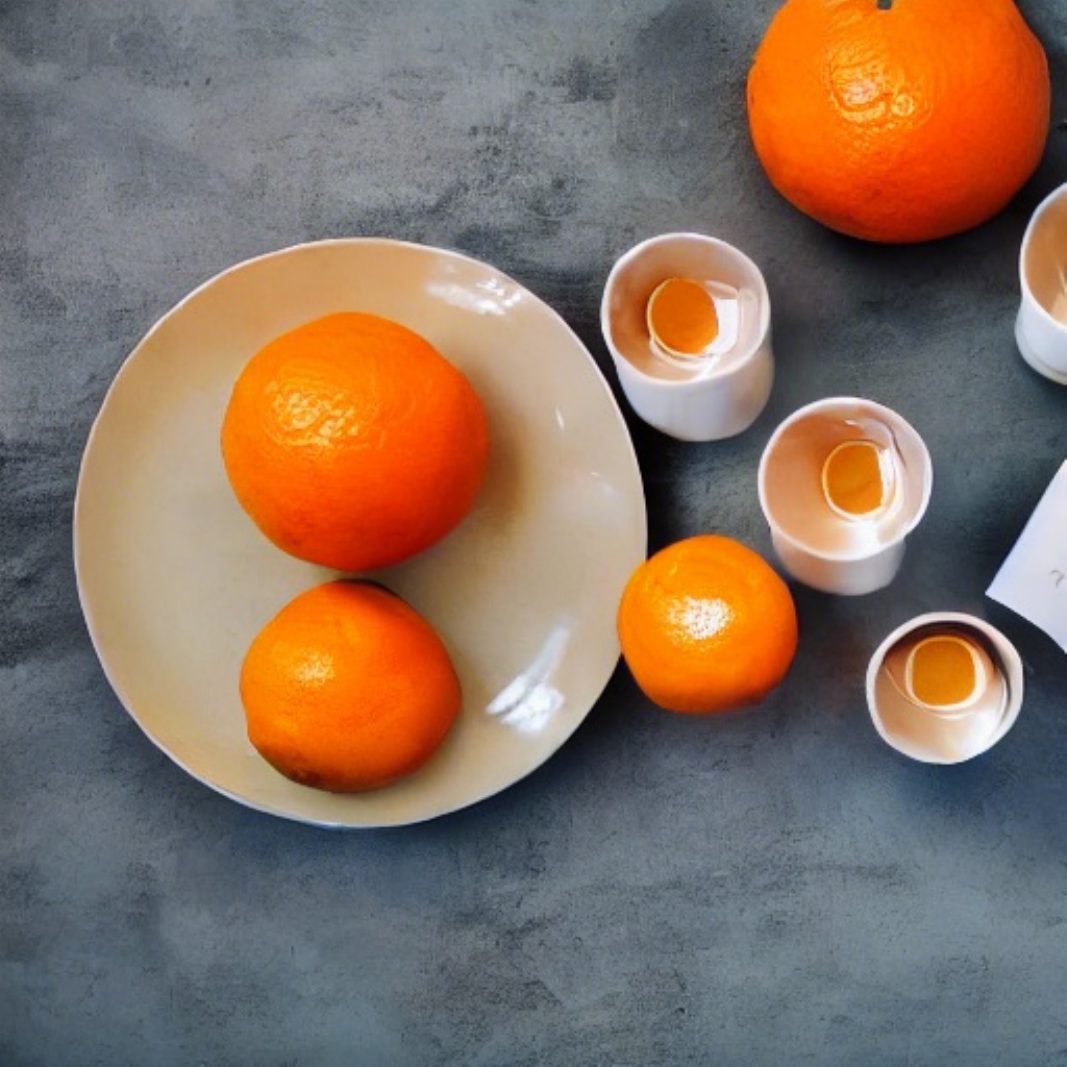} &
      \includegraphics[width=0.18\linewidth]{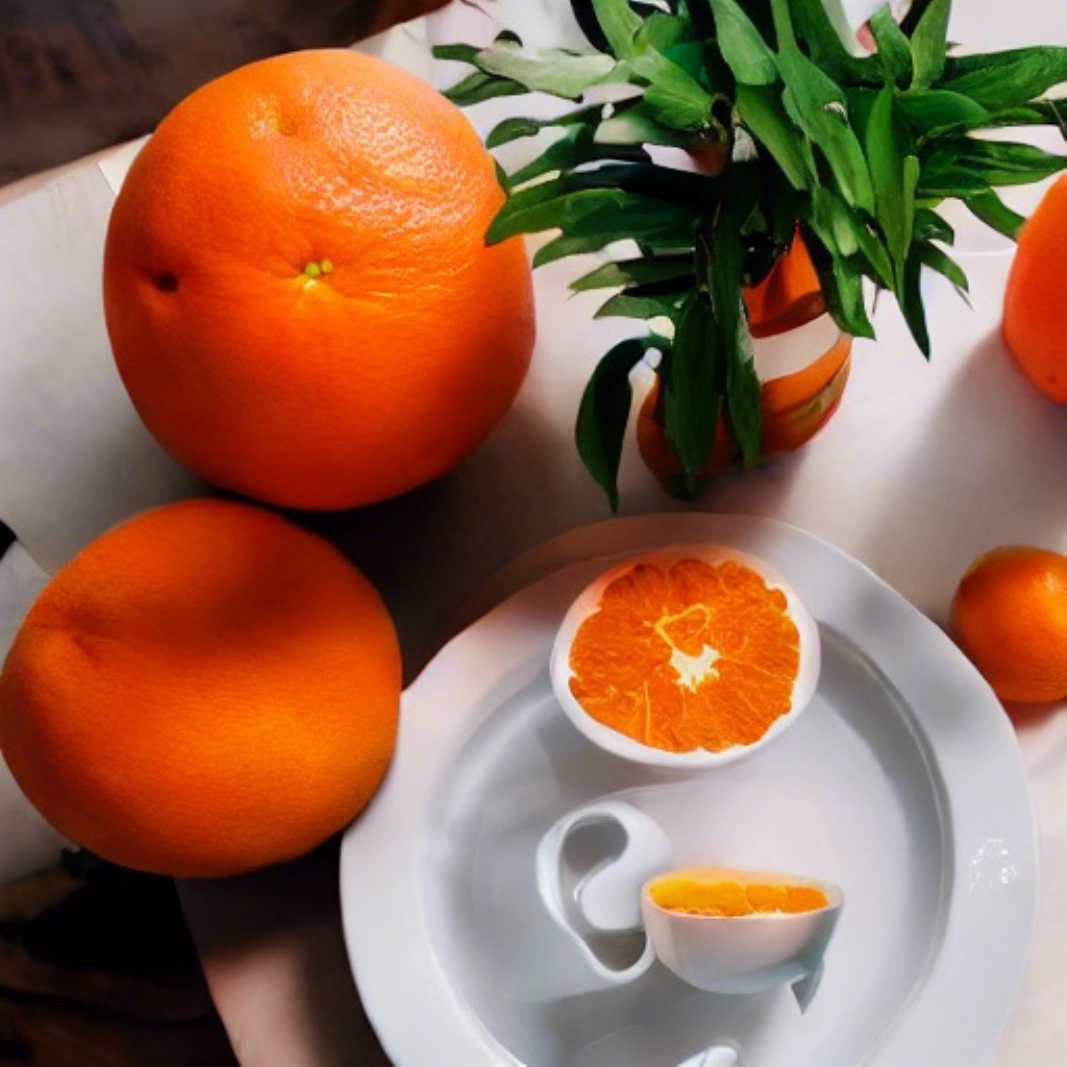} &
      \includegraphics[width=0.18\linewidth]{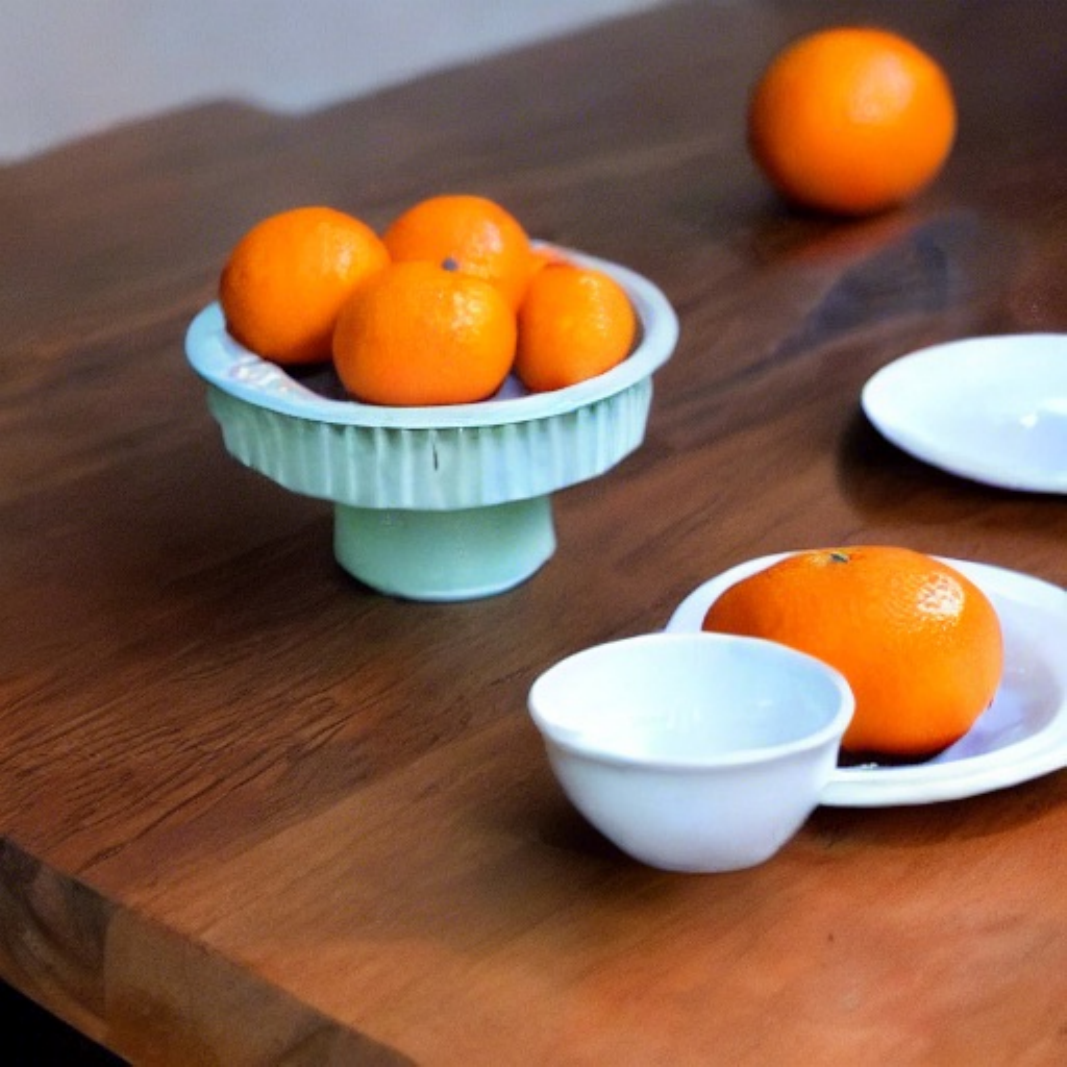} &
      \includegraphics[width=0.18\linewidth]{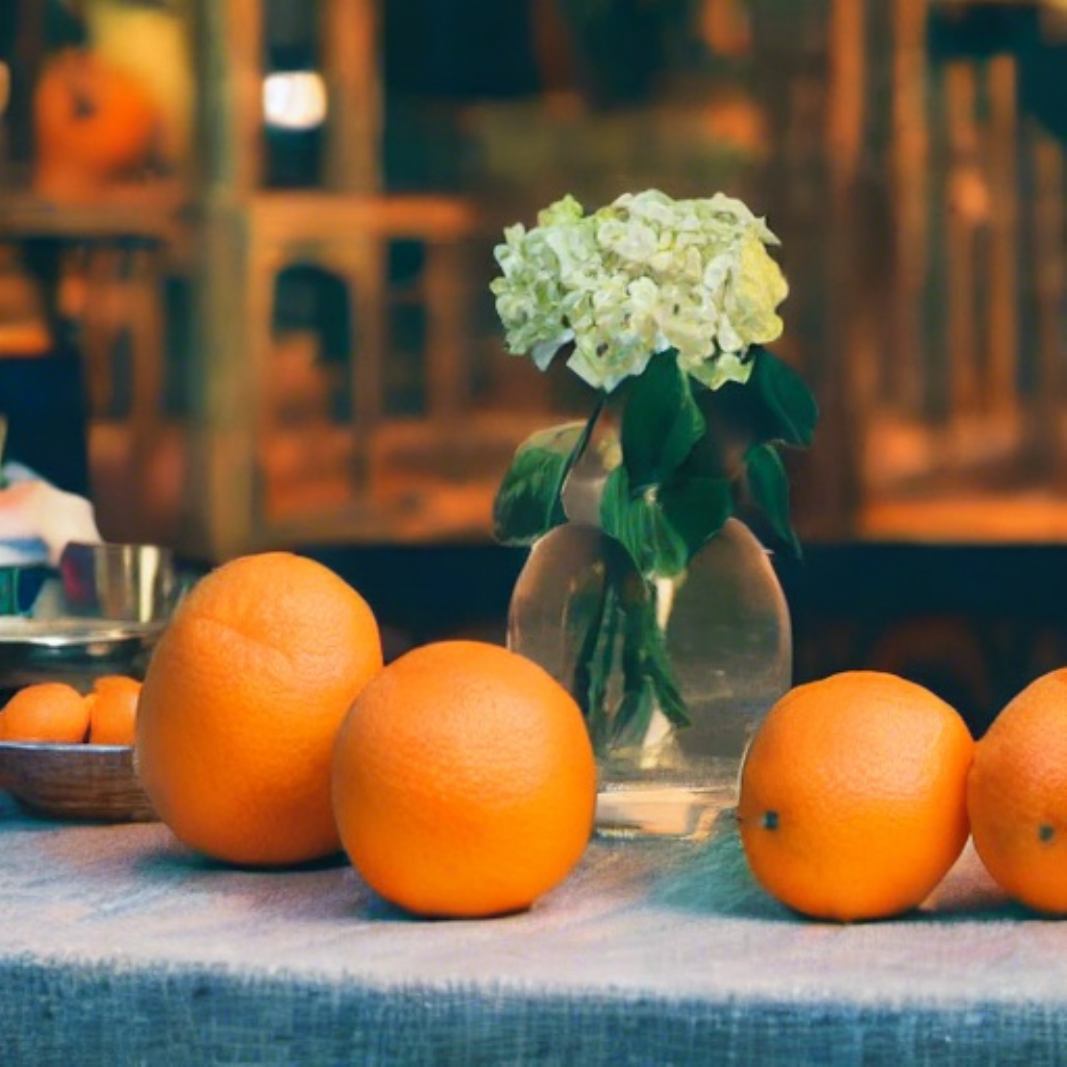} &
      \includegraphics[width=0.18\linewidth]{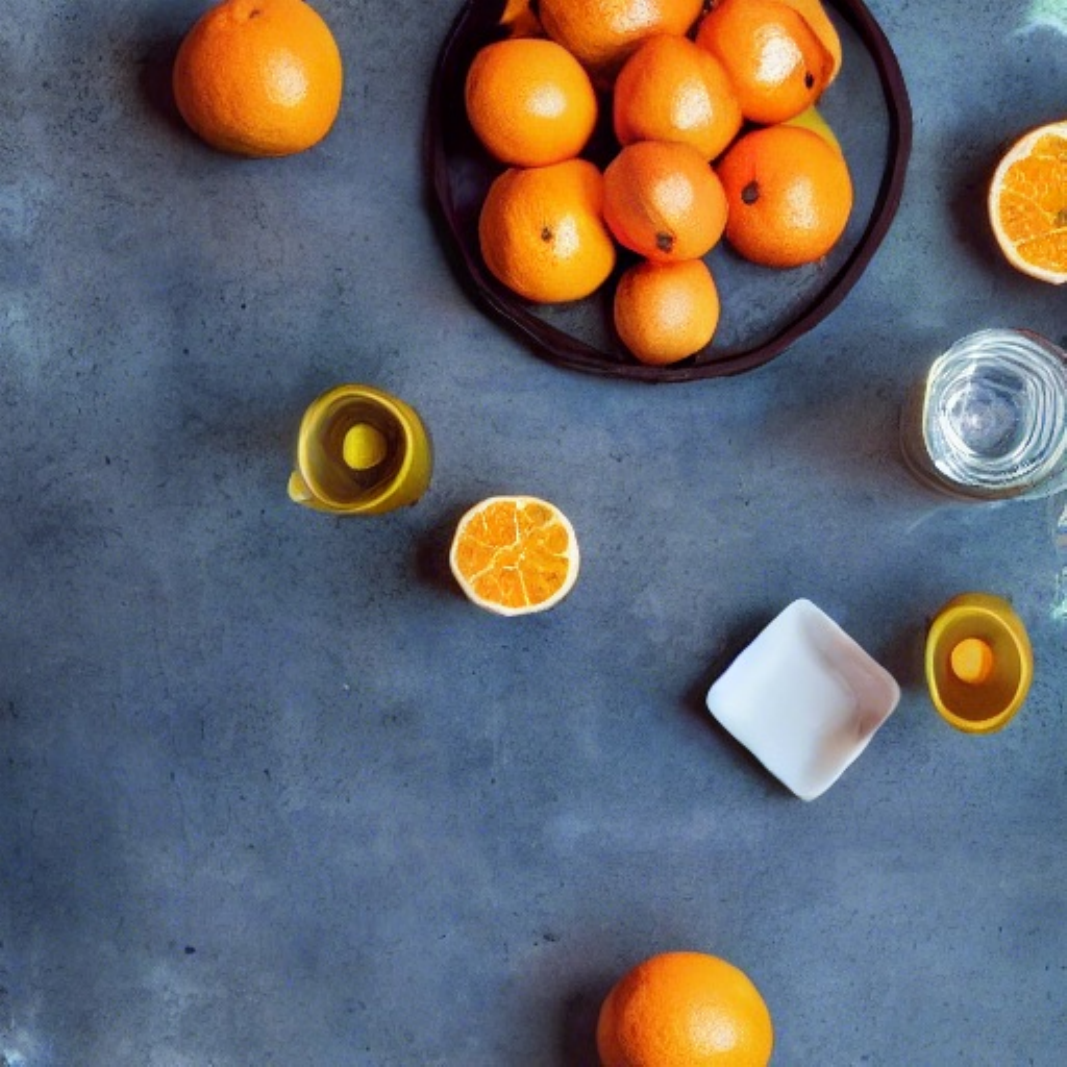}
    \end{tabular}
    \vspace{-2mm}
    \caption{\textit{"A table with some oranges and some cups."}}
  \end{subfigure}

  \vspace{2mm}
  \label{fig:appendix_instance_wise}
\end{figure*}

\begin{figure*}[p]\ContinuedFloat
  \centering
  \captionsetup[subfigure]{font=small,labelformat=empty,justification=centering}

  \begin{subfigure}[t]{\textwidth}
    \centering
    \begin{tabular}{@{}c@{\hspace{1mm}}c@{\hspace{1mm}}c@{\hspace{1mm}}c@{\hspace{1mm}}c@{}}
      \includegraphics[width=0.18\linewidth]{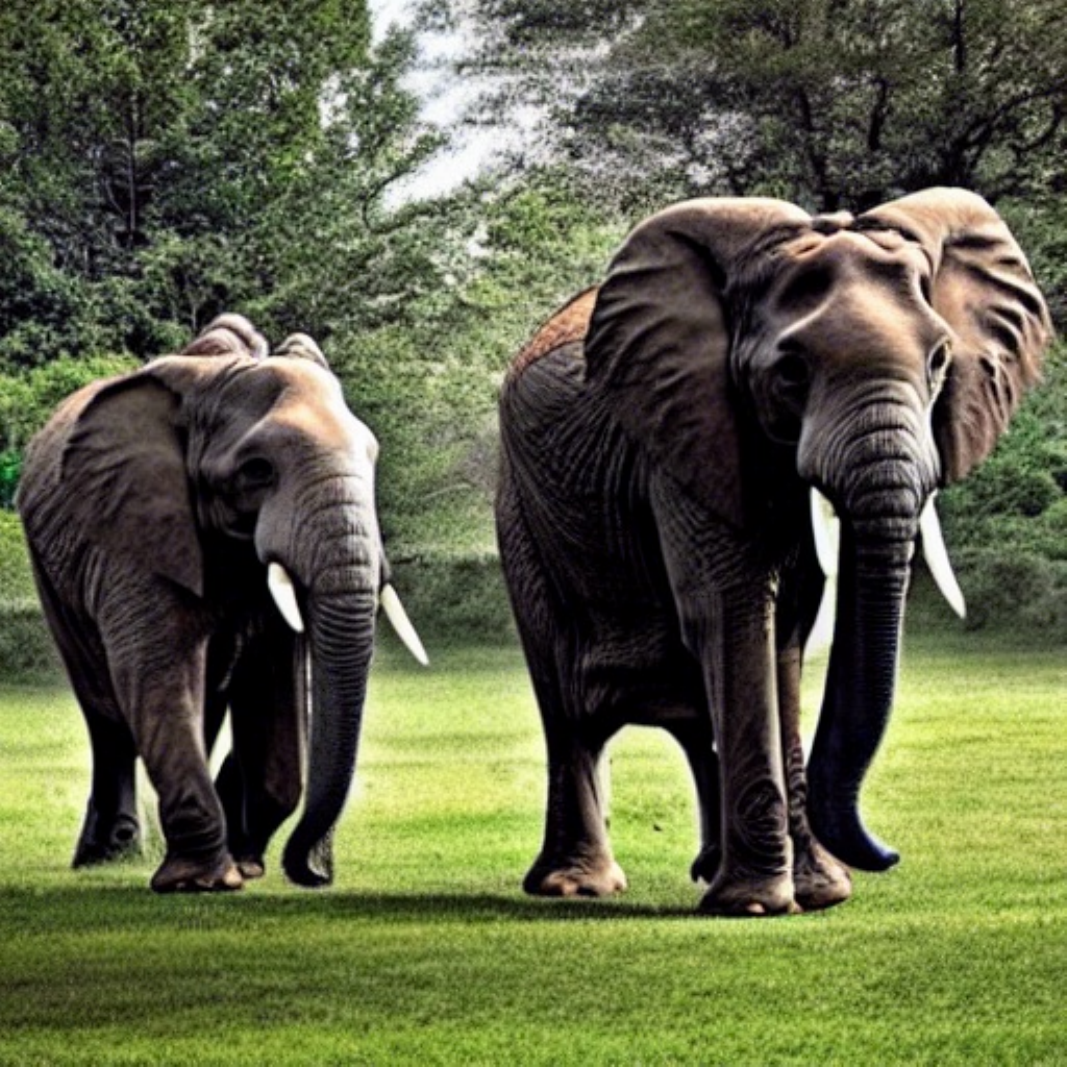} &
      \includegraphics[width=0.18\linewidth]{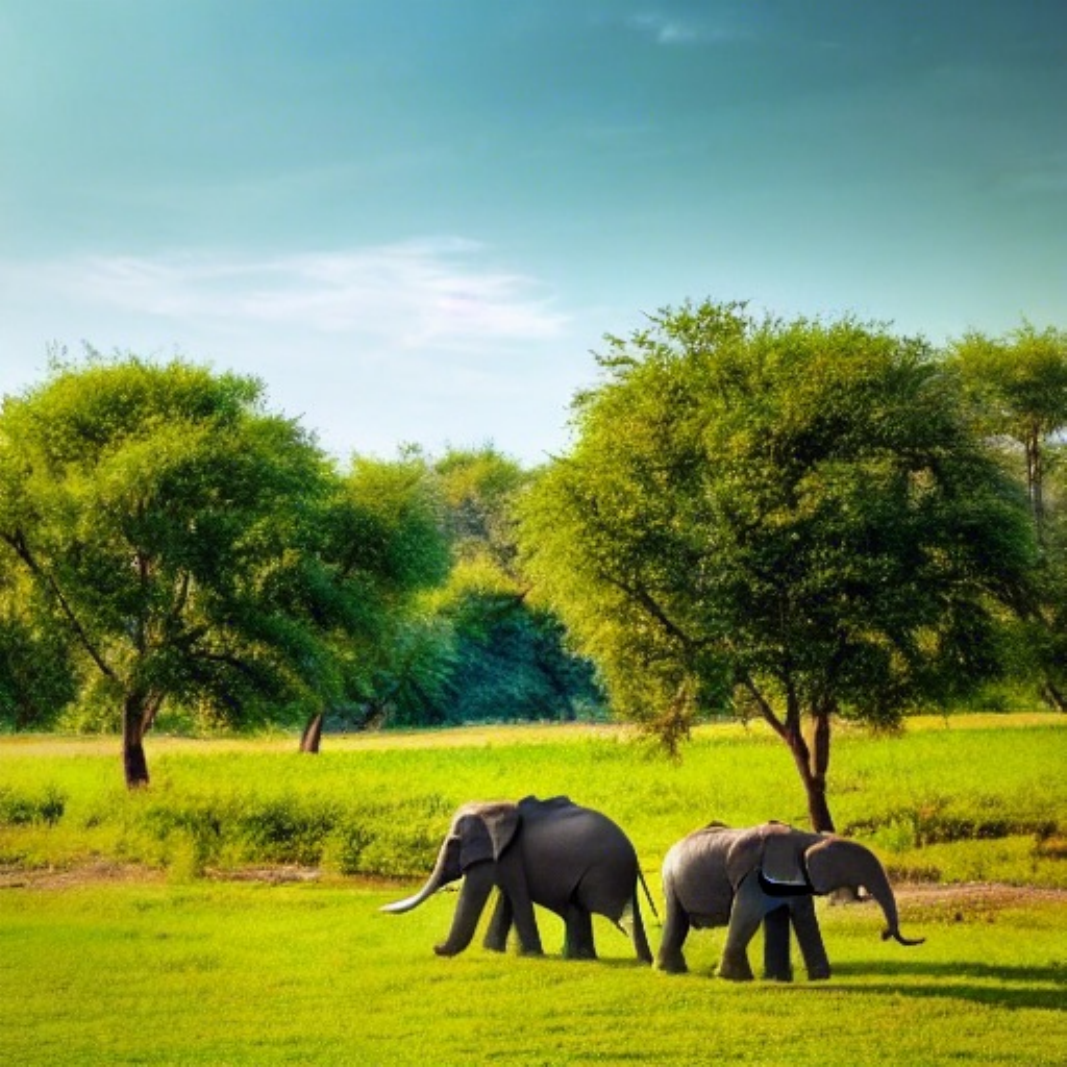} &
      \includegraphics[width=0.18\linewidth]{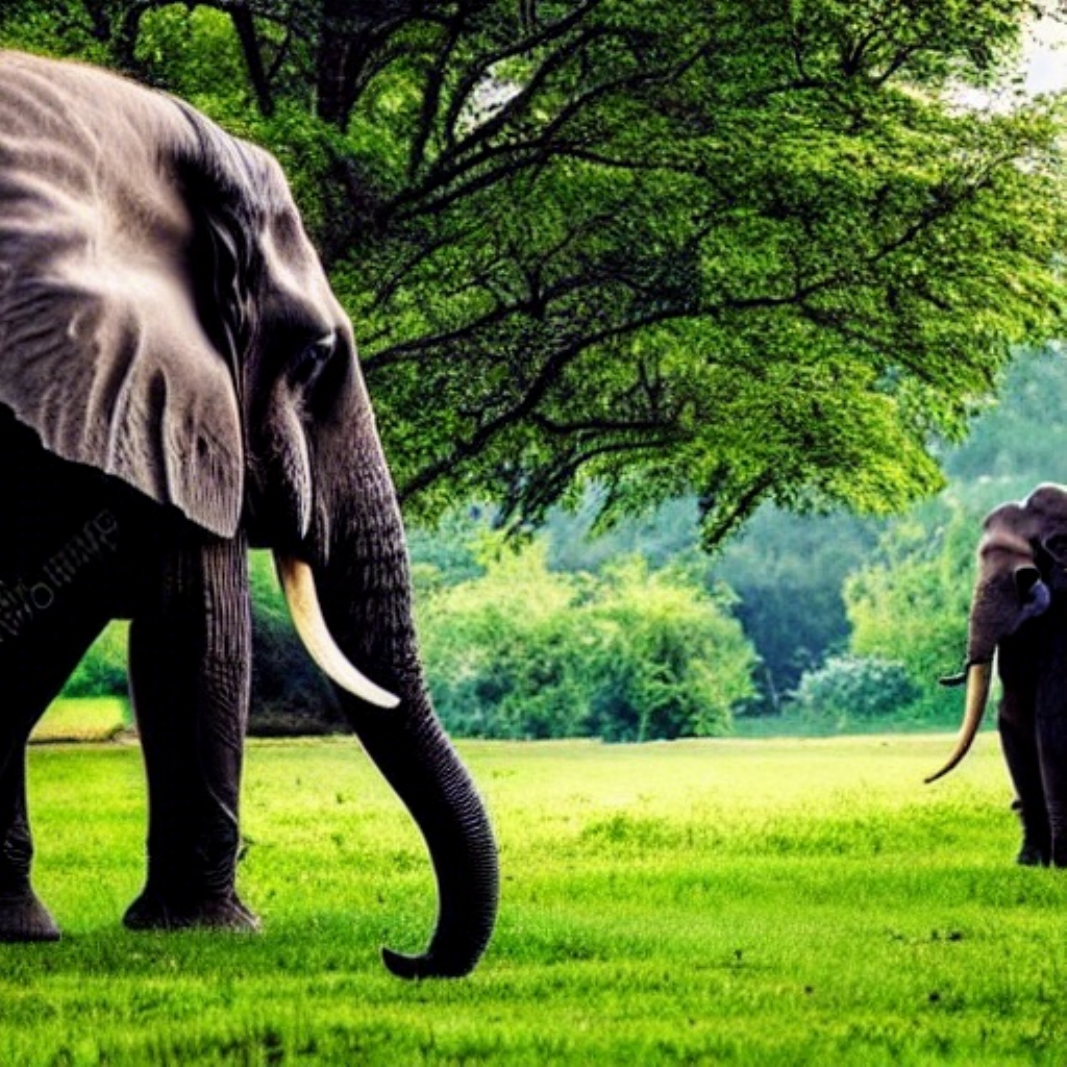} &
      \includegraphics[width=0.18\linewidth]{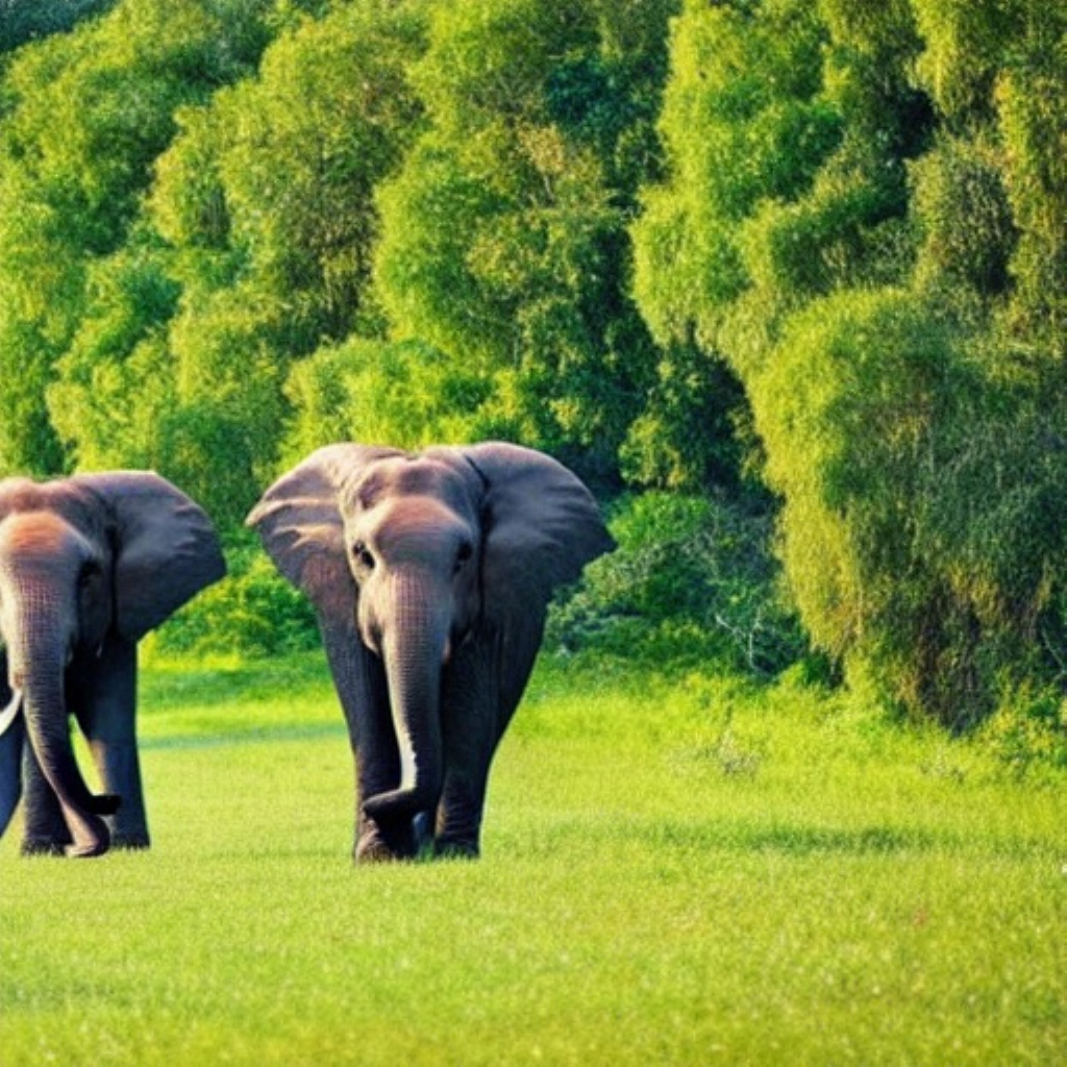} &
      \includegraphics[width=0.18\linewidth]{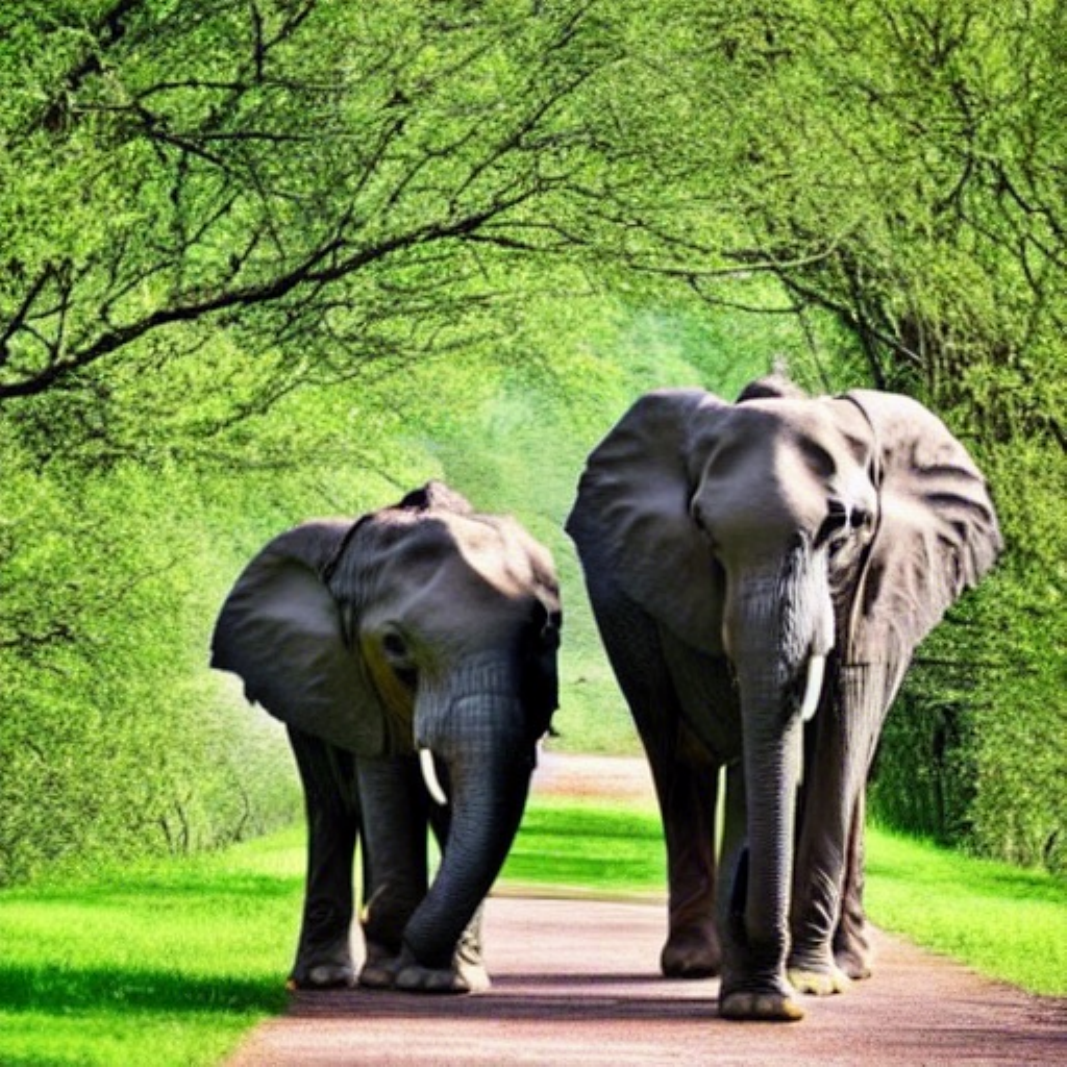} \\[-1pt]
      \includegraphics[width=0.18\linewidth]{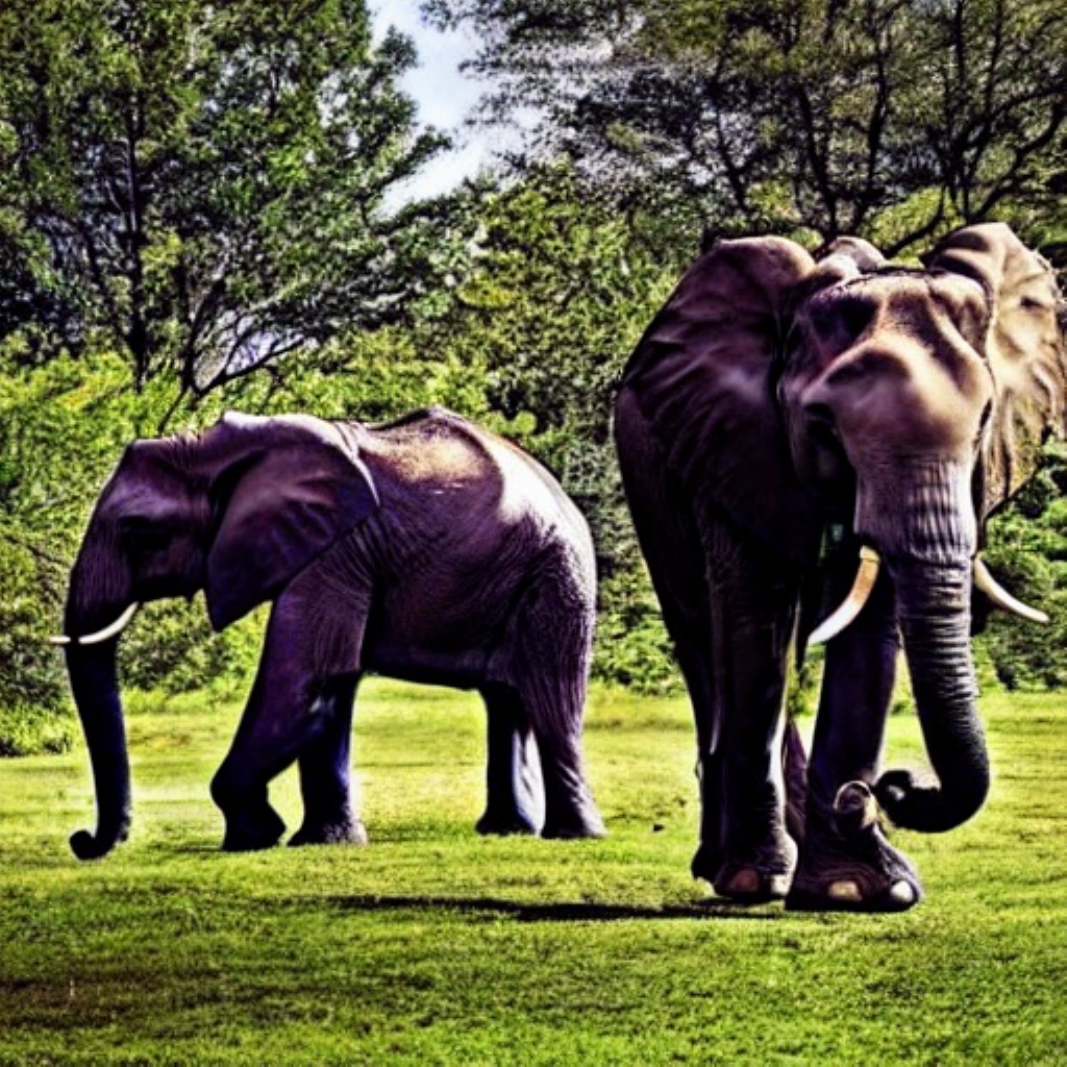} &
      \includegraphics[width=0.18\linewidth]{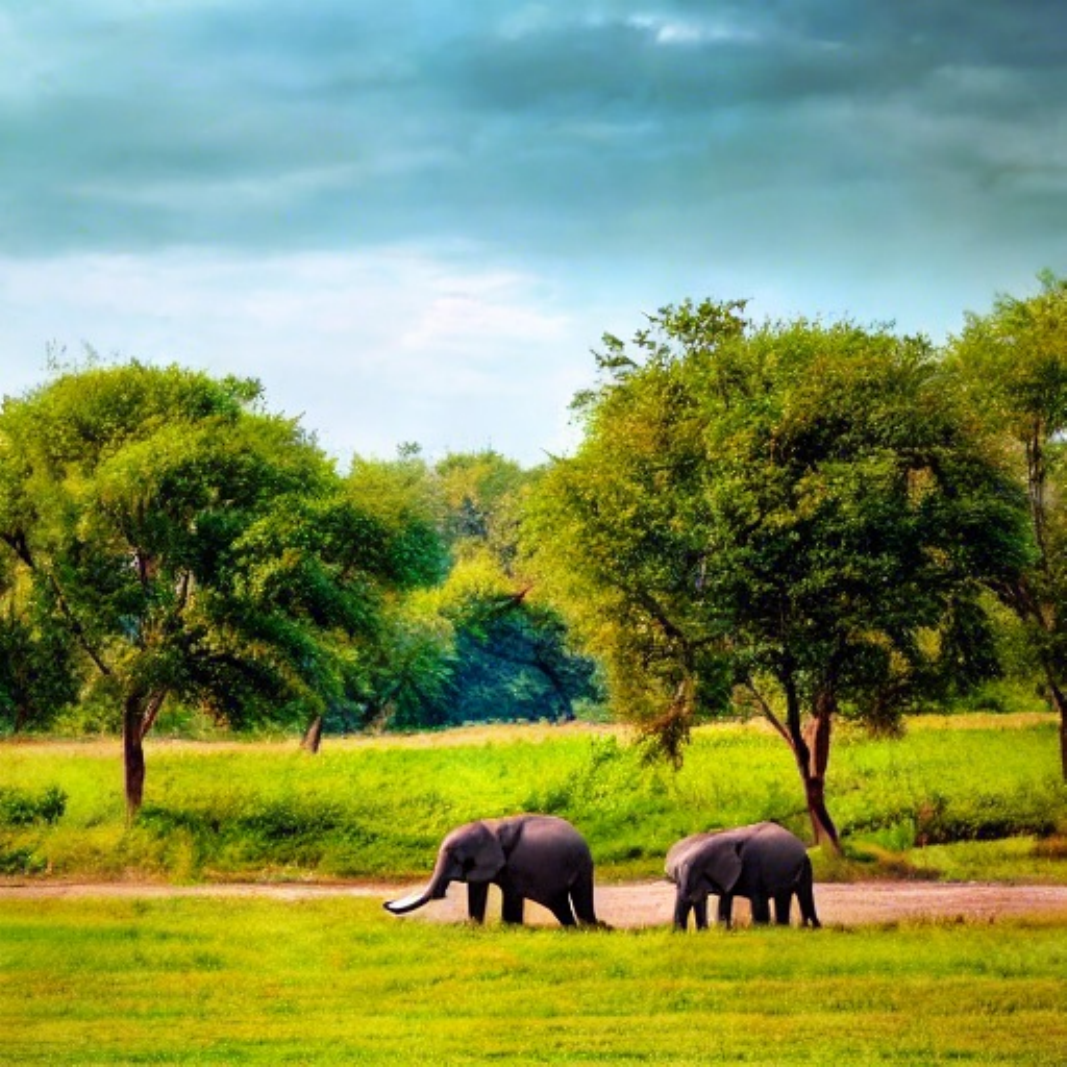} &
      \includegraphics[width=0.18\linewidth]{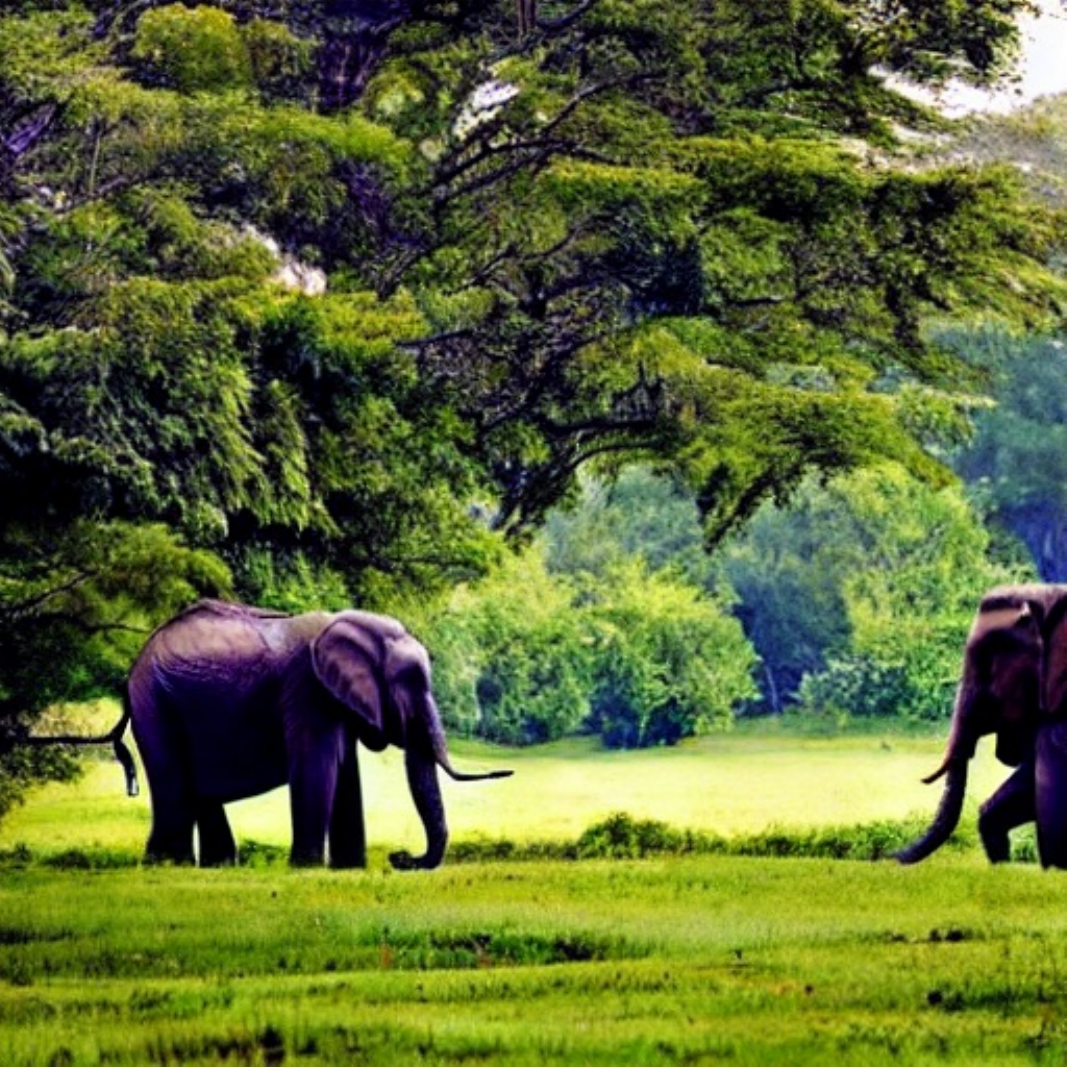} &
      \includegraphics[width=0.18\linewidth]{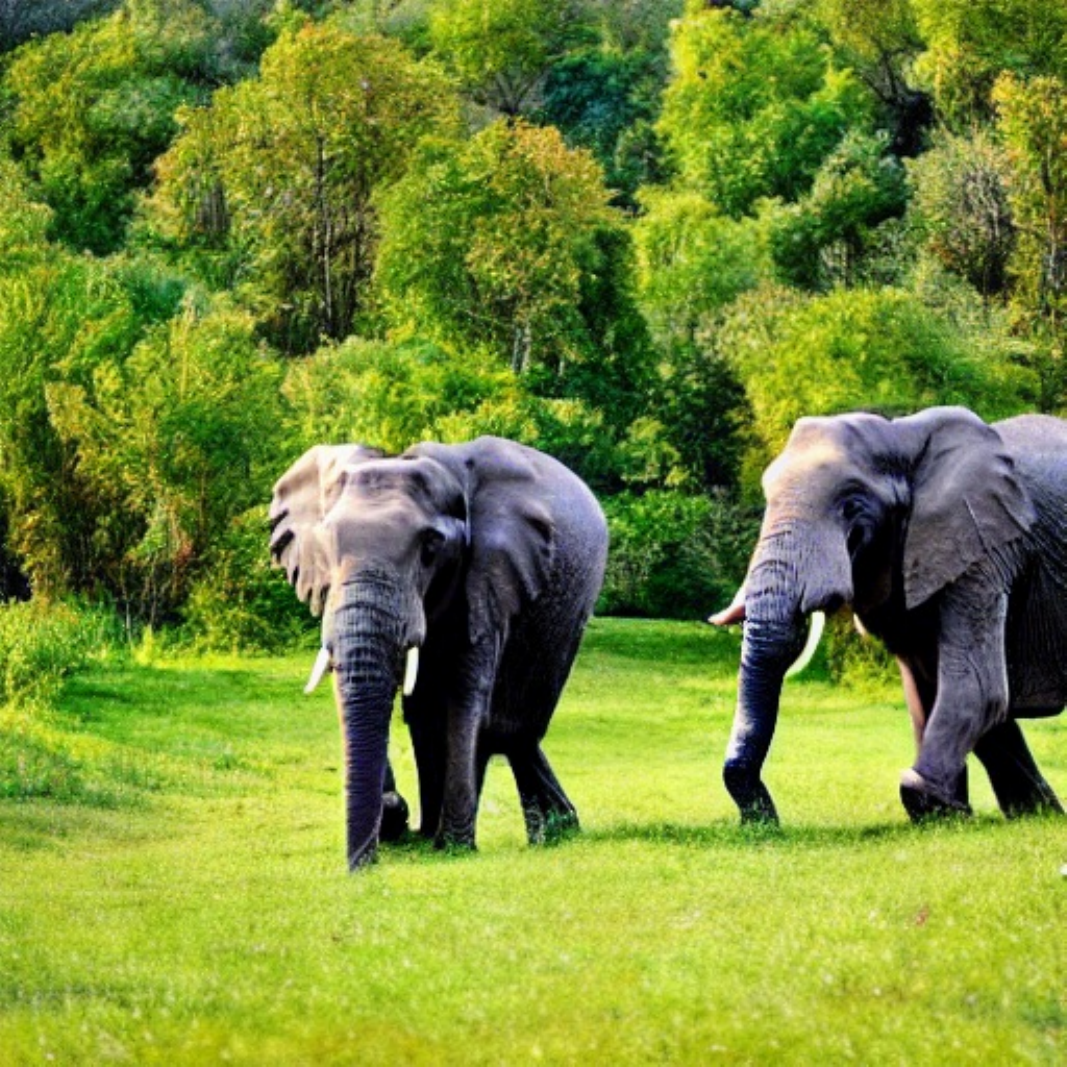} &
      \includegraphics[width=0.18\linewidth]{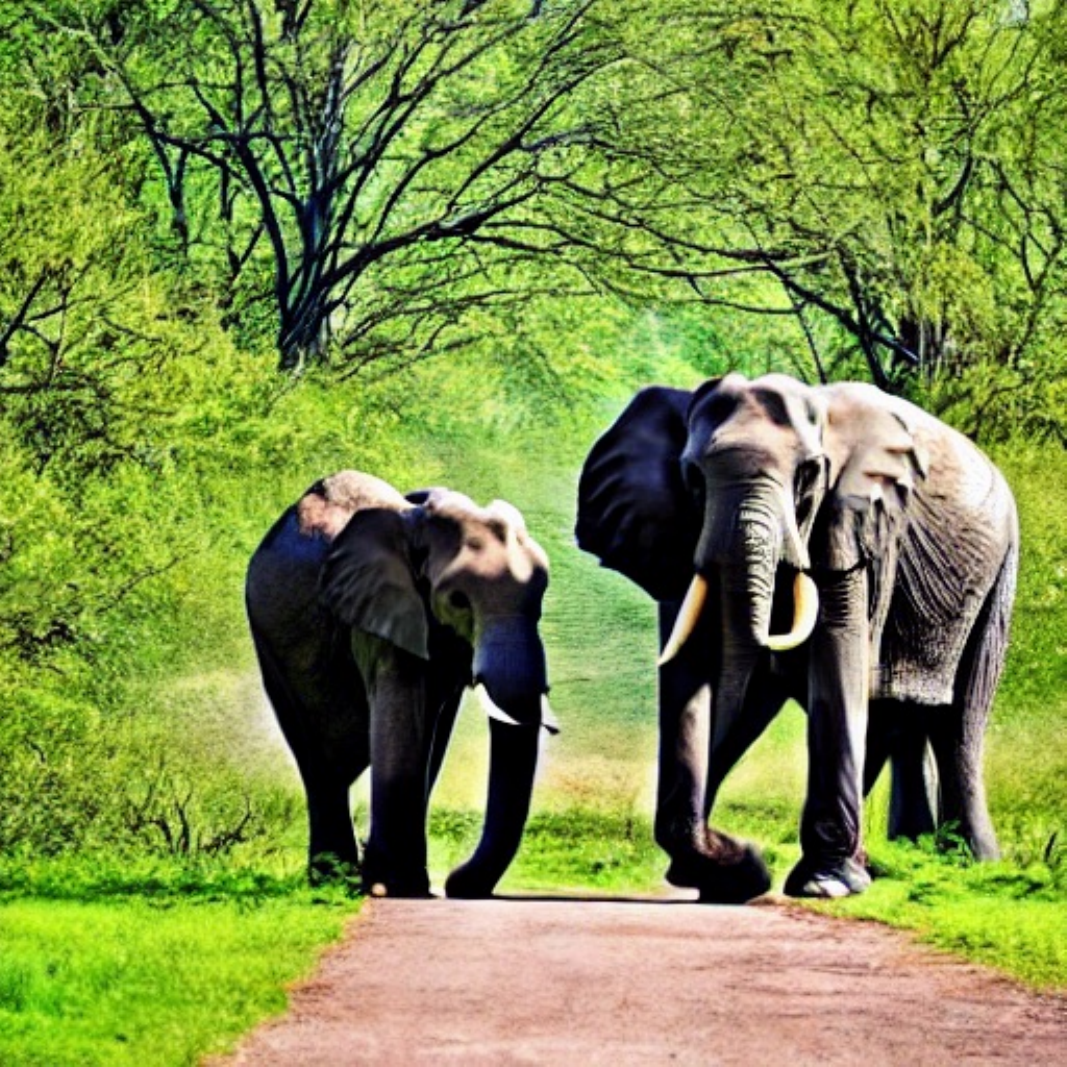} \\[-1pt]
      \includegraphics[width=0.18\linewidth]{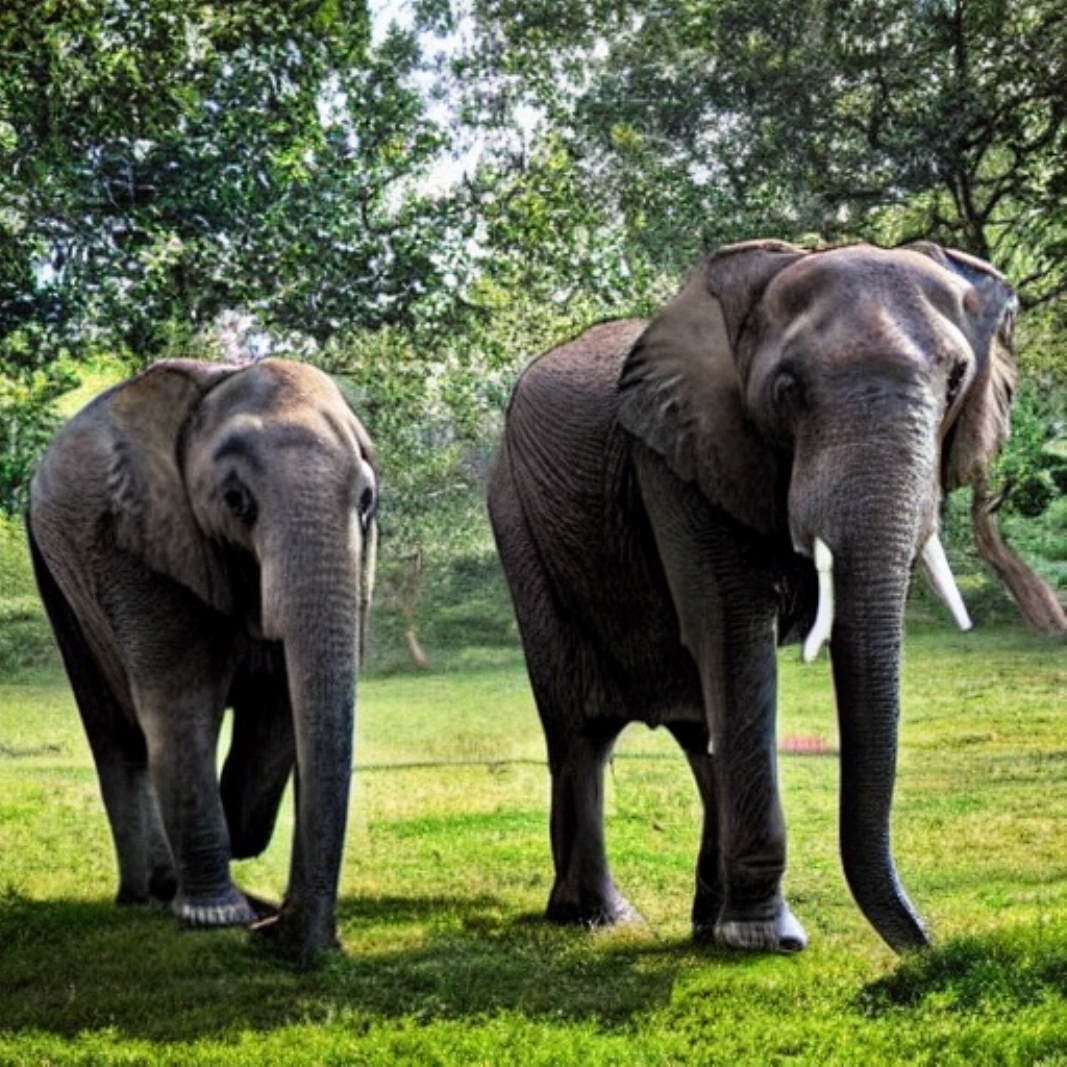} &
      \includegraphics[width=0.18\linewidth]{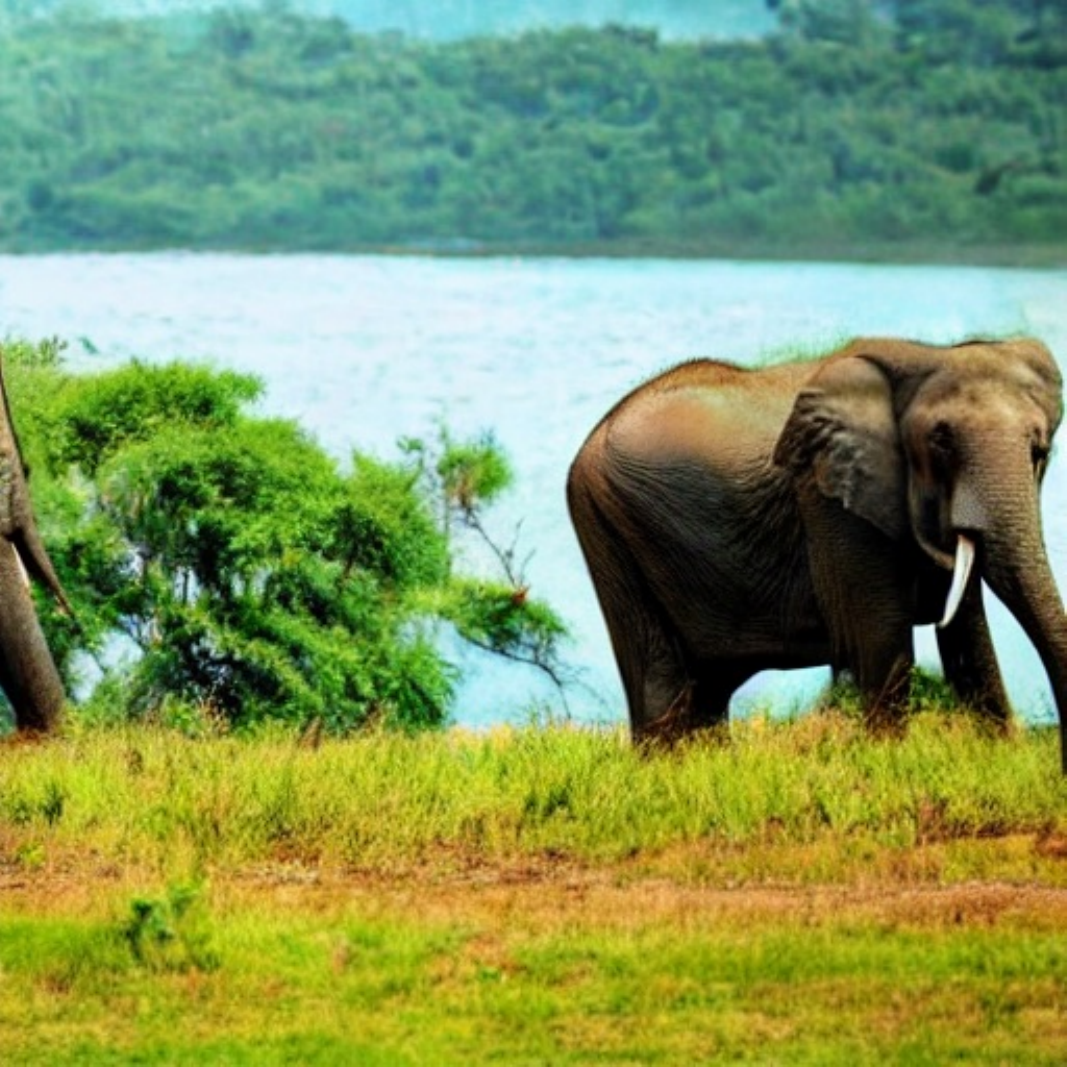} &
      \includegraphics[width=0.18\linewidth]{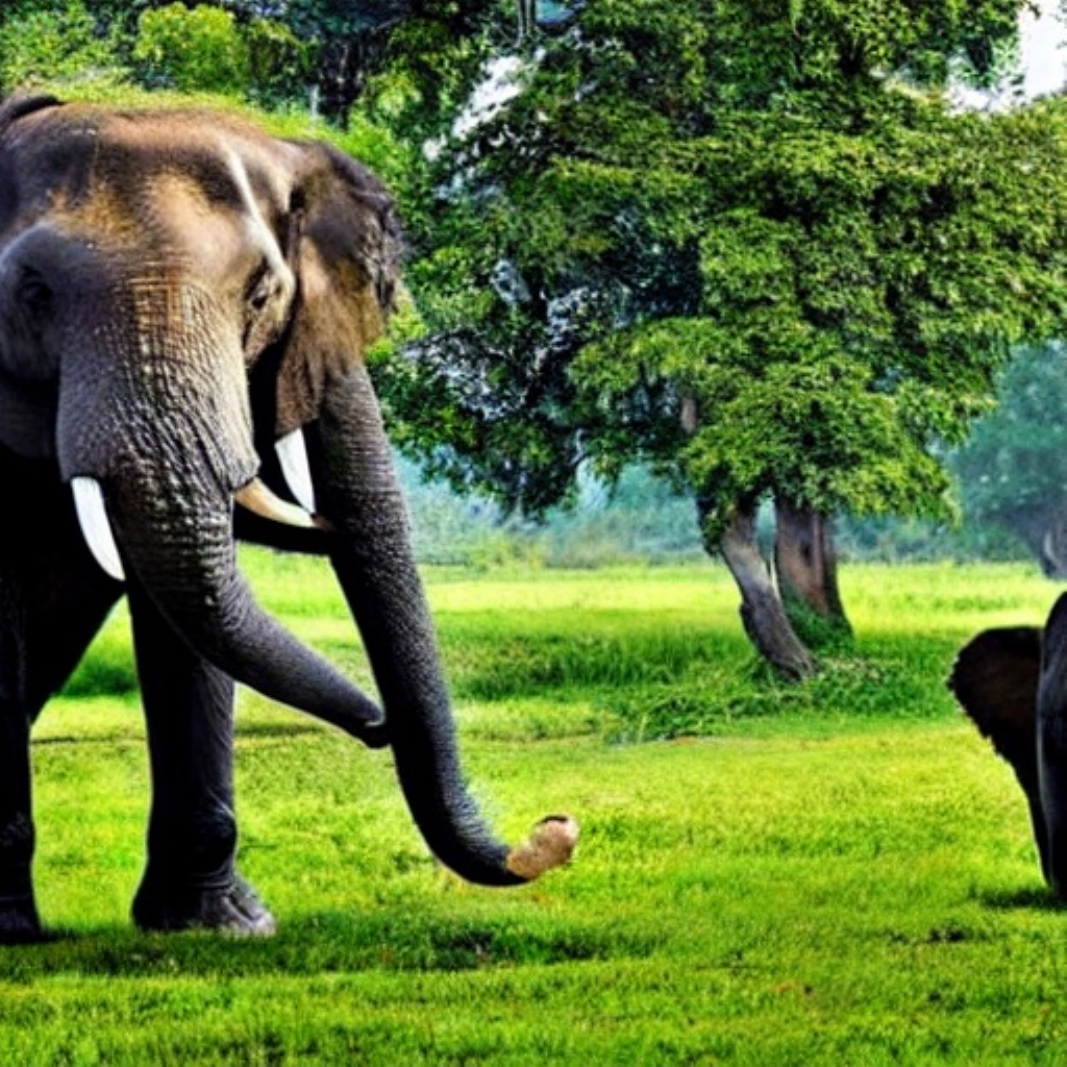} &
      \includegraphics[width=0.18\linewidth]{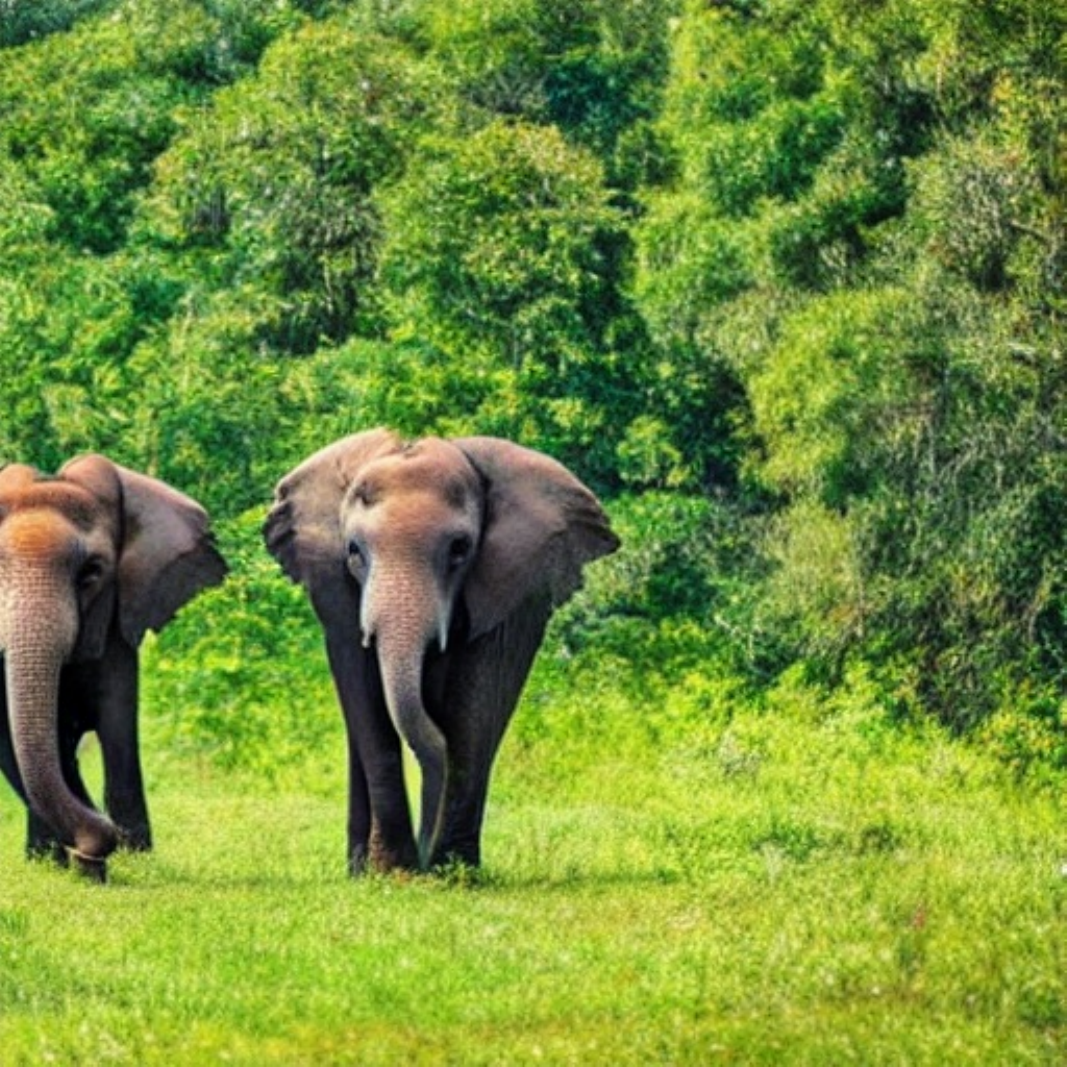} &
      \includegraphics[width=0.18\linewidth]{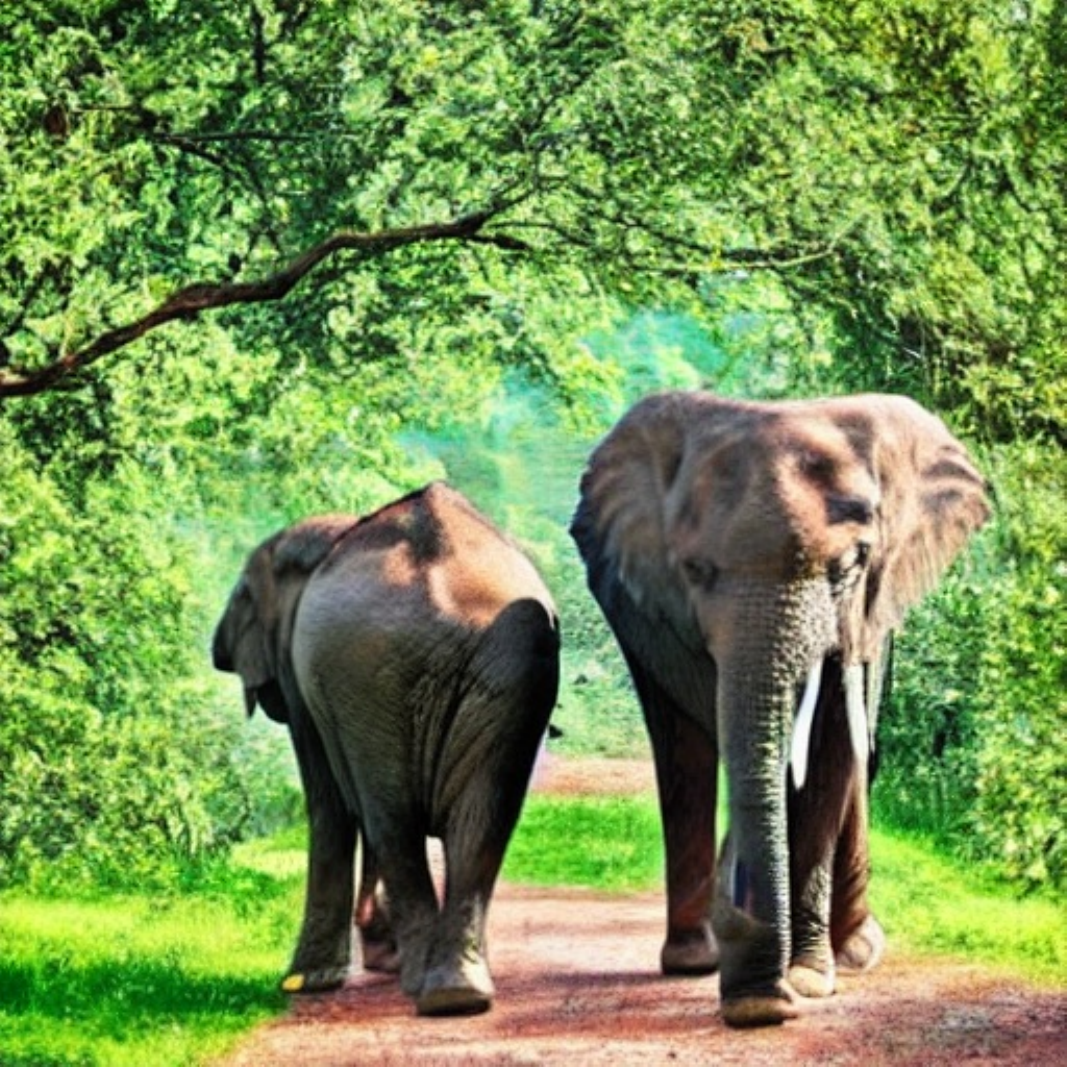} \\[-1pt]
      \includegraphics[width=0.18\linewidth]{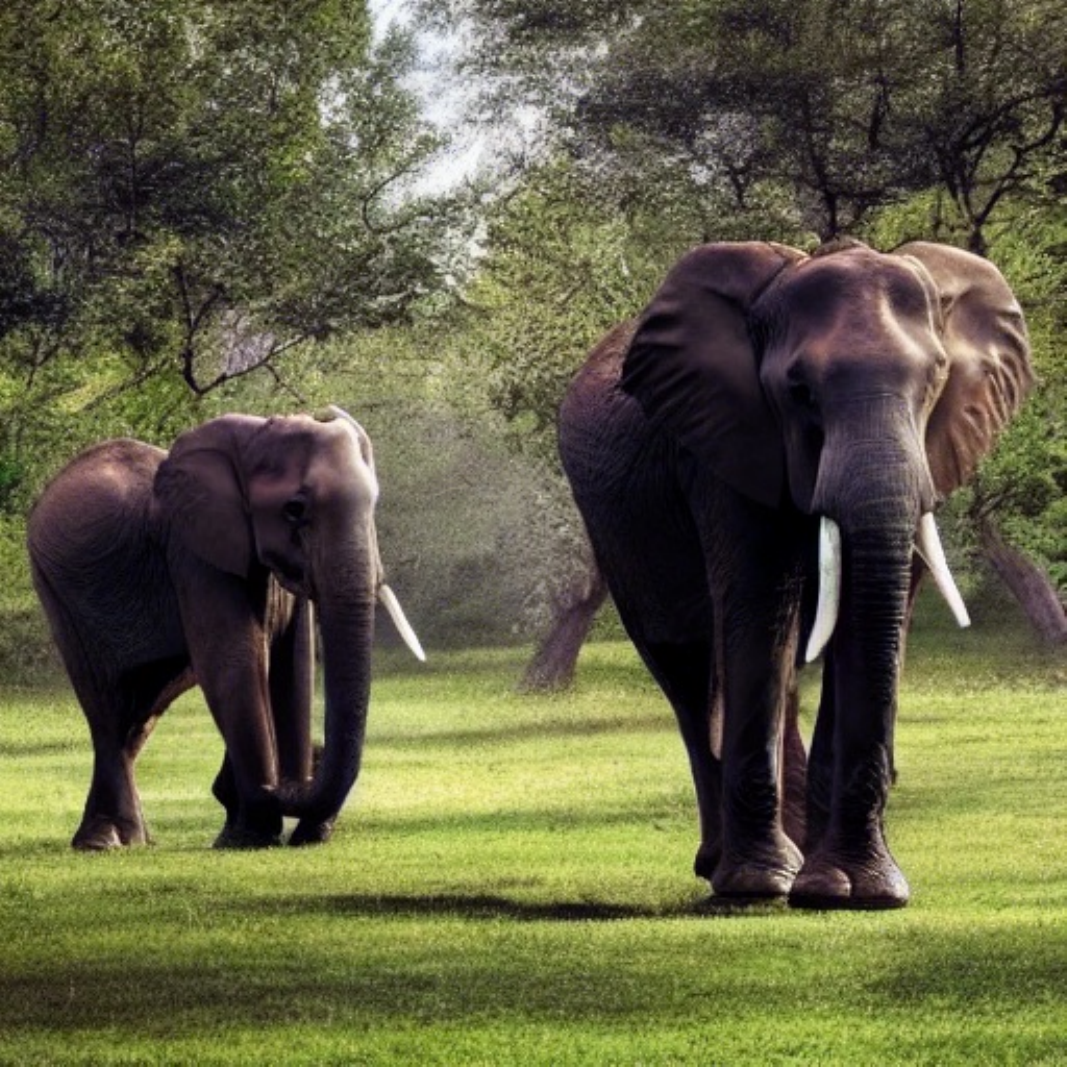} &
      \includegraphics[width=0.18\linewidth]{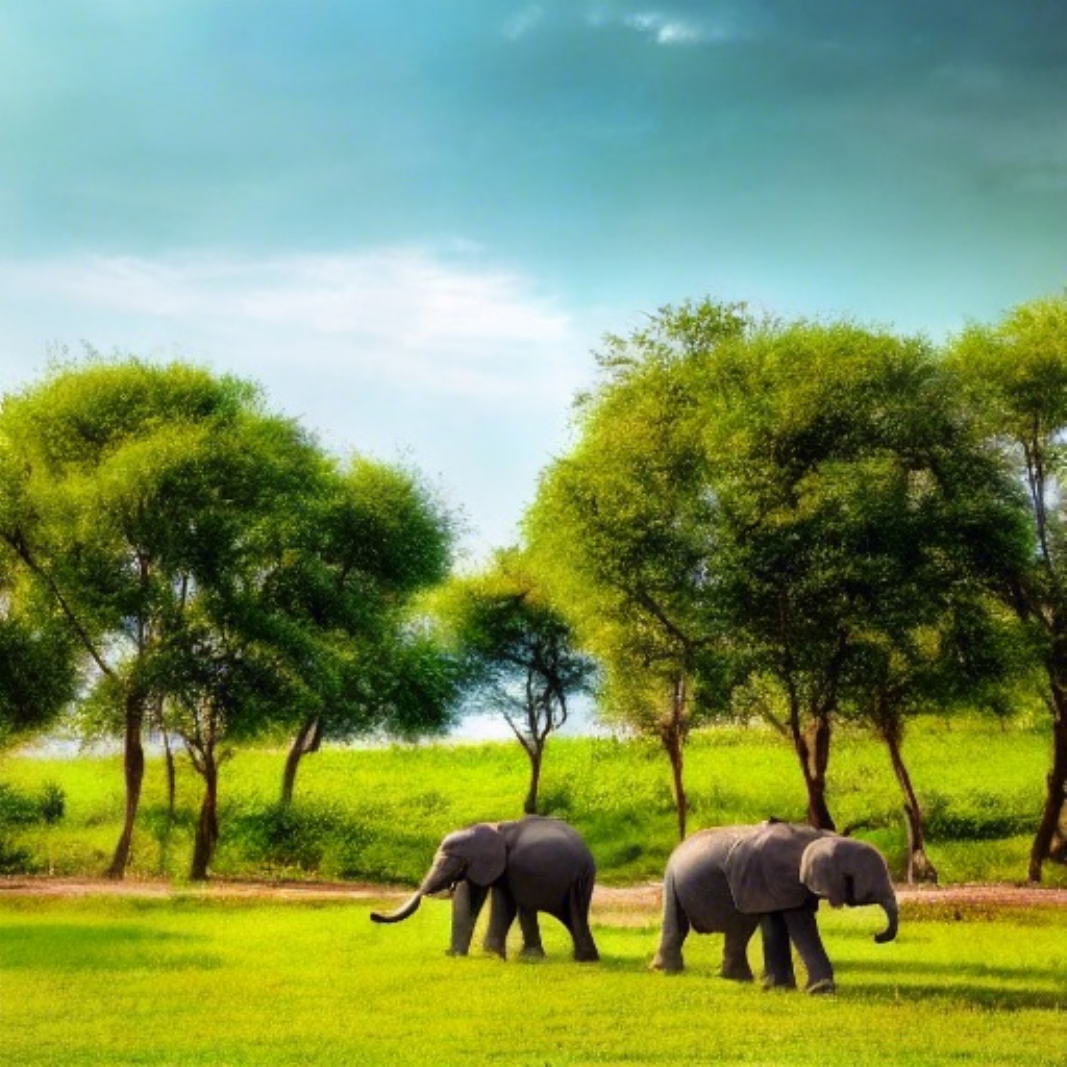} &
      \includegraphics[width=0.18\linewidth]{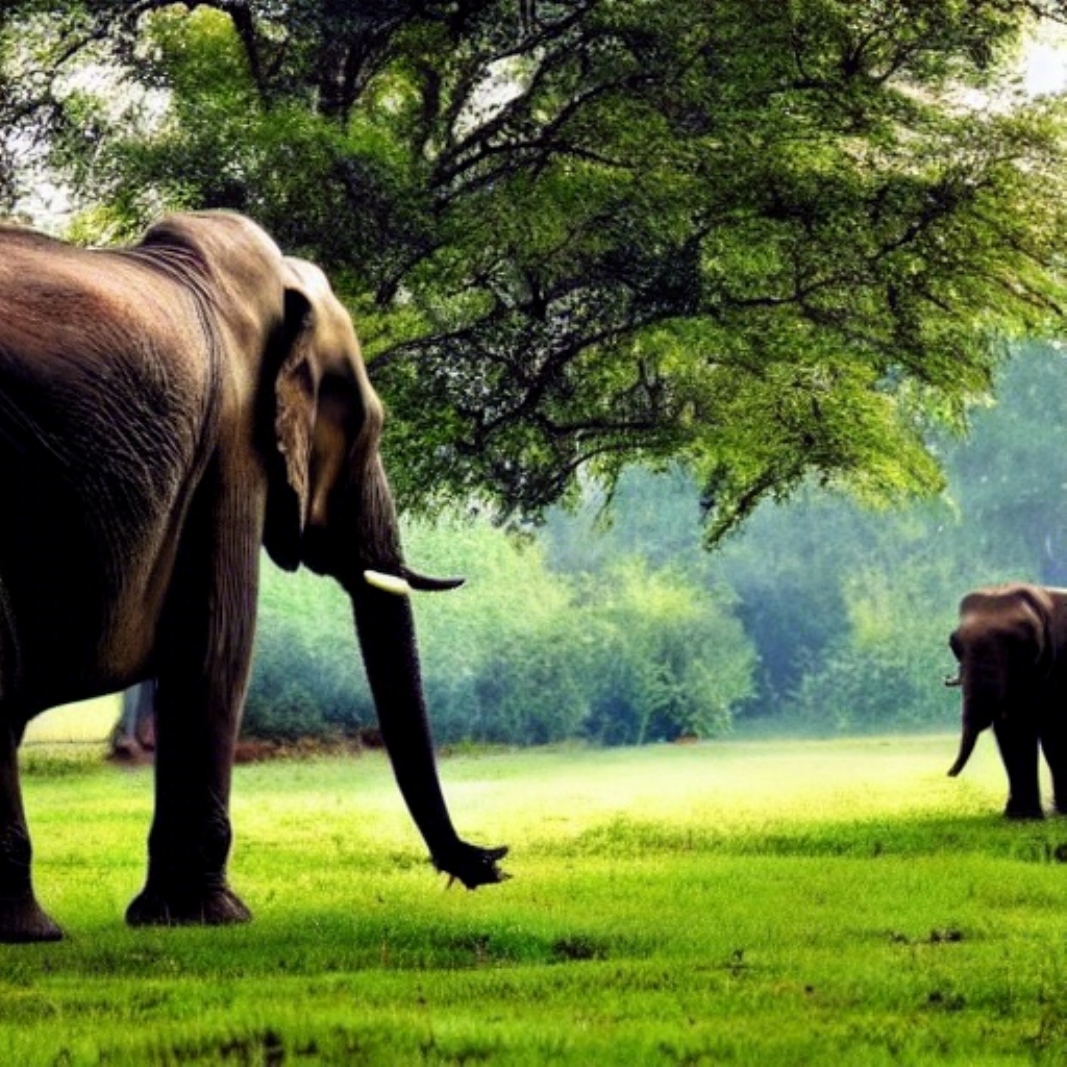} &
      \includegraphics[width=0.18\linewidth]{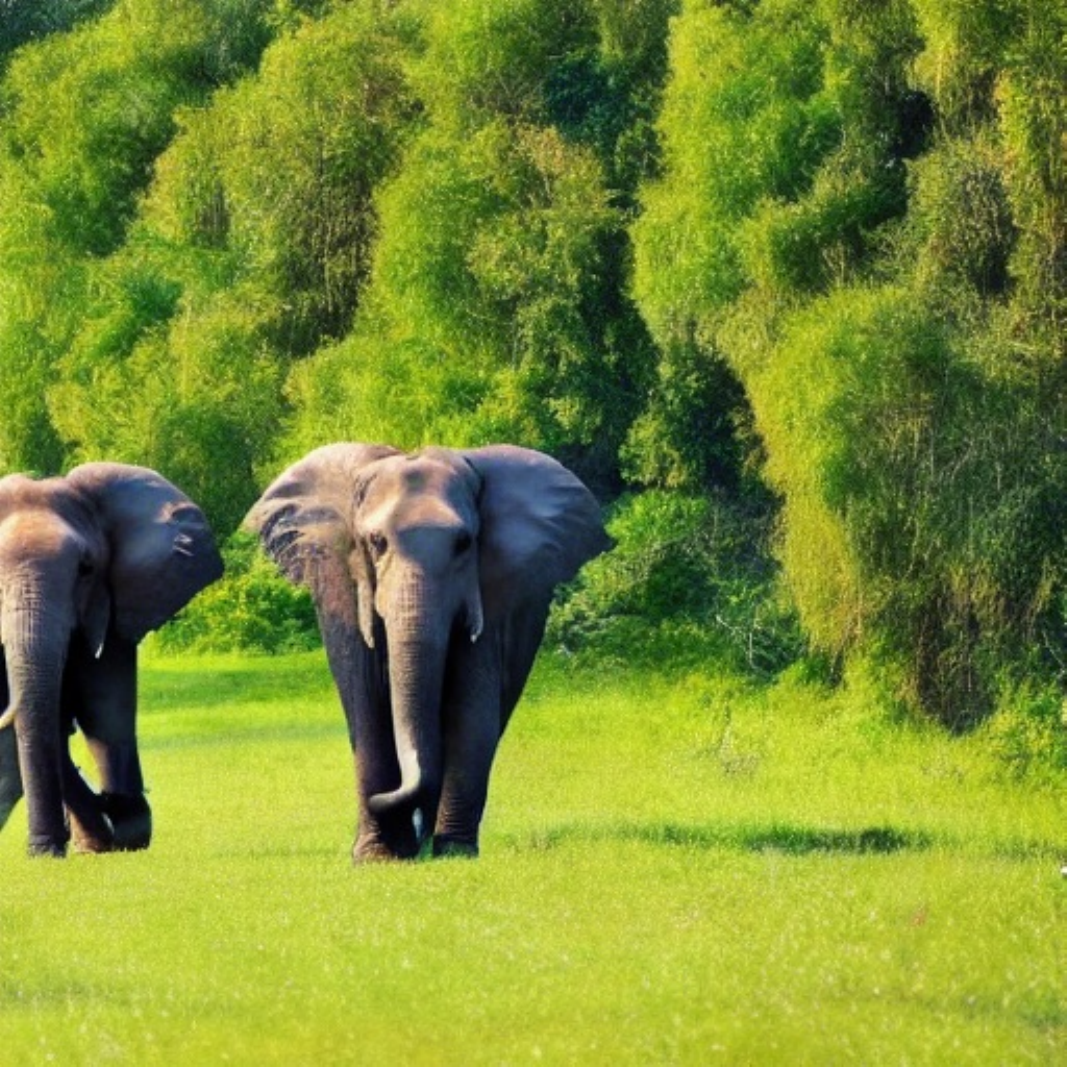} &
      \includegraphics[width=0.18\linewidth]{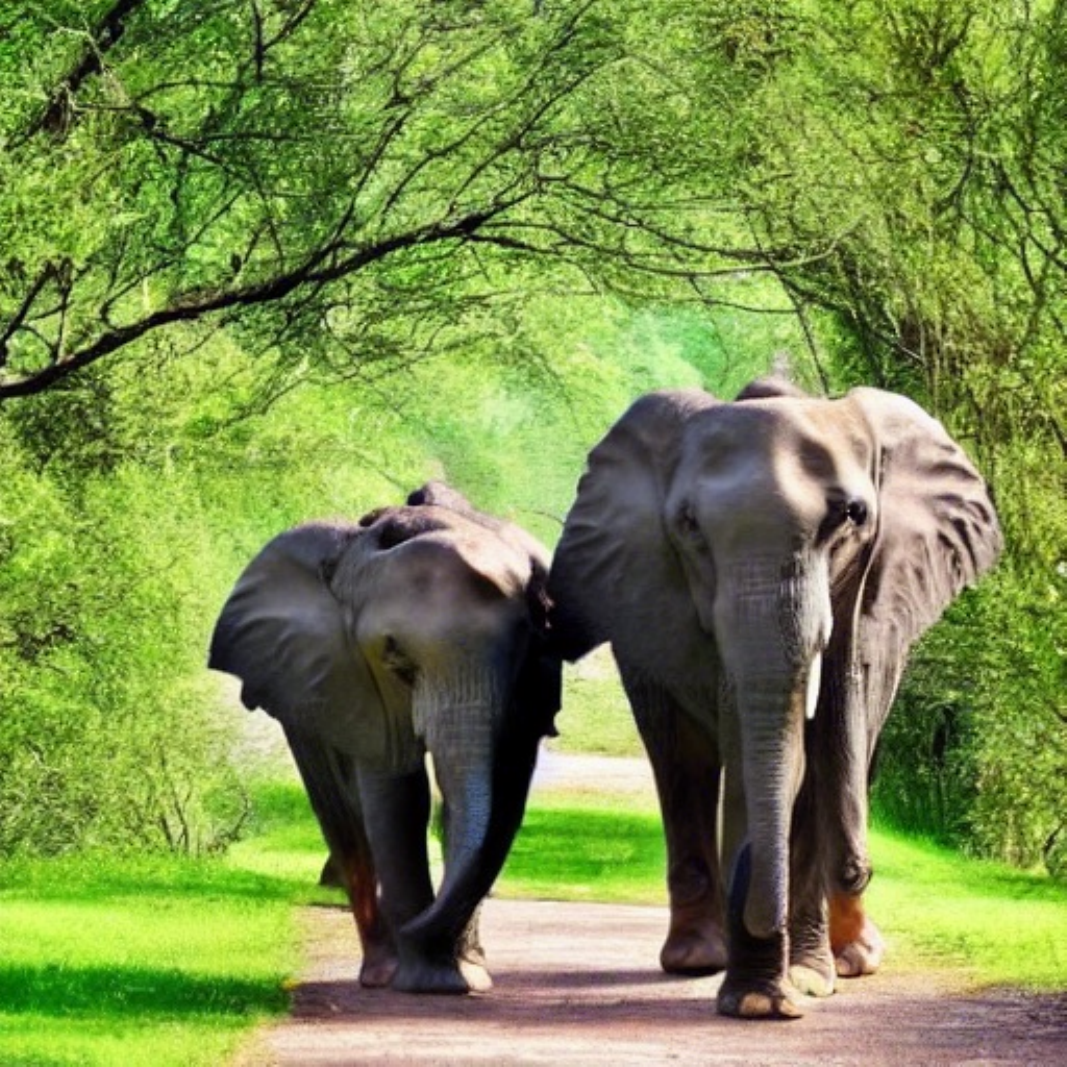}
    \end{tabular}
    \vspace{-2mm}
    \caption{\textit{"Two elephants outdoors walking on green grass with trees nearby."}}
  \end{subfigure}

  \vspace{2mm}

  \caption{Visual comparisons of generated images~(512$\times$512) conditioned on COCO validation prompts~\cite{lin2014microsoft}. We generate images by SD v1.4~\cite{rombach2022high} and its quantized versions using Q-Diffusion~\cite{li2023q} and AccuQuant under a 4/8-bit setting and PCR~\cite{tang2025post} in 4/8.4-bit setting. Each row corresponds to Full Precision, Q-Diffusion~\cite{li2023q}, PCR($\tau=0.2$)~\cite{tang2025post} and Ours.}
  \label{fig:appendix_sd_coco_prompts}
\end{figure*}

 \begin{figure*}[p]
   \captionsetup[subfigure]{font=small, labelformat=empty}
   \begin{center}
 
   \begin{subfigure}[c]{\linewidth}
     \centering
     \begin{tabular}{@{}c@{\hspace{1mm}}c@{\hspace{1mm}}c@{\hspace{1mm}}c@{\hspace{1mm}}c@{\hspace{1mm}}c@{}}
       \includegraphics[width=0.18\columnwidth]{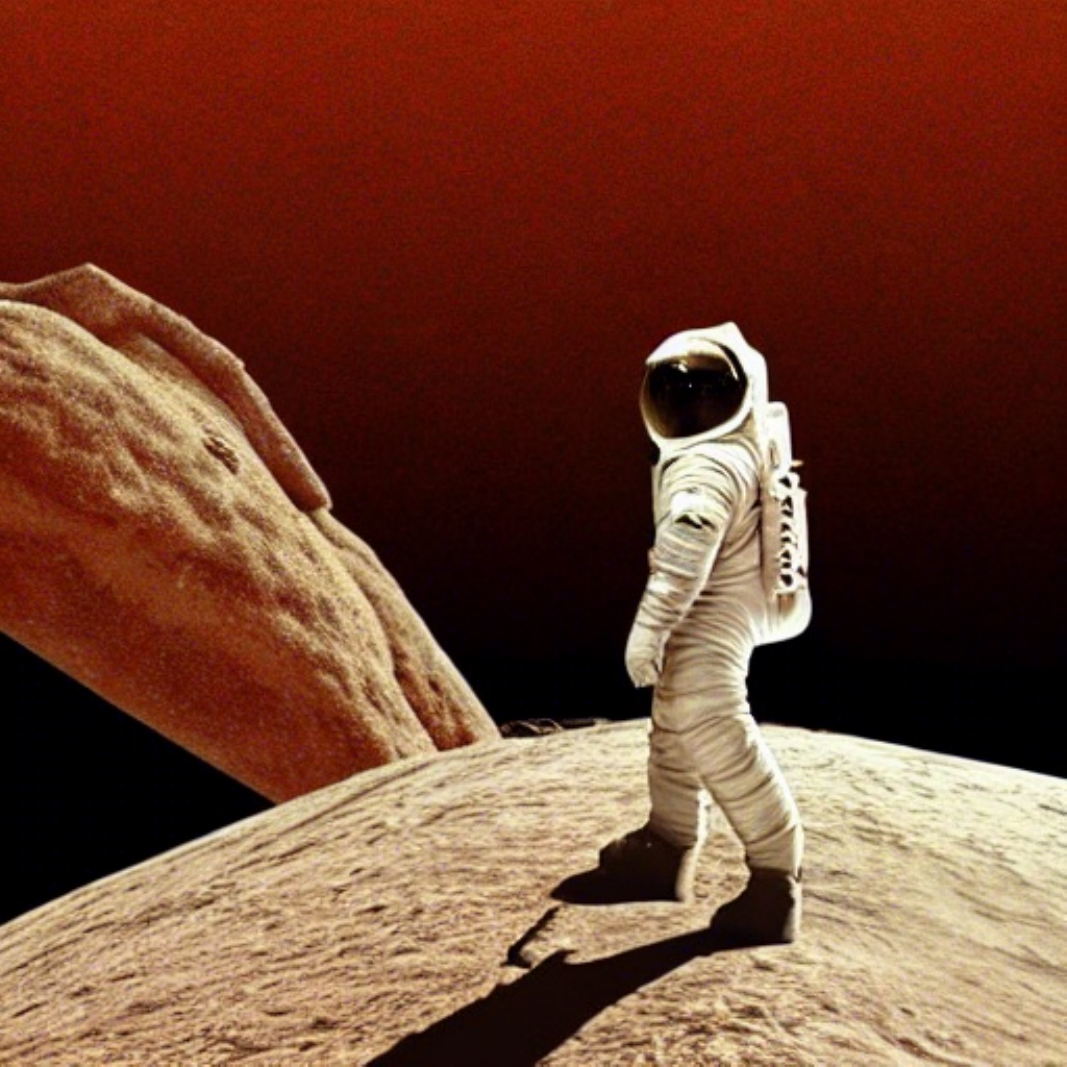}&
       \includegraphics[width=0.18\columnwidth]{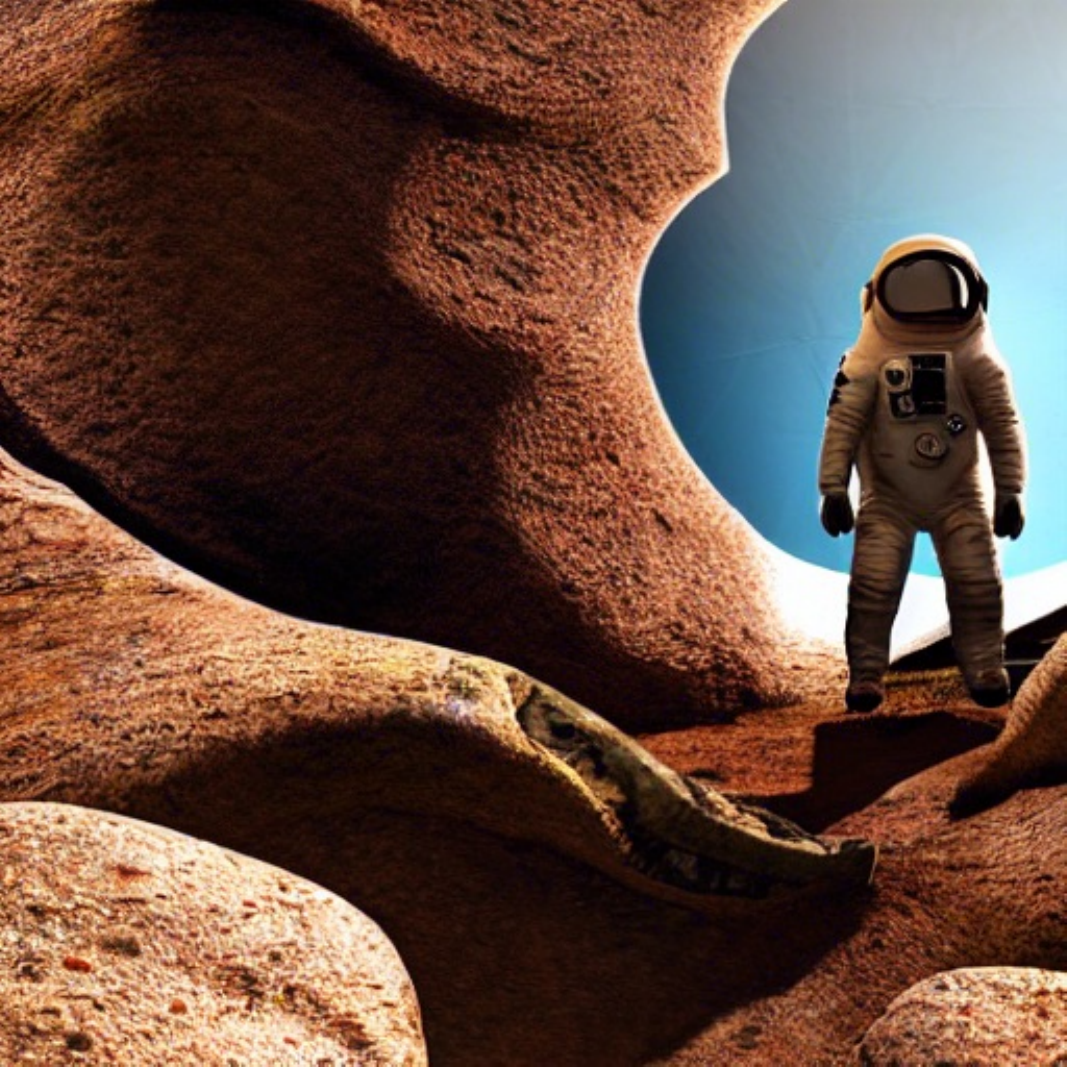}&
       \includegraphics[width=0.18\columnwidth]{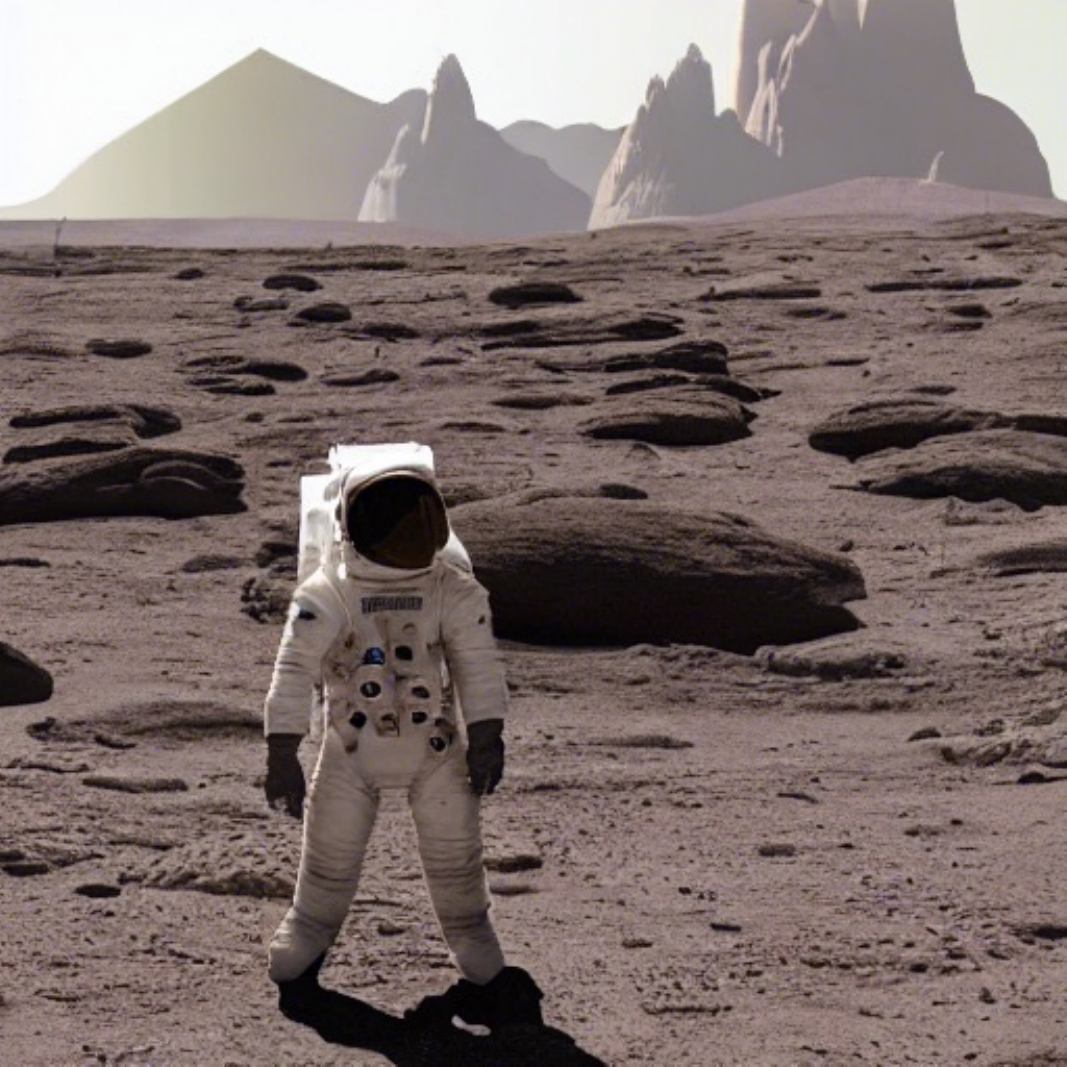}&
       \includegraphics[width=0.18\columnwidth]{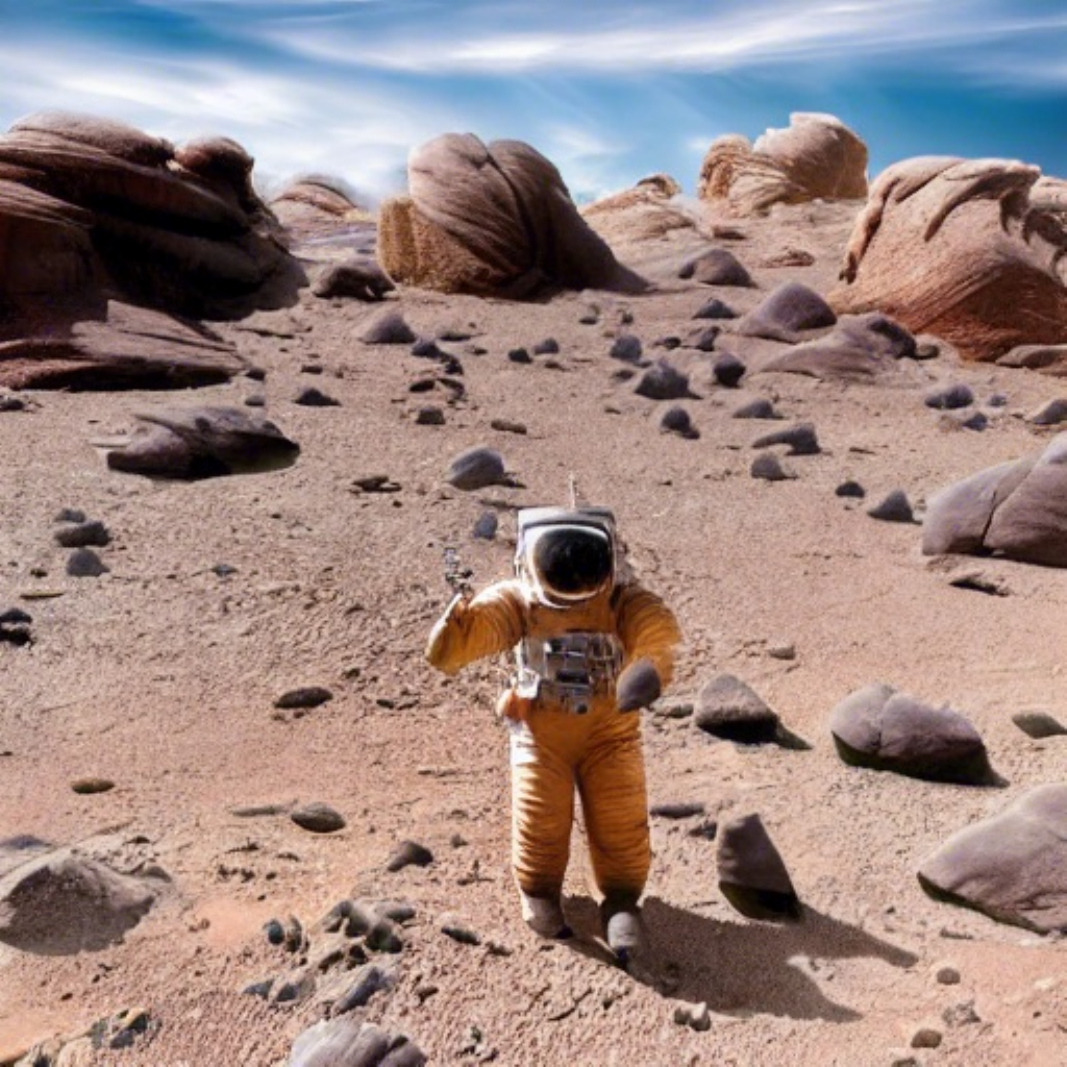}&
       \includegraphics[width=0.18\columnwidth]{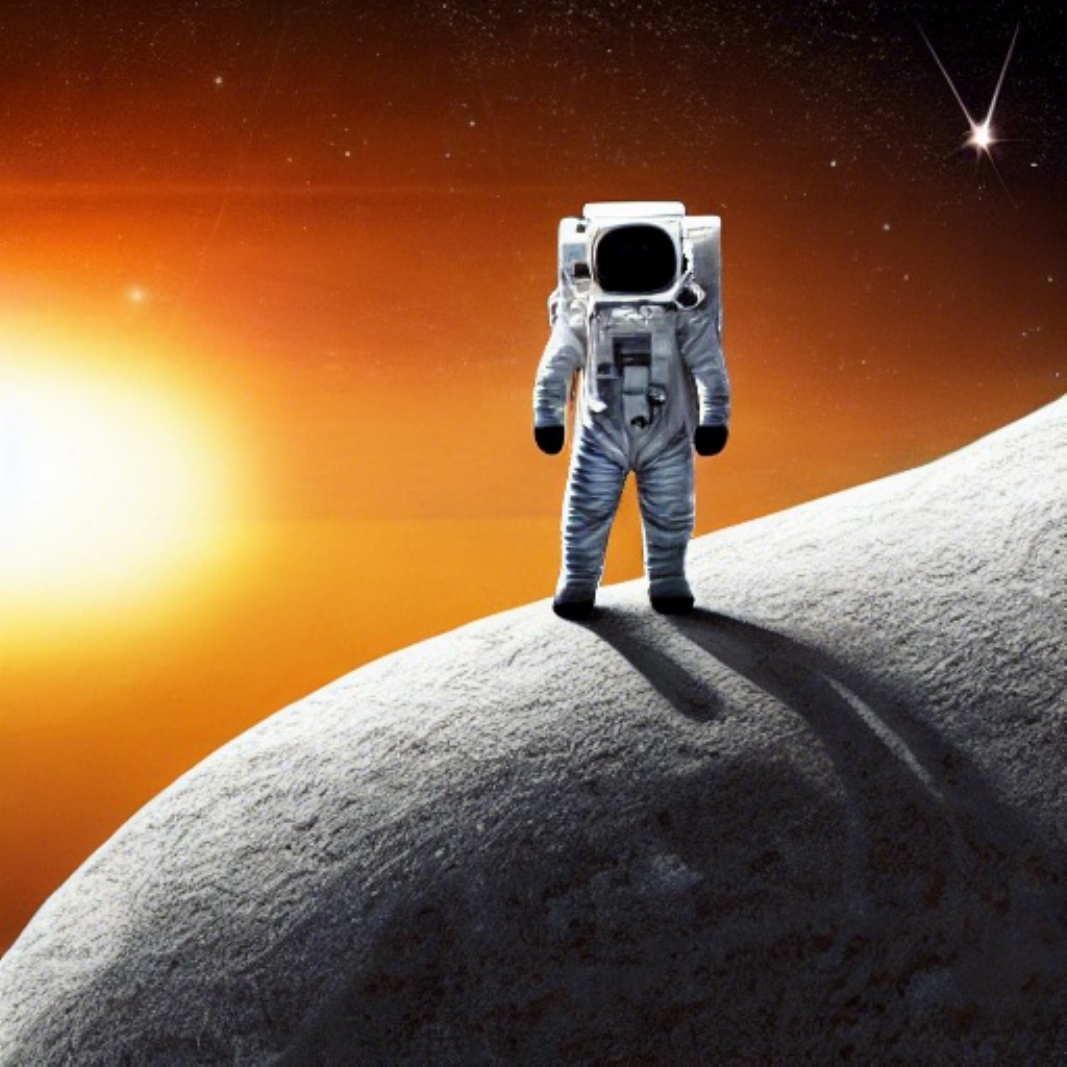}\\[-1pt]
       \includegraphics[width=0.18\columnwidth]{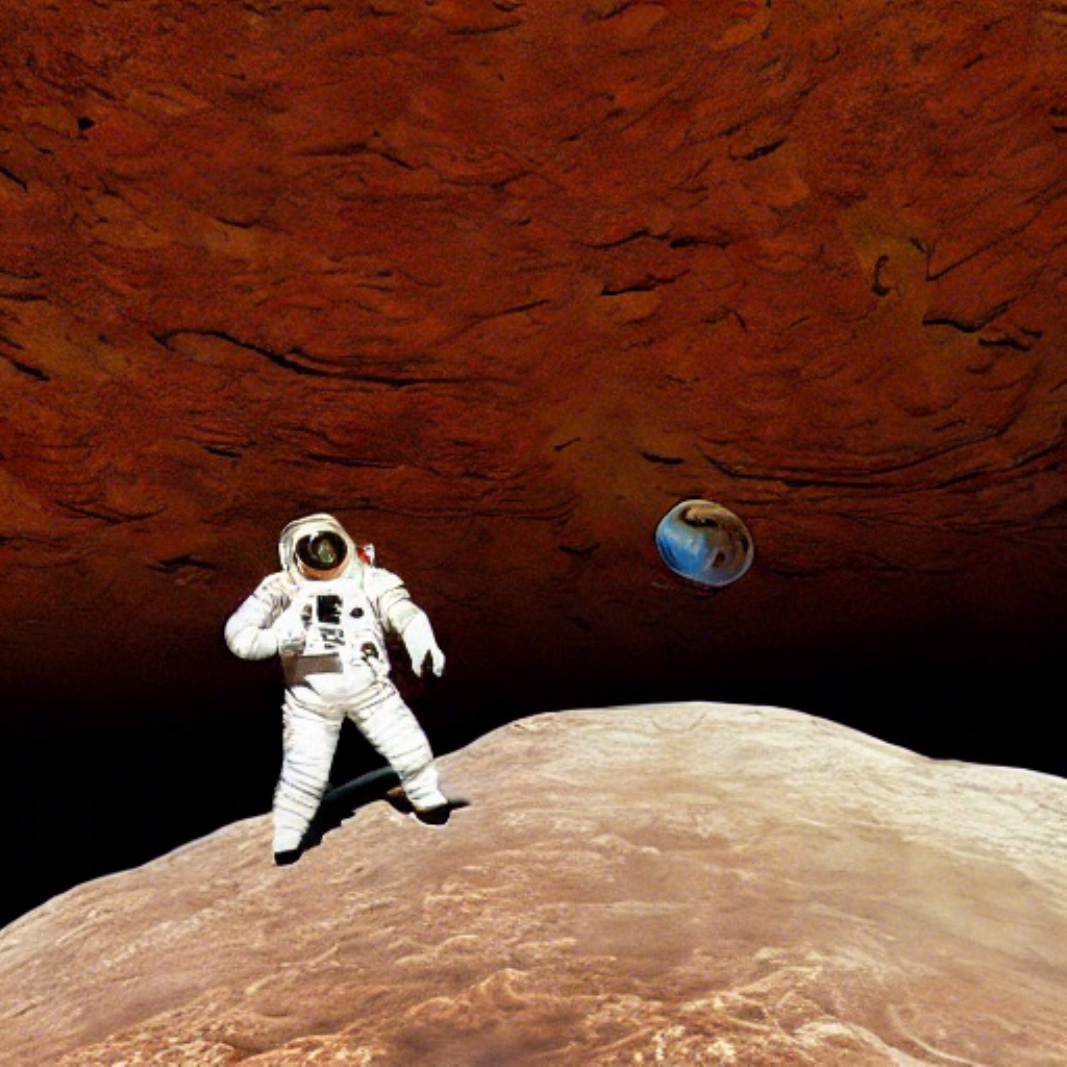}&
       \includegraphics[width=0.18\columnwidth]{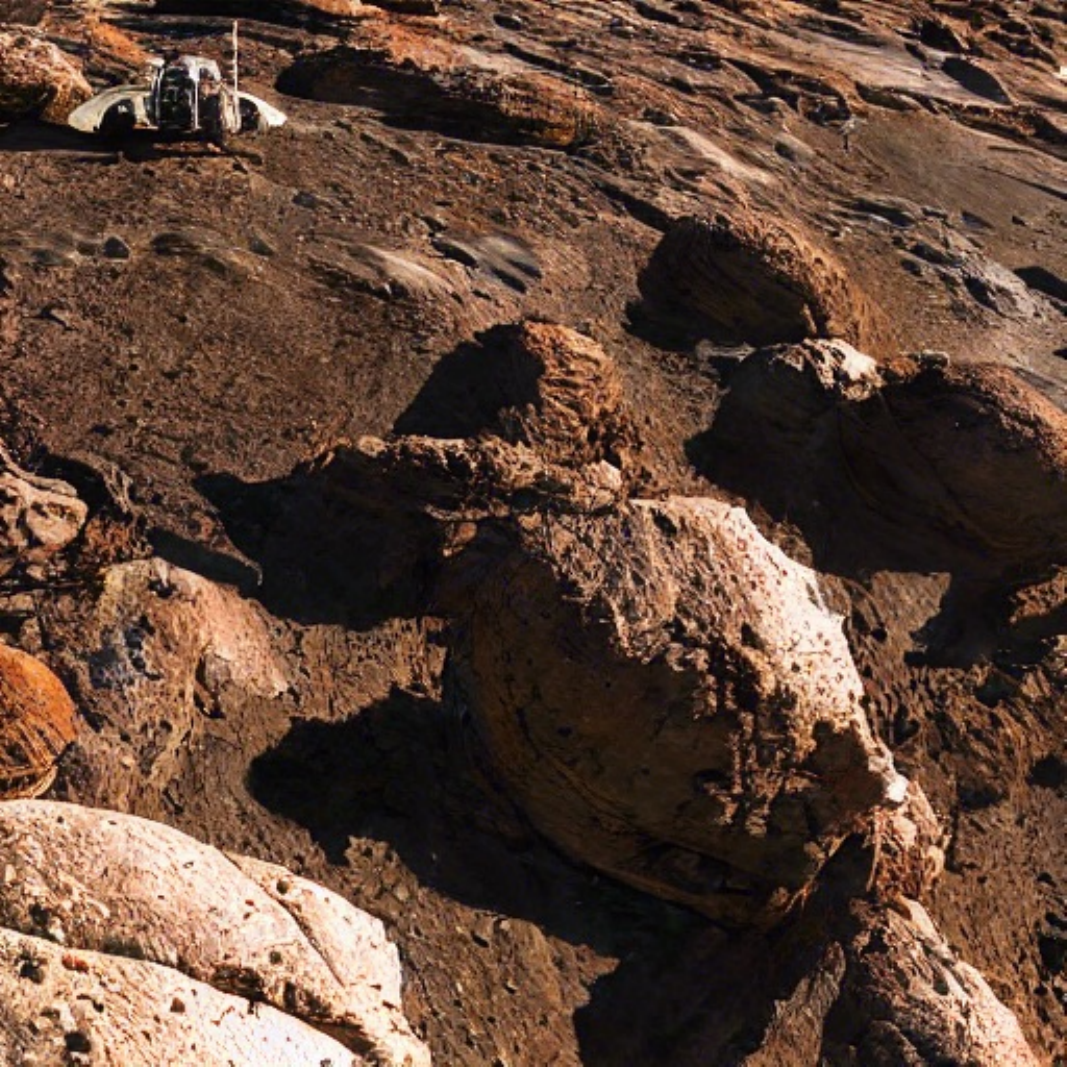}&
       \includegraphics[width=0.18\columnwidth]{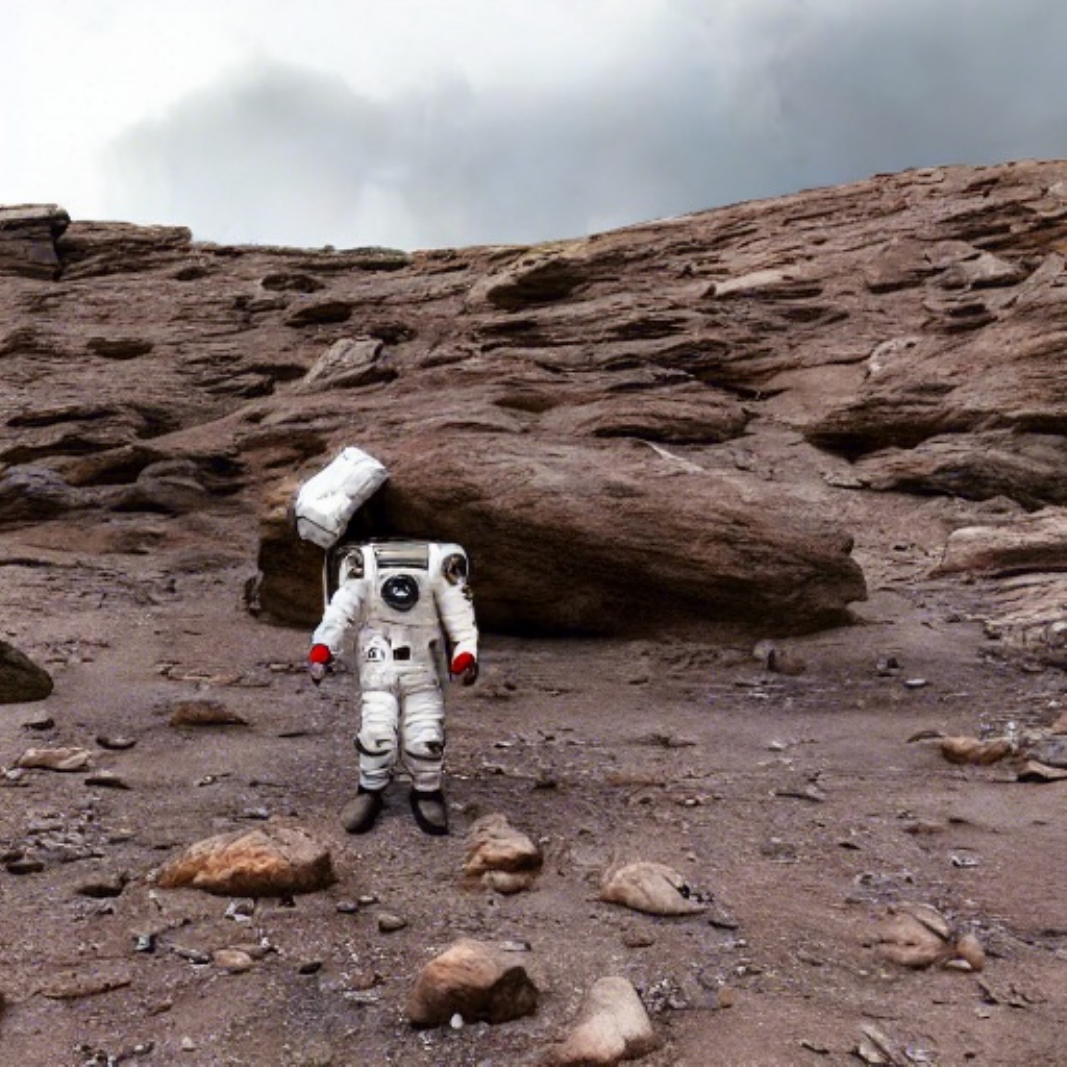}&
       \includegraphics[width=0.18\columnwidth]{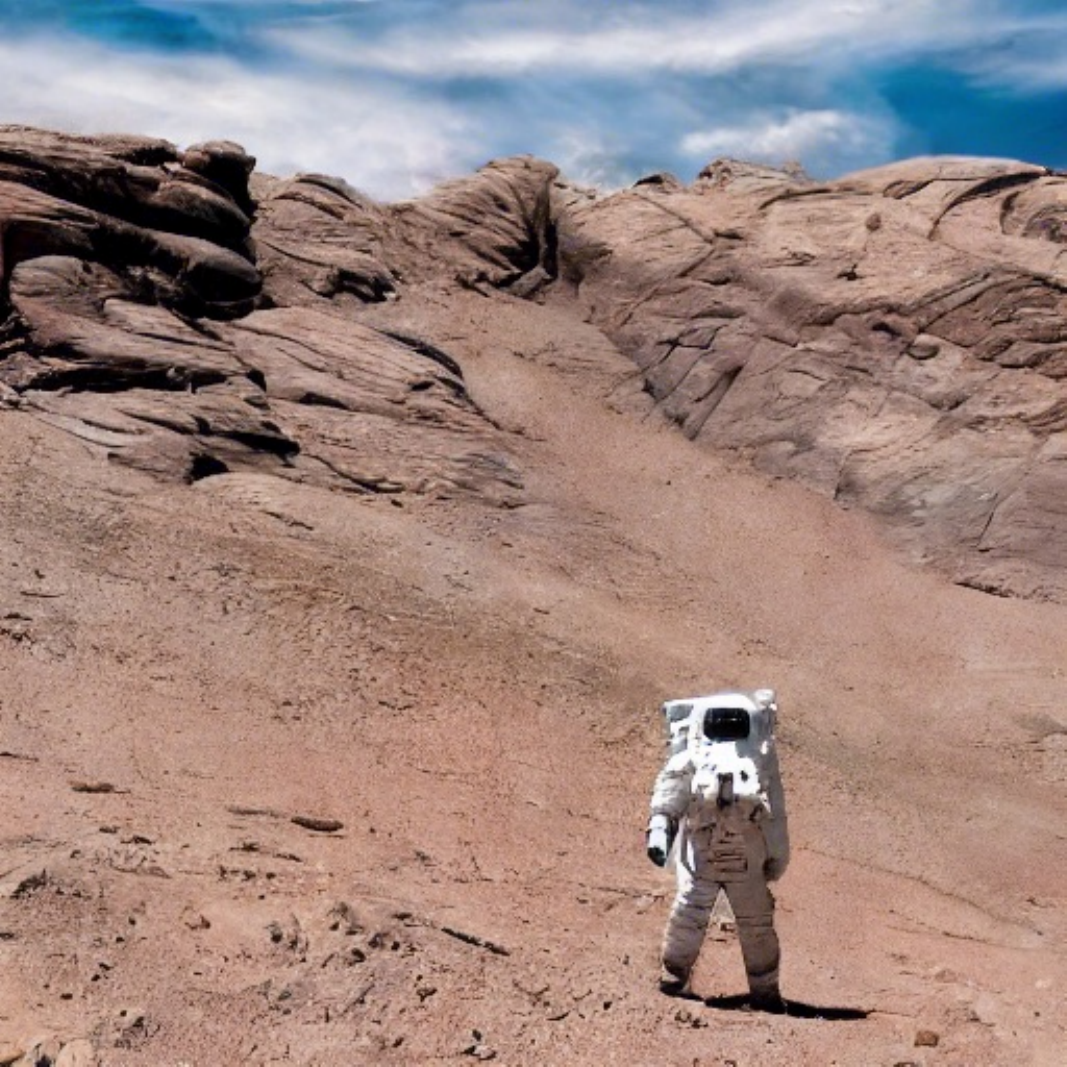}&
       \includegraphics[width=0.18\columnwidth]{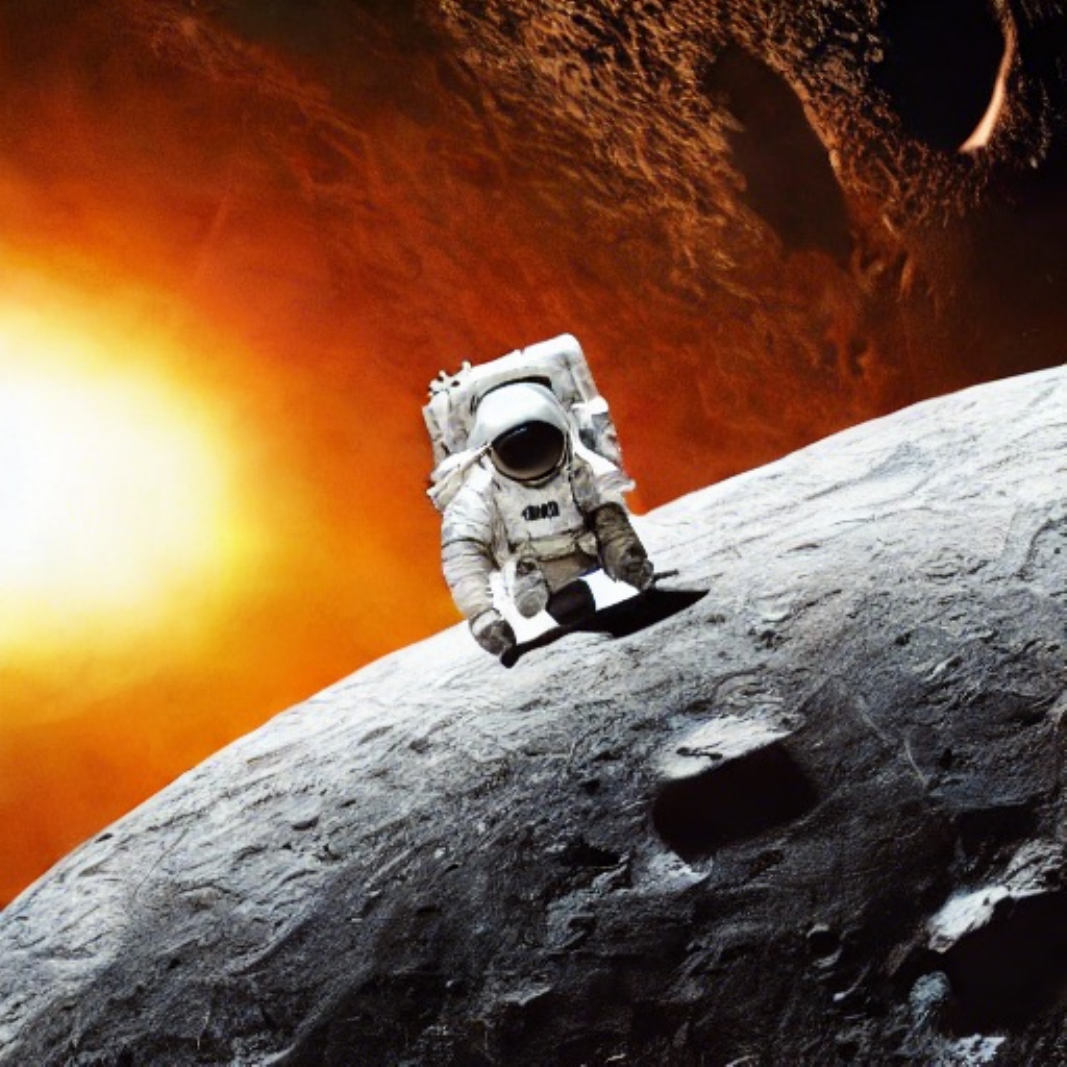}\\[-1pt]
       \includegraphics[width=0.18\columnwidth]{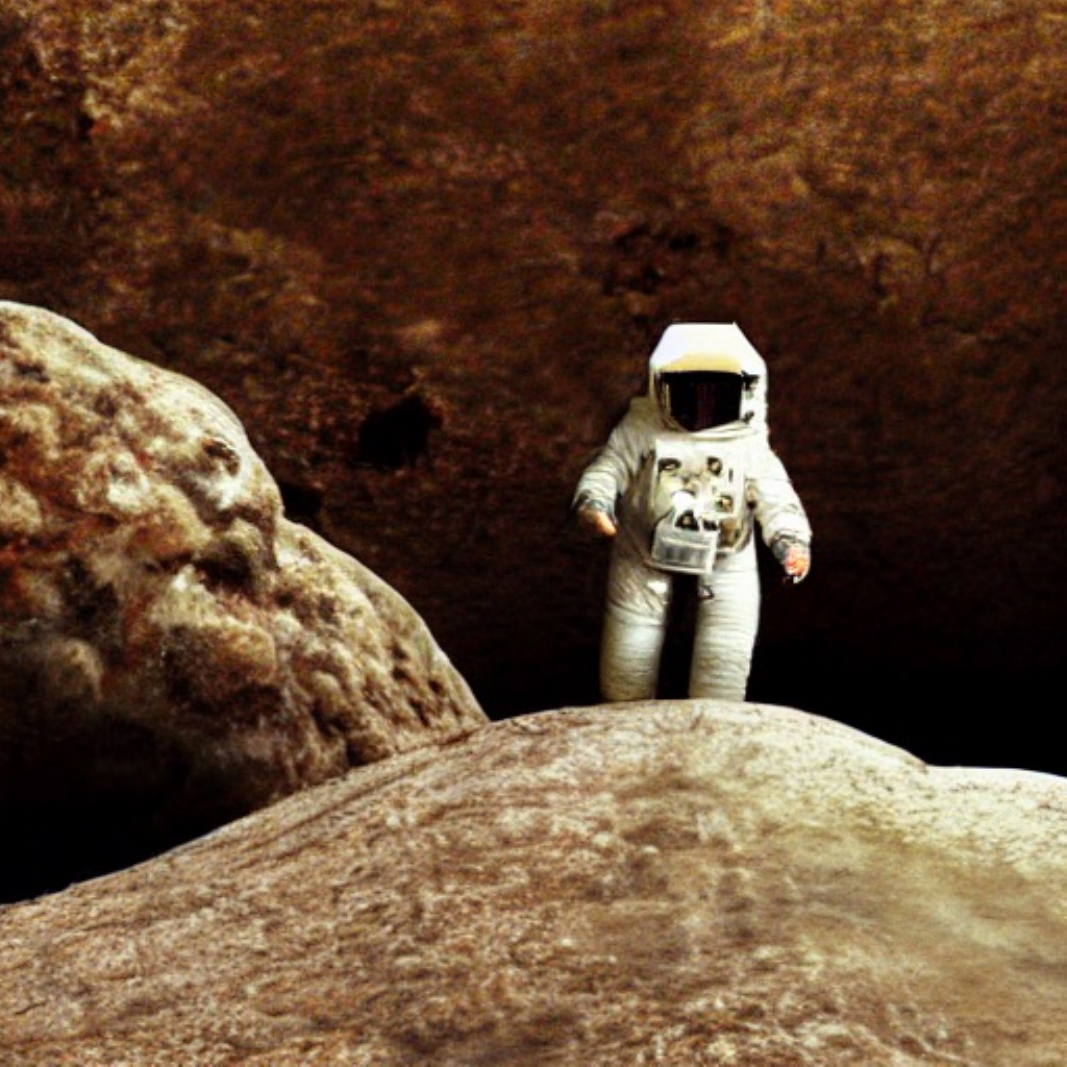}&
       \includegraphics[width=0.18\columnwidth]{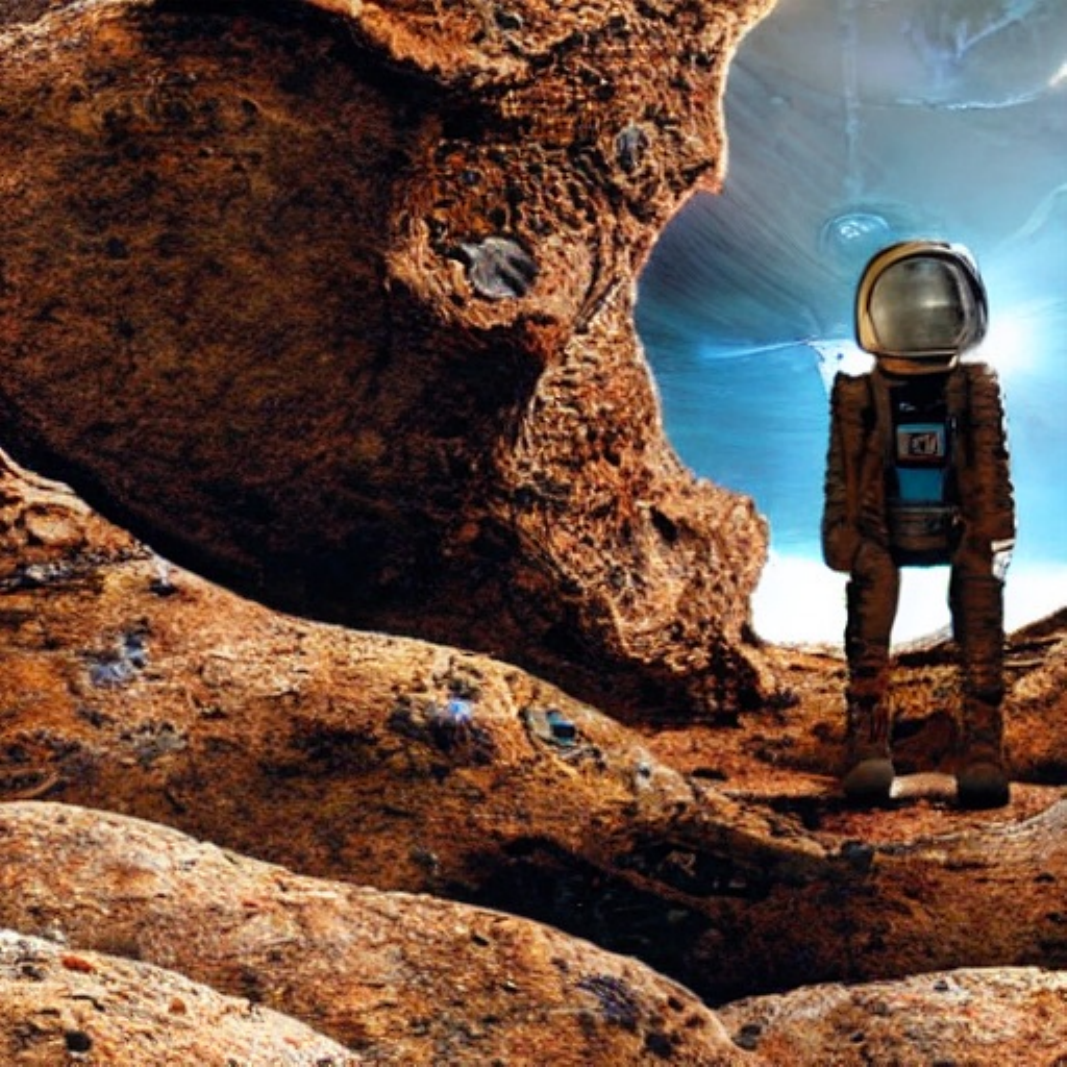}&
       \includegraphics[width=0.18\columnwidth]{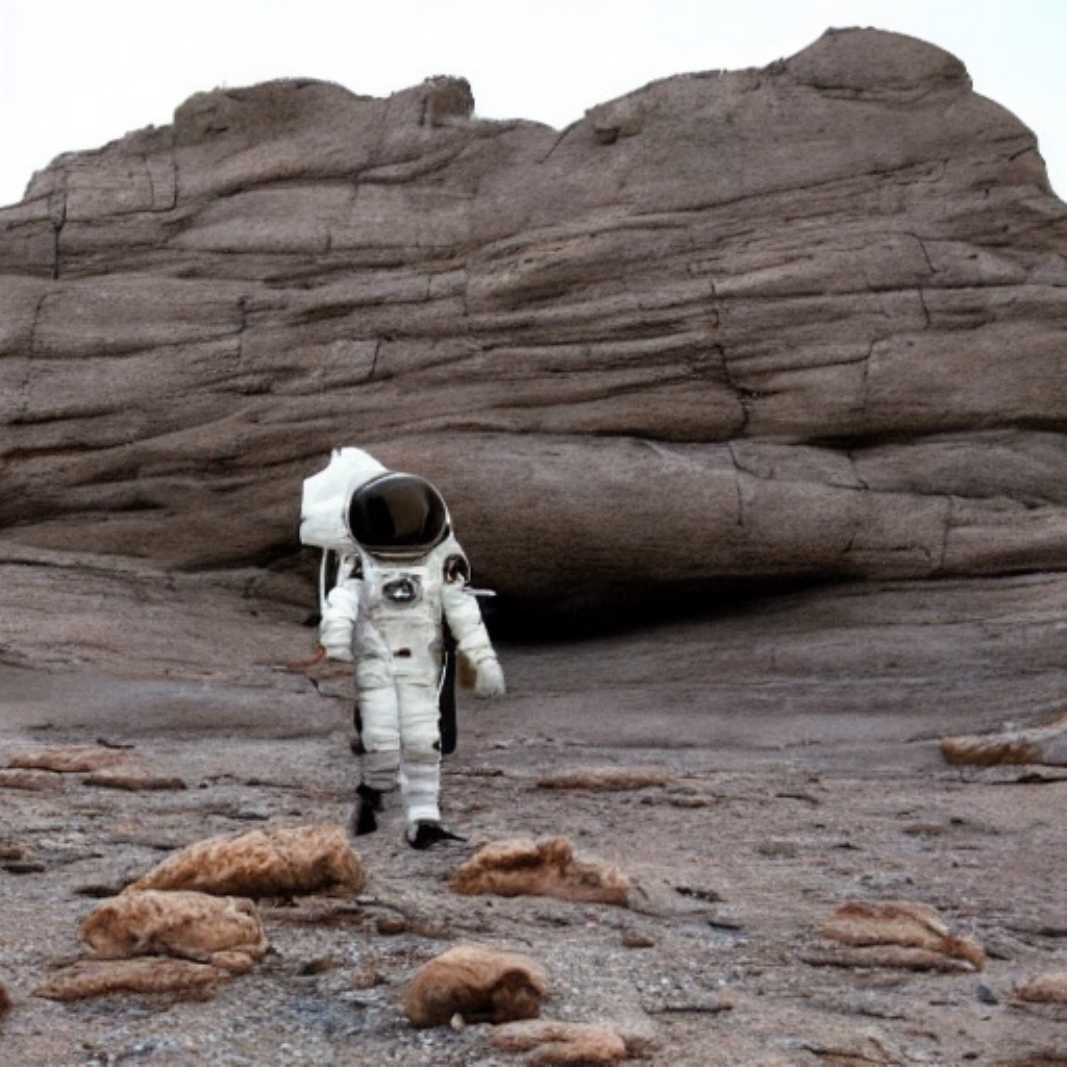}&
       \includegraphics[width=0.18\columnwidth]{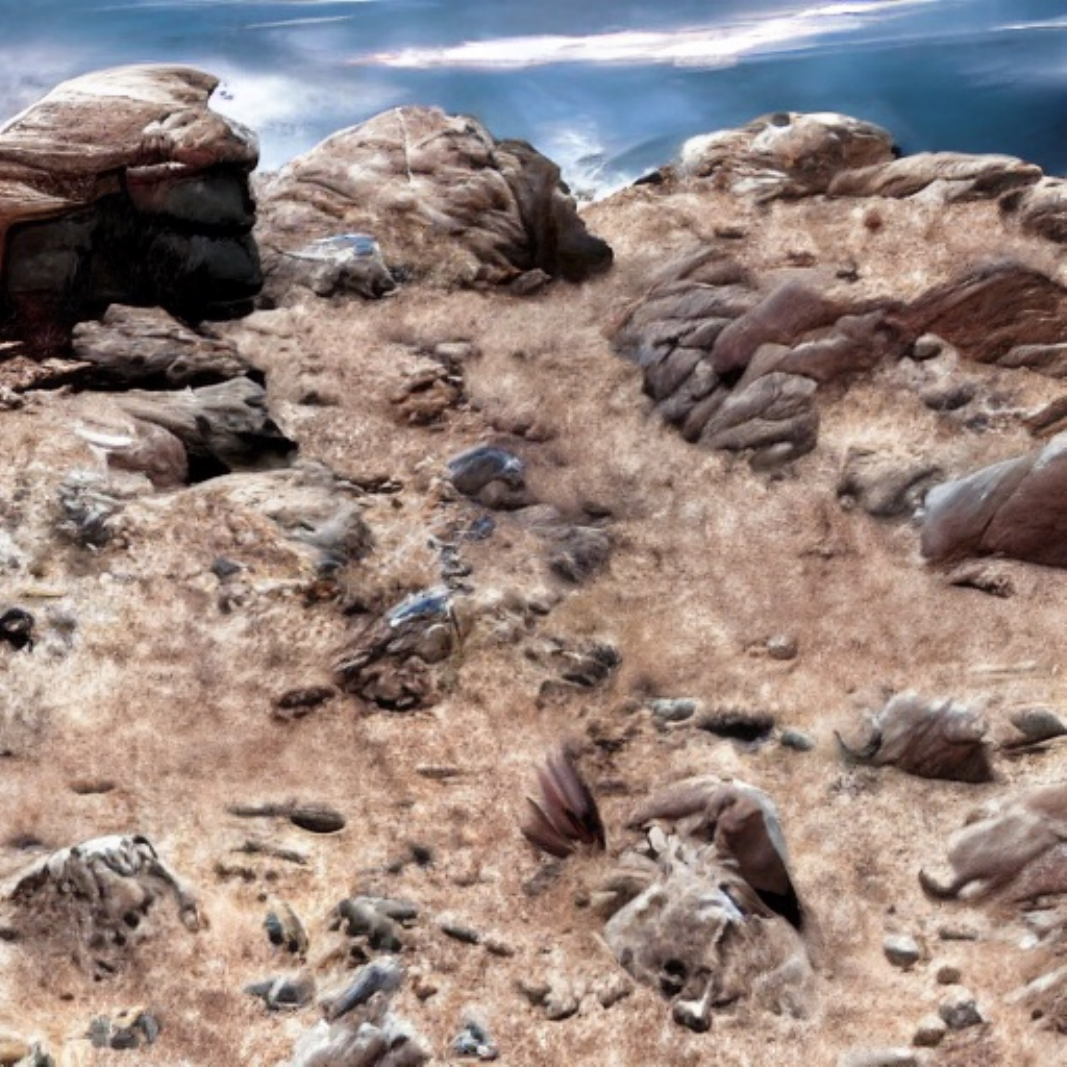}&
       \includegraphics[width=0.18\columnwidth]{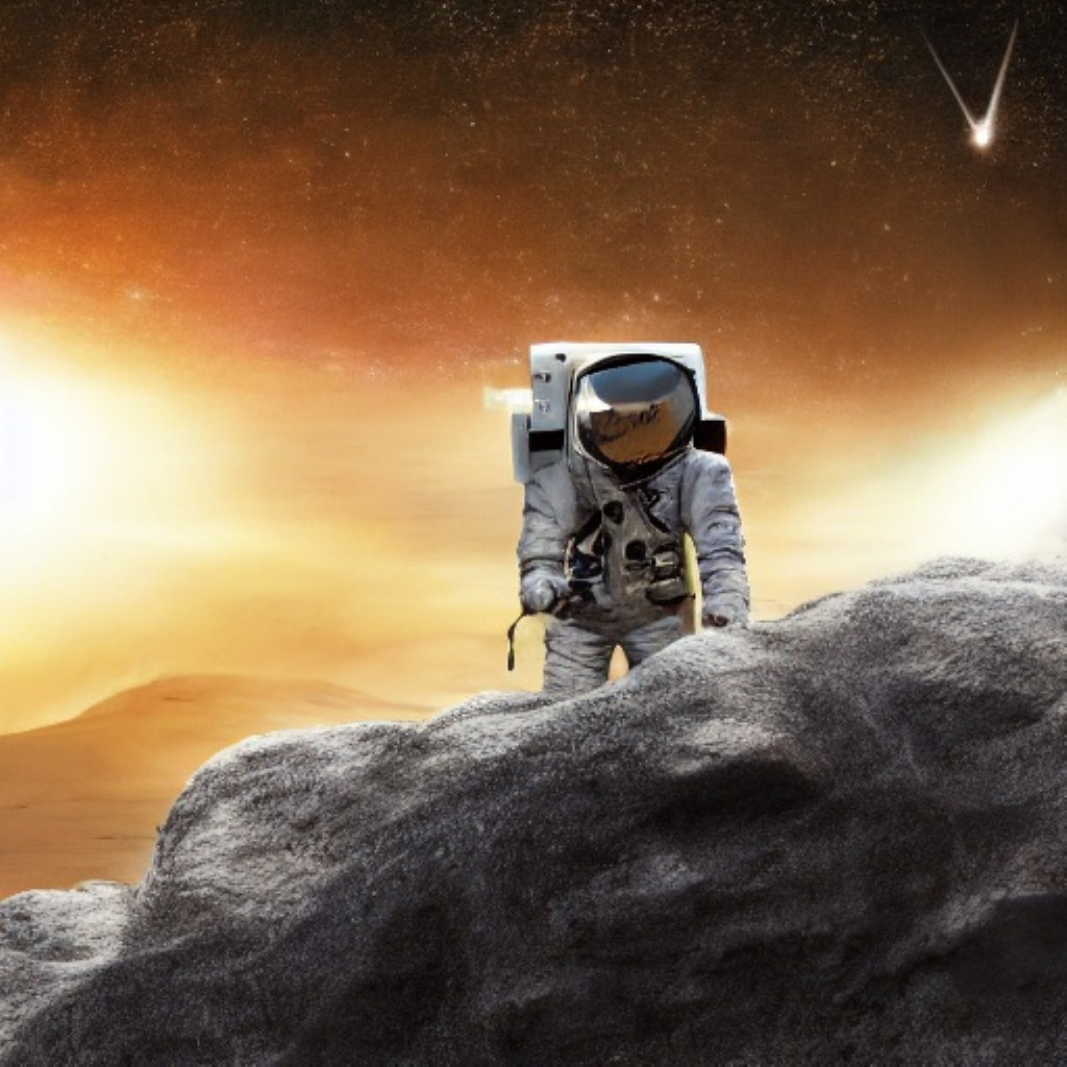}\\[-1pt]
       \includegraphics[width=0.18\columnwidth]{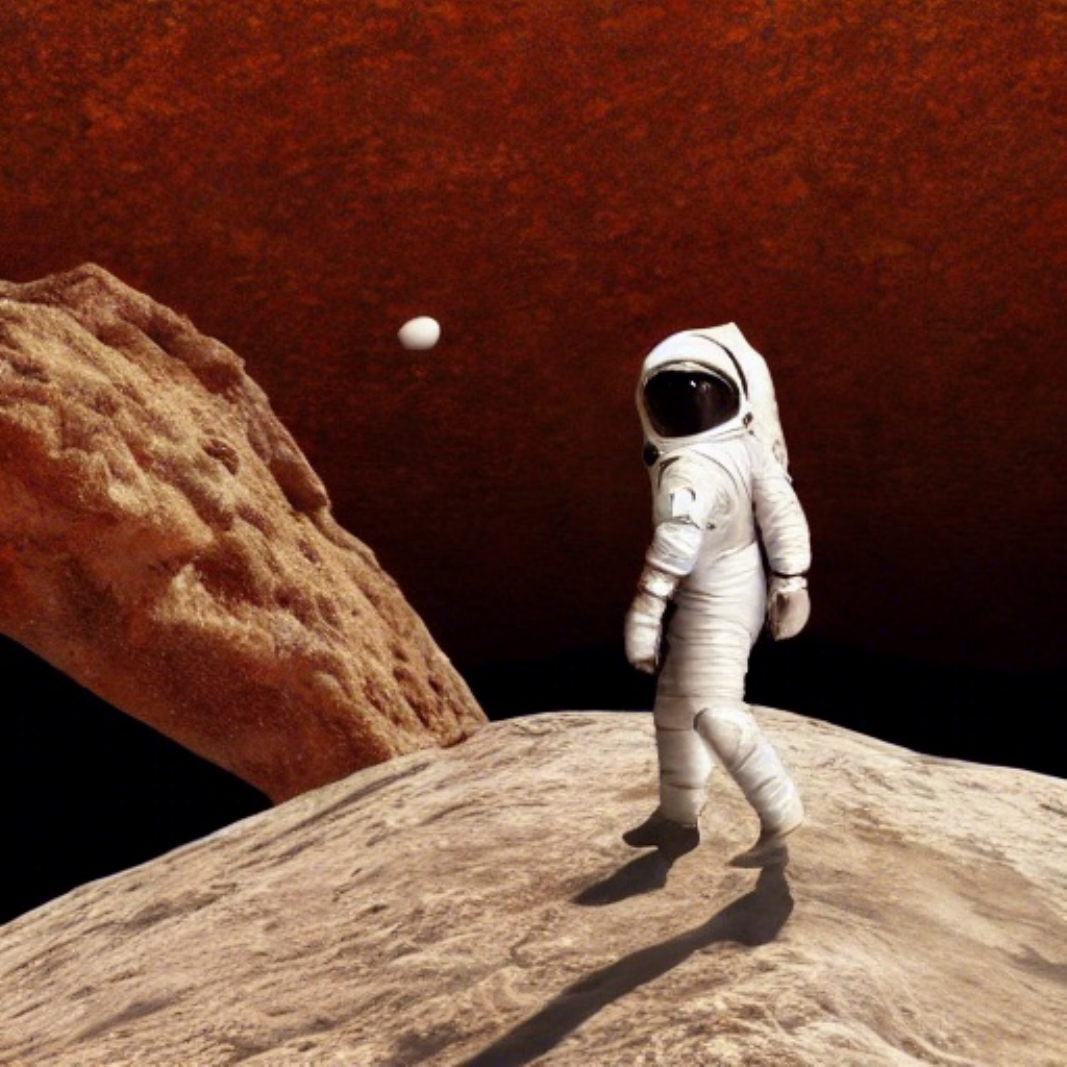}&
       \includegraphics[width=0.18\columnwidth]{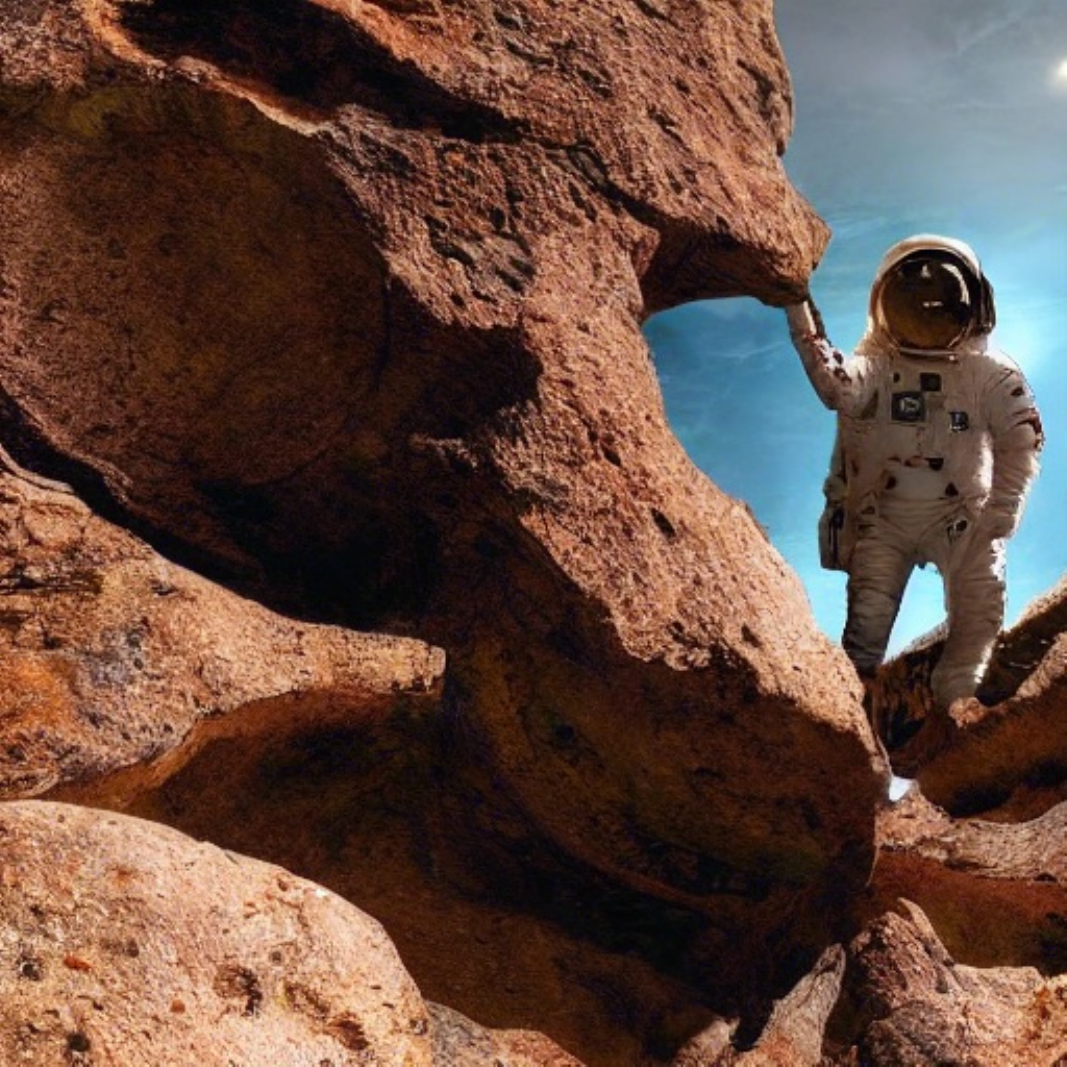}&
       \includegraphics[width=0.18\columnwidth]{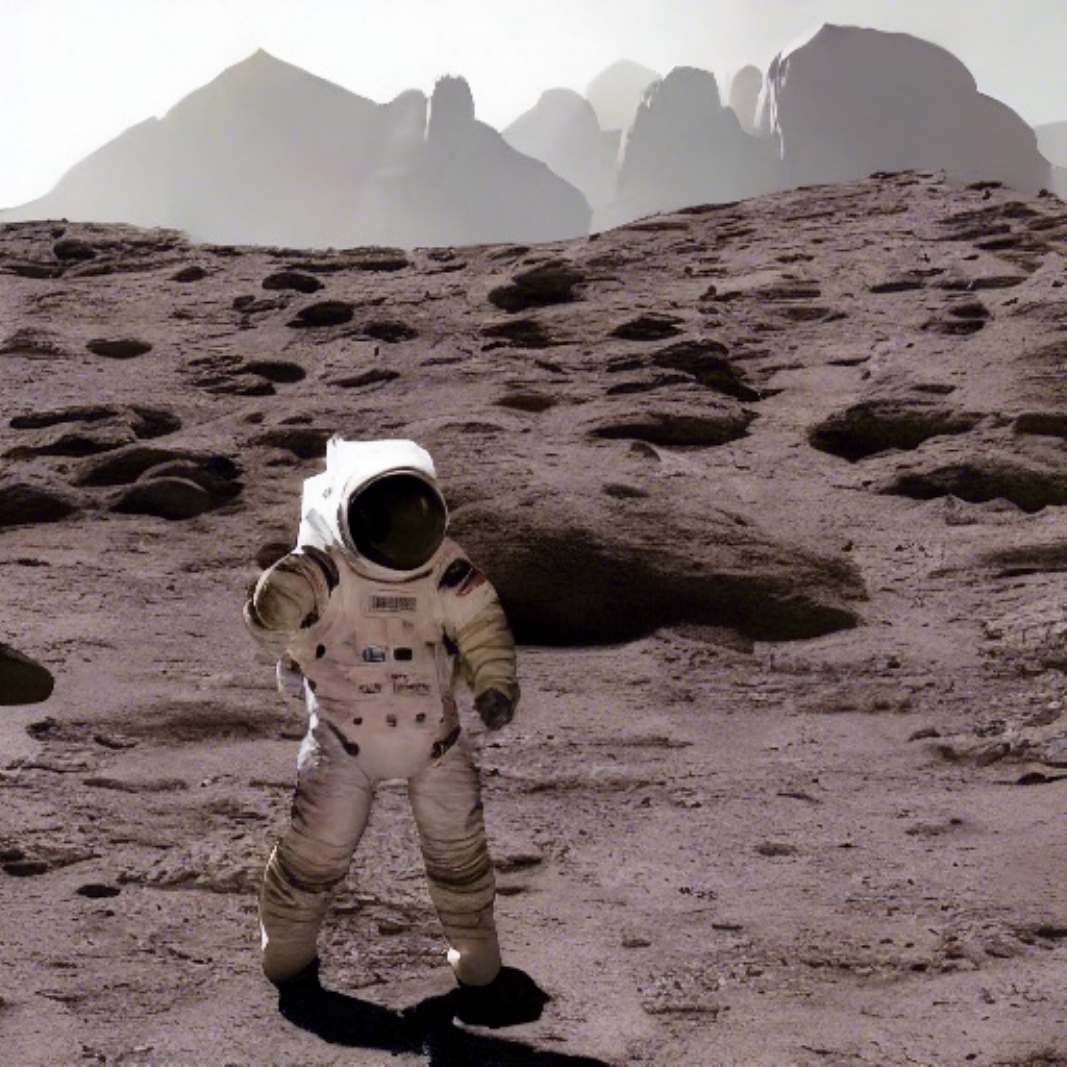}&
       \includegraphics[width=0.18\columnwidth]{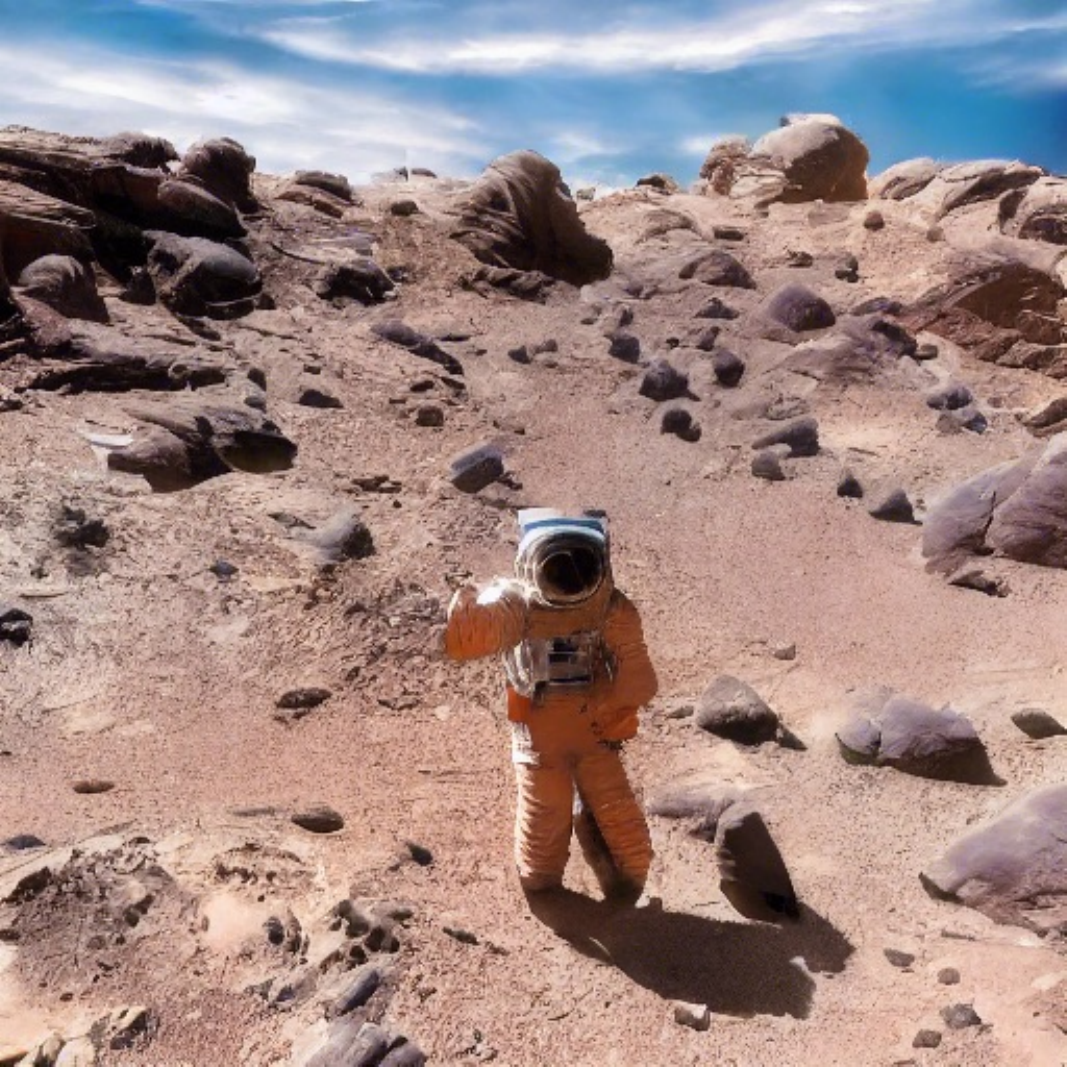}&
       \includegraphics[width=0.18\columnwidth]{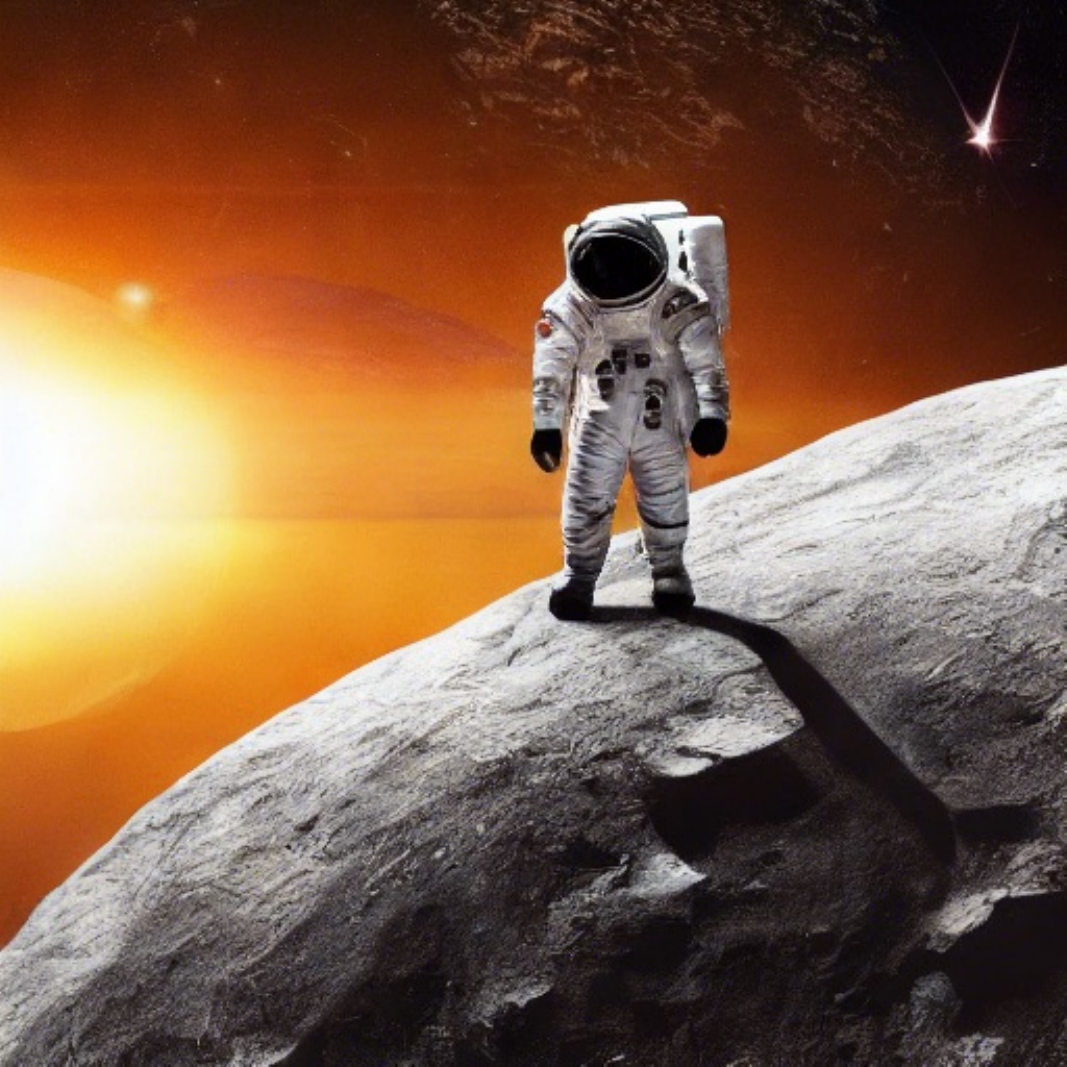}\\[-1pt]
     \end{tabular}
     \caption{\textit{``An astronaut exploring an alien planet with strange rock formations."}}
     \label{fig:astronaut}
   \end{subfigure}
 
   \vspace{5mm}

\begin{subfigure}[c]{\linewidth}
   \centering
   \begin{tabular}{@{}c@{\hspace{1mm}}c@{\hspace{1mm}}c@{\hspace{1mm}}c@{\hspace{1mm}}c@{}}
       \includegraphics[width=0.18\columnwidth]{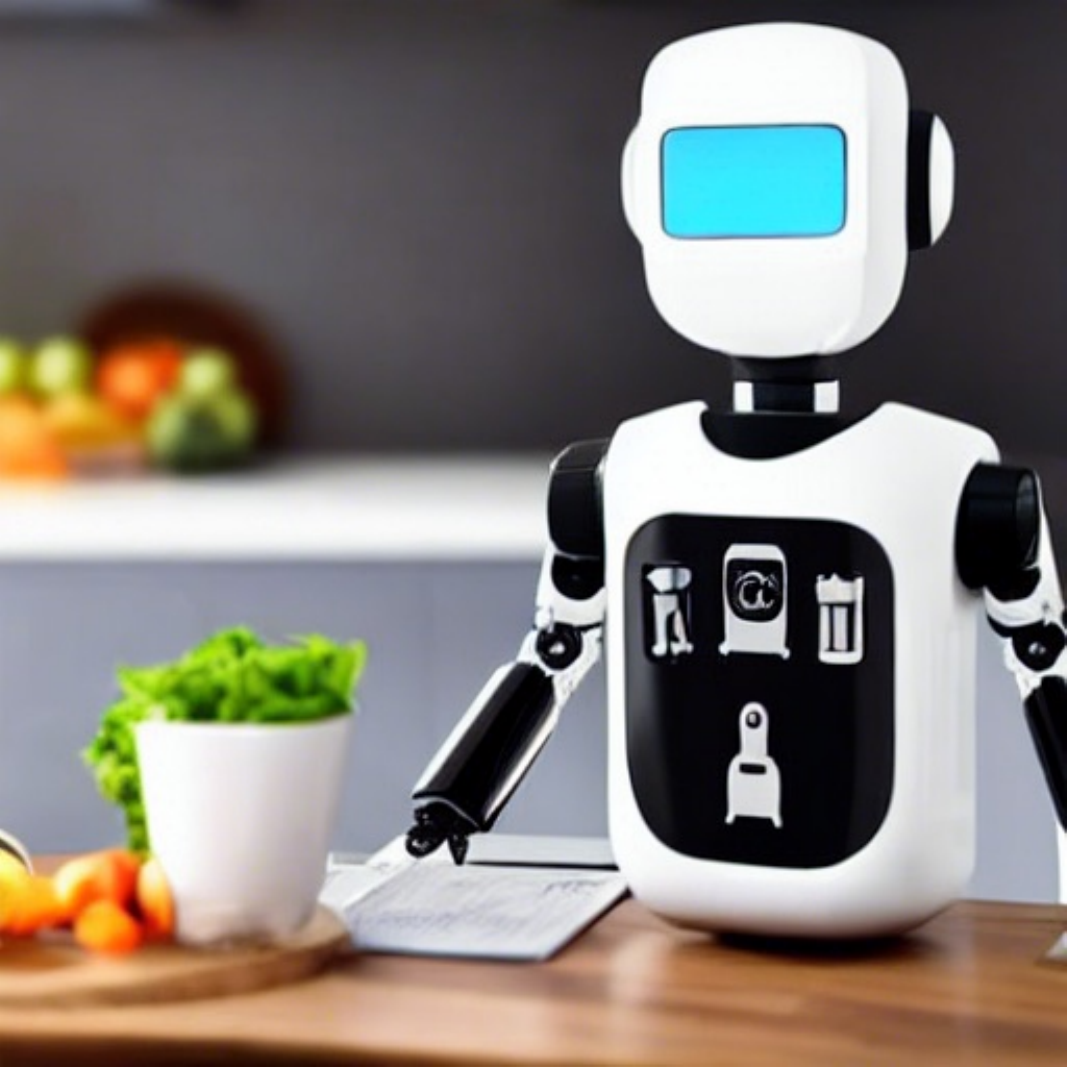}&
       \includegraphics[width=0.18\columnwidth]{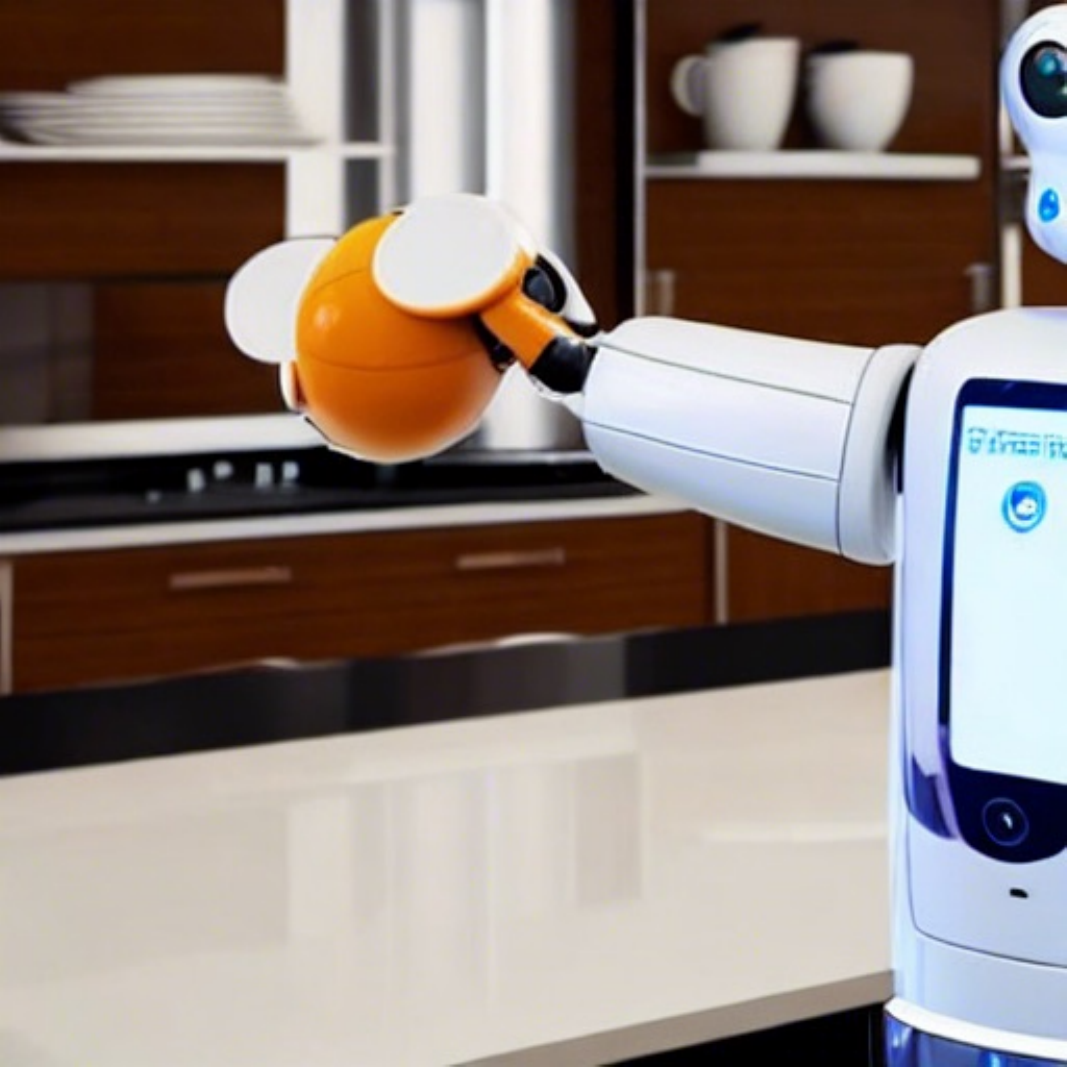}&
       \includegraphics[width=0.18\columnwidth]{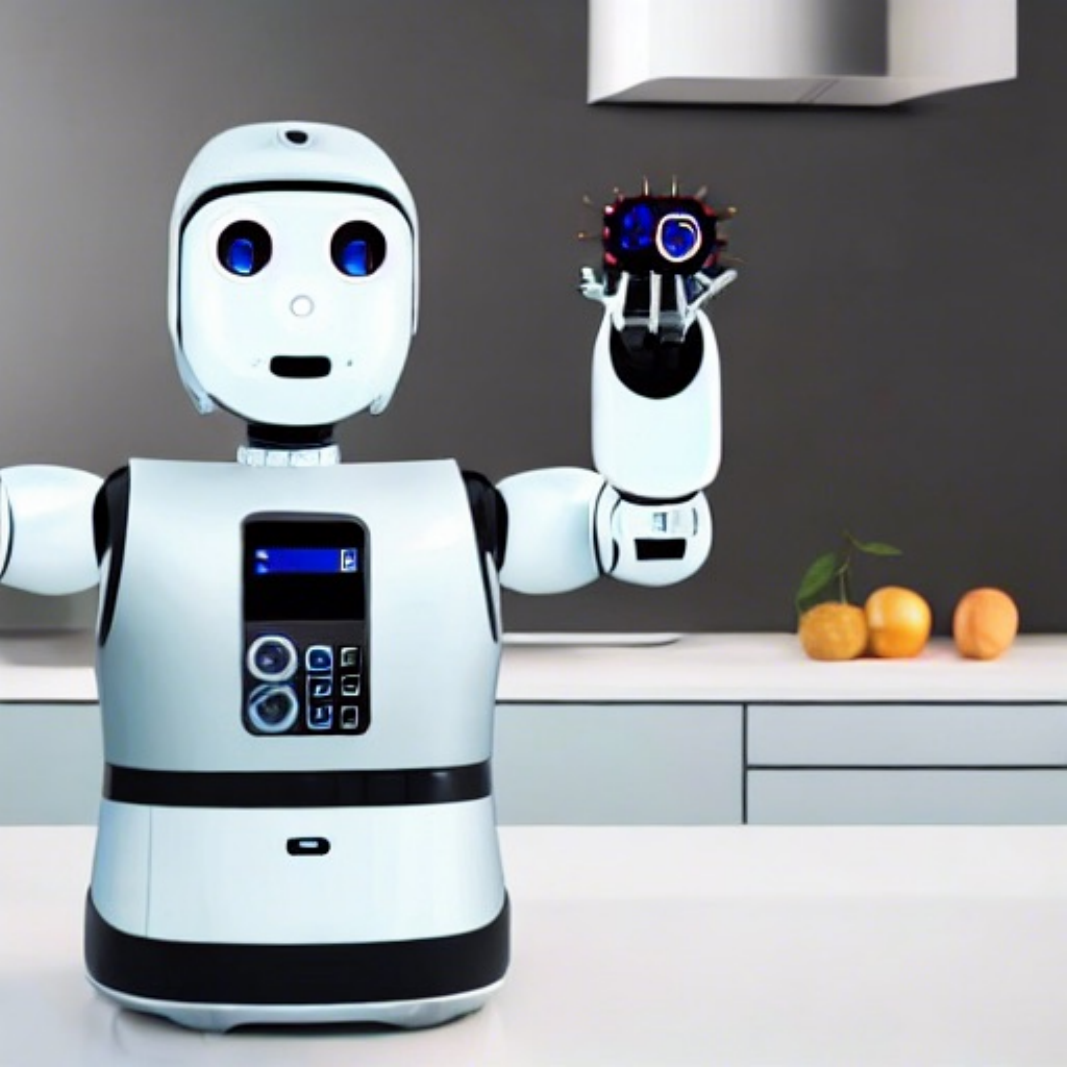}&
       \includegraphics[width=0.18\columnwidth]{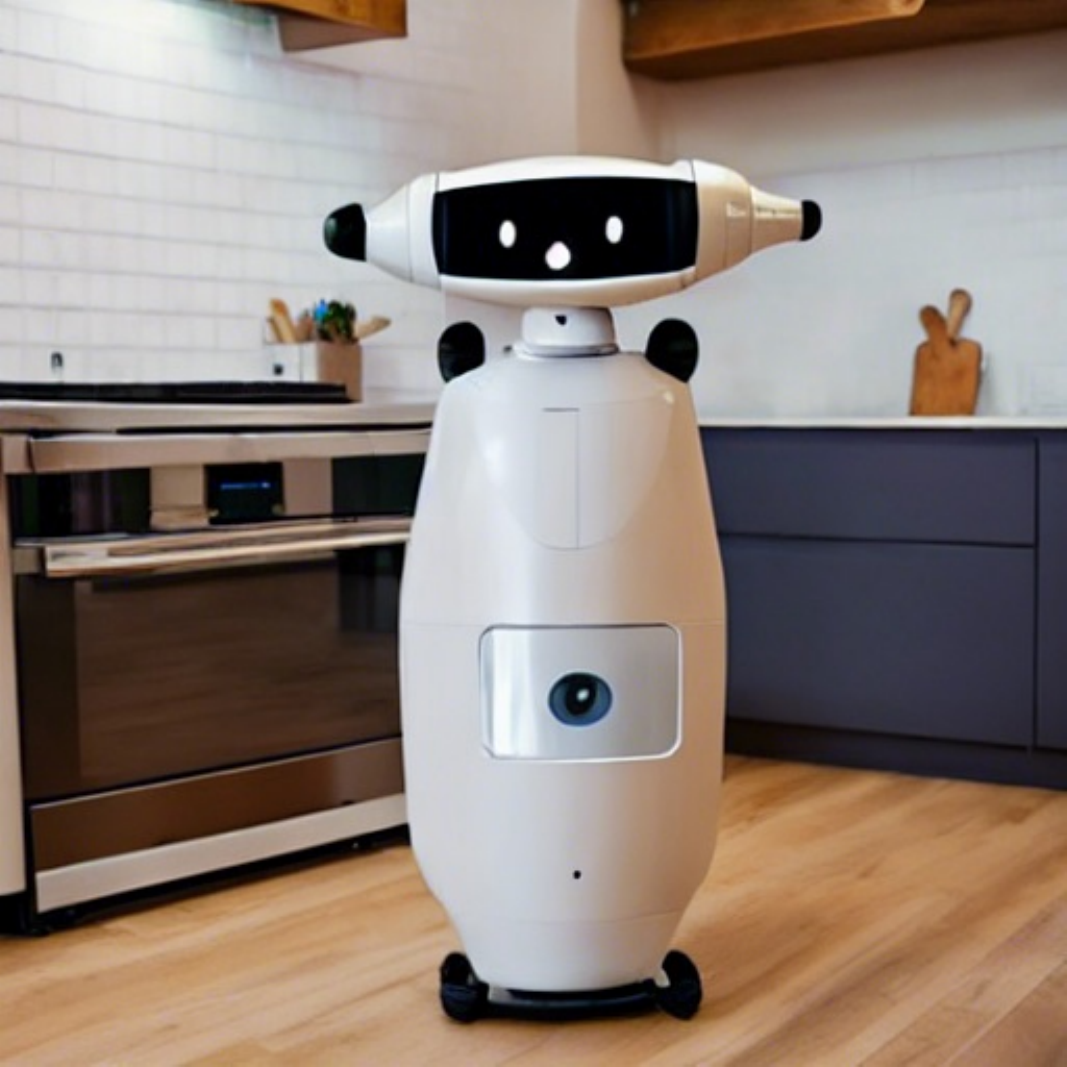}&
       \includegraphics[width=0.18\columnwidth]{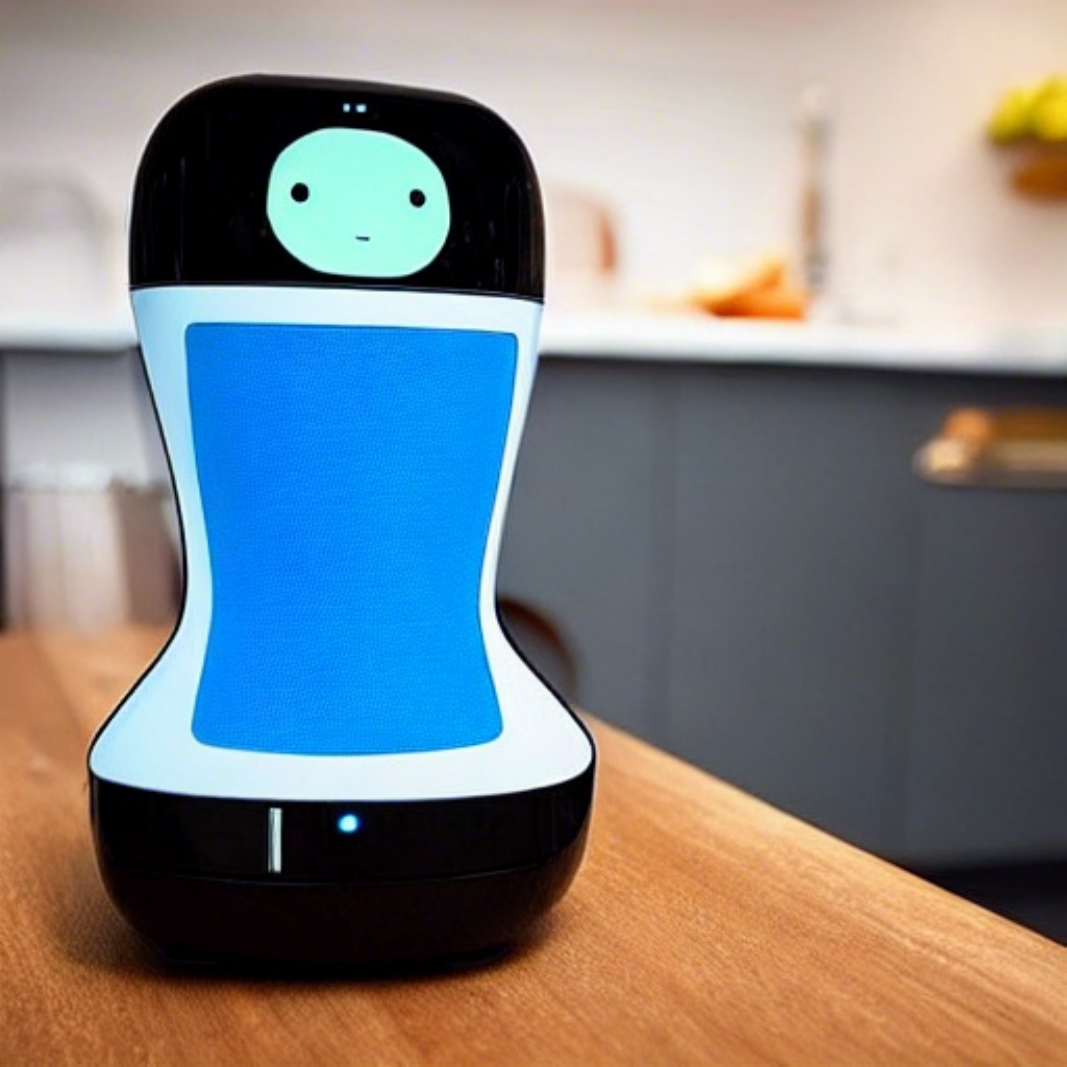}\\[-1pt]
       \includegraphics[width=0.18\columnwidth]{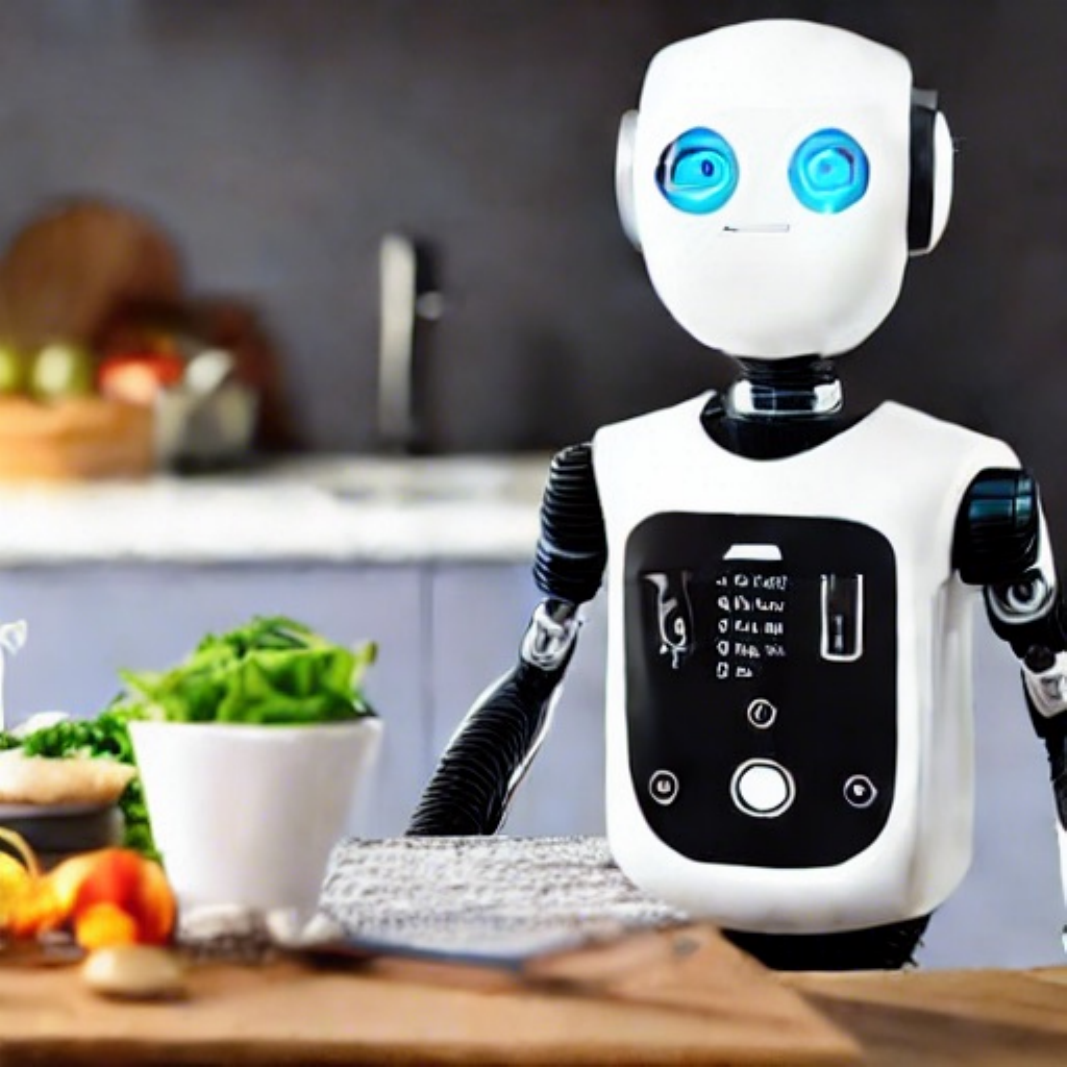}&
       \includegraphics[width=0.18\columnwidth]{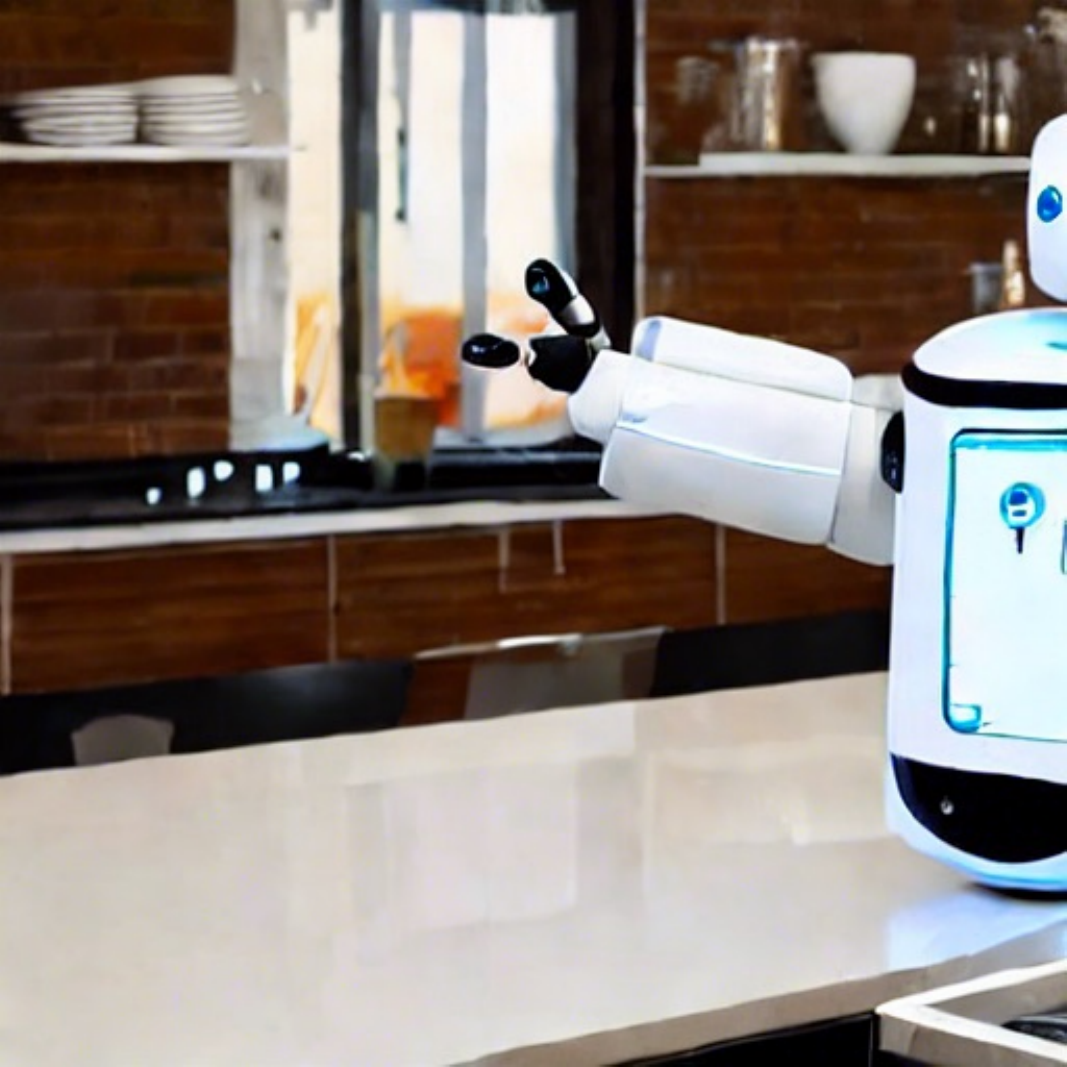}&
       \includegraphics[width=0.18\columnwidth]{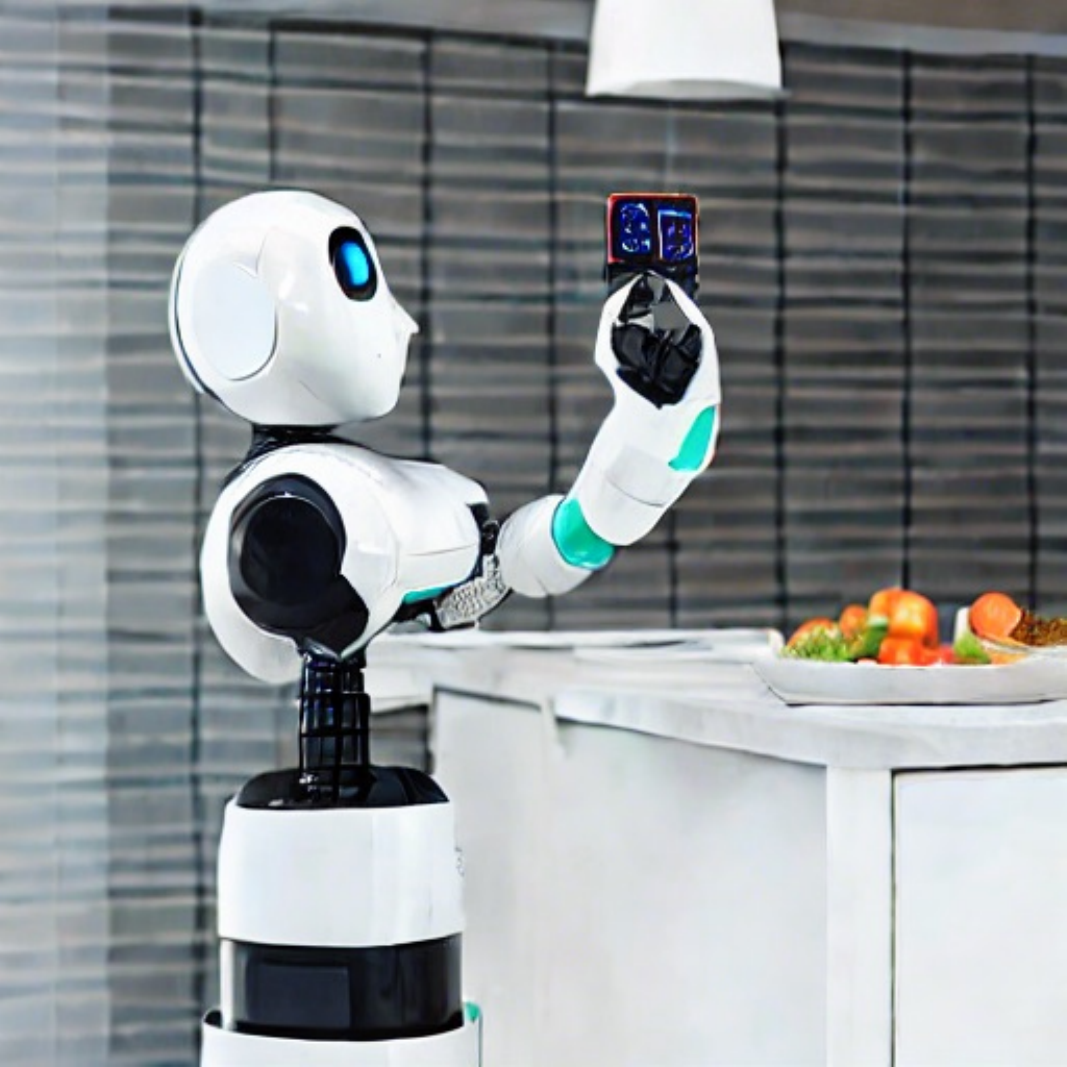}&
       \includegraphics[width=0.18\columnwidth]{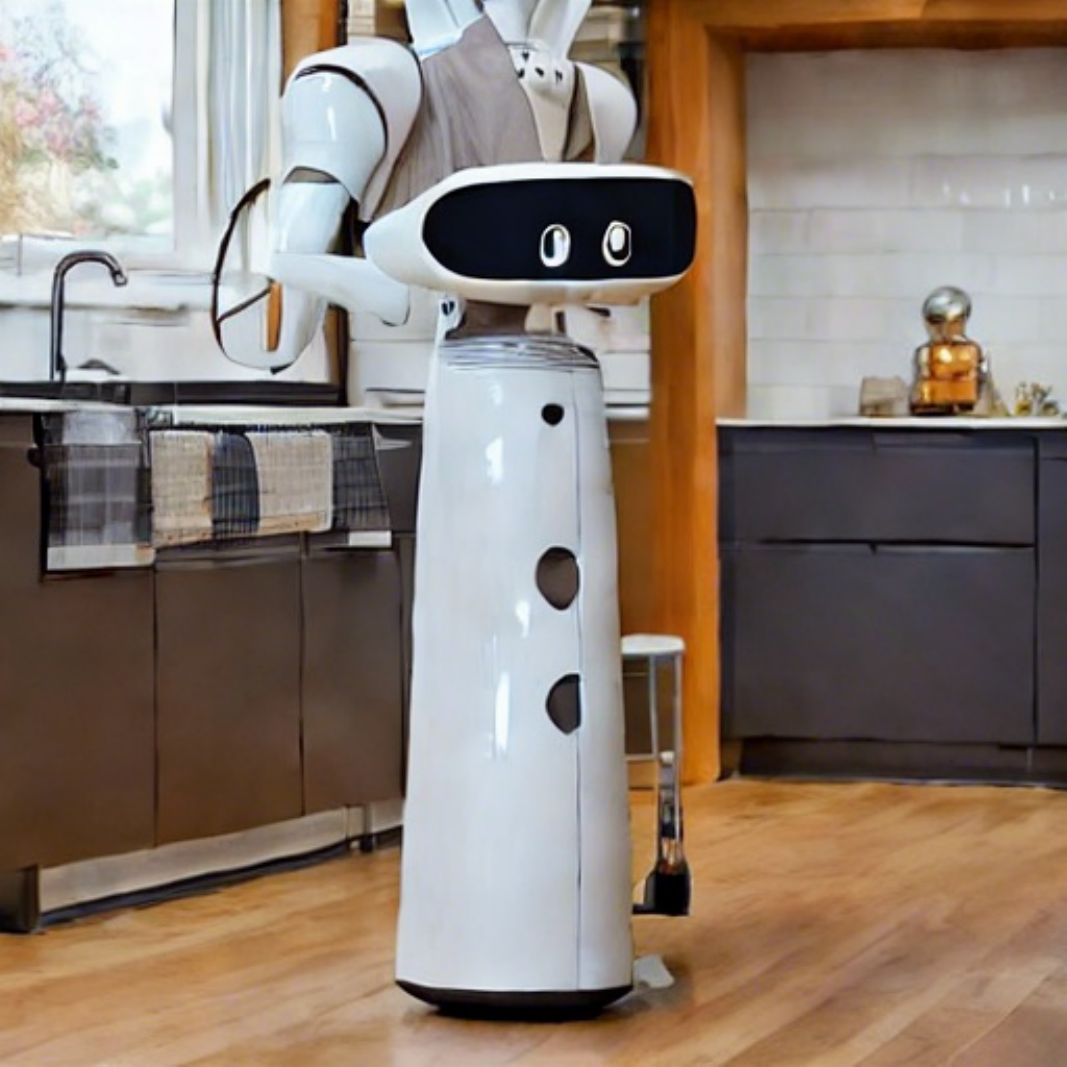}&
       \includegraphics[width=0.18\columnwidth]{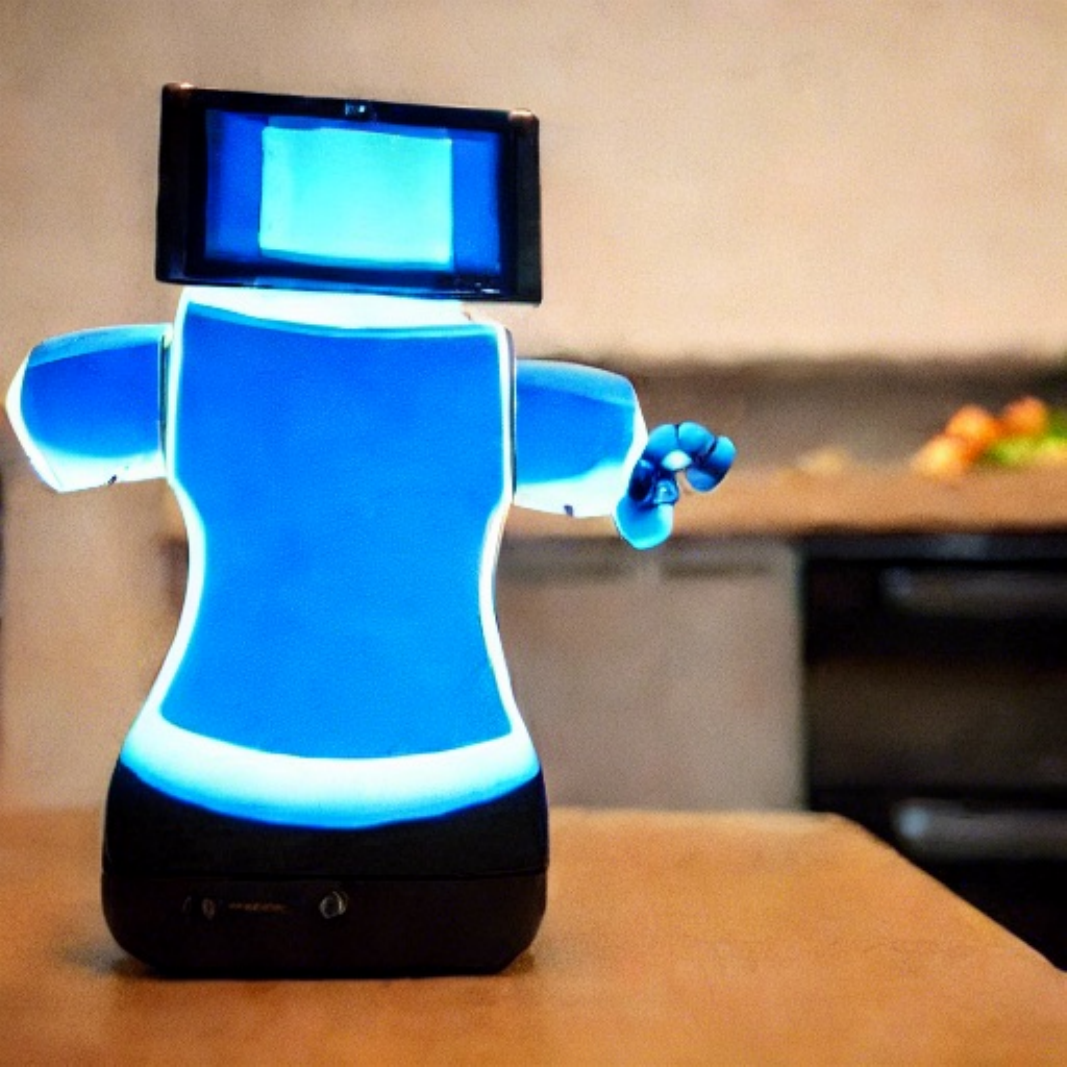}\\[-1pt]
       \includegraphics[width=0.18\columnwidth]{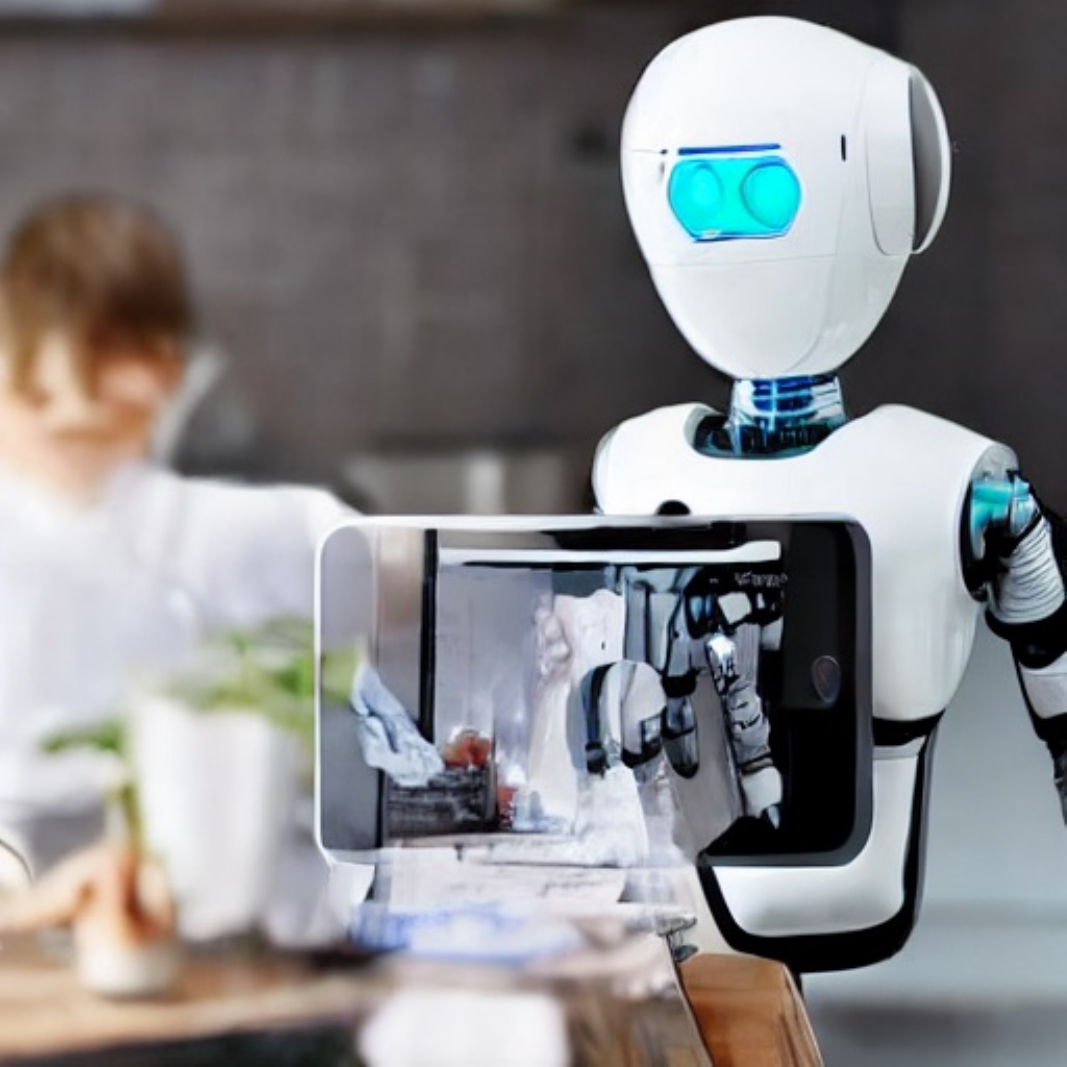}&
       \includegraphics[width=0.18\columnwidth]{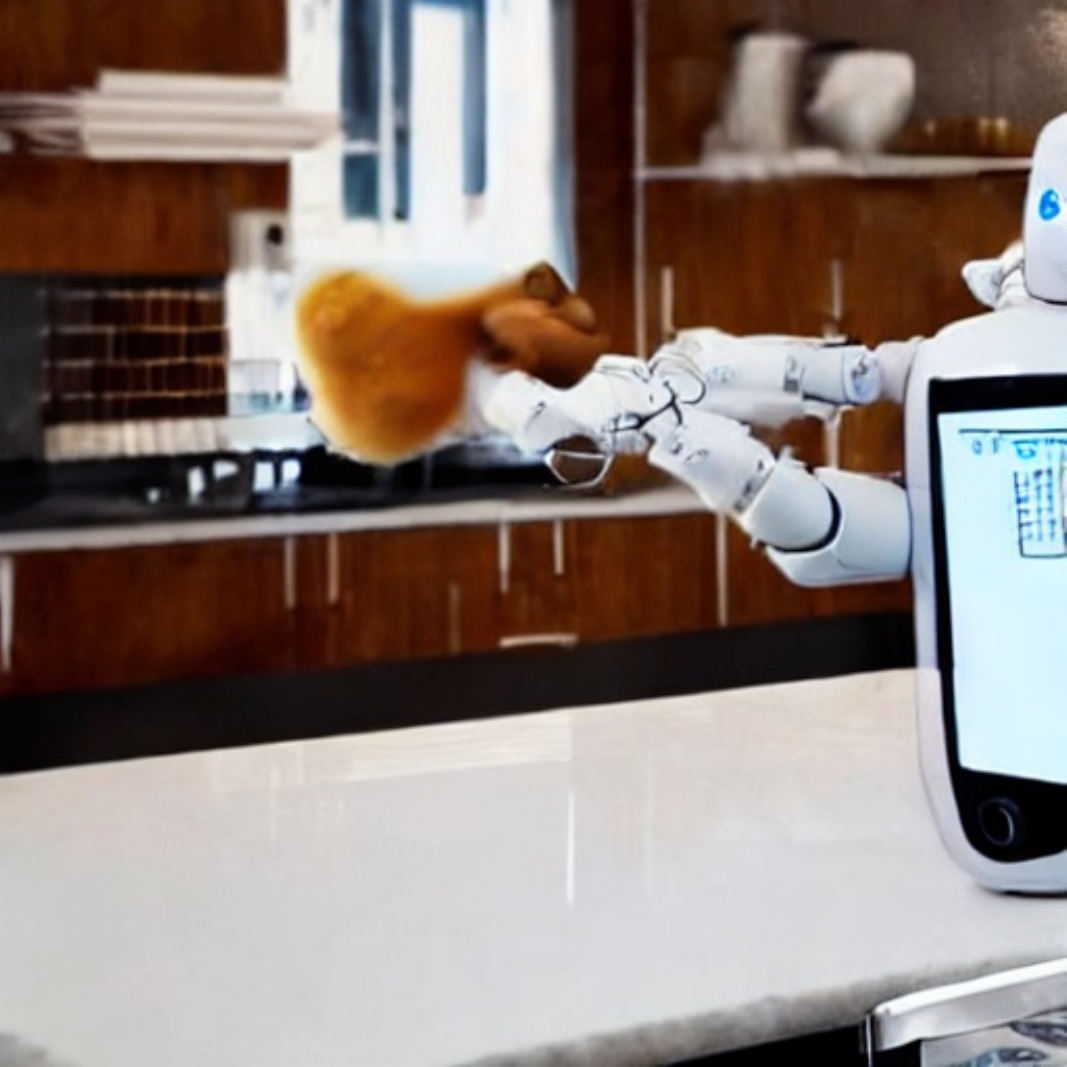}&
       \includegraphics[width=0.18\columnwidth]{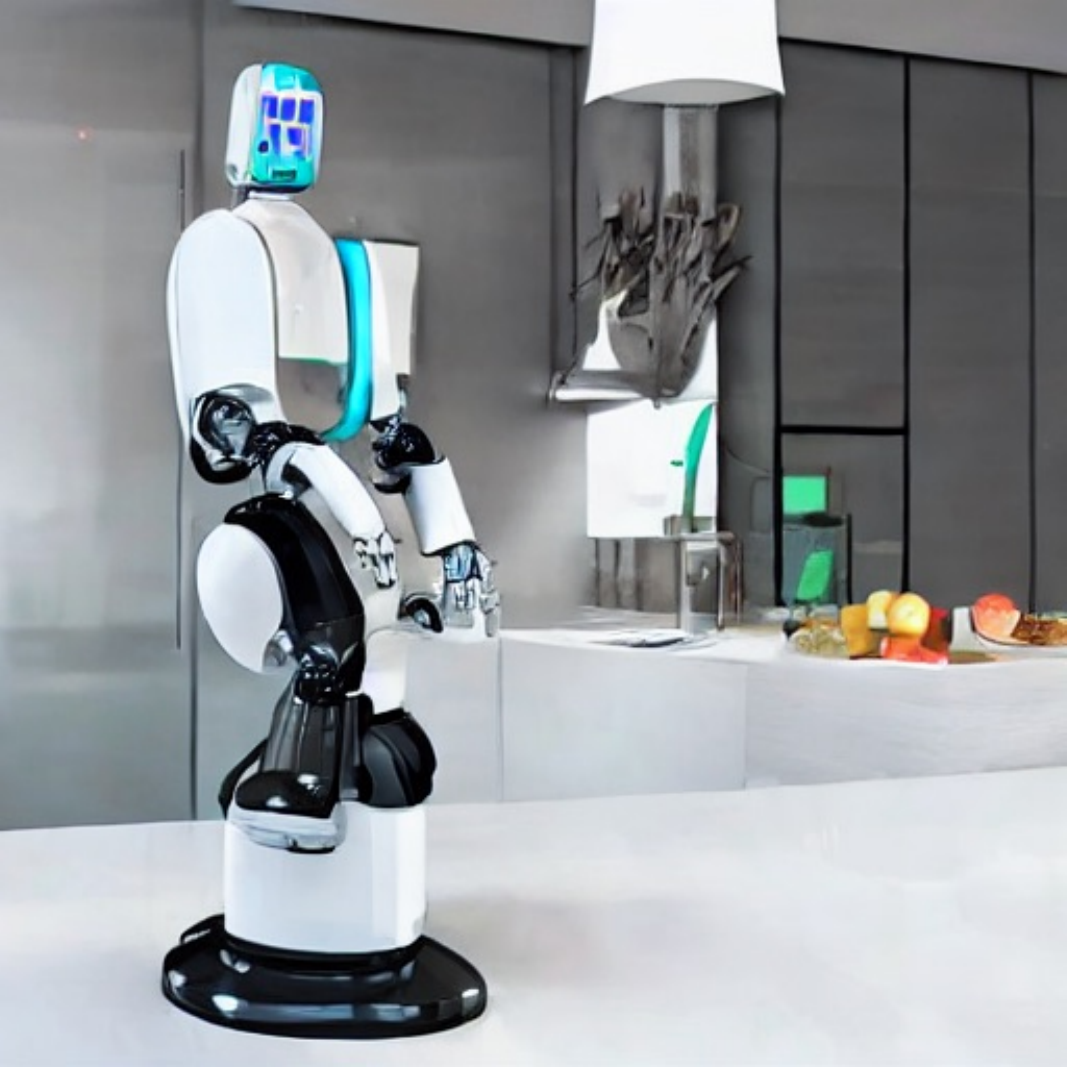}&
       \includegraphics[width=0.18\columnwidth]{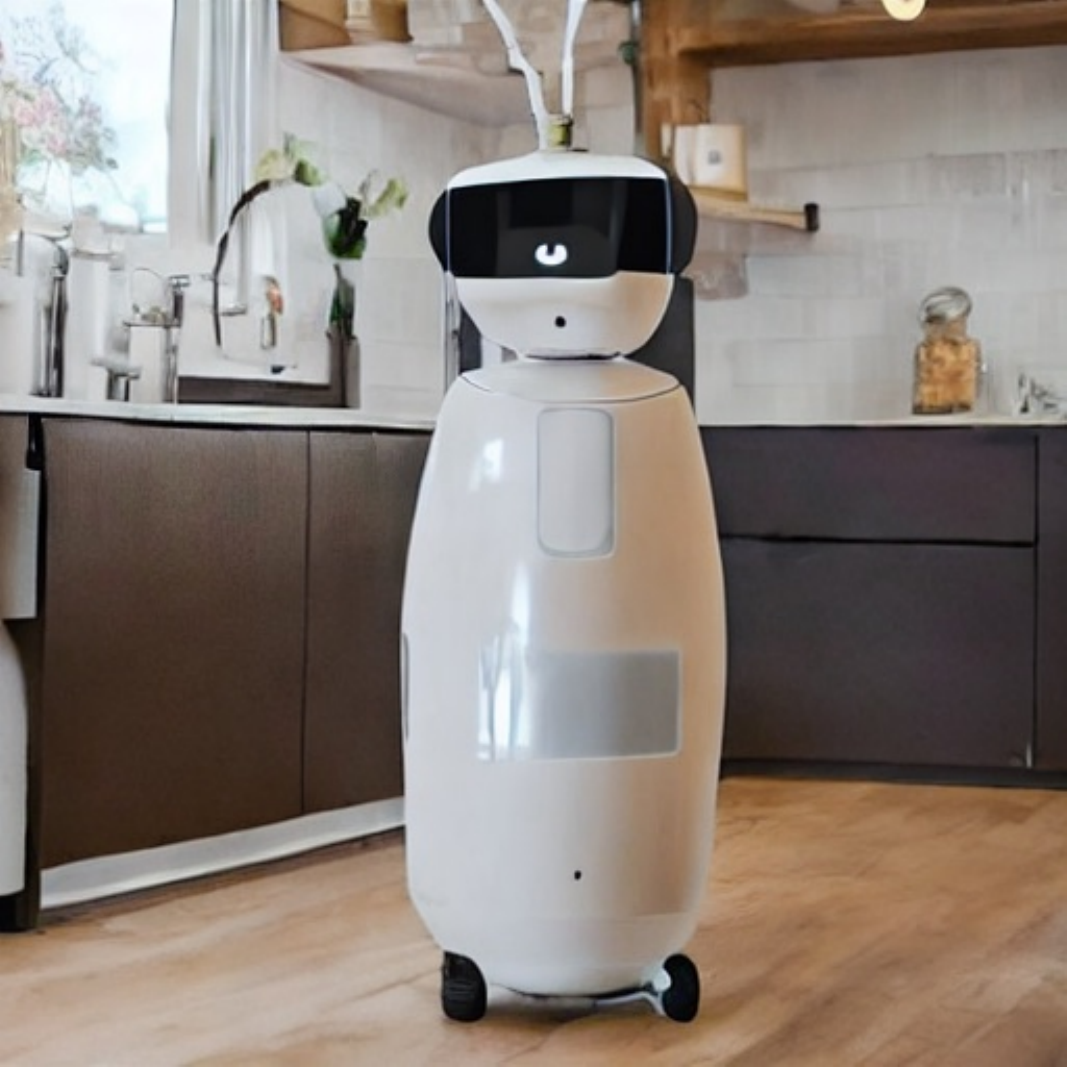}&
       \includegraphics[width=0.18\columnwidth]{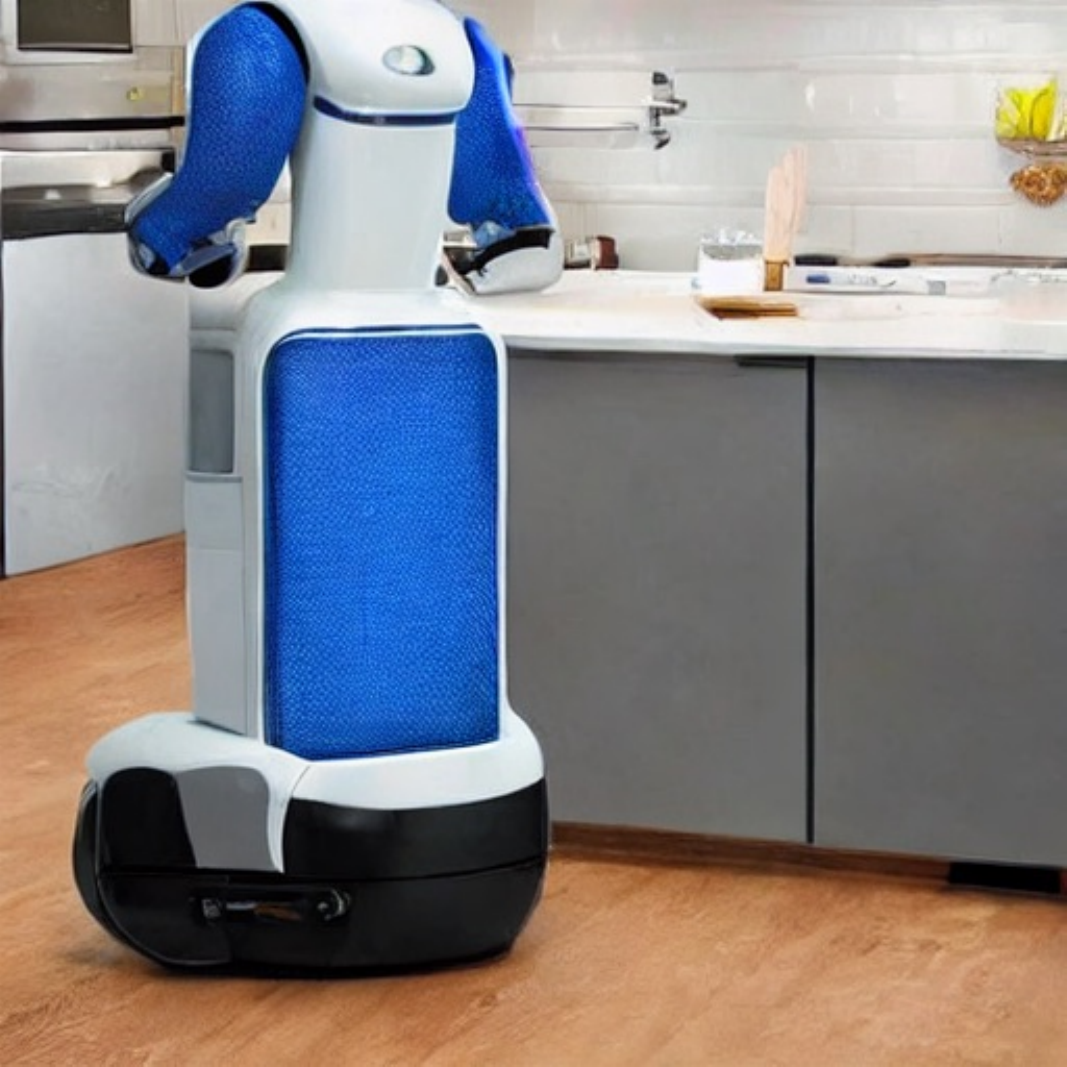}\\[-1pt]
       \includegraphics[width=0.18\columnwidth]{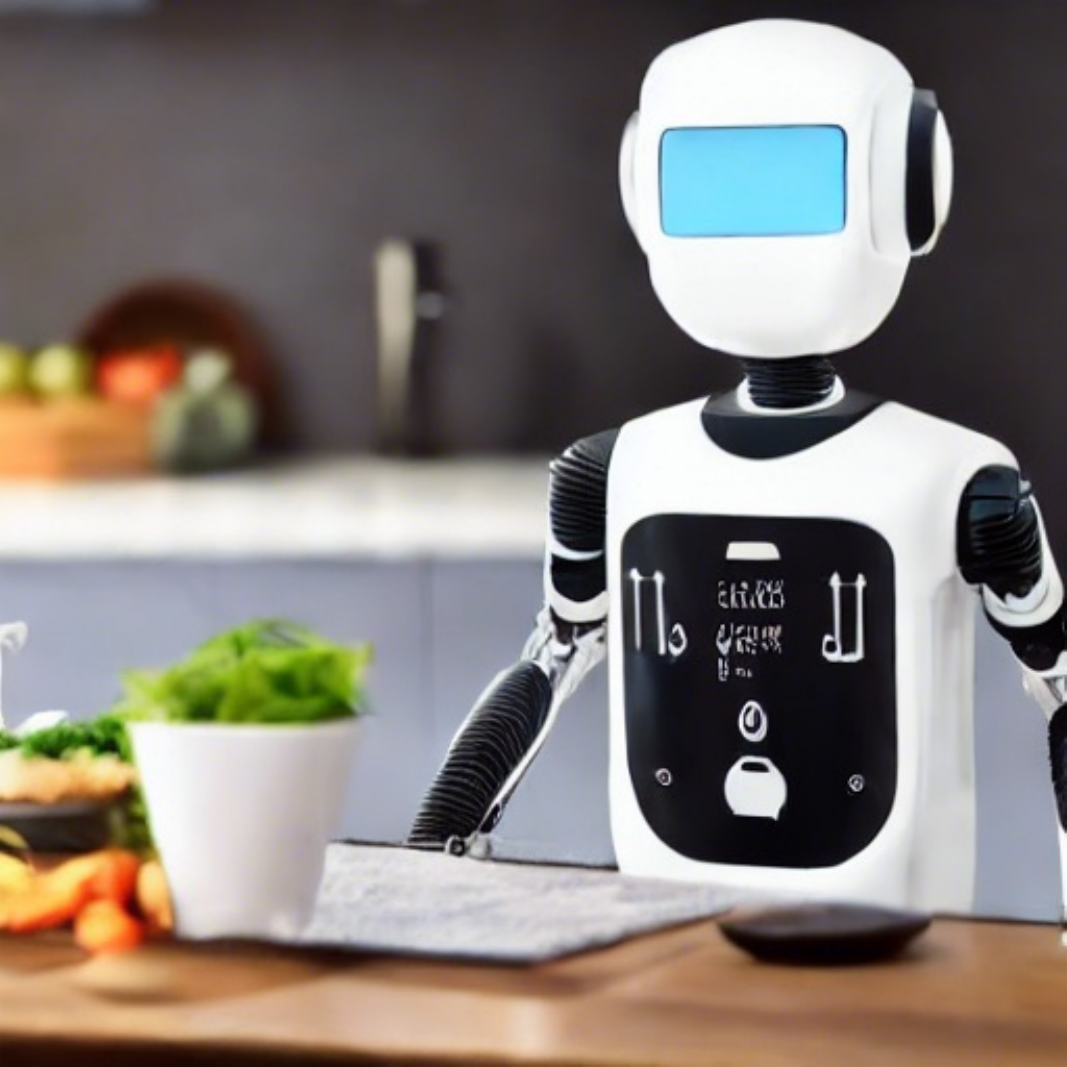}&
       \includegraphics[width=0.18\columnwidth]{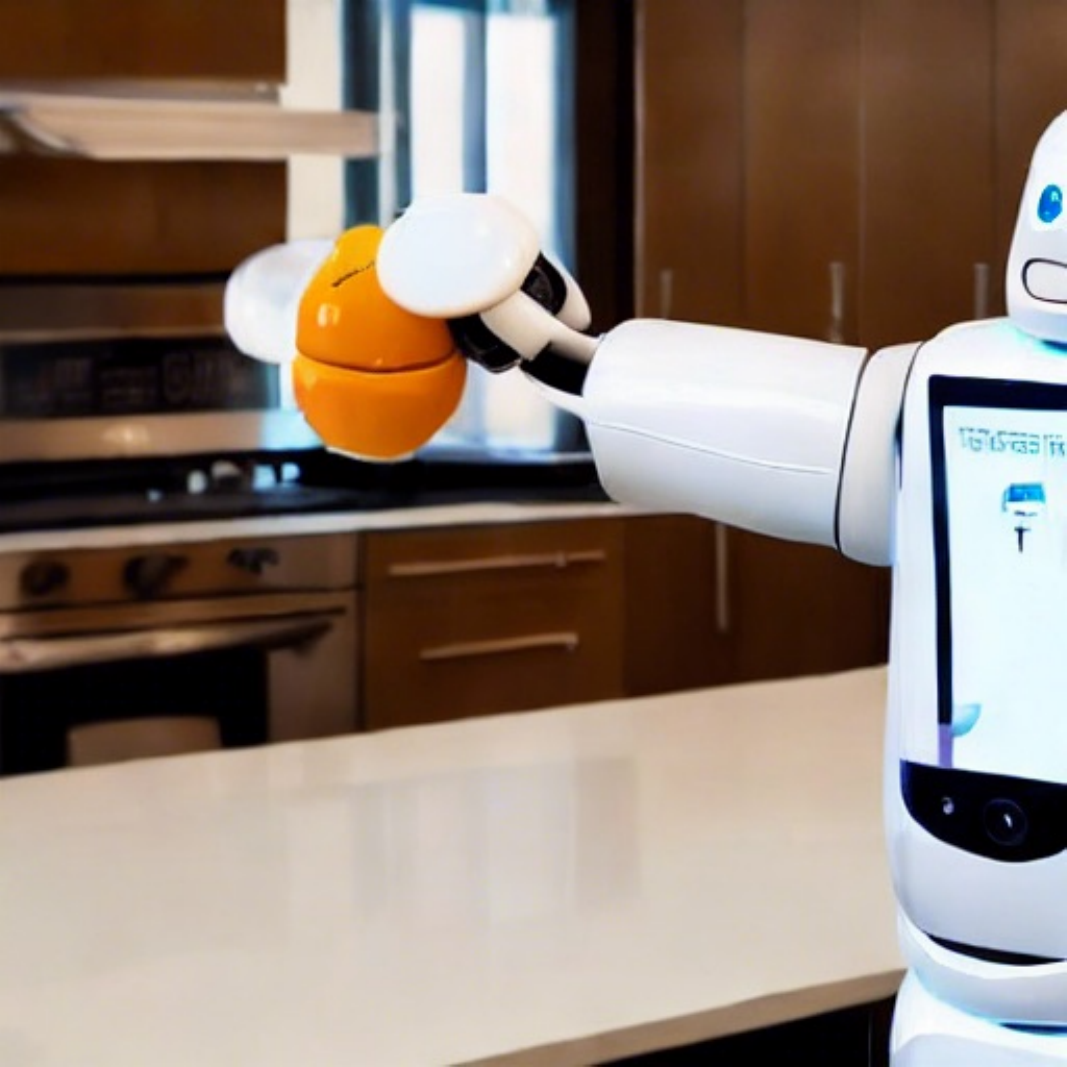}&
       \includegraphics[width=0.18\columnwidth]{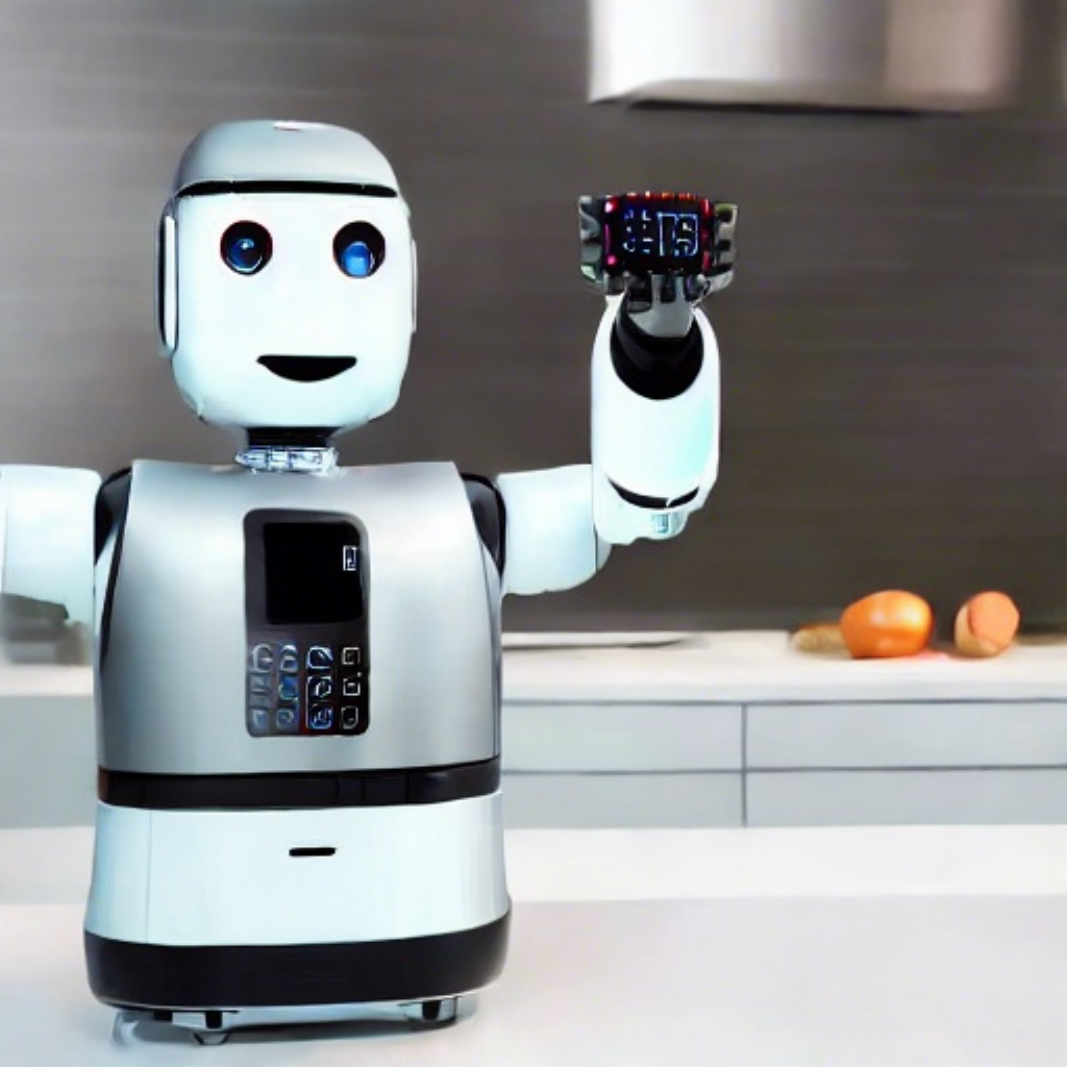}&
       \includegraphics[width=0.18\columnwidth]{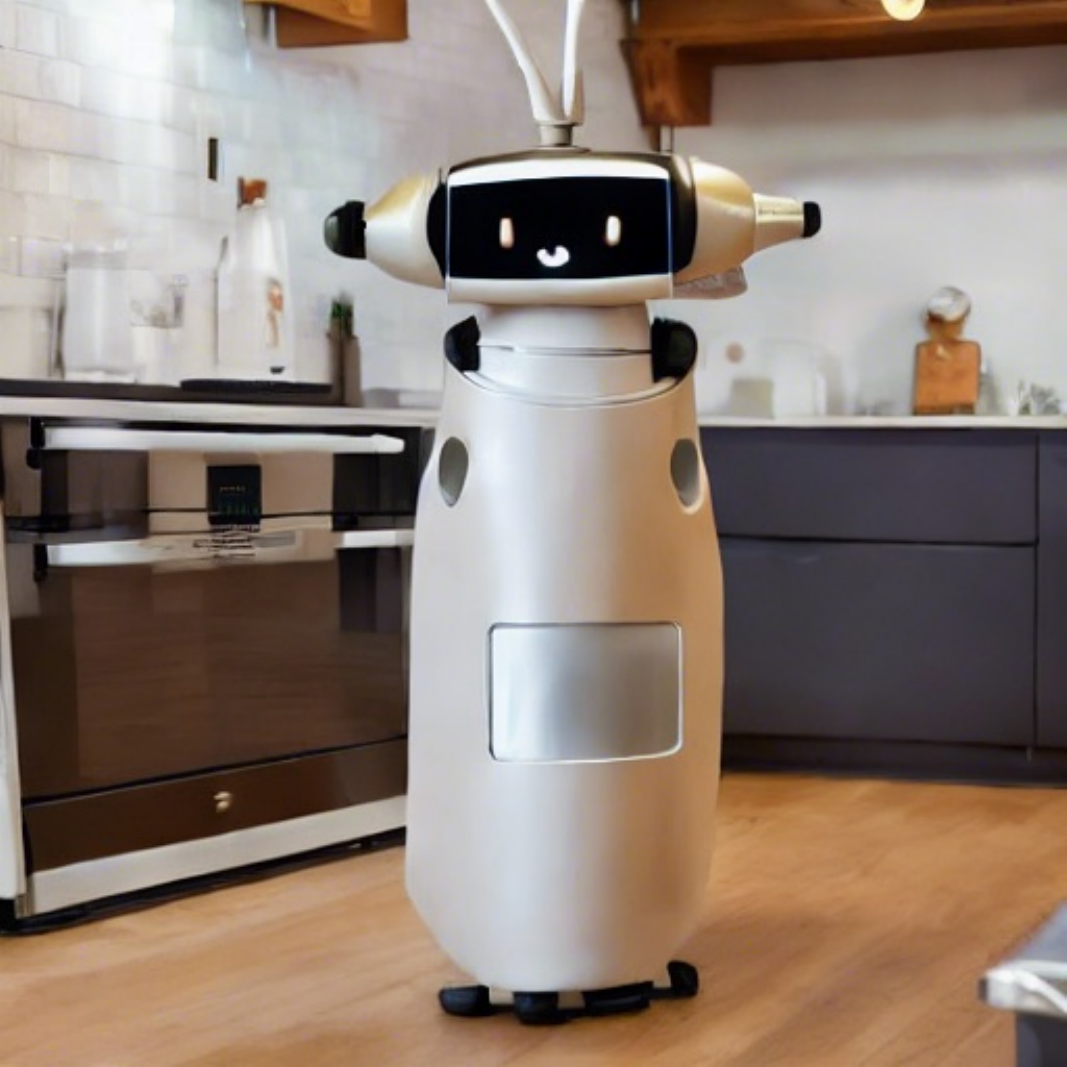}&
       \includegraphics[width=0.18\columnwidth]{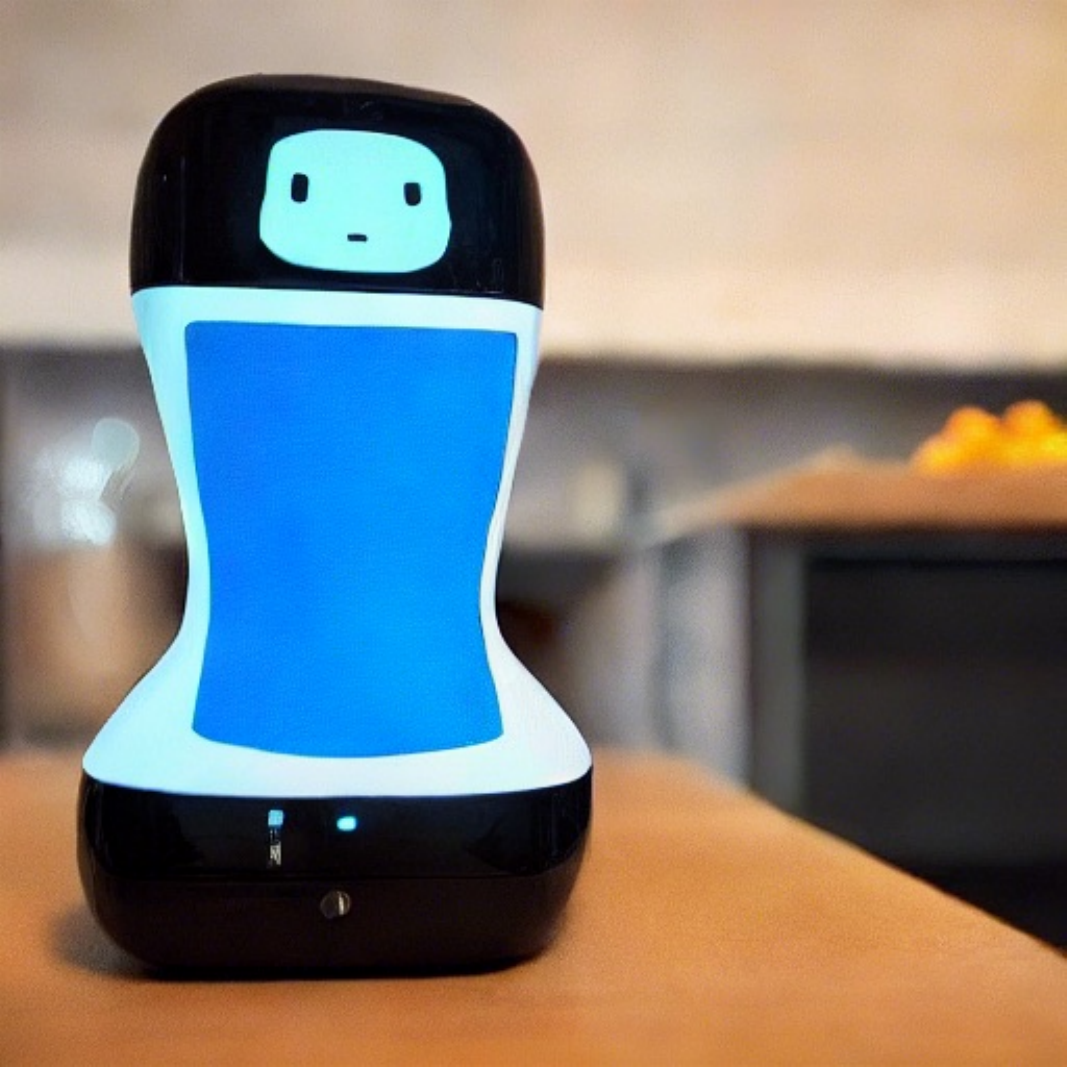}\\[-1pt]
     \end{tabular}
     \caption{\textit{``A robot assistant helping in a modern kitchen."}}
     \label{fig:robot}
   \end{subfigure}
   \end{center}
\end{figure*}
 
\clearpage
\begin{figure*}[p]\ContinuedFloat
   \centering
   \captionsetup[subfigure]{font=small, labelformat=empty,justification=centering}
   \vspace{5mm}
   \begin{subfigure}[c]{\linewidth}
      \centering
      \begin{tabular}{@{}c@{\hspace{1mm}}c@{\hspace{1mm}}c@{\hspace{1mm}}c@{\hspace{1mm}}c@{}}
        \includegraphics[width=0.18\columnwidth]{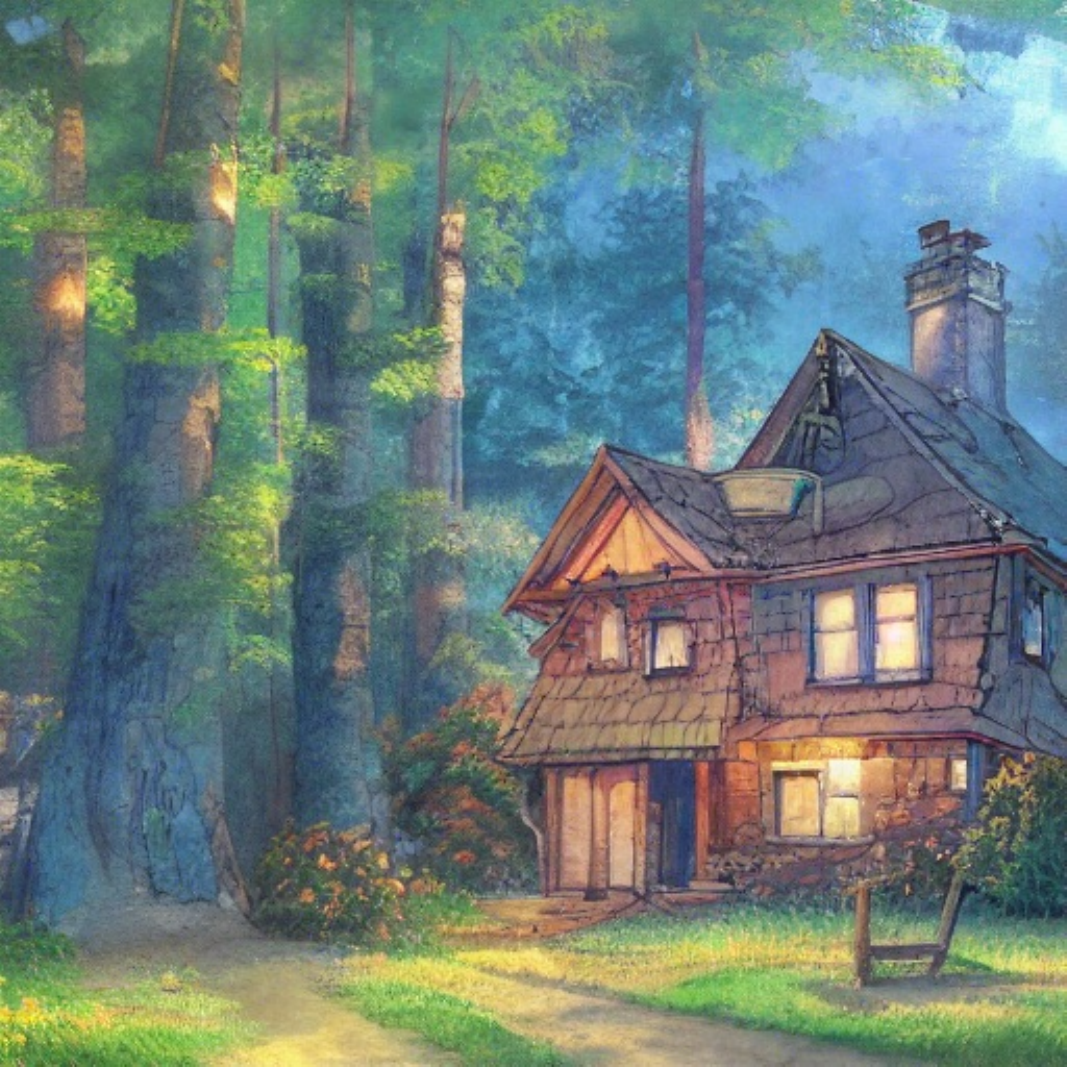}&
        \includegraphics[width=0.18\columnwidth]{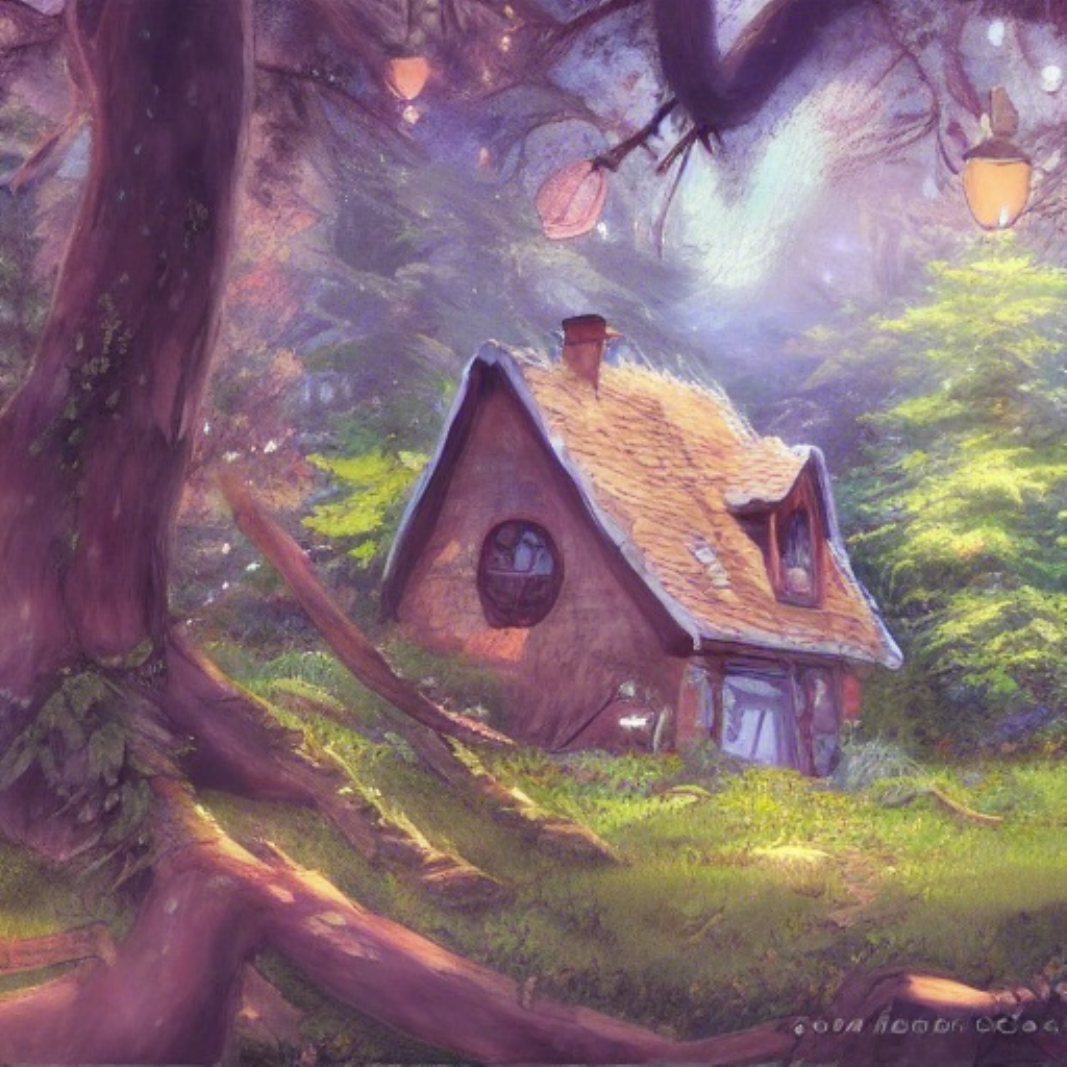}&
        \includegraphics[width=0.18\columnwidth]{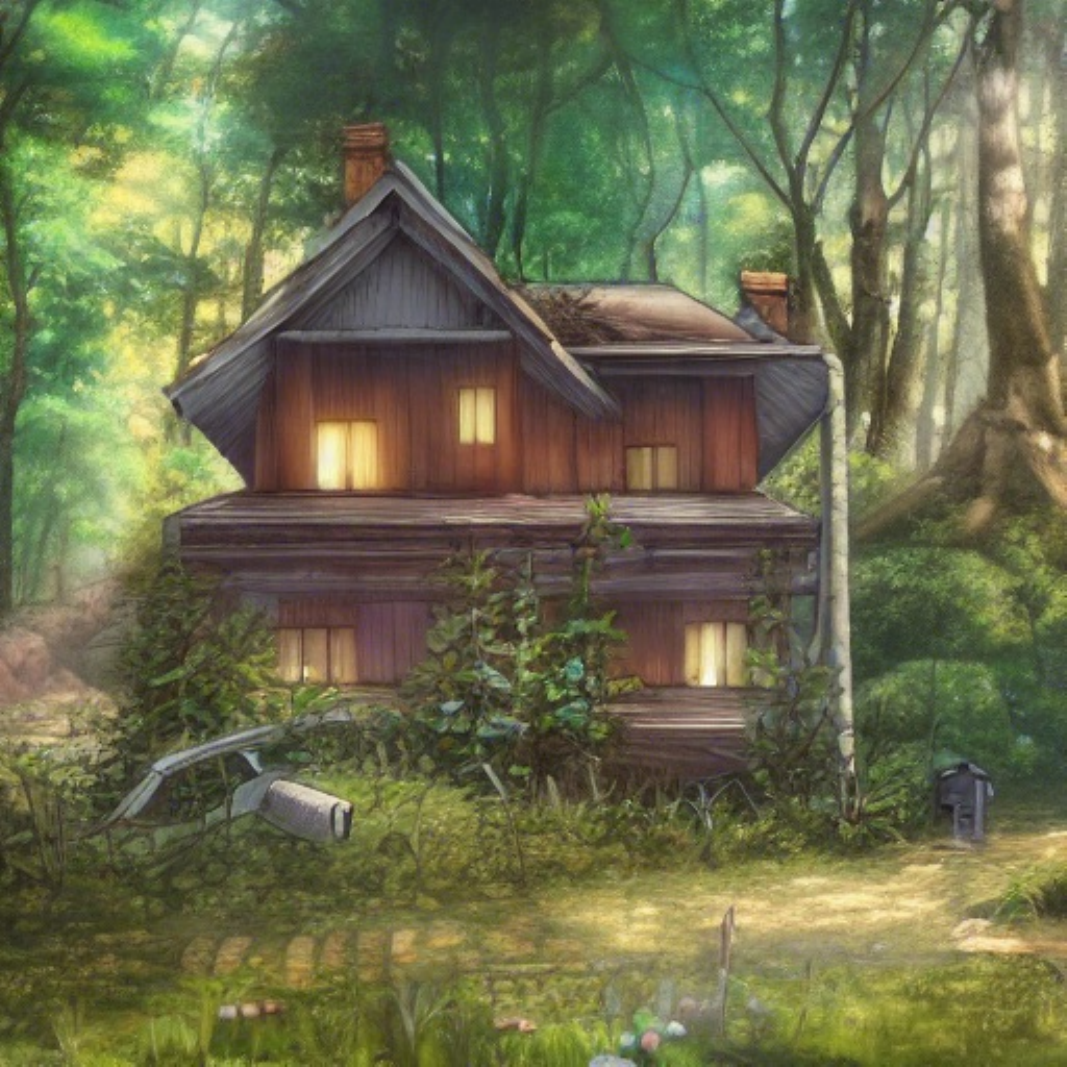}&
        \includegraphics[width=0.18\columnwidth]{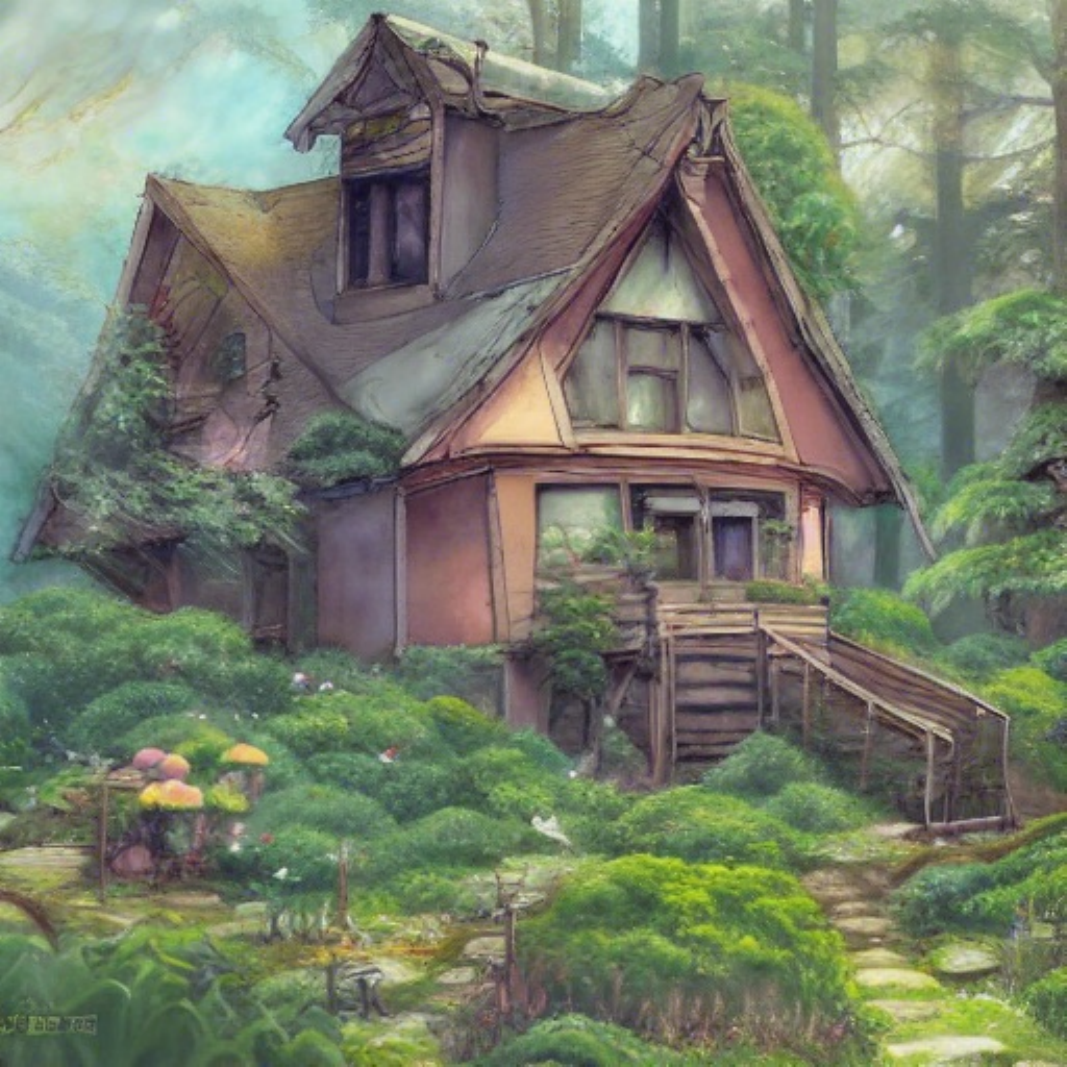}&
        \includegraphics[width=0.18\columnwidth]{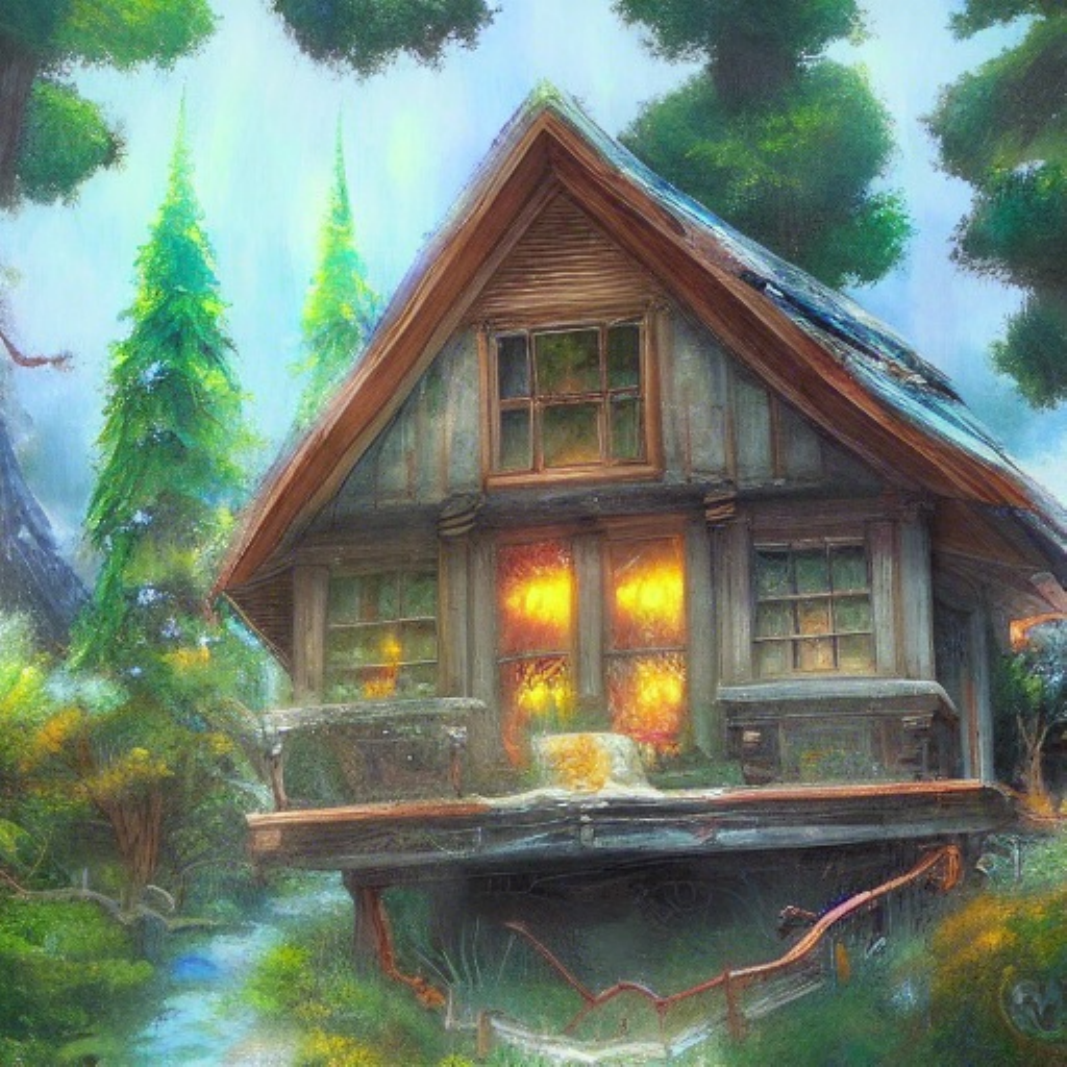}\\[-1pt]
        \includegraphics[width=0.18\columnwidth]{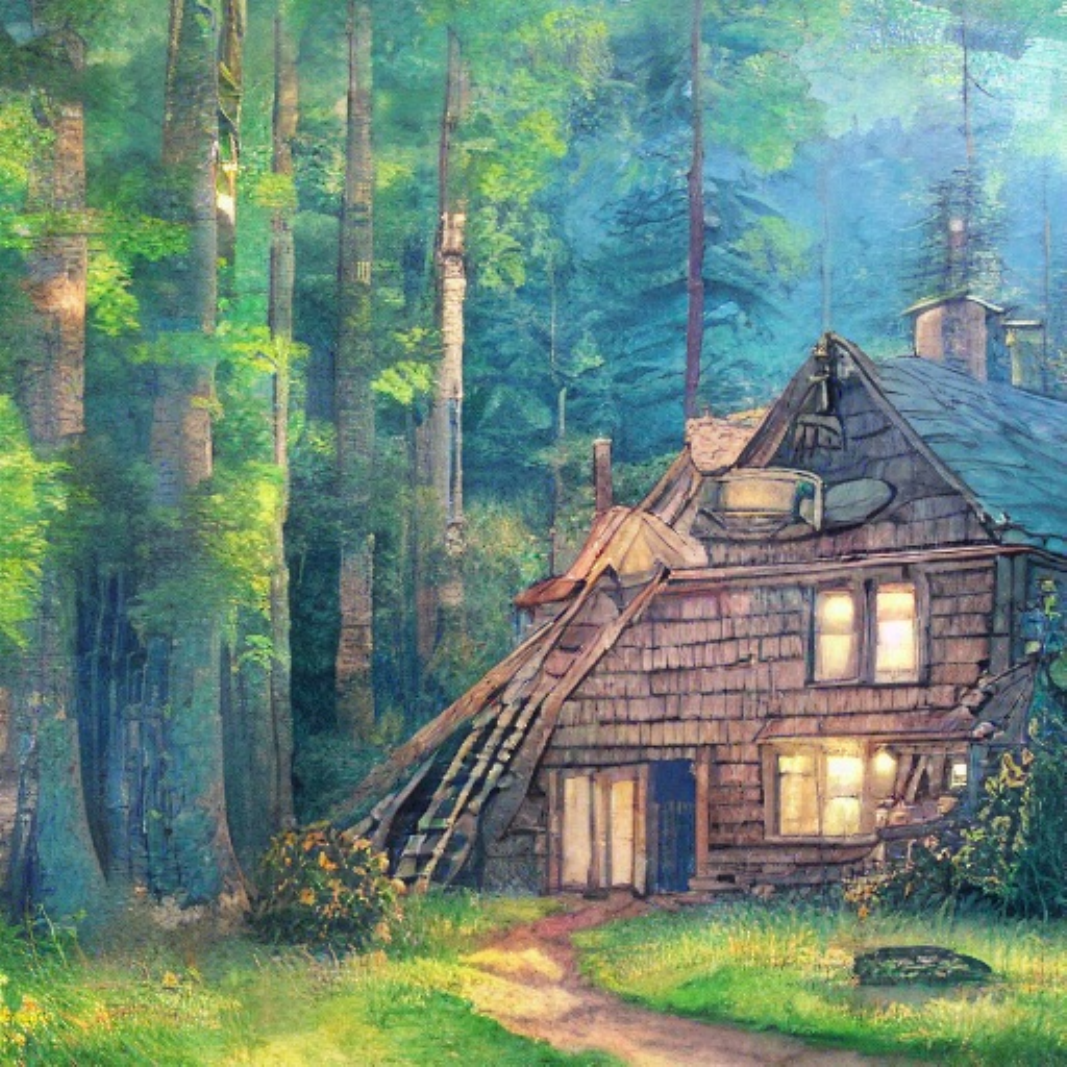}&
        \includegraphics[width=0.18\columnwidth]{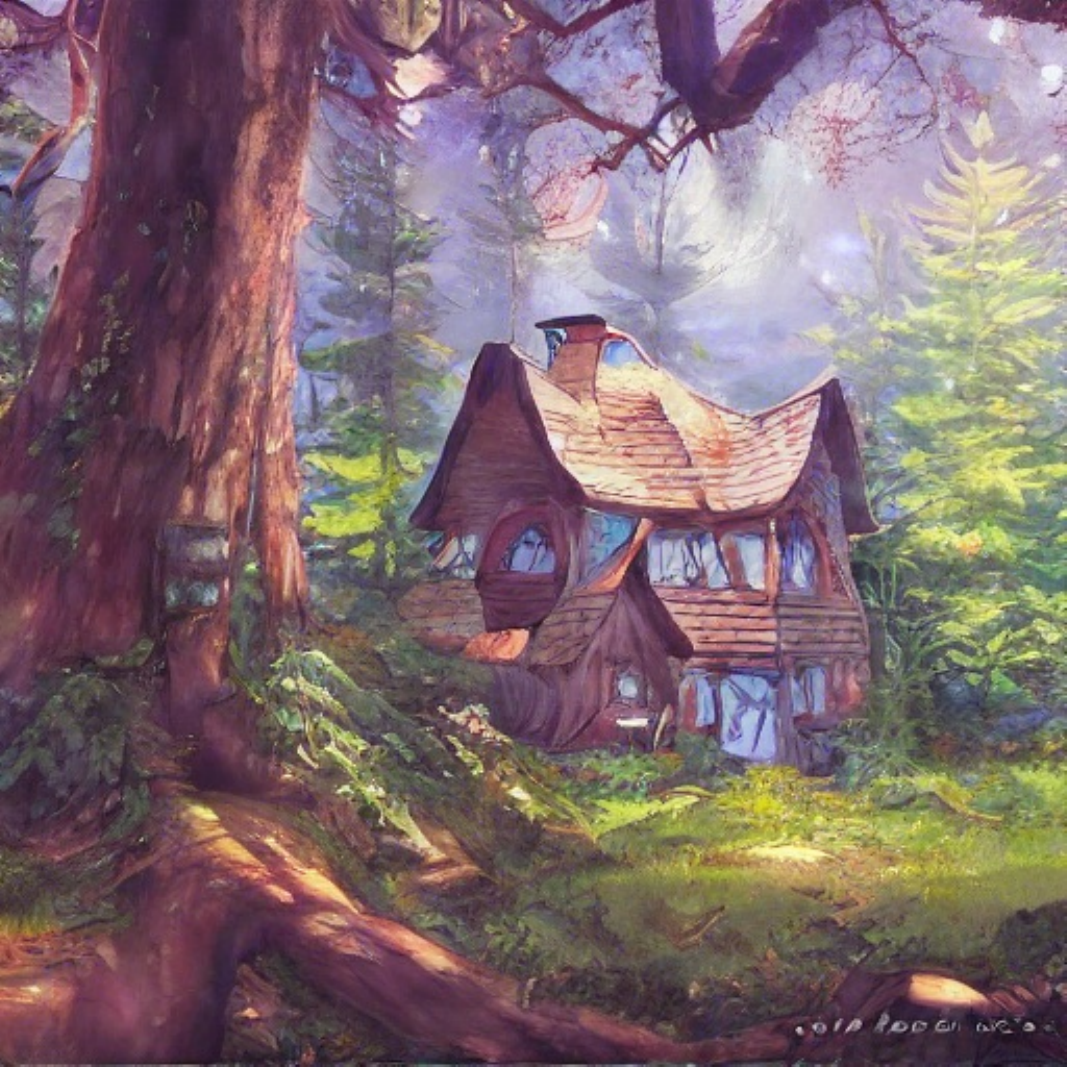}&
        \includegraphics[width=0.18\columnwidth]{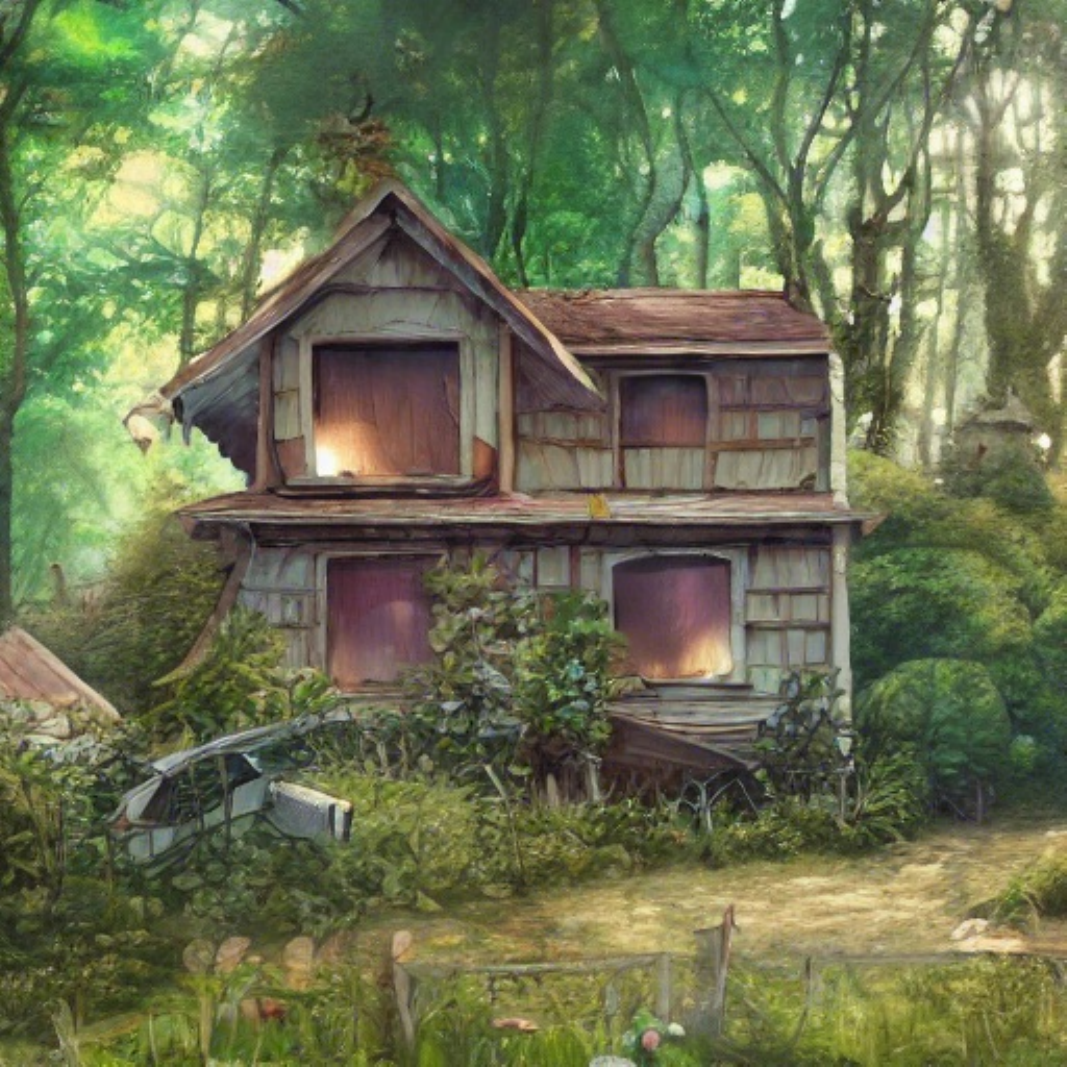}&
        \includegraphics[width=0.18\columnwidth]{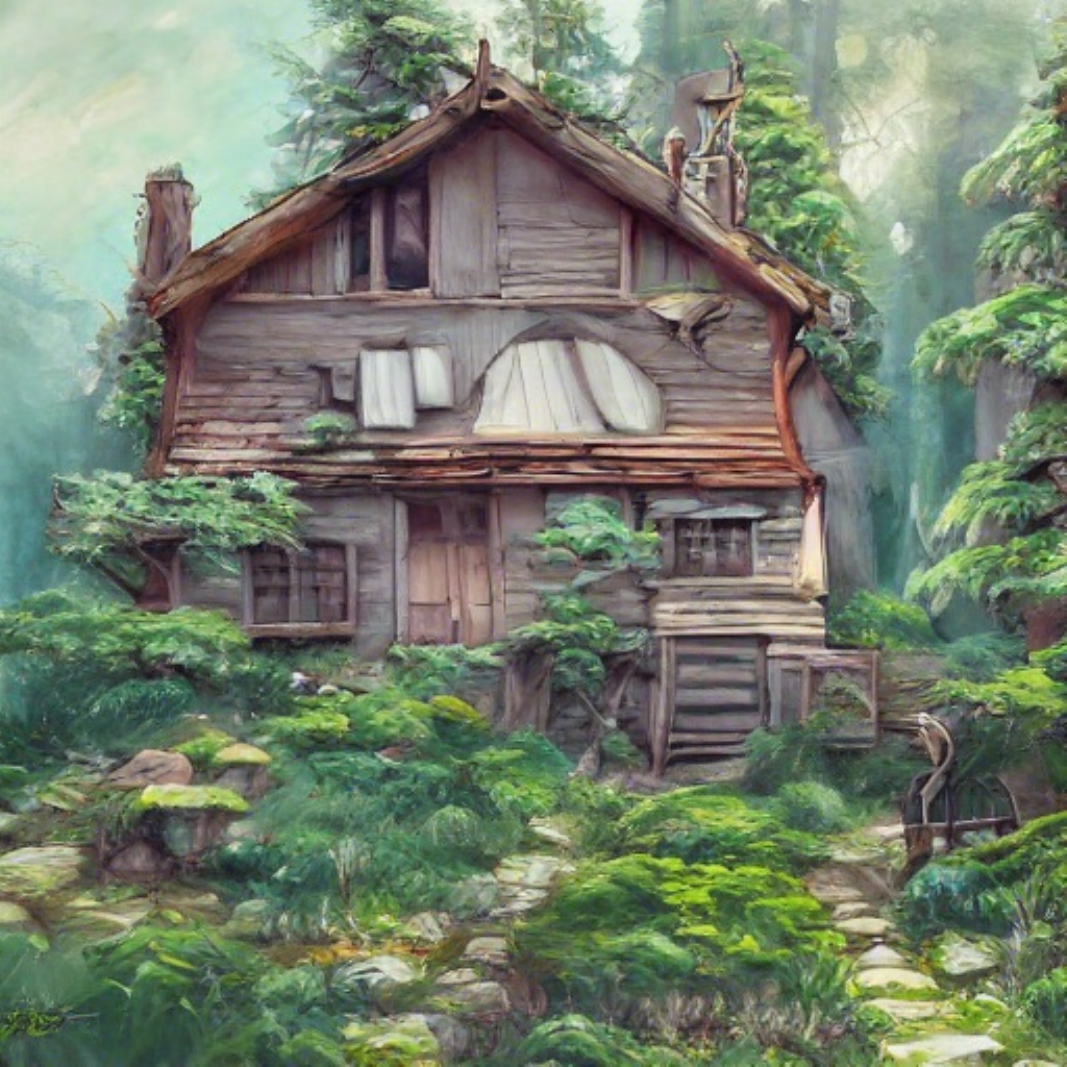}&
        \includegraphics[width=0.18\columnwidth]{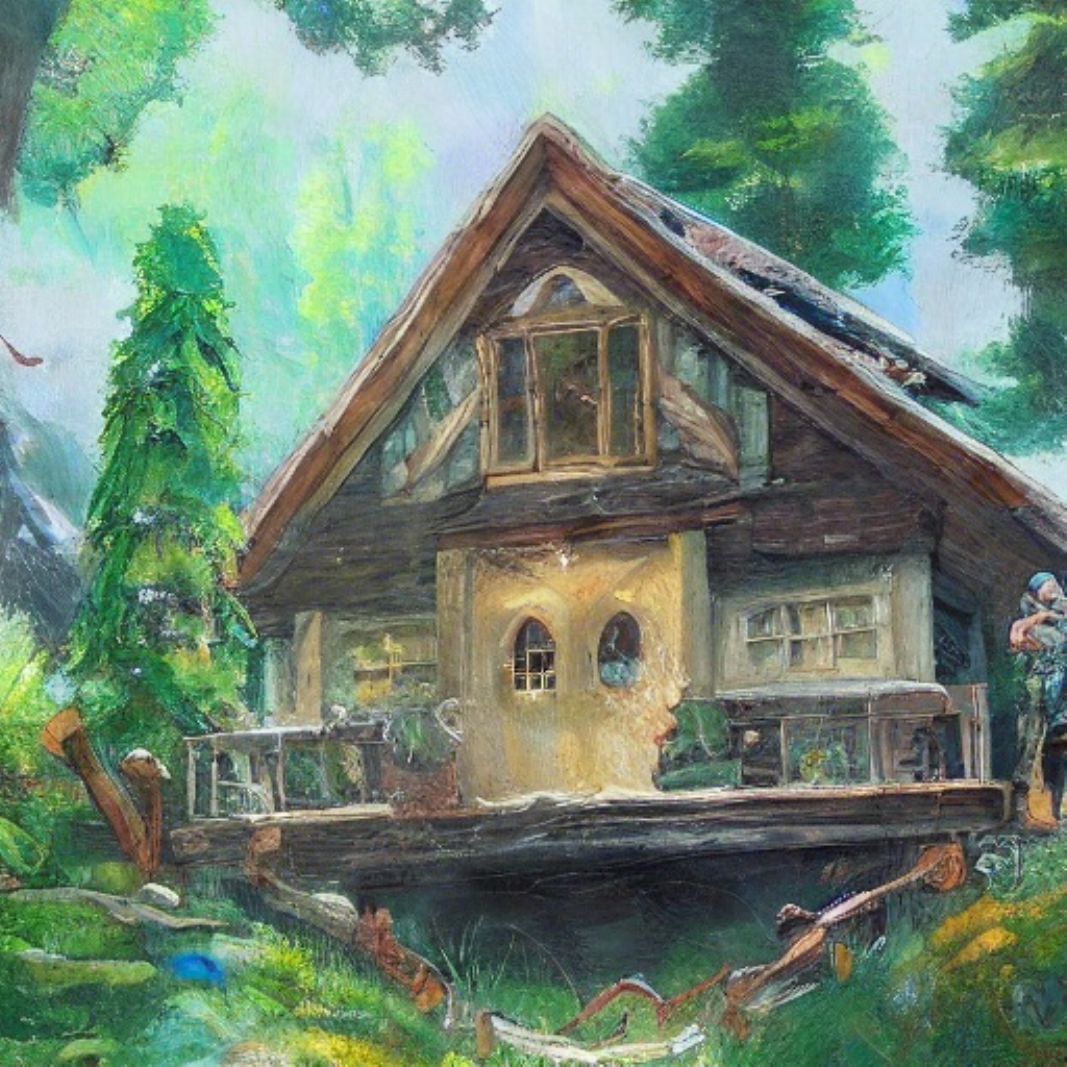}\\[-1pt]
        \includegraphics[width=0.18\columnwidth]{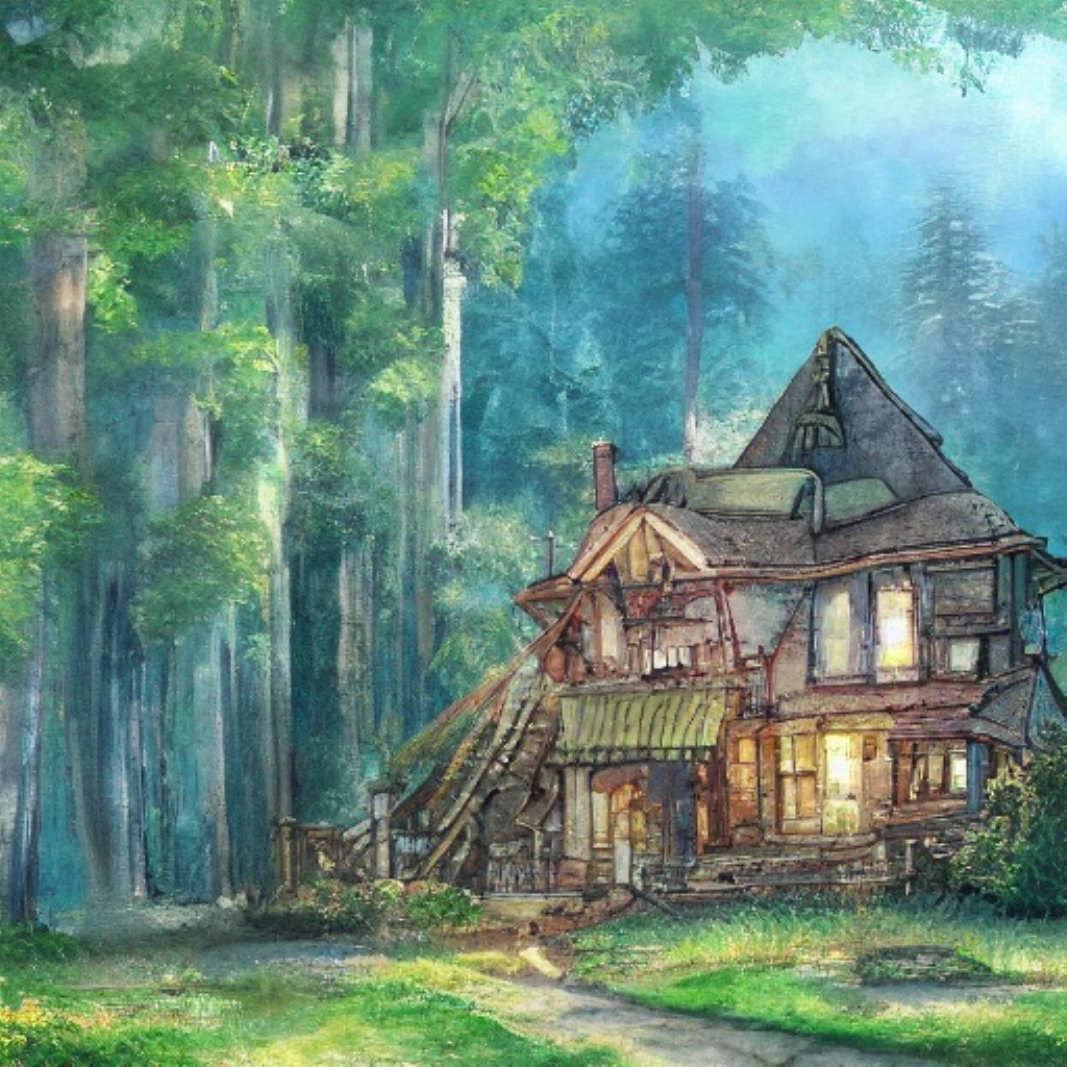}&
        \includegraphics[width=0.18\columnwidth]{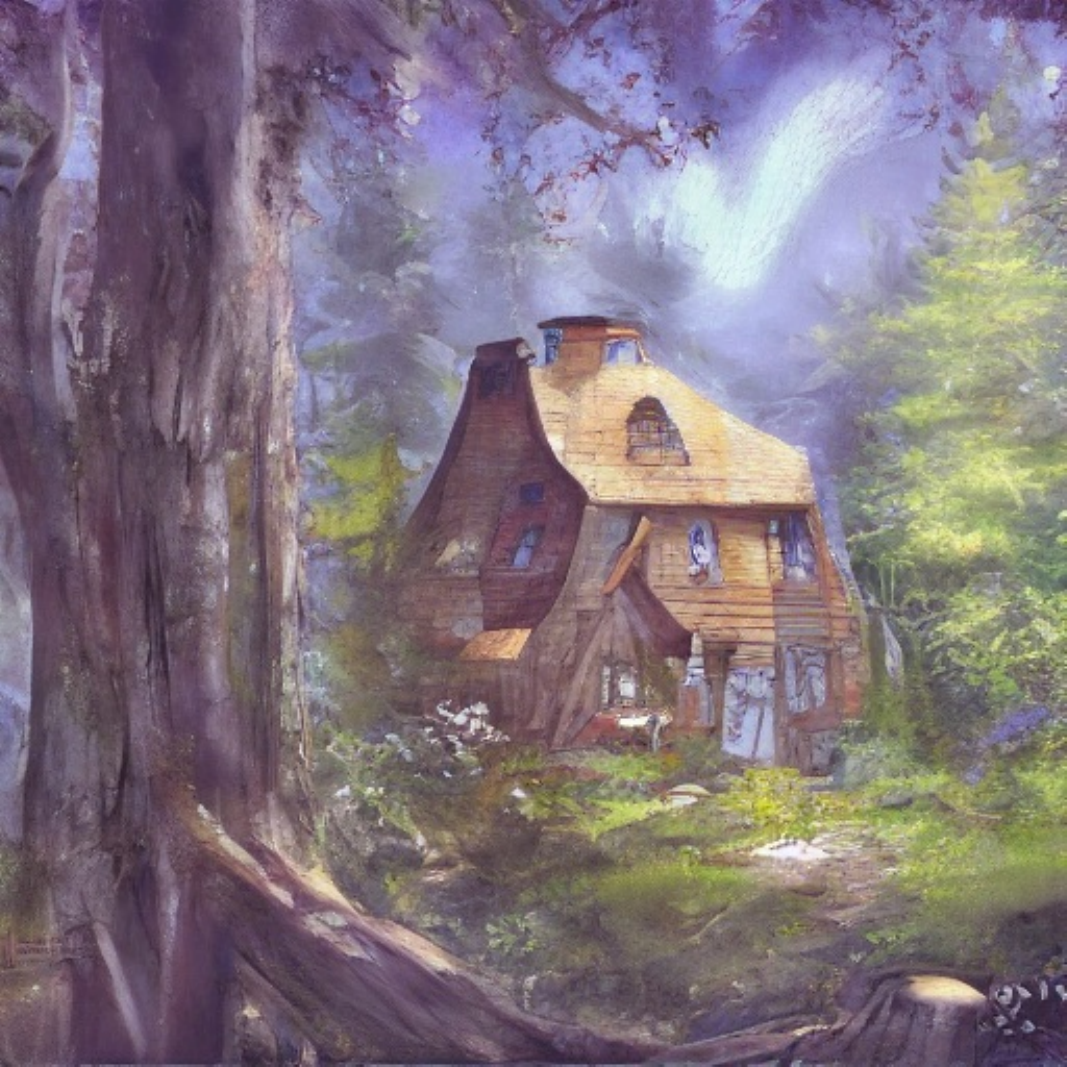}&
        \includegraphics[width=0.18\columnwidth]{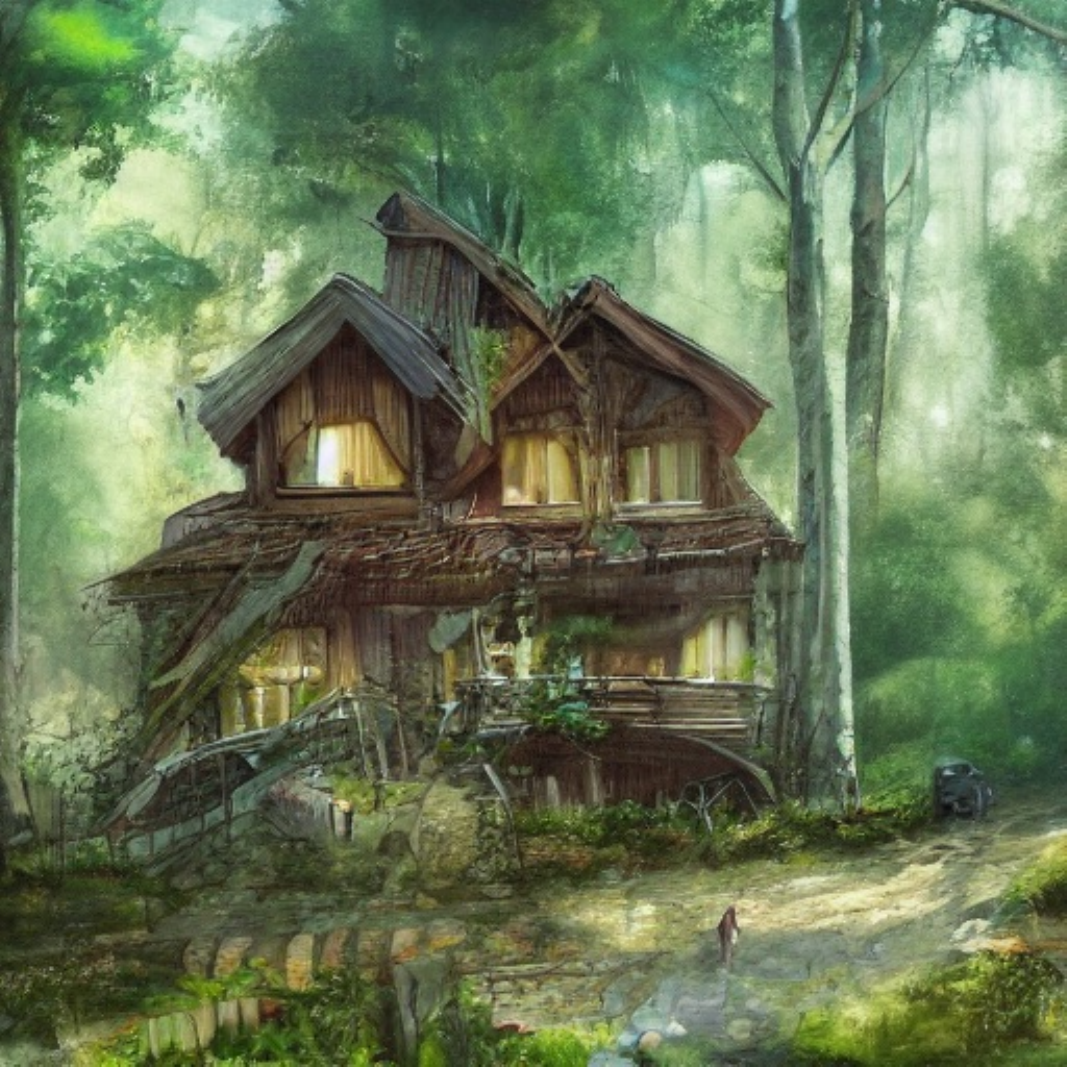}&
        \includegraphics[width=0.18\columnwidth]{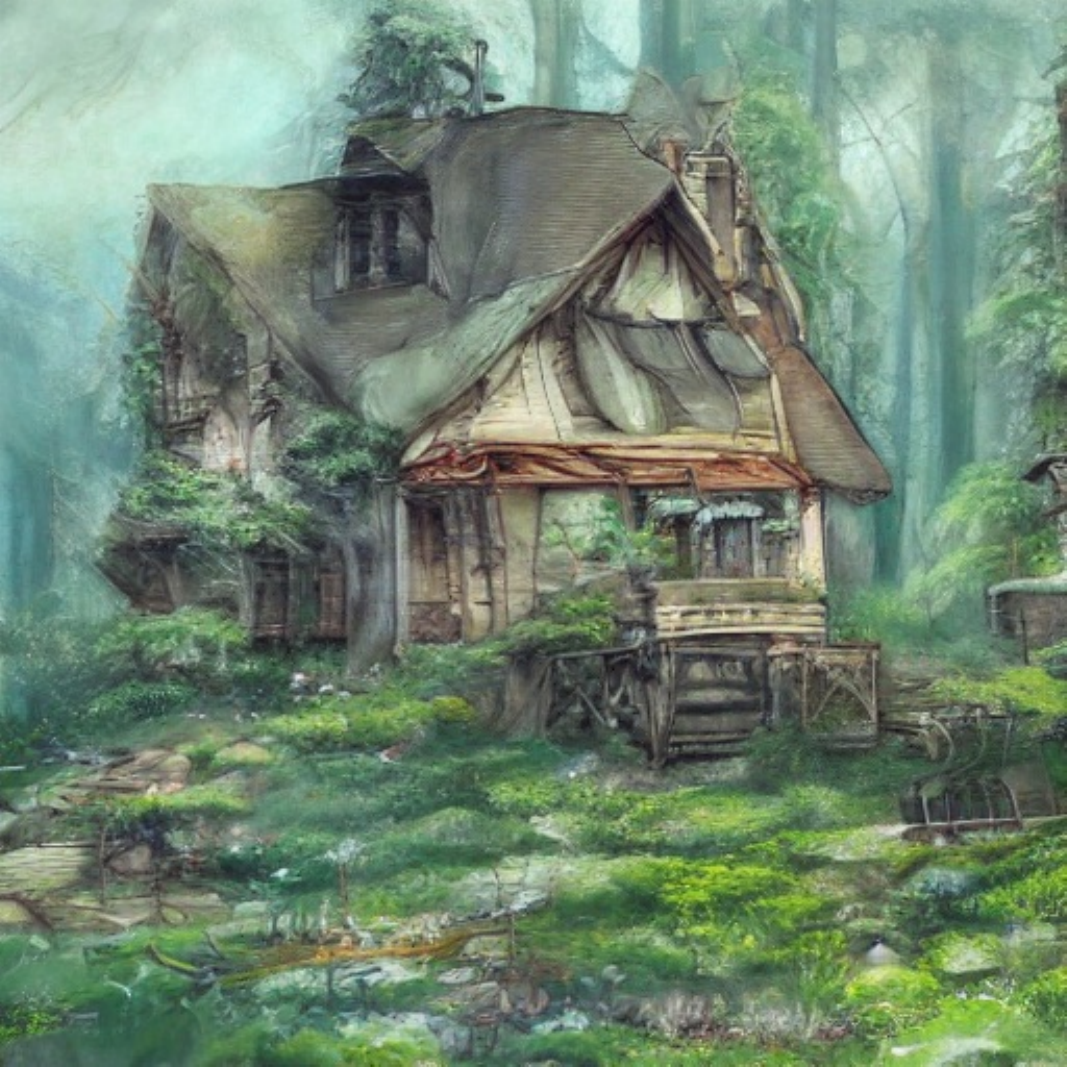}&
        \includegraphics[width=0.18\columnwidth]{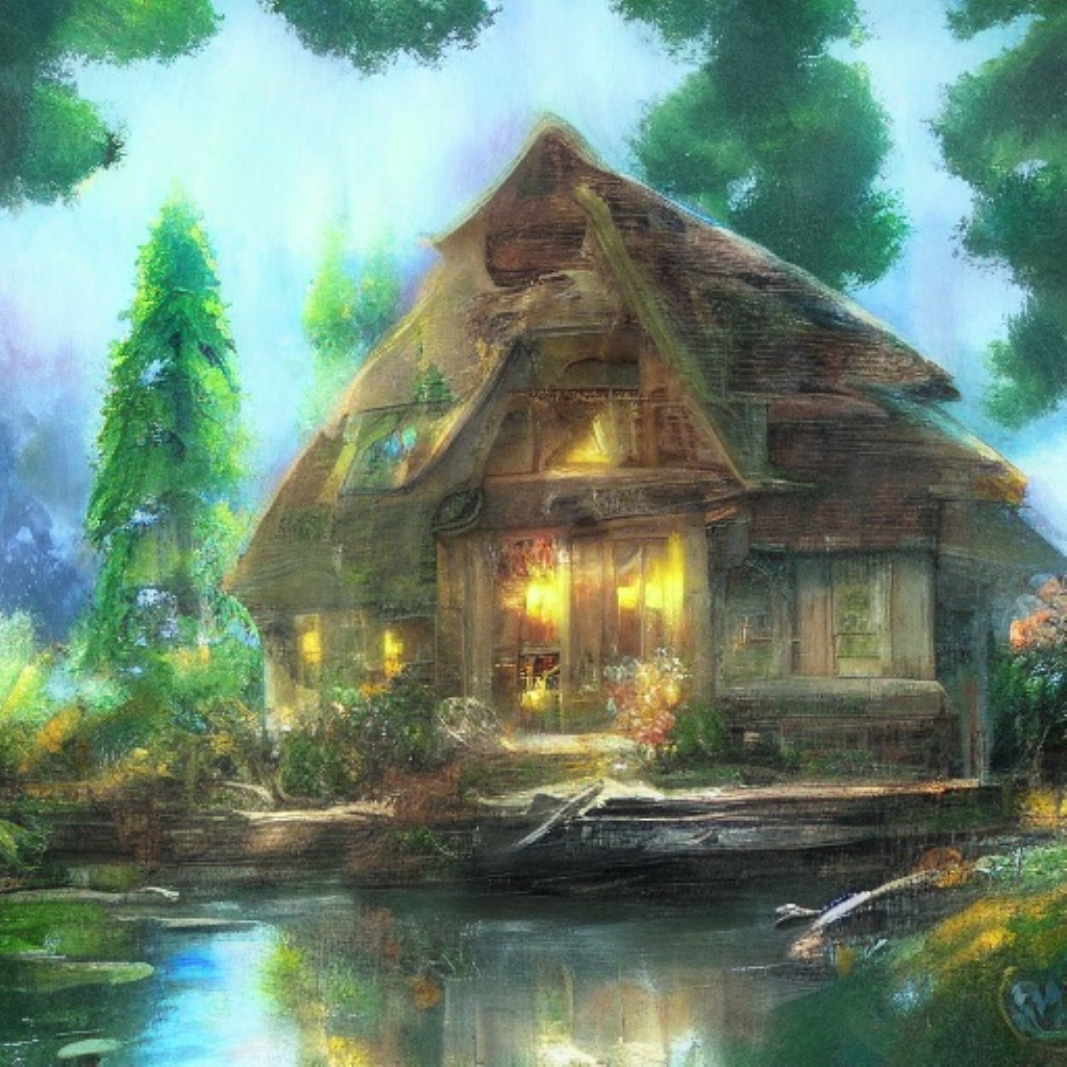}\\[-1pt]
        \includegraphics[width=0.18\columnwidth]{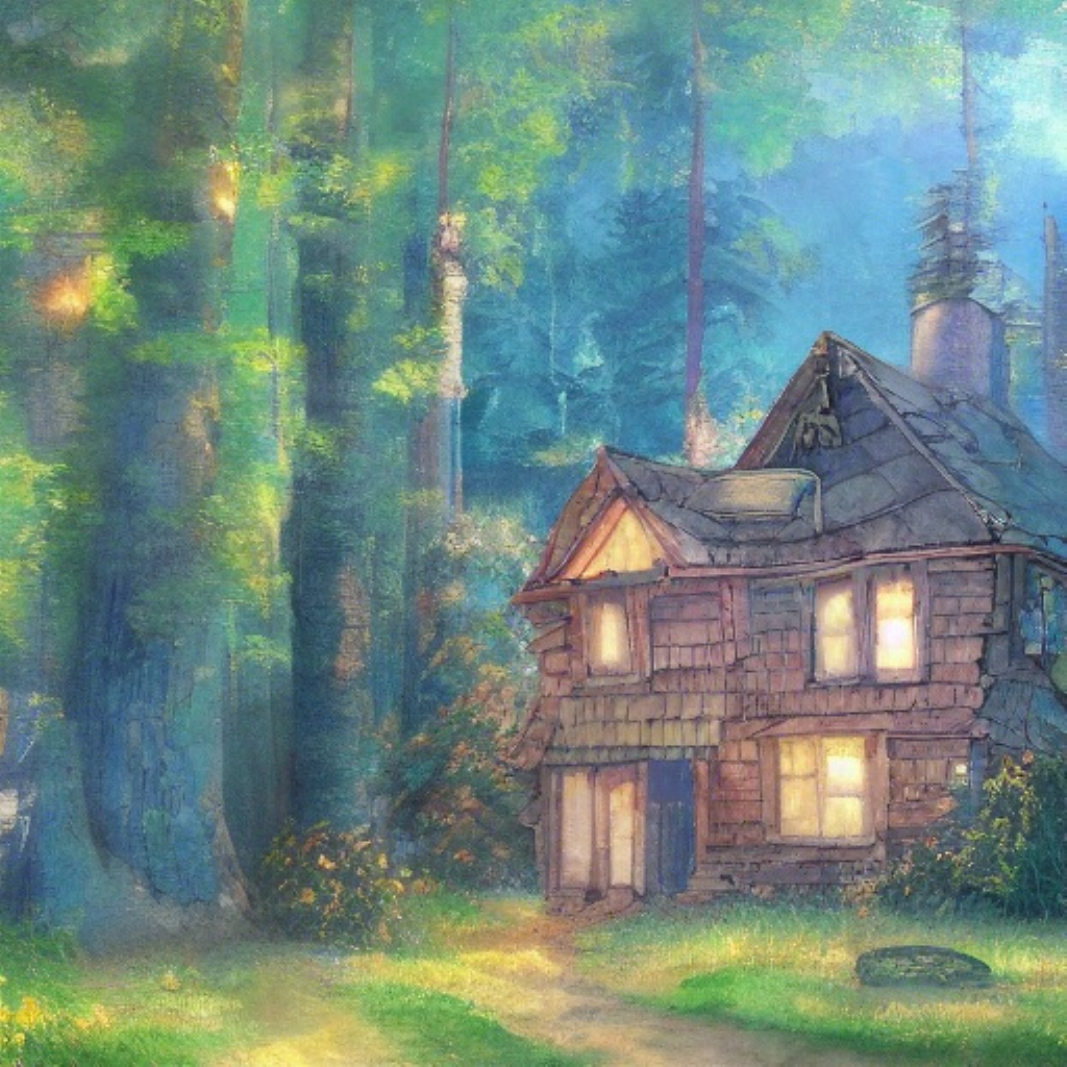}&
        \includegraphics[width=0.18\columnwidth]{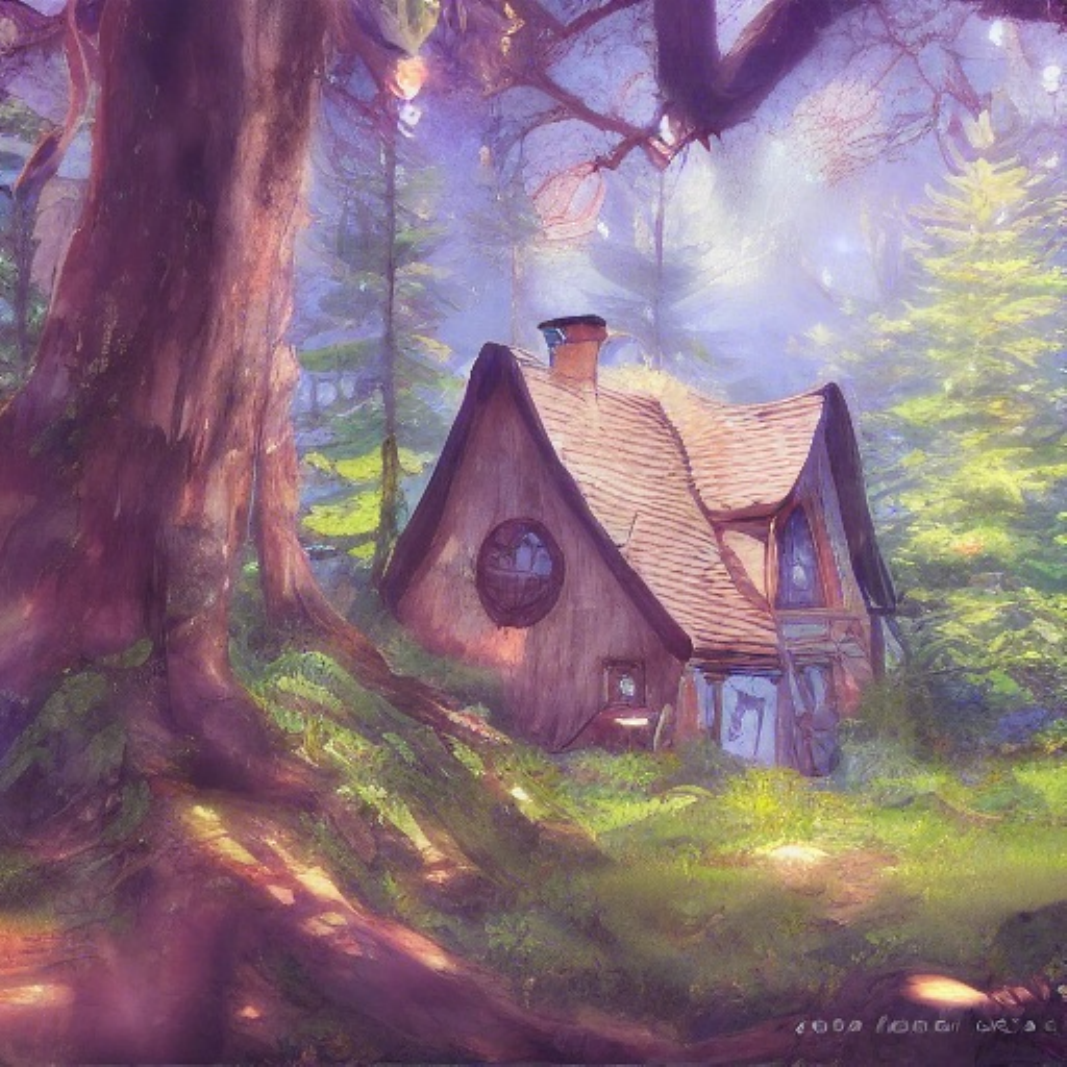}&
        \includegraphics[width=0.18\columnwidth]{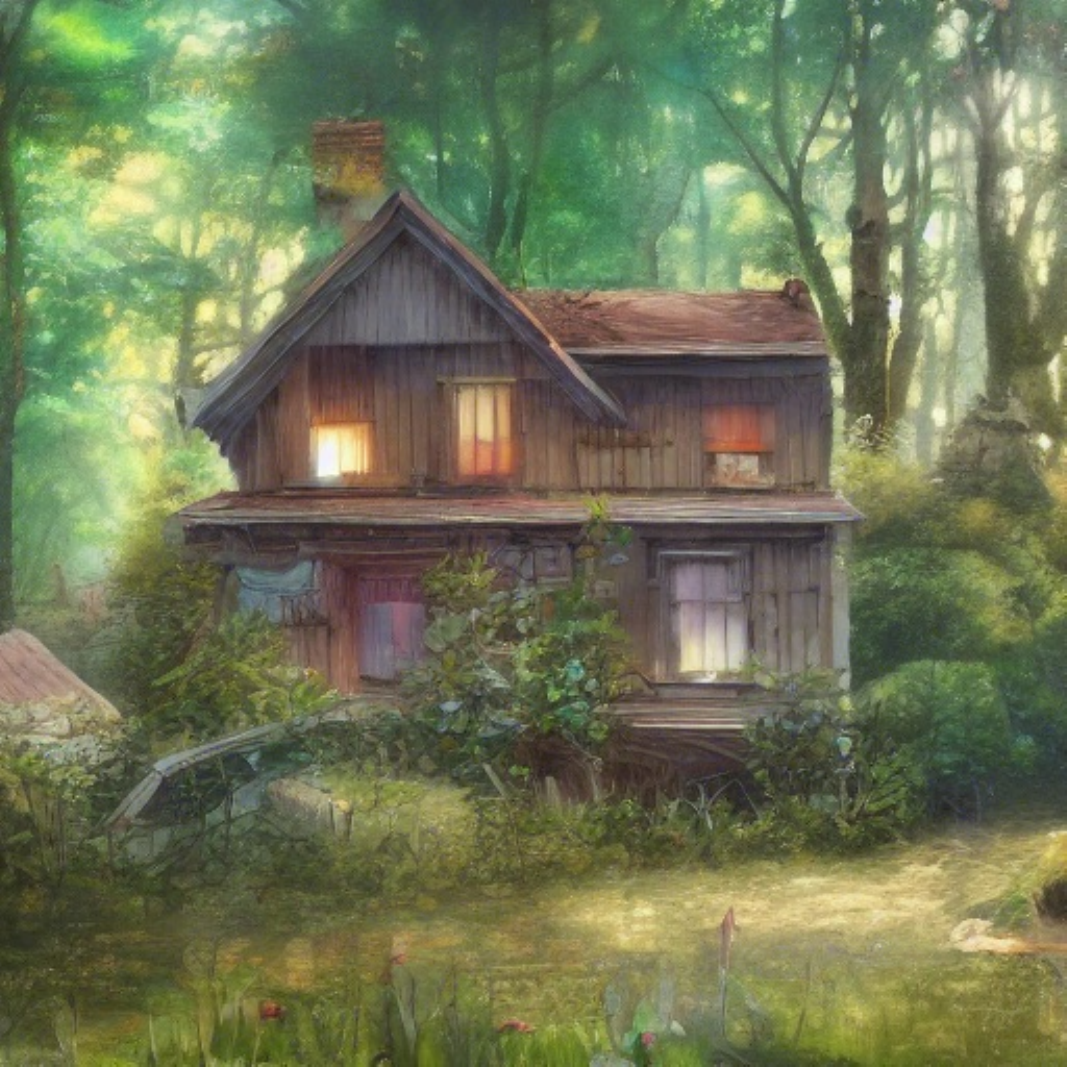}&
        \includegraphics[width=0.18\columnwidth]{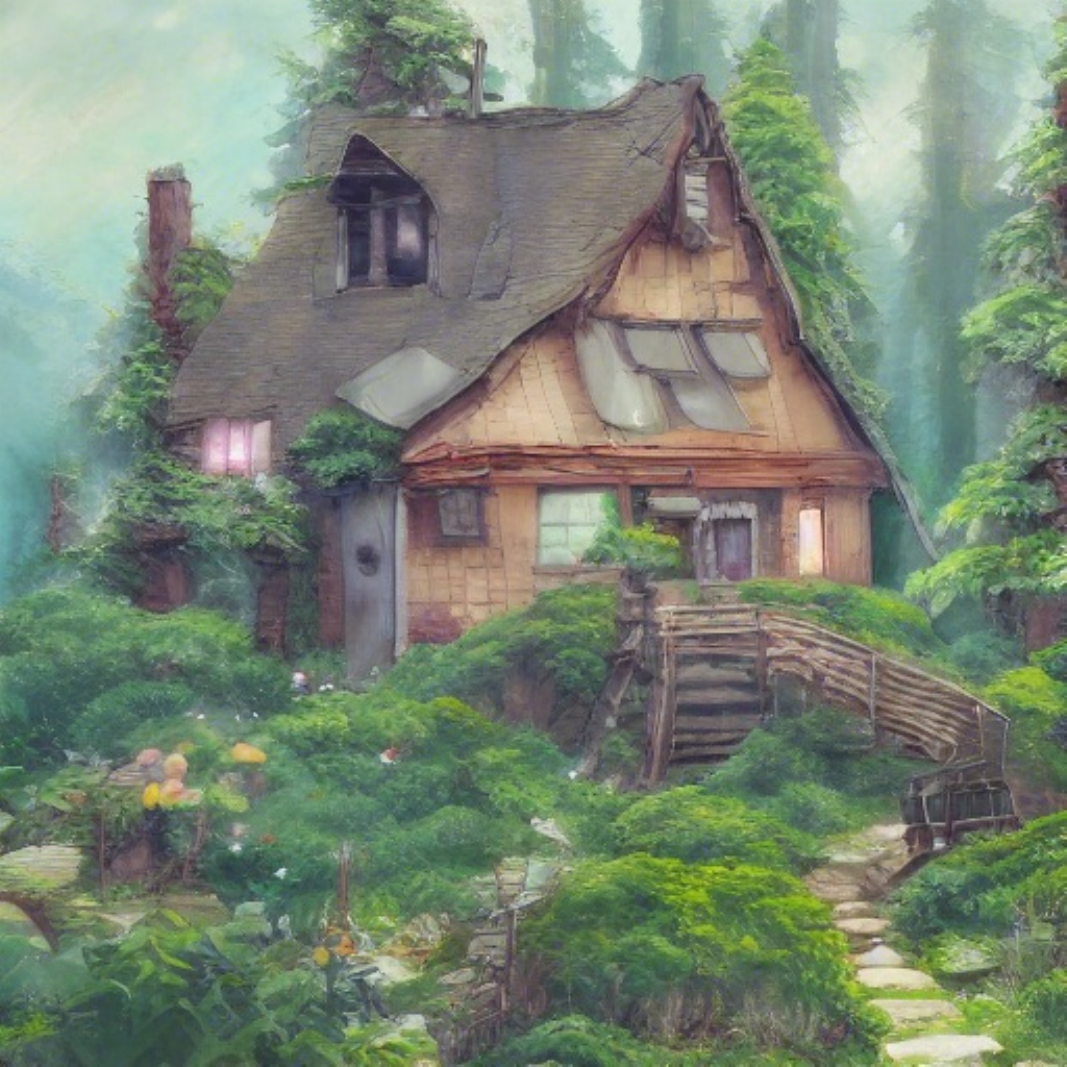}&
        \includegraphics[width=0.18\columnwidth]{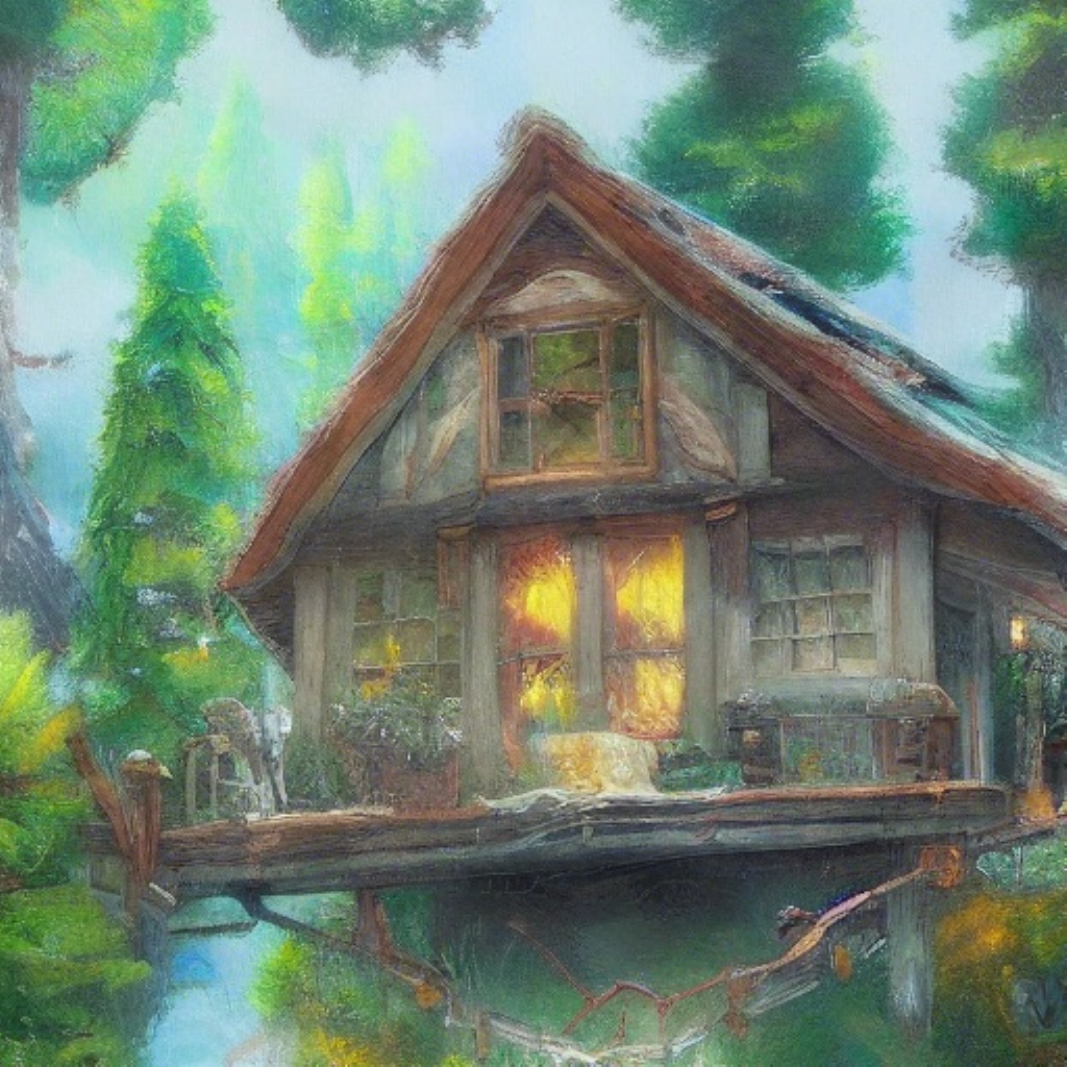}\\[-1pt]
      \end{tabular}
      \caption{\textit{``Cluttered house in the woods, anime, oil painting, high resolution, cottagecore, ghibli inspired"}}
      \label{fig:house}
    \end{subfigure}
 
   \caption{Visual comparisons of generated images~(512$\times$512) conditioned on user defined prompts. We generate images by SD v1.4~\cite{rombach2022high} and its quantized versions using Q-Diffusion~\cite{li2023q} and AccuQuant under a 4/8-bit setting and PCR~\cite{tang2025post} in 4/8.4-bit setting. Each row corresponds to Full Precision, Q-Diffusion~\cite{li2023q}, PCR($\tau=0.2$)~\cite{tang2025post} and Ours.}
   \label{fig:appendix_sd_user_prompts}
 \end{figure*}

\end{document}